\documentclass[smallcondensed]{svjour3}     
\RequirePackage{fix-cm}
\let\chapter\section
\usepackage{graphicx}
\usepackage{mathptmx}      
\usepackage{floatrow}
\usepackage{latexsym}
\usepackage{multirow}
\usepackage[cmex10]{amsmath}
\usepackage{amsfonts}
\usepackage[ruled,vlined,noend,linesnumbered]{algorithm2e}
\usepackage{array}
\usepackage{mdwmath}
\usepackage{mdwtab}
\usepackage[caption=false,font=footnotesize]{subfig}
\usepackage{stfloats}
\usepackage{url}
\usepackage{natbib}
\usepackage{wrapfig}
\usepackage{bm}
\usepackage{lipsum}
\newcommand{\fancym}{\mathcal{M}}
\usepackage{enumitem}
\setlist{parsep=0pt,listparindent=\parindent}

\begin{document}

\title{Minimum message length estimation of mixtures of multivariate 
Gaussian and von Mises-Fisher distributions}
\titlerunning{Mixture modelling using MML}      

\author{Parthan Kasarapu \and
        Lloyd Allison
}
\institute{P. Kasarapu  \at
           Faculty of Information Technology, Monash University, VIC 3800, Australia \\
              \email{parthan.kasarapu@monash.edu}           
           \and
           L. Allison \at
           Faculty of Information Technology, Monash University, VIC 3800, Australia \\
           \email{lloyd.allison@monash.edu}           
}

\date{ }

\maketitle

\begin{abstract} 
Mixture modelling involves explaining some observed evidence using a combination of
probability distributions.
The crux of the problem is the inference of an optimal number of mixture components and
their corresponding parameters.  This paper discusses 
unsupervised learning of mixture models using the Bayesian
Minimum Message Length (MML) criterion. To demonstrate the effectiveness of
search and inference of mixture parameters using the proposed approach, we 
select two key probability distributions, each handling fundamentally 
different types of data: the multivariate Gaussian distribution to address 
mixture modelling of data distributed in Euclidean space, and the multivariate 
von Mises-Fisher (vMF) distribution to address mixture modelling of 
directional data distributed on a unit hypersphere. The key contributions of this paper,
in addition to the general 
search and inference methodology, include the derivation of MML expressions for
encoding the data using multivariate Gaussian and von Mises-Fisher distributions,
and the analytical derivation of the MML estimates of the 
parameters of the two distributions.
Our approach is tested on simulated and real world data sets.
For instance, we infer vMF 
mixtures that concisely explain experimentally determined 
three-dimensional protein conformations, providing an effective \emph{null model}
description of protein structures that is central to many inference problems in 
structural bioinformatics. The 
experimental results demonstrate that the performance of our proposed search 
and inference method along with the encoding schemes improve on the state of the art
mixture modelling techniques.
\keywords{mixture modelling \and minimum message length \and multivariate Gaussian \and von Mises-Fisher
\and directional data}
\end{abstract}

\section{Introduction} \label{sec:intro}
Mixture models are common tools
in statistical pattern recognition \citep{mclachlan1988mixture}.
They offer a mathematical basis to explain data in fields as diverse as
astronomy, biology, ecology, engineering, and economics, amongst many others 
\citep{mclachlan2000finite}. A mixture model is composed 
of component probabilistic models;
a component may variously correspond to
a subtype, kind, species, or subpopulation of the observed data.
These models aid in the identification 
of hidden patterns in the data through sound probabilistic formalisms.
Mixture models have been extensively used in machine learning tasks
such as classification and unsupervised learning 
\citep{titterington1985statistical,mclachlan2000finite,jain2000statistical}.

Formally, mixture modelling involves representing a 
distribution of data as a weighted sum of individual probability distributions.
Specifically, the problem we consider here is to model the observed
data using a mixture $\fancym$ of probability distributions of the form: 
\begin{equation}
  \Pr(\mathbf{x};\fancym) =
  \sum_{j=1}^M w_j f_j(\mathbf{x};\Theta_j) \label{eqn:mixture} 
\end{equation}
where $\mathbf{x}$ is a $d$-dimensional datum, $M$ is the number of 
mixture components, $w_j$
and $f_j(\mathbf{x};\Theta_j)$ are the weight 
and probability density of the $j^{\text{th}}$ component respectively;
the weights are positive and sum to one. 

The problem of modelling some observed data using a mixture distribution involves
determining the number of components, $M$, and estimating the mixture
parameters. Inferring the optimal number of mixture 
components involves the difficult problem of balancing the trade-off
between two conflicting objectives: low \textit{hypothesis complexity} as
determined by the number of components \textit{and} their respective parameters, versus
good quality of \textit{fit} to the observed data. Generally a hypothesis with more free 
parameters can fit observed data better than a hypothesis with fewer 
free parameters.
A number of strategies have been used to control this balance as discussed in Section 
\ref{sec:mixture_existing_methods}. These methods provide varied formulations
to assess the mixture components and their ability to explain
the data. Methods using the minimum message length criterion 
\citep{wallace68}, a Bayesian method of inductive 
inference, have been proved to be effective in achieving a reliable 
balance between these conflicting aims \citep{wallace68,oliver1996unsupervised,roberts,figueiredo2002unsupervised}.

Although much of the literature concerns the theory and application
of Gaussian mixtures \citep{mclachlan2000finite,Jain:1988}, mixture 
modelling using other probability distributions has been widely used. 
Some examples are:
Poisson models \citep{wang1996mixed,wedel1993latent},
exponential mixtures \citep{seidel2000cautionary},
Laplace \citep{jones1990laplace},
t-distribution \citep{peel2000robust},
Weibull \citep{patra1999multivariate},
Kent \citep{peel2001fitting},
von Mises-Fisher \citep{Banerjee:clustering-hypersphere}, and many more. 
\cite{mclachlan2000finite} provide a comprehensive summary of finite
mixture models and their variations.
The use of Gaussian mixtures
in several research disciplines has been partly motivated by its computational 
tractability \citep{mclachlan2000finite}.
For datasets where 
the \textit{direction} of the constituent vectors is important, Gaussian mixtures 
are inappropriate and distributions such as the von Mises-Fisher may be used
\citep{Banerjee:clustering-hypersphere,mardia1979multivariate}. 
In any case, whatever the kind of distribution used for an individual component,
one needs to estimate the parameters of the mixture,
and provide a sound justification for selecting the 
appropriate number of mixture components.

Software for mixture modelling relies on the following elements:
\begin{enumerate}
\item An \emph{estimator} of the parameters of each component of a mixture,
\item An \emph{objective function}, that is a cost or score, that can be used to compare
two hypothetical mixtures and decide which is better, and
\item A \emph{search strategy} for the best number of components and their weights.
\end{enumerate}
Traditionally, parameter estimation is done by strategies such as maximum likelihood (ML)
or Bayesian maximum \emph{a posteriori} probability (MAP) estimation. 
In this work, we use the Bayesian minimum message length (MML) principle.
Unlike MAP, MML estimators are invariant under 
non-linear transformations of the data \citep{oliver1994mml}, and unlike ML,
MML considers the number and precision of a model's parameters.
It has been used in the inference of several probability distributions \citep{WallaceBook}.
MML-based inference operates by considering the 
problem as encoding first the parameter estimates and then the data given those 
estimates. The parameter values that result in the least \emph{overall} message length
to explain the whole data
are taken as the MML estimates for an inference problem. 
The MML scheme thus incorporates the cost of stating parameters into model selection.
It is self evident that a continuous parameter value can
only be stated to some finite precision;
the cost of encoding a parameter is determined by its prior and the precision.
ML estimation ignores the cost of stating a parameter and MAP based
estimation uses the probability \textit{density} of a parameter instead of its
probability measure.
In contrast, the MML inference process 
calculates the optimal precision to which parameters should be stated 
and a probability value of the corresponding parameters is then computed.
This is used in the computation of the message 
length corresponding to the parameter estimates. Thus, models with varying
parameters are evaluated based on their resultant total message lengths.
We use this characteristic property of MML to evaluate mixtures containing 
different numbers of components.  

Although there have been several attempts to address the challenges 
of mixture modelling, 
the existing methods are shown to have some limitations in their formalisms 
(see Section \ref{sec:mixture_existing_methods}). In particular, some
methods based on MML are incomplete
in their formulation. We aim to rectify these drawbacks by proposing a 
comprehensive MML formulation and develop a search 
heuristic that selects the number of mixture components
based on the proposed formulation.
To demonstrate the effectiveness of the proposed
search and parameter estimation, we first consider 
modelling problems using
Gaussian mixtures and extend the work to include relevant discussion on 
mixture modelling of directional data using von Mises-Fisher distributions. 

The importance of Gaussian mixtures in practical applications is well 
established.
For a \textit{given} number of components, the conventional method 
of estimating the parameters of a mixture relies on the 
expectation-maximization (EM) algorithm \citep{dempster1977maximum}.
The standard EM is a local optimization method, is sensitive to 
initialization, and in certain cases may converge to the boundary of 
parameter space \citep{krishnan1997algorithm,figueiredo2002unsupervised}.
Previous attempts to infer Gaussian mixtures based on the MML framework
have been undertaken using simplifying assumptions,
such as the covariance matrices being diagonal \citep{oliver1996unsupervised}, or
coarsely approximating the probabilities of mixture
parameters \citep{roberts,figueiredo2002unsupervised}.
Further, the search heuristic adopted in some of these methods is to run
the EM several times for different numbers of components, $M$, and
select the $M$ with the best EM outcome \citep{oliver1996unsupervised,roberts,icl}. 
A search method based on iteratively \emph{deleting} components has
been proposed by \cite{figueiredo2002unsupervised}. It begins by
assuming a very large number of components and selectively eliminates
components deemed redundant; there is no provision for recovering
from deleting a component in error.

In this work, we propose a search method which selectively \emph{splits, deletes}, 
or \emph{merges} components
depending on improvement to the MML objective function. The operations,
combined with EM steps, result in a sensible redistribution of data between
the mixture components.
As an example, a component may be split into two children, and at a later
stage, one of the children may be merged with another component.
Unlike the method of \cite{figueiredo2002unsupervised}, our method starts with
a one-component mixture and alters the number of components in 
subsequent iterations. This avoids the overhead of dealing with a large
number of components unless required.

The proposed search heuristic can be used with probability distributions
for which the MML expressions to
calculate message lengths for estimates 
and for data given those estimates are known.
As an instance of this, Section \ref{sec:search_method} discusses
modelling directional data using the von Mises-Fisher distributions. 
Directional statistics 
have garnered support recently in real data mining problems where
the \textit{direction}, not  magnitude, of  vectors is significant.
Examples of such scenarios are found in 
earth sciences, meteorology, physics, biology, and other fields 
\citep{mardia-book}. The statistical properties of directional data have 
been studied using several types of distributions \citep{fisher1953dispersion,watson1956construction,fisher1993statistical,mardia-book}, 
often described on surfaces of compact manifolds, such as the sphere, 
ellipsoid, torus \textit{etc}. The most fundamental of these is the von 
Mises-Fisher (vMF) distribution which is analogous to a symmetric 
multivariate Gaussian distribution, wrapped around a unit hypersphere 
\citep{watson1956construction}. 
The probability density function of a vMF distribution with parameters 
$\Theta=(\boldsymbol{\mu},\kappa)\equiv$ (mean direction, concentration 
parameter) for a random unit vector $\mathbf{x}\in\mathbb{R}^d$ on a 
$(d-1)$- dimensional hypersphere $\mathbb{S}^{d-1}$ is given by: 
\begin{equation}
f(\mathbf{x};\boldsymbol{\mu},\kappa) = C_d(\kappa) e^{\kappa \boldsymbol{\mu}^T\mathbf{x}} \label{eqn:vmf_density}
\end{equation}
where $C_d(\kappa) = \dfrac{\kappa^{d/2-1}}{(2\pi)^{d/2} I_{{d/2}-1}(\kappa)}$
is the normalization constant and $I_v$ is a modified Bessel function of 
the first kind and order $v$.  

The estimation of the parameters of the vMF distribution is often done using
maximum likelihood. However, the complex nature of the mathematical form 
presents difficulty in estimating the concentration parameter $\kappa$. 
This has lead to researchers using many different approximations, as discussed
in Section~\ref{sec:vmf_existing_methods}. Most of these methods 
perform well when the amount of data is large. At smaller sample sizes, they 
result in inaccurate estimates of $\kappa$, and are thus
unreliable. We demonstrate this by the experiments conducted on 
a range of sample sizes. The problem is particularly evident when the 
dimensionality of the data is large, also affecting the application in 
which it is used, such as mixture modelling. We aim to rectify this issue 
by using MML estimates for $\kappa$. Our experiments section
demonstrates that the MML estimate of $\kappa$ provides a more 
reliable answer and is an improvement on the current state of the art.
These MML estimates are subsequently used in
mixture modelling of vMF distributions
(see Sections \ref{sec:search_method} and  \ref{sec:vmf_applications}).

Previous studies have established the importance of
von Mises circular (two-dimensional) and
von Mises-Fisher (three-dimensional and higher) mixtures, and
demonstrated applications to clustering of protein dihedral angles 
\citep{mardia2007protein,dowe1996circular}, large-scale text clustering 
\citep{Banerjee:generative-clustering}, and gene expression 
analyses \citep{Banerjee:clustering-hypersphere}. The merit of using 
cosine based similarity metrics, which are closely related to the vMF,
for clustering high dimensional text data 
has been investigated in \cite{strehl2000impact}. For text clustering, 
there is evidence that vMF mixture models have a superior performance
compared to other statistical distributions such as normal, multinomial, 
and Bernoulli \citep{salton1988term,salton1983introduction,zhong2003comparative,Banerjee:clustering-hypersphere}. 
movMF is a widely used package to perform clustering using vMF 
distributions \citep{hornik2013movmf}.\\ 

\noindent\textbf{Contributions:} The main contributions of this paper are 
as follows:
\begin{itemize}
\item We derive the analytical estimates of the parameters of a
multivariate Gaussian distribution with full covariance 
matrix, using the MML principle \citep{wallace-87}. 

\item We derive the expression to infer the concentration parameter 
$\kappa$ of a generic $d$-dimensional vMF distribution using MML-based 
estimation. We demonstrate, through a series of experiments, that this
estimate outperforms the previous ones, therefore making it a reliable 
candidate to be used in mixture modelling.

\item A generalized MML-based search heuristic is proposed to infer the 
optimal number of mixture components that would best explain the observed data;
it is based on the search used in various versions of
the 'Snob' classification program \citep{wallace68,wallace1986improved,jorgensen2008wallace}.
We compare it with the work of \cite{figueiredo2002unsupervised} 
and demonstrate its effectiveness.

\item The search implements a generic approach to mixture modelling and allows,
in this instance, the use of $d$-dimensional
Gaussian and vMF distributions under the MML framework.
It infers the optimal number of mixture components, and their corresponding parameters.

\item Further, we demonstrate the effectiveness of MML mixture modelling 
through its application to high dimensional text clustering and clustering 
of directional data that arises out of protein conformations. 
\end{itemize}

The rest of the paper is organized as follows: 
Sections \ref{sec:gaussian_estimates} and \ref{sec:vmf_existing_methods} 
describe the respective estimators of Gaussian and vMF distributions
that are commonly used. Section \ref{sec:mml_framework} introduces the MML
framework for parameter estimation. Section \ref{sec:mml_est_derivations} 
outlines the derivation of the MML parameter estimates 
of multivariate Gaussian and vMF distributions. 
Section \ref{sec:mml_mixture_modelling} describes the formulation of a 
mixture model using MML and the estimation of the mixture parameters
under the framework.
Section~\ref{sec:mixture_existing_methods} reviews the 
existing methods for selecting the mixture components.
Section \ref{sec:search_method} 
describes our proposed approach to determine the number of mixture 
components. Section \ref{sec:gaussian_experiments} depicts the competitive 
performance of the proposed MML-based search through experiments conducted 
with Gaussian mixtures. Section \ref{sec:vmf_experiments} 
presents the results for MML-based vMF parameter estimation and mixture 
modelling followed by results supporting its applications to text 
clustering and protein structural data in Section \ref{sec:vmf_applications}.

\section{Existing methods of estimating the parameters of a Gaussian distribution} 
\label{sec:gaussian_estimates}
The probability density function $f$ of a $d$-variate Gaussian 
distribution is given as
\begin{equation}
f(\mathbf{x};\boldsymbol{\mu},\mathbf{C}) = 
\frac{1}{(2\pi)^{\frac{d}{2}} |\mathbf{C}|^{\frac{1}{2}}}
e^{-\frac{1}{2}(\mathbf{x}-\boldsymbol{\mu})^T \mathbf{C}^{-1}(\mathbf{x}-\boldsymbol{\mu})}
\label{eqn:gaussian_density}
\end{equation}
where $\boldsymbol{\mu}$, $\mathbf{C}$ are the respective mean, 
(symmetric) covariance matrix of the distribution, and $|\mathbf{C}|$ is 
the determinant of the covariance matrix. 
The traditional method to estimate the parameters of a Gaussian distribution
is by maximum likelihood. Given data 
$D=\{\mathbf{x}_1,\ldots,\mathbf{x}_N\}$, 
where $\mathbf{x}_i\in\mathbb{R}^{d}$, the log-likelihood $\mathcal{L}$ is 
given by
\begin{equation}
\mathcal{L}(D|\boldsymbol{\mu},\mathbf{C}) = 
-\frac{Nd}{2} \log (2\pi) - \frac{N}{2}\log|\mathbf{C}|
- \frac{1}{2}\sum_{i=1}^N (\mathbf{x}_i-\boldsymbol{\mu})^T \mathbf{C}^{-1}(\mathbf{x}_i-\boldsymbol{\mu}) \label{eqn:gaussian_negloglikelihood}
\end{equation}
To compute the maximum likelihood estimates, Equation \eqref{eqn:gaussian_negloglikelihood}
needs to be \emph{maximized}. This is achieved by computing the gradient of the
log-likelihood function with respect to the parameters
and solving the resultant equations. 
The \emph{gradient vector} of $\mathcal{L}$ with respect to $\boldsymbol{\mu}$ 
and the \emph{gradient matrix} of $\mathcal{L}$ with respect to $\mathbf{C}$ 
are given below.
\begin{align}
\nabla_{\boldsymbol{\mu}}\mathcal{L} = \frac{\partial \mathcal{L}}{\partial \boldsymbol{\mu}} &= 
\sum_{i=1}^N \mathbf{C}^{-1} (\mathbf{x}_i - \boldsymbol{\mu}) \label{eqn:gradient_mu}\\
 \nabla_{\mathbf{C}}\mathcal{L} = \frac{\partial \mathcal{L}}{\partial \mathbf{C}} &= -\frac{N}{2}\mathbf{C}^{-1} +
\frac{1}{2} \sum_{i=1}^N \mathbf{C}^{-1}(\mathbf{x}_i-\boldsymbol{\mu}) (\mathbf{x}_i-\boldsymbol{\mu})^T\mathbf{C}^{-1} \label{eqn:gradient_cov}
\end{align}
The maximum likelihood estimates are then computed by
solving 
$\nabla_{\boldsymbol{\mu}}\mathcal{L} = 0$ and $\nabla_{\mathbf{C}}\mathcal{L} = 0$
and are given as:
\begin{equation}
\boldsymbol{\hat{\mu}} = \frac{1}{N} \sum_{i=1}^N \mathbf{x}_i 
\quad \text{and} \quad
\hat{\mathbf{C}}_{\text{ML}} = \frac{1}{N}\sum_{i=1}^N (\mathbf{x}_i-\boldsymbol{\hat{\mu}}) (\mathbf{x}_i-\boldsymbol{\hat{\mu}})^T
\label{eqn:gaussian_ml_est}
\end{equation}

$\hat{\mathbf{C}}_{\text{ML}}$ is known to be a biased estimate of the 
covariance matrix \citep{ml_bias1,ml_bias2,ml_bias3,ml_bias4} and issues
related with its use in mixture modelling have been documented
in \cite{ml_mixture_bias1} and \cite{ml_mixture_bias2}. An unbiased estimator of 
$\mathbf{C}$ was proposed by \cite{ml_bias1} and is given below.
\begin{equation}
\hat{\mathbf{C}}_{\text{unbiased}} = \dfrac{1}{N-1}\sum_{i=1}^N (\mathbf{x}_i-\boldsymbol{\hat{\mu}}) (\mathbf{x}_i-\boldsymbol{\hat{\mu}})^T
\label{eqn:gaussian_ml_est_unbiased}
\end{equation}

In addition to the maximum likelihood estimates, Bayesian inference 
of Gaussian parameters
involving conjugate priors over the parameters has also been 
dealt with in the literature \citep{bishop2006pattern}.
However, the unbiased estimate of the covariance matrix, as determined
by the sample covariance (Equation \eqref{eqn:gaussian_ml_est_unbiased}),
is typically used in the analysis of Gaussian distributions.

\section{Existing methods of estimating the parameters of a von Mises-Fisher distribution} 
\label{sec:vmf_existing_methods}
For a von Mises-Fisher (vMF) distribution $f$ characterized
by Equation \eqref{eqn:vmf_density}, and given
data $D=\{\mathbf{x}_1,\ldots,\mathbf{x}_N\}$, such that 
$\mathbf{x}_i\in\mathbb{S}^{d-1}$,
the log-likelihood $\mathcal{L}$ is 
given by
\begin{equation}
\mathcal{L}(D|\boldsymbol{\mu},\kappa) = N\log C_d(\kappa) + \kappa \boldsymbol{\mu}^T \mathbf{R} \label{eqn:vmf_negloglikelihood}
\end{equation}
where $N$ is the sample size and $\mathbf{R} = \displaystyle\sum_{i=1}^N \mathbf{x}_i$ 
(the vector sum). Let $R$ denote the magnitude of the resultant vector 
$\mathbf{R}$ and let $\hat{\boldsymbol{\mu}}$ and $\hat{\kappa}$ be the 
maximum likelihood estimators of $\mu$ and $\kappa$ respectively. 
Under the condition that $\hat{\boldsymbol{\mu}}$ is a unit vector,
the maximum likelihood estimates are obtained by maximizing $\mathcal{L}$ as follows:
\begin{equation}
\boldsymbol{\hat{\mu}} = \frac{\mathbf{R}}{R}, \quad 
{\hat{\kappa}} = A_d^{-1}(\bar{R}) \quad
\text{where}\quad A_d(\hat{\kappa}) = -\frac{C_d'(\hat{\kappa})}{C_d(\hat{\kappa})} = \frac{R}{N} = \bar{R} \label{eqn:ml_estimates}
\end{equation}
Solving the non-linear equation: $F(\kappa) \equiv A_d(\hat{\kappa}) - \bar{R} = 0$ yields 
the corresponding maximum likelihood estimate where
\begin{equation}
A_d(\kappa) = \frac{I_{d/2}(\kappa)}{I_{d/2-1}(\kappa)} \label{eqn:ratio_bessels}
\end{equation}
represents the ratio of Bessel functions. 
Because of the difficulties in analytically solving
Equation (\ref{eqn:ml_estimates}), there have been several approaches to
approximating $\hat{\kappa}$ \citep{mardia-book}.
Each of these methods is an improvement over their respective predecessors. 
\cite{tanabe2007} is an improvement over the estimate proposed by 
\cite{Banerjee:clustering-hypersphere}. \cite{sra2012short}
is an improvement over \cite{tanabe2007} and \cite{song2012high} fares better when compared to
\cite{sra2012short}. The methods are summarized below.

\subsection{\emph{\cite{Banerjee:clustering-hypersphere}}}
The approximation given by Equation (\ref{eqn:banerjee_approx}) is due to 
\cite{Banerjee:clustering-hypersphere} and provides an easy to use expression for $\hat{\kappa}$.
The formula is very appealing as it eliminates the need to evaluate complex 
Bessel functions. \cite{Banerjee:clustering-hypersphere} demonstrated
that this approximation yields better results compared to the ones suggested in
\cite{mardia-book}. It is an empirical approximation which can be used as a starting point
in estimating the root of Equation (\ref{eqn:ml_estimates}).
\begin{equation}
\kappa_B = \frac{\bar{R}(d-\bar{R}^2)}{1-\bar{R}^2} \label{eqn:banerjee_approx}
\end{equation}

\subsection{\emph{\cite{tanabe2007}}}
The approximation given by Equation (\ref{eqn:tanabe_approx}) is due to \cite{tanabe2007}.
The method utilizes the properties of Bessel functions to determine the lower
and upper bounds for $\hat{\kappa}$ and uses a fixed point iteration function 
in conjunction with linear interpolation to approximate $\hat{\kappa}$.
The bounds for $\hat{\kappa}$ are given by
\begin{equation*}
\kappa_l = \frac{\bar{R}(d-2)}{1-\bar{R}^2} \le \hat{\kappa} \le \kappa_u = \frac{\bar{R}d}{1-\bar{R}^2} 
\end{equation*}
\cite{tanabe2007} proposed to use a fixed point iteration function
defined as $\phi_{2d}(\kappa) = \bar{R}\kappa A_d(\kappa)^{-1}$ and used this to approximate $\hat{\kappa}$ as
\begin{equation}
\kappa_T = \frac{\kappa_l \phi_{2d}(\kappa_u) - \kappa_u \phi_{2d}(\kappa_l)}{(\phi_{2d}(\kappa_u)-\phi_{2d}(\kappa_l))-(\kappa_u - \kappa_l)} \label{eqn:tanabe_approx}
\end{equation}

\subsection{\emph{\cite{sra2012short}} : Truncated Newton approximation}
This a heuristic approximation provided by \cite{sra2012short}. It involves refining
the approximation
given by \cite{Banerjee:clustering-hypersphere} (Equation (\ref{eqn:banerjee_approx}))
by performing two iterations of Newton's method.
\cite{sra2012short} demonstrate that this approximation fares well when compared to the approximation
proposed by \cite{tanabe2007}. The following two iterations result in $\kappa_N$, 
the approximation proposed by \cite{sra2012short}:
\begin{equation}
\kappa_1 = \kappa_B - \frac{F(\kappa_B)}{F'(\kappa_B)} \quad\text{and}\quad
\kappa_N = \kappa_1 - \frac{F(\kappa_1)}{F'(\kappa_1)} \label{eqn:sra_approx}
\end{equation}
\begin{equation}
\text{where}\quad F'(\kappa) = A_d'(\kappa) = 1 - A_d(\kappa)^2 -\frac{(d-1)}{\kappa} A_d(\kappa) \label{eqn:ratio_first_derivative}
\end{equation}

\subsection{\emph{\cite{song2012high}} : Truncated Halley approximation}
This approximation provided by \cite{song2012high} uses Halley's 
method which is the second order expansion of Taylor's series of a given function $F(\kappa)$.
The higher order approximation results in a more accurate estimate
as demonstrated by \cite{song2012high}. The iterative Halley's method is truncated
after iterating through two steps of the root finding algorithm 
(similar to that done by \cite{sra2012short}). The following two iterations 
result in $\kappa_H$, the approximation proposed by \cite{song2012high}:
\begin{equation}
\kappa_1 = \kappa_B - \frac{2 F(\kappa_B)F'(\kappa_B)}{2 F'(\kappa_B)^2 - F(\kappa_B) F''(\kappa_B)}\quad\text{and}\quad
\kappa_H = \kappa_1 - \frac{2 F(\kappa_1)F'(\kappa_1)}{2 F'(\kappa_1)^2 - F(\kappa_1) F''(\kappa_1)}\label{eqn:song_approx}
\end{equation}
\begin{equation}
\text{where}\quad F''(\kappa) = A_d''(\kappa) = 2 A_d(\kappa)^3 + \frac{3(d-1)}{\kappa} A_d(\kappa)^2 + \frac{(d^2-d-2\kappa^2)}{\kappa^2}A_d(\kappa)-\frac{(d-1)}{\kappa} \label{eqn:ratio_second_derivative}
\end{equation}

The common theme in all these methods is that they 
try to approximate the maximum likelihood estimate governed by Equation (\ref{eqn:ml_estimates}).
It is to be noted that the maximum likelihood estimators (of concentration parameter $\kappa$) 
have considerable bias \citep{schou1978estimation,best1981bias,cordeiro1999theory}.
To counter this effect, we explore the minimum message length based estimation procedure.
This Bayesian method of estimation not only results in an unbiased estimate but also
provides a framework to choose from several competing models \citep{wallace-87,classification_mml}.
Through a series of empirical tests, we demonstrate that the MML estimate
is more reliable than any of the contemporary methods. \cite{vmf_mmlestimate} have
demonstrated the superior performance of the MML estimate for a three-dimensional
vMF distribution. We extend their work to derive the MML estimators for a generic
$d$-dimensional vMF distribution and compare its performance with the existing methods.

\section{Minimum Message Length (MML) Inference}   
\label{sec:mml_framework}
\subsection{Model selection using Minimum Message Length}
\cite{wallace68} developed the first practical criterion
for model selection to be based on information theory.
The resulting framework
provides a rigorous means to objectively compare two
competing hypotheses and, hence, choose the best one. As per Bayes's theorem,
\[\Pr(H\&D) = \Pr(H) \times \Pr(D|H) = \Pr(D) \times \Pr(H|D)\]
where $D$ denotes some observed data, and $H$ some
hypothesis about that data. Further, $\Pr(H\&D)$ is the joint probability of
data $D$ and hypothesis $H$, $\Pr(H)$ is the prior probability of
hypothesis $H$, $\Pr(D)$ is the prior probability of data $D$, $\Pr(H|D)$
is the posterior probability of $H$ given $D$, and $\Pr(D|H)$ is the
likelihood.  

MML uses the following result from information theory
\citep{shannon1948}: given an event or outcome $E$ whose probability is
$\Pr(E)$, the length of the optimal lossless code $I(E)$ to represent that
event requires $I(E) = -\log_2 (\Pr(E))$ bits.  
Applying Shannon's insight to
Bayes's theorem, \cite{wallace68} got the following relationship between conditional
probabilities in terms of optimal message lengths: 
\begin{equation} 
I(H\&D) = I(H) + I(D|H) = I(D) + I(H|D) \label{eqn:shannon_bayes_msg}
\end{equation}
As a result, given two competing hypotheses $H$
and $H^\prime$, 
\begin{equation*}
\Delta I = I(H\&D) - I(H^\prime\&D) = I(H|D) - I(H^\prime|D) = \log_2\left(\frac{\Pr(H^\prime|D)}{\Pr(H|D)}\right) \quad\text{bits.}
\end{equation*}
\begin{equation}
\frac{\Pr(H^\prime|D)}{\Pr(H|D)} = 2^{\Delta I}\label{eqn:compare_models}
\end{equation}
gives the posterior log-odds ratio between the two competing hypotheses.
Equation (\ref{eqn:shannon_bayes_msg}) can be intrepreted as the \emph{total} cost to encode a
message comprising the hypothesis $H$ and data $D$. This message is composed over
into two parts:
\begin{enumerate}
\item \emph{First part:} the hypothesis $H$, which takes $I(H)$ bits,
\item \emph{Second part:} the observed data $D$ using knowledge of $H$, which takes $I(D|H)$ bits.
\end{enumerate}

Clearly, the message length can vary depending on the complexity of $H$ and
how well it can explain $D$. A more complex $H$ may fit (i.e., explain) $D$
better but take more bits to be stated itself.  The trade-off comes
from the fact that (hypothetically) transmitting the message requires the encoding of both
the hypothesis and the data given the hypothesis, that is, the model
complexity $I(H)$ and the goodness of fit $I(D|H)$.

\subsection{Minimum message length based parameter estimation}  \label{subsec:mml_parameter_estimation}
Our proposed method of parameter estimation uses the MML inference 
paradigm. It is a Bayesian method which has been applied to infer the 
parameters of several statistical distributions \citep{WallaceBook}. We 
apply it to infer the parameter estimates of multivariate Gaussian and vMF
distributions. 
\cite{wallace-87} introduced a generalized scheme to estimate a vector of 
parameters $\Theta$ of any distribution $f$ given data $D$. The method
involves choosing a reasonable prior $h(\Theta)$ on the hypothesis and 
evaluating the
\textit{determinant} of the Fisher information matrix $|\mathcal{F}(\Theta)|$ of the 
\textit{expected} second-order partial derivatives of the negative 
log-likelihood function, $-\mathcal{L}(D|\Theta)$. The parameter vector $\Theta$ that minimizes 
the message length expression (Equation (\ref{eqn:two_part_msg}))
is the MML estimate according to \cite{wallace-87}. 
\begin{equation}
I(\Theta,D) = \underbrace{\frac{p}{2}\log q_p -\log\left(\frac{h(\Theta)}{\sqrt{|\mathcal{F}(\Theta)|}}\right)}_{\mathrm{I(\Theta)}} - \underbrace{\mathcal{L}(D|\Theta) + \frac{p}{2}}_{\mathrm{I(D|\Theta)}}\label{eqn:two_part_msg}
\end{equation}
where $p$ is the number of free parameters in the model, and $q_p$ is the 
lattice quantization constant \citep{conwaySloane84} in $p$-dimensional 
space. The total message length $I(\Theta,D)$ in MML framework is composed 
of two parts: 
\begin{enumerate}
\item the statement cost of encoding the parameters, $I(\Theta)$ and 
\item the cost of encoding the data given the parameters, $I(D|\Theta)$.
\end{enumerate}
A concise description of the MML method is presented in \cite{oliver1994mml}.

We note here a few key differences between MML and ML/MAP based estimation methods.
In maximum likelihood estimation, the statement cost of parameters is ignored, in effect considered constant,
and minimizing the message length corresponds to minimizing the negative
log-likelihood of the data (the second part).
In MAP based estimation, a probability \textit{density} 
rather than the probability measure is used.
Continuous parameters can necessarily only be stated only to finite precision.
MML incorporates this in the framework
by determining the region of uncertainty in which the parameter is located.
The value of $\dfrac{q_p^{-p/2}}{\sqrt{|\mathcal{F}(\Theta)|}}$ gives a measure of the volume
of the region of uncertainty in which the parameter $\Theta$ is centered.
This multiplied by the probability density $h(\Theta)$ gives the 
\emph{probability} of a particular $\Theta$
and is \emph{proportional} to $\dfrac{h(\Theta)}{\sqrt{|\mathcal{F}(\Theta)|}}$.
This probability is used to compute the message length associated with
encoding the continuous valued parameters (to a finite precision).

\section{Derivation of the MML parameter estimates of Gaussian and von Mises-Fisher distributions}
\label{sec:mml_est_derivations}
Based on the MML inference process discussed in Section \ref{sec:mml_framework},
we now proceed to formulate the message length expressions and derive the 
parameter estimates of Gaussian and von Mises-Fisher distributions.

\subsection{MML-based parameter estimation of a multivariate Gaussian distribution} 
\label{subsec:mml_gaussian_est}
The MML framework requires the statement of parameters
to a finite precision. The optimal precision is related to the Fisher information 
and in conjunction with a reasonable prior, the probability of parameters
is computed. 

\subsubsection{Prior probability of the parameters}
A flat prior is usually chosen on each of the $d$ dimensions
of $\boldsymbol{\mu}$ \citep{roberts,oliver1996unsupervised} and a 
conjugate inverted Wishart prior is chosen for the covariance matrix $\mathbf{C}$ 
\citep{gaussian_map,agusta2003unsupervised,bishop2006pattern}.
The joint prior density of the parameters is then given as
\begin{equation}
h(\boldsymbol{\mu},\mathbf{C}) \propto |\mathbf{C}|^{-\frac{d+1}{2}} \label{eqn:gaussian_joint_prior}
\end{equation}

\subsubsection{Fisher information of the parameters}
The computation of the Fisher information requires the evaluation of the
second order partial derivatives of $-\mathcal{L}(D|\boldsymbol{\mu},\mathbf{C})$. 
Let $|\mathcal{F}(\boldsymbol{\mu},\mathbf{C})|$ 
represent the determinant of the Fisher information matrix. This is 
approximated as the product of $|\mathcal{F}(\boldsymbol{\mu})|$
and $|\mathcal{F}(\mathbf{C})|$
\citep{oliver1996unsupervised,roberts}, where 
$|\mathcal{F}(\boldsymbol{\mu})|$ and $|\mathcal{F}(\mathbf{C})|$
are the respective determinants of Fisher information matrices due to the 
parameters $\boldsymbol{\mu}$ and $\mathbf{C}$. 
Differentiating the gradient vector in Equation \eqref{eqn:gradient_mu}
with respect to $\boldsymbol{\mu}$, we have:
\begin{align}
-\nabla^2_{\boldsymbol{\mu}}\mathcal{L} &= N\mathbf{C}^{-1} \notag\\
\text{Hence,}\quad|\mathcal{F}(\boldsymbol{\mu})| &= N^d |\mathbf{C}|^{-1}
\label{eqn:gaussian_fisher_mu}
\end{align}
To compute $|\mathcal{F}(\mathbf{C})|$, \cite{magnus1988matrix} derived an
analytical expression using the theory of matrix derivatives based on 
matrix vectorization \citep{dwyer1967some}. Let $\mathbf{C} = [c_{ij}]\,
\forall 1 \leq i,j \leq d$ 
where $c_{ij}$ denotes the element corresponding to the $i^{th}$ row and
$j^{th}$ column of the covariance matrix. Let $v(\mathbf{C}) = 
(c_{11},\ldots,c_{1d},c_{22},\ldots,c_{2d},\ldots,c_{dd})$ be the vector
containing the $\dfrac{d(d+1)}{2}$ free parameters that completely describe
the symmetric matrix $\mathbf{C}$. Then, the Fisher information due to 
the vector of parameters $v(\mathbf{C})$ is equal to 
$|\mathcal{F}(\mathbf{C})|$ and is given by
Equation \eqref{eqn:gaussian_fisher_cov} \citep{magnus1988matrix,bozdogan1990information,drton2009lectures}.
\begin{equation}
|\mathcal{F}(\mathbf{C})| = N^{\frac{d(d+1)}{2}} 2^{-d} |\mathbf{C}|^{-(d+1)} \label{eqn:gaussian_fisher_cov}
\end{equation}
Multiplying Equations \eqref{eqn:gaussian_fisher_mu} and \eqref{eqn:gaussian_fisher_cov}, we have 
\begin{equation}
|\mathcal{F}(\boldsymbol{\mu},\mathbf{C})| = N^{\frac{d(d+3)}{2}} 2^{-d} |\mathbf{C}|^{-(d+2)} \label{eqn:gaussian_fisher}
\end{equation}

\subsubsection{Message length formulation}
To derive the message length expression to encode data
using a certain $\boldsymbol{\mu},\mathbf{C}$, substitute Equations
 \eqref{eqn:gaussian_negloglikelihood}, \eqref{eqn:gaussian_joint_prior}, 
and \eqref{eqn:gaussian_fisher}, in Equation \eqref{eqn:two_part_msg}
using the number of free parameters of the distribution as 
$p = \dfrac{d(d+3)}{2}$. Hence,
\begin{equation}
I(\boldsymbol{\mu},\mathbf{C},D) = \frac{(N-1)}{2}\log|\mathbf{C}| + 
\frac{1}{2}\sum_{i=1}^N (\mathbf{x}_i-\boldsymbol{\mu})^T \mathbf{C}^{-1}(\mathbf{x}_i-\boldsymbol{\mu}) + \text{constant}\label{eqn:gaussian_msglen}
\end{equation}

To obtain the MML estimates of $\boldsymbol{\mu}$ and $\mathbf{C}$, 
Equation \eqref{eqn:gaussian_msglen}
needs to be minimized. The MML estimate of $\boldsymbol{\mu}$ is same as
the maximum likelihood estimate (given in Equation \eqref{eqn:gaussian_ml_est}). To compute the MML estimate of $\mathbf{C}$, we need to compute the
gradient matrix of $I(\boldsymbol{\mu},\mathbf{C},D)$ with respect to 
$\mathbf{C}$ and is given by Equation \eqref{eqn:gradient_msglen} 
\begin{equation}
\nabla_{\mathbf{C}} I = \frac{\partial I}{\partial \mathbf{C}} =
\frac{(N-1)}{2}\mathbf{C}^{-1} -
\frac{1}{2} \sum_{i=1}^N \mathbf{C}^{-1}(\mathbf{x}_i-\boldsymbol{\mu}) (\mathbf{x}_i-\boldsymbol{\mu})^T\mathbf{C}^{-1} 
\label{eqn:gradient_msglen}
\end{equation}
The MML estimate of $\mathbf{C}$ is obtained by solving $\nabla_{\mathbf{C}} I = 0$ (given in Equation\eqref{eqn:gaussian_mml_est}).
\begin{equation}
\nabla_{\mathbf{C}} I = 0 \implies 
\hat{\mathbf{C}}_{\text{MML}} = \frac{1}{N-1}\sum_{i=1}^N (\mathbf{x}_i-\boldsymbol{\hat{\mu}}) (\mathbf{x}_i-\boldsymbol{\hat{\mu}})^T
\label{eqn:gaussian_mml_est}
\end{equation}
We observe that the MML estimate $\hat{\mathbf{C}}_{\text{MML}}$ is same
as the \emph{unbiased} estimate of the covariance matrix $\mathbf{C}$, 
thus, lending credibility for its preference over the traditional ML 
estimate (Equation \eqref{eqn:gaussian_ml_est}).

\subsection{MML-based parameter estimation of a von Mises-Fisher distribution} 
\label{subsec:mml_vmf_est}
Parameter estimates for two and three-dimensional vMF have been explored 
previously \citep{wallace1994estimation,dowe1996bayesian,vmf_mmlestimate}.
MML estimators of three-dimensional vMF were explored in
\cite{vmf_mmlestimate}, where they demonstrate that the MML-based 
inference compares favourably against the traditional ML and MAP based 
estimation methods.
We use the \cite{wallace-87} method to formulate the objective function 
(Equation \eqref{eqn:two_part_msg}) corresponding to a generic vMF 
distribution. 

\subsubsection{Prior probability of the parameters}
Regarding choosing a reasonable prior (in the absence of any supporting evidence) for the parameters
$\Theta=(\boldsymbol{\mu},\kappa)$ of a vMF distribution,
\cite{wallace1994estimation} and  \cite{vmf_mmlestimate}
suggest the use of the following ``\emph{colourless} prior that is uniform in direction,
normalizable and locally uniform at the Cartesian origin in $\kappa$":
\begin{equation}
h(\boldsymbol{\mu},\kappa) \propto \frac{\kappa^{d-1}}{(1+\kappa^2)^{\frac{d+1}{2}}} \label{eqn:vmf_joint_prior}
\end{equation}

\subsubsection{Fisher information of the parameters}
Regarding evaluating the Fisher information, \cite{vmf_mmlestimate} argue 
that in the general $d$-dimensional case, 
\begin{equation}
|\mathcal{F}(\boldsymbol{\mu},\kappa)| = (N\kappa A_d(\kappa))^{d-1} \times N A_d'(\kappa) \label{eqn:det_fisher}
\end{equation}
where $A_d(\kappa)$ and $A_d'(\kappa)$ are described by Equations 
\eqref{eqn:ratio_bessels} and \eqref{eqn:ratio_first_derivative} 
respectively. 

\subsubsection{Message length formulation}
Substituting Equations \eqref{eqn:vmf_negloglikelihood}, 
\eqref{eqn:vmf_joint_prior} and \eqref{eqn:det_fisher} in Equation 
\eqref{eqn:two_part_msg} with number of free parameters $p = d$, 
we have the net message length expression:
\begin{equation}
I(\boldsymbol{\mu},\kappa,D) = \frac{(d-1)}{2}\log\frac{A_d(\kappa)}{\kappa} + \frac{1}{2}\log A_d'(\kappa) 
+ \frac{(d+1)}{2} \log(1+\kappa^2) - N \log C_d(\kappa) - \kappa \boldsymbol{\mu}^T \mathbf{R} + \text{constant}\label{eqn:vmf_msglen}
\end{equation}
To obtain the MML estimates of $\boldsymbol{\mu}$ and $\kappa$, Equation 
\eqref{eqn:vmf_msglen}
needs to be minimized. The estimate for $\boldsymbol{\mu}$ is same as
the maximum likelihood estimate (Equation \eqref{eqn:ml_estimates}). 
The resultant equation in $\kappa$ that needs to be minimized is then given by:
\begin{equation}
I(\kappa) = \frac{(d-1)}{2}\log\frac{A_d(\kappa)}{\kappa} + \frac{1}{2}\log A_d'(\kappa) 
+ \frac{(d+1)}{2} \log(1+\kappa^2) - N \log C_d(\kappa) - \kappa R + \text{constant}\label{eqn:I_kappa}
\end{equation}
To obtain the MML estimate of $\kappa$, we need to differentiate Equation (\ref{eqn:I_kappa}) and set it to zero. 
\begin{equation}
\text{Let}\quad G(\kappa) \equiv \frac{\partial I}{\partial \kappa}
= -\frac{(d-1)}{2\kappa} + \frac{(d+1)\kappa}{1+\kappa^2} + \frac{(d-1)}{2}\frac{A_d'(\kappa)}{A_d(\kappa)} + \frac{1}{2}\frac{A_d''(\kappa)}{A_d'(\kappa)} + N A_d(\kappa) - R  \label{eqn:I_first_derivative}
\end{equation}
The non-linear equation: $G(\kappa) = 0$ does not have a closed form solution.
We try both the Newton and Halley's method to find an approximate solution.
We discuss both variants and comment on the effects of the two approximations
in the experimental results.
To be fair and consistent with \cite{sra2012short} and \cite{song2012high},
we use the initial guess of the root as $\kappa_B$ (Equation (\ref{eqn:banerjee_approx}))
and iterate twice to obtain the MML estimate.
\begin{enumerate}
\item \emph{Approximation using Newton's method: }
\begin{equation}
\kappa_1 = \kappa_B - \frac{G(\kappa_B)}{G'(\kappa_B)} \quad\text{and}\quad
\kappa_{\text{MN}} = \kappa_1 - \frac{G(\kappa_1)}{G'(\kappa_1)} \label{eqn:mml_newton_approx}
\end{equation}
\item \emph{Approximation using Halley's method: }
\begin{equation}
\kappa_1 = \kappa_B - \frac{2 G(\kappa_B)G'(\kappa_B)}{2 G'(\kappa_B)^2 - G(\kappa_B) G''(\kappa_B)}\quad\text{and}\quad
\kappa_{\text{MH}} = \kappa_1 - \frac{2 G(\kappa_1)G'(\kappa_1)}{2 G'(\kappa_1)^2 - G(\kappa_1) G''(\kappa_1)}\label{eqn:mml_halley_approx}
\end{equation}
The details of evaluating $G'(\kappa)$ and $G''(\kappa)$ are discussed in Appendix~\ref{subsec:appendix_derivations}.
\end{enumerate}
Equation (\ref{eqn:mml_newton_approx}) gives the MML estimate ($\kappa_{MN}$) using Newton's method
and Equation (\ref{eqn:mml_halley_approx}) gives the MML estimate ($\kappa_{MH}$) using Halley's method.
We used these values of MML estimates in mixture modelling using vMF distributions.
\label{sec:mml_gaussian_est}

\section{Minimum Message Length Approach to Mixture Modelling} 
\label{sec:mml_mixture_modelling}
Mixture modelling involves representing an observed distribution of data as a 
weighted sum of individual probability density functions. Specifically, 
the problem we consider here is to model the mixture distribution $\fancym$ 
as defined in
Equation \eqref{eqn:mixture}. 
For some observed data $D = \{\mathbf{x}_1,\ldots,\mathbf{x}_N\}$
($N$ is the sample size), and a mixture $\fancym$, the log-likelihood 
using the mixture distribution is as follows:
\begin{equation}
  \mathcal{L}(D|\boldsymbol{\Phi}) = \sum_{i=1}^N \log \sum_{j=1}^M w_j f_j(\mathbf{x}_i;\Theta_j) \label{eqn:loglike_mixture}
\end{equation}
where $\boldsymbol{\Phi} = \{w_1,\cdots,w_M,\Theta_1,\cdots,\Theta_M\}$,
$w_j$ and $f_j(\mathbf{x};\Theta_j)$ are the weight 
and probability density of the $j^{\text{th}}$ component respectively.
For a fixed $M$, the mixture parameters $\boldsymbol{\Phi}$ are traditionally 
estimated using a standard \textit{expectation-maximization}(EM) algorithm 
\citep{dempster1977maximum,krishnan1997algorithm}. This
is briefly discussed below.

\subsection{Standard EM algorithm to estimate mixture parameters} \label{subsec:em_ml}
The standard EM algorithm is based on maximizing the log-likelihood 
function of the data (Equation \eqref{eqn:loglike_mixture}).
The maximum likelihood estimates are then given 
as $\boldsymbol{\Phi}_{ML} = \underset{\boldsymbol{\Phi}}{\text{arg\,max}} \,\,\mathcal{L}(D|\boldsymbol{\Phi})$.
Because of the absence of a closed form solution for $\boldsymbol{\Phi}_{ML}$, 
a gradient descent method is employed where the parameter estimates are 
iteratively updated until convergence to some local optimum is achieved
\citep{dempster1977maximum,mclachlan1988mixture,xu1996convergence,krishnan1997algorithm,mclachlan2000finite}.
The EM method consists of two steps:
\begin{itemize}
  \item \textit{E-step}: Each datum $\mathbf{x}_i$ has fractional 
  membership to each of the mixture components. These partial memberships of the data points 
  to each of the components are defined using the \textit{responsibility matrix}
  \begin{equation}
    r_{ij} = \frac{w_j f(\mathbf{x}_i;\Theta_j)}{\sum_{k=1}^M w_k f(\mathbf{x}_i;\Theta_k)}, \quad\forall \, 1\le i\le N, 1\le j\le M \label{eqn:responsibility}
  \end{equation}
  where $r_{ij}$ denotes the conditional probability of a datum 
  $\mathbf{x}_i$ belonging to the $j^{\text{th}}$ component. 
  The effective membership associated with each component is then given by
  \begin{equation}
    n_j = \sum_{i=1}^N r_{ij} \quad\text{and}\quad \sum_{j=1}^M n_j = N
    \label{eqn:comp_eff_mshp}
  \end{equation}

  \item \textit{M-step}: Assuming $\boldsymbol{\Phi}^{(t)}$
  be the estimates at some iteration $t$, the expectation of the 
  log-likelihood using $\boldsymbol{\Phi}^{(t)}$ and the partial memberships 
  is then \emph{maximized} which is tantamount to computing 
  $\boldsymbol{\Phi}^{(t+1)}$, the updated maximum likelihood estimates for 
  the next iteration $(t+1)$. The weights are updated as 
  $w_j^{(t+1)} = \dfrac{n_j^{(t)}}{N}$.
\end{itemize}
The above sequence of steps are repeated until a certain convergence 
criterion is satisfied.
At some intermediate iteration $t$, the mixture parameters are updated using
the corresponding ML estimates and are given below.
\begin{itemize}
\item \emph{Gaussian:} The ML updates of the mean and covariance matrix are
\begin{equation*}
\hat{\boldsymbol{\mu}}_j^{(t+1)} = \frac{1}{n_j^{(t)}} \sum_{i=1}^N r_{ij}^{(t)} \mathbf{x}_i
\quad\text{and}\quad
\hat{\mathbf{C}}_j^{(t+1)} = \dfrac{1}{n_j^{(t)}}\sum_{i=1}^N r_{ij}^{(t)} 
\left(\mathbf{x}_i-\boldsymbol{\hat{\mu}}_j^{(t+1)}\right) \left(\mathbf{x}_i-\boldsymbol{\hat{\mu}}_j^{(t+1)}\right)^T
\end{equation*}
\item \emph{von Mises-Fisher:} The resultant vector sum is updated as 
$\mathbf{R}_j^{(t+1)} = \displaystyle\sum_{i=1}^N r_{ij}^{(t)} \mathbf{x}_i$. If $R_j^{(t+1)}$ represents
the magnitude of vector $\mathbf{R}_j^{(t+1)}$, then the updated mean and
concentration parameter are 
\begin{equation*}
\hat{\boldsymbol{\mu}}_j^{(t+1)} = \frac{\mathbf{R}_j^{(t+1)}}{R_j^{(t+1)}},\quad
\bar{R}_j^{(t+1)} = \frac{R_j^{(t+1)}}{n_j^{(t)}},\quad
\hat{\kappa}_j^{(t+1)} = A_d^{-1}\left(\bar{R}_j^{(t+1)}\right) 
\end{equation*}
\end{itemize}

\subsection{EM algorithm to estimate mixture parameters using MML} \label{subsec:em_mml}
We will first describe the methodology involved in formulating the MML-based objective
function. We will then discuss how EM is applied in this context.
 
\subsubsection{Encoding a mixture model using MML}\label{susubsec:message_format}
We refer to the discussion in~\cite{WallaceBook} to briefly describe the
intuition behind mixture modelling using MML. Encoding of a message using 
MML requires the encoding of (1) the model parameters and then (2) the data using 
the parameters. The statement costs for encoding the mixture model and the 
data can be decomposed into:
\begin{enumerate}
  \item Encoding the \emph{number of components} $M$: 
  In order to encode the message losslessly, it is required to initially
  state the number of components. In the absence of background
  knowledge, one would like to model the
  prior belief in such a way that the probability 
  decreases for increasing number of components.
  If $h(M) \propto 2^{-M}$, then $I(M) = M \log 2 + \text{constant}$. The prior reflects 
  that there is a difference of one bit in encoding the \emph{numbers}
  $M$ and $M+1$. Alternatively, one could assume a uniform prior over
  $M$ within some predefined range.
  The chosen prior has little effect as its contribution is minimal
  when compared to the magnitude of the total message length \citep{WallaceBook}.

  \item Encoding  the \emph{weights} $w_1,\cdots,w_M$ which are treated as
  parameters of a multinomial distribution with sample size $n_j,~\forall 1\le j\le M$. 
  The length of encoding all the weights is then given by the expression 
  \citep{mml_multistate}:\\
  \begin{equation}
    I(\mathbf{w})=\frac{(M-1)}{2}\log N -\frac{1}{2}\sum\limits_{j=1}^M \log w_j - (M-1)!
  \end{equation}

  \item Encoding each of the \emph{component parameters} $\Theta_j$ 
  as given by $I(\Theta_j)=-\log\dfrac{h(\Theta_j)}{\sqrt{|\mathcal{F}(\Theta_j)|}}$ 
  (discussed in Section~\ref{subsec:mml_parameter_estimation}).

  \item Encoding the \emph{data}: 
  each datum $\mathbf{x}_i$ can be stated to a finite precision which
  is dictated by the accuracy of measurement\footnote{We note that 
  $\epsilon$ is a constant value and has no effect on the overall 
  inference process. It is used in order to maintain the theoretical
  validity when making the distinction between \emph{probability} and 
  \emph{probability density}.}.  If the precision to which
  each element of a $d$-dimensional vector can be stated is $\epsilon$, 
  then the \emph{probability} of a datum $\mathbf{x}_i \in \mathbb{R}^d$
  is given as $\Pr(\mathbf{x}_i) = \epsilon^d \Pr(\mathbf{x}_i|\fancym)$
  where $\Pr(\mathbf{x}_i|\fancym)$ is the \emph{probability density}
  given by Equation~\eqref{eqn:mixture}. Hence, the \emph{total} length of
  its encoding is given by
  \begin{equation}
    I(\mathbf{x}_i) = -\log\Pr(\mathbf{x}_i) = -d\log\epsilon -\log\sum_{j=1}^M w_j f_j(\mathbf{x}_i|\Theta_j)
  \end{equation}
  The entire data $D$ can now be encoded as:
  \begin{equation}
    I(D|\boldsymbol{\Phi}) = -Nd\log\epsilon -\sum_{i=1}^N \log \sum_{j=1}^M w_j f_j(\mathbf{x}_i;\Theta_j)
  \end{equation}
\end{enumerate}
Thus, the total message length of a $M$ component mixture is given by 
Equation~\eqref{eqn:mixture_msglen}. 
\begin{align}
I(\boldsymbol{\Phi},D) &= I(M) + I(\mathbf{w}) + \sum_{j=1}^M I(\Theta_j) + I(D|\boldsymbol{\Phi}) + \text{constant}\notag\\
&= I(M) + I(\mathbf{w}) + \left( -\sum_{j=1}^M  \log\,h(\Theta_j) + \frac{1}{2} \sum_{j=1}^M \log \,|\mathcal{F}(\Theta_j)| \right) + I(D|\boldsymbol{\Phi}) + \text{constant}
\label{eqn:mixture_msglen}
\end{align}

Note that the \textit{constant} term includes the lattice quantization 
constant (resulting from stating all the model parameters) in a 
$p$-dimensional space, where $p$ is equal to the number of free 
parameters in the mixture model.

\subsubsection{Estimating the mixture parameters} 
\label{subsubsec:mml_mixture_est}
The parameters of the mixture model are those that \emph{minimize} 
Equation \eqref{eqn:mixture_msglen}. To achieve this we use the standard
EM algorithm (Section \ref{subsec:em_ml}), where, iteratively, the 
parameters are updated using their respective \emph{MML estimates}. The 
component weights are obtained by differentiating Equation
\eqref{eqn:mixture_msglen} with respect to $w_j$ under the constraint
$\sum_{j=1}^M w_j = 1$. The derivation of the MML updates of the 
weights is shown in Appendix~\ref{subsec:wts_mml} and are given as:
\begin{equation}
  w_j^{(t+1)} = \frac{n_j^{(t)} + \frac{1}{2}}{N+\frac{M}{2}} \label{eqn:mml_weights}
\end{equation}

\noindent The parameters of the $j^{\text{th}}$ component are updated 
using $r_{ij}^{(t)}$ and $n_j^{(t)}$ (Equations \eqref{eqn:responsibility}, 
\eqref{eqn:comp_eff_mshp}), the partial memberships assigned to the 
$j^{\text{th}}$ component at some intermediate iteration $t$ and and are 
given below.
\begin{itemize}
\item \emph{Gaussian:} The MML updates of the mean and covariance matrix are
\begin{equation}
\hat{\boldsymbol{\mu}}_j^{(t+1)} = \frac{1}{n_j^{(t)}} \sum_{i=1}^N r_{ij}^{(t)} \mathbf{x}_i
\quad\text{and}\quad
\hat{\mathbf{C}}_j^{(t+1)} = \frac{1}{n_j^{(t)}-1}\sum_{i=1}^N r_{ij}^{(t)} 
\left(\mathbf{x}_i-\boldsymbol{\hat{\mu}}_j^{(t+1)}\right) \left(\mathbf{x}_i-\boldsymbol{\hat{\mu}}_j^{(t+1)}\right)^T
\label{eqn:mml_gaussian_updates}
\end{equation}
\item \emph{von Mises-Fisher:} The resultant vector sum is updated as 
$\mathbf{R}_j^{(t+1)} = \sum_{i=1}^N r_{ij}^{(t)} \mathbf{x}_i$. If $R_j^{(t+1)}$ represents
the magnitude of vector $\mathbf{R}_j^{(t+1)}$, then the updated mean is given by 
Equation \eqref{eqn:mml_vmf_mean_update}.
\begin{equation}
\hat{\boldsymbol{\mu}}_j^{(t+1)} = \frac{\mathbf{R}_j^{(t+1)}}{R_j^{(t+1)}}
\label{eqn:mml_vmf_mean_update}
\end{equation}
The MML update of the concentration parameter $\hat{\kappa}_j^{(t+1)}$ is 
obtained by solving $G(\hat{\kappa}_j^{(t+1)}) = 0$ after substituting 
$N \rightarrow n_j^{(t)}$ and $R \rightarrow R_j^{(t+1)}$ in Equation 
\eqref{eqn:I_first_derivative}.
\end{itemize}

The EM is terminated when the change in the total message length 
(improvement rate) between successive iterations is less than some predefined 
threshold.
The difference between the two variants of standard EM discussed above 
is firstly the objective function that is being optimized. In 
Section~\ref{subsec:em_ml}, the log-likelihood function is
\emph{maximized} which corresponds to $I(D|\boldsymbol{\Phi})$ term 
in Section~\ref{subsec:em_mml}. Equation (\ref{eqn:mixture_msglen})
includes additional terms that correspond to the cost associated
with stating the mixture parameters. Secondly, in the M-step, in 
Section~\ref{subsec:em_ml}, the components are updated using their 
ML estimates whereas in Section~\ref{subsec:em_mml},
the components are updated using their MML estimates.

\subsection{Issues arising from the use of EM algorithm}
The standard EM algorithms outlined above can be used only when the
number of mixture components $M$ is fixed or known \emph{a priori}. 
Even when the number of components are fixed, EM has potential pitfalls.
The method is sensitive to the initialization conditions.
To overcome this, some reasonable start state for the EM
may be determined by initially clustering the data 
\citep{krishnan1997algorithm,mclachlan2000finite}. Another strategy is to 
run the EM a few times and choose the best amongst all the trials. 
\cite{figueiredo2002unsupervised} point out that, in the case of
Gaussian mixture modelling, EM can converge to the 
boundary of the parameter space when the corresponding covariance matrix is
nearly singular or when there are few initial 
members assigned to that component.

\section{Existing methods of inferring the number of mixture components}
\label{sec:mixture_existing_methods} 
Inferring the ``right" number of mixture
components for unlabelled data has proven to be a thorny issue \citep{mclachlan2000finite}
and there have been numerous approaches proposed that attempt to tackle 
this problem \citep{aic,bic,rissanen1978modeling,icomp,oliver1996unsupervised,roberts,icl,figueiredo2002unsupervised}.
Given some observed data, there are infinitely many mixtures
that one can fit to the data. Any method that aims to selectively determine
the optimal number of components should be able to factor the cost 
associated with the mixture parameters. To this end, several methods based 
on information theory have been proposed where there is some form of 
penalty associated with choosing a certain parameter value
\citep{wallace68,aic,bic,wallace-87,rissanen1989stochastic}.
We briefly review some of these methods and discuss the state of the 
art and then proceed to explain our proposed method.

\subsection{\textbf{AIC} \citep{aic} \& \textbf{BIC} \citep{bic}} 
AIC in the simplest form adds the \emph{number} of free parameters $p$
to the negative log-likelihood expression.
There are some variants of AIC suggested \citep{bozdogan1983determining,burnham2002model}.
However, these variants introduce the same penalty constants for each 
additional parameter:
\begin{equation*}
  \text{AIC}(p) = p - \mathcal{L}(D|\boldsymbol{\Phi}) 
\end{equation*}
BIC, similar to AIC, adds a 
constant multiple of $\frac{1}{2}\log N$ ($N$ being the sample size), for each free
parameter in the model.
\begin{equation*}
  \text{BIC}(p) = \frac{p}{2}\log N - \mathcal{L}(D|\boldsymbol{\Phi})
\end{equation*}
\cite{rissanen1978modeling} formulated minimum description length (MDL)
which formally coincides with BIC \citep{oliver1996unsupervised,figueiredo2002unsupervised}. 

\subsubsection{Formulation of the scoring functions}
AIC and BIC/MDL serve as scoring functions to evaluate a model and its corresponding
fit to the data. The formulations suggest that
the parameter cost associated with adopting a model is dependent only on
the number of free parameters and \emph{not} on the parameter values themselves. In other words, the 
criteria consider
all models of a particular type (of probability distribution) to have the same statement cost
associated with the parameters. For example, a generic $d$-dimensional 
Gaussian distribution has $p = \frac{d(d+3)}{2}$ free parameters.
All such distributions will have the same parameter costs regardless of 
their characterizing means and covariance matrices, which is an oversimplifying
assumption which can hinder proper inference.

The criteria can be interpreted under the MML framework wherein the first part 
of the message is a constant multiplied by the number of free parameters.
AIC and BIC formulations can be obtained as approximations to the two-part 
MML formulation governed by Equation \eqref{eqn:two_part_msg}
\citep{figueiredo2002unsupervised}.
It has been argued that for tasks such as mixture 
modelling, where the number of free parameters potentially grows in proportion to
the data, MML
is known in theory to give consistent results as compared to AIC and BIC
\citep{wallace1986improved,Wallace99minimummessage}.

\subsubsection{Search method to determine the optimal number of mixture components}
To determine the optimal number of mixture components $M$, the AIC or
BIC scores are computed for mixtures with varying values of $M$. The mixture model
with the least score is selected as per these criteria.

A $d$-variate Gaussian mixture with $M$ number of components has 
$p=\dfrac{Md(d+3)}{2}+(M-1)$ free parameters. All mixtures with a set
number of components have the same cost associated with their parameters
using these criteria.
The mixture complexity is therefore treated as independent of the constituent
mixture parameters.
In contrast, the MML formulation incorporates the statement cost
of losslessly encoding mixture parameters by calculating their relevant probabilities
as discussed in Section \ref{sec:mml_mixture_modelling}.

\subsection{\textbf{MML Unsupervised} \citep{oliver1996unsupervised}} 
\subsubsection{Formulation of the scoring function}
A MML-based scoring function akin to the one shown in Equation \eqref{eqn:mixture_msglen} 
was used to model Gaussian mixtures. However,
the authors only consider the specific case of Gaussians 
with diagonal covariance matrices, and fail to provide a general 
method dealing with full covariance matrices. 

\subsubsection{Search method to determine the optimal number of mixture components}
A rigorous treatment on the selection of number of mixture
components $M$ is lacking. \cite{oliver1996unsupervised} experiment with different values of 
$M$ and choose the one which results in the minimum message length. 
For each $M$, the standard EM algorithm (Section~\ref{subsec:em_ml}) 
was used to attain local convergence.

\subsection{\textbf{Approximate Bayesian} \citep{roberts}} 
The method, also referred to as \emph{Laplace-empirical criterion} (LEC) 
\citep{mclachlan2000finite}, uses a scoring function  
derived using Bayesian inference 
and serves to provide a tradeoff between model complexity and the 
quality of fit. The parameter estimates $\boldsymbol{\Phi}$ are those 
that result in the minimum value of the following scoring function.
\begin{equation}
  -\log P(D,\boldsymbol{\Phi}) = -\mathcal{L}(D|\boldsymbol{\Phi}) + M d \log (2 \alpha\beta \sigma^2_{p}) - \log (M-1)! - \frac{N_d}{2} \log (2\pi) + \frac{1}{2} \log \,|H(\boldsymbol{\Phi})|  \label{eqn:approx_bayesian}
\end{equation} 
where $D$ is the dataset, $-\mathcal{L}(D|\boldsymbol{\Phi})$ is the 
negative 
log-likelihood given the mixture parameters $\boldsymbol{\Phi}$, $M$ is 
the number of mixture components, $d$ the dimensionality of the data, 
$\alpha, \beta$ are hyperparameters (which are set to 1 in their 
experiments), $\sigma_p$ is a pre-defined constant or is pre-computed 
using the entire data, $H (\boldsymbol{\Phi})$ is the Hessian matrix
which is equivalent to the empirical Fisher matrix for the set of component 
parameters, and $p$ is the number of free parameters in the model.

\subsubsection{Formulation of the scoring function}
The formulation in Equation \eqref{eqn:approx_bayesian} can be obtained as an
approximation to the message length expression in
Equation \eqref{eqn:mixture_msglen} by identifying the following related 
terms in both equations.
\begin{enumerate}
  \item $I(\mathbf{w}) \rightarrow - \log (M-1)!$
  \item For a $d$-variate Gaussian with mean $\boldsymbol{\mu}$ and 
  covariance matrix $\mathbf{C}$, the joint prior $h(\boldsymbol{\mu},\mathbf{C})$ 
  is calculated as follows:
  \begin{itemize}
    \item \emph{Prior on $\boldsymbol{\mu}$: } Each of the $d$ 
    parameters of the mean direction are assumed to be have uniform priors
    in the range $(-\alpha\sigma_p,\alpha\sigma_p)$, so that the prior 
    density of the mean is $h(\boldsymbol{\mu}) = \dfrac{1}{(2\alpha\sigma_p)^d}$.
    \item \emph{Prior on $\mathbf{C}$: } It is assumed that the prior 
    density is dependent only on the diagonal elements in $\mathbf{C}$.
    Each diagonal covariance element is assumed to have a prior
    in the range $(0,\beta\sigma_p)$ so that the prior on $\mathbf{C}$ 
    is considered to be $h(\mathbf{C}) = \dfrac{1}{(\beta\sigma_p)^d}$
  \end{itemize}
   The joint prior, is therefore, assumed to be $h(\boldsymbol{\mu},\mathbf{C}) = \dfrac{1}{(2\alpha\beta\sigma_p^2)^d}$.\\
   Thus, $-\sum_{j=1}^M  \log\,h(\Theta_j) \rightarrow M d \log (2 \alpha\beta \sigma^2_{p})$ \\

  \item $\frac{1}{2} \sum_{j=1}^M \log \,|\mathcal{F}(\Theta_j)| \rightarrow \frac{1}{2} \log \,|H|$ \\

  \item $\text{constant} \rightarrow   \frac{p}{2} \log (2\pi)$
\end{enumerate} 
Although the formulation is an improvement over the previously 
discussed methods, there are some limitations 
due to the assumptions made
while proposing the scoring function: 
\begin{itemize} 
  \item While computing the prior density of the covariance matrix, the 
  off-diagonal elements are ignored.
  \item The computation of the determinant of the Fisher matrix is 
  approximated by computing the Hessain $|H|$. It is to be noted that
  while the Hessian is the \emph{observed information} (data dependent),
  the Fisher information is the \emph{expectation} of the observed
  information. MML formulation requires the use of the expected value.
  \item Further, the approximated Hessian was derived for Gaussians with 
  diagonal covariances. For Gaussians with full covariance matrices, the 
  Hessian was approximated by replacing the diagonal elements with
  the corresponding eigen values in the Hessian expression. The 
  empirical Fisher computed in this form does not guarantee the charactersitic 
  invariance property of the classic MML method \citep{oliver1994mml}.
\end{itemize}

\subsubsection{Search method to determine the optimal number of mixture components}
The search method used to select the optimal number of components is rudimentary.
The optimal number of mixture components is chosen by running 
the EM 10 times for every value of $M$
within a given range. An optimal $M$ is selected as the one 
for which the best of the 10 trials results in the least value of the scoring function.

\subsection{\textbf{Integrated Complete Likelihood (ICL)} \citep{icl}} 
The ICL criterion \emph{maximizes} the \emph{complete log-likelihood} (CL) given by
\begin{equation*}
  CL(D,\boldsymbol{\Phi}) = \mathcal{L}(D|\boldsymbol{\Phi}) 
                          - \sum_{i=1}^N \sum_{j=1}^M z_{ij} \log r_{ij}
\end{equation*}
where $\mathcal{L}(D|\boldsymbol{\Phi})$ is the log-likelihood (Equation
\eqref{eqn:loglike_mixture}), $r_{ij}$ is the responsibility term 
(Equation~\eqref{eqn:responsibility}), and $z_{ij} = 1$ if $\mathbf{x}_i$
arises from component $j$ and zero otherwise. The term 
$\displaystyle\sum_{i=1}^N \sum_{j=1}^M z_{ij} \log r_{ij}$ is explained as the 
estimated mean entropy.

\subsubsection{Formulation of the scoring function}
The ICL criterion is then defined as: 
$
  ICL(\boldsymbol{\Phi},M) =  CL(\boldsymbol{\Phi}) - \dfrac{p}{2} \log N,
$
where $p$ is the number of free parameters in the model. We observe that similar to BIC, 
the ICL scoring function penalizes each free parameter by a constant value
and does not account for the model parameters.

\subsubsection{Search method to determine the optimal number of mixture components}
The search method adopted in this work is similar to the one used
by \citep{roberts}. The EM algorithm is initiated 20 times for each
value of $M$ with random starting points and the best amongst those
is chosen.

\subsection{\textbf{Unsupervised Learning of Finite Mixtures} \citep{figueiredo2002unsupervised}}
\label{subsec:fj}
The method uses the MML criterion to formulate the scoring function given by Equation \eqref{eqn:fj}.
The formulation can be intrepreted as a
two-part message for encoding the model parameters and the observed data.
\begin{equation}
I(D,\boldsymbol{\Phi}) = \underbrace{\frac{N_p}{2} \sum_{j=1}^M \log \left(\frac{Nw_j}{12} \right)
+ \frac{M}{2} \log\frac{N}{12} + \frac{M(N_p+1)}{2}}_{\text{first part}} \underbrace{- \mathcal{L}(D|\boldsymbol{\Phi})}_{\text{second part}} \label{eqn:fj}
\end{equation}
where $N_p$ is the \emph{number} of free parameters per component and $w_j$ is 
the component weight. 

\subsubsection{Formulation of the scoring function}
The scoring function is derived from Equation \eqref{eqn:mixture_msglen}
by assuming the prior density of the component parameters to be a Jeffreys
prior. If $\Theta_j$ is the vector of parameters describing the 
$j^{\text{th}}$ component, then the prior density 
$h(\Theta_j) \propto \sqrt{|\mathcal{F}(\Theta_j)|}$ \citep{jeffreys1946invariant}. 
Similarly, a prior for weights would result in 
$h(w_1,\ldots,w_M) \propto (w_1 \ldots w_M)^{-1/2}$. These 
assumptions are used in the encoding of the parameters which correspond to the
first part of the message.

We note that the scoring function is consistent with the MML scheme of 
encoding parameters and the data using those parameters. However, the 
formulation can be improved by amending the assumptions as detailed in
in Section \ref{sec:mml_est_derivations}. Further, the assumptions
made in \cite{figueiredo2002unsupervised} have the following side effects:
\begin{itemize}
  \item The value of $-\log\dfrac{h(\Theta_j)}{\sqrt{|\mathcal{F}(\Theta_j)|}}$
  gives the cost of encoding the component parameters.
  By assuming 
  $h(\Theta_j) \propto \sqrt{|\mathcal{F}(\Theta_j)|}$, 
  the message length associated with using any vector of parameters
  $\Theta_j$ is essentially treated the same. To avoid 
  this, the use of independent uniform priors over non-informative 
  Jeffreys's priors was advocated previously \citep{oliver1996unsupervised,lee1994bayesian,roberts}.
  The use of Jeffreys prior offers certain advantages, for example,
  not having to compute the Fisher information \citep{jeffreys1946invariant}. However, this is 
  crucial and cannot be ignored as it dictates the \emph{precision of 
  encoding the parameter vector}.
  \cite{WallaceBook} states that 
  ``Jeffreys, while noting the interesting properties of the
  prior formulation did not advocate its use as a genuine expression
  of prior knowledge."

  By making this assumption, \cite{figueiredo2002unsupervised}
  ``\emph{sidestep}" the difficulty associated with explicitly computing the
  Fisher information associated with the component parameters. Hence,
  for encoding the parameters of the entire mixture, \emph{only} the 
  cost associated with encoding the component weights is considered.

  \item The code length to state each $\Theta_j$ is,
  therefore, greatly simplified as $(N_p/2)\log(Nw_j)$ (notice the 
  sole dependence on weight $w_j$). \cite{figueiredo2002unsupervised}
  interpret this as being similar to a MDL
  formulation because $Nw_j$ gives the expected number of data points
  generated by the $j^{\text{th}}$ component. This is equivalent to 
  the BIC criterion discussed earlier. We note that MDL/BIC are highly
  simplified versions of MML formulation and therefore, Equation
  \eqref{eqn:fj} does not capture the entire essence of complexity
  and goodness of fit accurately.
%
\end{itemize}

\subsubsection{Search method to determine the optimal number of mixture components}
\label{subsubsec:fj_search_drawbacks}
The method begins by assuming a large number of 
components and updates the weights iteratively in the EM steps as
\begin{equation}
  w_j = \frac{\text{max}\left\{0,n_j-\frac{N_p}{2}\right\}}
                  {\sum_{j=1}^M \text{max}\left\{0,n_j-\frac{N_p}{2}\right\}}  
  \label{eqn:fj_weight_update}
\end{equation}
where $n_j$ is the effective membership of data points in $j^{\text{th}}$
component (Equation \eqref{eqn:comp_eff_mshp}). A component is 
annihilated when its weight becomes zero and consequently the number of
mixture components decreases.
We note that the search method proposed by \cite{figueiredo2002unsupervised}
using the MML criterion is an improvement over
the methods they compare against.
However, we make the following remarks about their search method.
\begin{itemize}
  \item The method updates the weights as given by Equation
  \eqref{eqn:fj_weight_update}. During any iteration,
  if the amount of data allocated to a component is less than $N_p/2$,
  its weight is updated as zero and this component is ignored in
  subsequent iterations. This imposes a lower bound on the amount of
  data that can be assigned to each component. As an example, for a
  Gaussian mixture
  in 10-dimensions, the number of free parameters per component
  is $N_p = 65$, and hence the lower bound is 33. Hence, in this 
  exampe, if a component has $\sim 30$ data, the mixture size is 
  reduced and these data are assigned to some other component(s).
  Consider a scenario where there are 50 observed 10 dimensional 
  data points originally generated by a mixture with two components 
  with equal mixing proportions. The method would always infer that 
  there is only one component regardless of the separation between the
  two components. This is clearly a wrong inference! (see Section 
  \ref{subsec:fj_weight_updates_exp2b} for the relevant experiments).

  \item Once a component is discarded, the mixture size decreases by
  one, and it cannot be recovered. Because the memberships $n_j$ are 
  updated iteratively using an EM algorithm and because EM might
  not always lead to global optimum, it is conceivable that the 
  updated values need not always be optimal. This might lead to
  situations where a component is deleted owing to its low
  prominence. There is no provision to increase the mixture size in 
  the subsequent stages of the algorithm to account for such 
  behaviour.

  \item The method assumes a large number of initial components in an
  attempt to be robust with respect to EM initialization. However,
  this places a significant overhead on the computation due to 
  handling several components.
\end{itemize}

\noindent\emph{Summary:} We observe that while all these methods
(and many more) work well within 
their defined scope, they are incomplete in achieving the true objective
that is to rigorously score models and their ability to fit the data.
The methods discussed above
can be seen as different approximations to the MML framework.
They adopted various simplifying assumptions and approximations.
To avoid such limitations, we developed a classic MML formulation, giving
the complete message length formulations for Gaussian and 
von Mises-Fisher distributions in Section \ref{sec:mml_est_derivations}.

Secondly, in most of these methods, the search for the optimal number of 
mixture components is achieved by selecting the mixture that results in the 
best EM outcome out of many trials \citep{aic,bic,oliver1996unsupervised,roberts,icl}.
This is not an elegant solution and \cite{figueiredo2002unsupervised}
proposed a search heuristic which integrates estimation and model selection.
A comparative study of these methods is presented in \cite{mclachlan2000finite}.
Their analysis suggested the superior performance of ICL \citep{icl}
and LEC \citep{roberts}. Later, \cite{figueiredo2002unsupervised}
demonstrated that their proposed method outperforms the contemporary
methods based on ICL and LEC and is regarded as the current state of the 
art. We, therefore, compare our method against that of
\cite{figueiredo2002unsupervised} and demonstrate its effectiveness.

With this background, we formulate an alternate search heuristic 
to infer the optimal number of 
mixture components which aims to address the above limitations.

\section{Proposed approach to infer an optimal mixture}
\label{sec:search_method}
The space of candidate mixture models to explain the given data is infinitely large.
As per the MML criterion (Equation~\eqref{eqn:mixture_msglen}),
the goal is to search for the mixture that has the smallest overall message length.
We have seen in Section~\ref{subsec:em_mml} that if the number of mixture
components are fixed, then the EM algorithm can be used to estimate the
mixture parameters, namely the component weights and the parameters of each component.
However, here it is required to search for the optimal \emph{number} 
of mixture components along with the corresponding mixture parameters.

Our proposed search heuristic extends the MML-based Snob program 
\citep{wallace68,wallace1986improved,jorgensen2008wallace} for unsupervised learning.
We define three operations, namely \emph{split, delete,} and \emph{merge}
that can be applied to any component in the mixture. 

\subsection{The complete algorithm}
\begin{algorithm}[htb]\label{algm}
\DontPrintSemicolon
\caption{Achieve an optimal mixture model}
  $current \gets \textrm{one-component-mixture}$\;
  \While{$true$} {
    $components \gets current \,\,\textrm{mixture components}$\;
    $M \gets \textrm{number of} components$\;
    \For(\tcc*[f]{exhaustively split all components})
    {$i \gets 1$ \textbf{ to } $M$} { 
       $splits[i] \gets \textrm{Split}(current,components[i])$\;
    }
    $BestSplit \gets best(splits)$\tcc*[r]{remember the best split}
    \If{$M > 1$} {
      \For(\tcc*[f]{exhaustively delete all components})
      {$i \gets 1 \textrm{ to } M$} {
         $deletes[i] \gets \textrm{Delete}(current,components[i])$\;
      }
      $BestDelete \gets best(deletes)$\tcc*[r]{remember the best deletion}
    }
    \For(\tcc*[f]{exhaustively merge all components})
    {$i \gets 1 \textbf{ to } M$} {
       $j \gets \textrm{closest-component}(i)$\;
       $merges[i] \gets \textrm{Merge}(current,i,j)$\;
    }
    $BestMerge \gets best(merges)$\tcc*[r]{remember the best merge}
    $BestPerturbation \gets best(BestSplit,BestDelete,BestMerge)$\tcc*[r]{select the best perturbation}
    $\Delta I \gets \textrm{message\_length}(BestPerturbation) - \textrm{message\_length}(current)$\tcc*[r]{check for improvement}
    \eIf{$\Delta I < 0$} {
      $current \gets BestPerturbation$\;
      $continue$\;
    } {
      $break$\;
    }
  }
  \Return{$current$}\;
\end{algorithm}
The pseudocode of our search method is presented in Algorithm \ref{algm}.
The basic idea behind the search strategy is to \emph{perturb} a mixture
from its current suboptimal state to obtain a new state 
(if the perturbed mixture results in a smaller message length).
In general, if a (current) mixture has $M$ components, it is
perturbed using a series of \emph{Split, Delete}, and \emph{Merge} operations
to check for improvement. Each component is split
and the new $(M+1)$-component mixture is re-estimated. 
If there is an improvement (\textit{i.e.,} if there is a decrease in 
message length with respect to the current mixture), the new $(M+1)$-component 
mixture is retained. There are $M$ splits 
possible and the one that results in the greatest improvement is recorded
(lines 5 -- 7 in Algorithm~\ref{algm}).
A component is first split into two sub-components (children) which are 
locally optimized by the EM algorithm on the data that belongs to that sole 
component. The child components are then integrated with the others and the 
mixture is then optimized to generate a $M+1$ component mixture. The reason
for this is, rather than use random initial values for the EM, it is better if we 
start from some already optimized state to reach to a better state. Similarly, 
each of the components is then deleted, one after the other, and the $(M-1)$-component mixture is 
compared against the current mixture. There are $M$ possible deletions and 
the best amongst these is recorded (lines 8 -- 11 in Algorithm~\ref{algm}).
Finally, the components in the current 
mixture are merged with their closest matches (determined by calculating the KL-divergence) 
and each of the resultant 
$(M-1)$-component mixtures are evaluated against the $M$ component mixture. 
The best among these merged mixtures is then retained
(lines 12 -- 15 in Algorithm~\ref{algm}).

We initially start by assuming a 
one component mixture. This component is split into two children which are 
locally optimized. If the split results in a better model, it is retained.
For any given $M$-component mixture, there might be improvement due to 
splitting, deleting and/or merging its components. We select the perturbation
that best improves the current mixture. 
This process is repeated until there is no further improvement possible. 
and the algorithm is continued. The notion of \emph{best} or improved
mixture is based on the amount of reduction of message length that the perturbed 
mixture provides.
In the current state, the observed data have partial memberships in each of the $M$
components. Before the execution of each operation, these memberships need 
to be adjusted and a EM is subsequently carried out to achieve an optimum
with a different number of components.
We will now examine each operation in detail and see how the memberships
are affected after each operation.

\subsection{Strategic operations employed to determine an optimal mixture model}
\label{subsec:search_operations}
Let $R=[r_{ij}]$ be the $N\times M$
responsibility (membership) matrix and $w_j$ be the weight of $j^{\text{th}}$ component 
in mixture $\fancym$. 
\begin{enumerate}
  \item \emph{Split (Line 6 in Algorithm~\ref{algm}):} As an example, assume a component with index 
  $\alpha \in \{1,M\}$ and  
  weight $w_{\alpha}$ in the current mixture $\fancym$ is split to generate two child 
  components. The goal is to find two distinct clusters amongst the data 
  associated with component $\alpha$. It is to be noted that the data have 
  fractional memberships to component $\alpha$. The EM is therefore, carried
  out \emph{within} the component $\alpha$ assuming a \emph{two-component 
  sub-mixture} with the data weighted as per their current memberships 
  $r_{i\alpha}$. The remaining $M-1$ components are untouched. 
  An EM is carried out to optimize the
  two-component sub-mixture. The initial state and the subsequent updates 
  in the Maximization-step are described below. \\

  \noindent\emph{Parameter initialization of the two-component sub-mixture:} The goal is 
  to identify two distinct clusters within the component $\alpha$. 
  For \emph{Gaussian} mixtures, to provide
  a reasonable starting point, we compute the direction of maximum variance
  of the parent component and locate two points which are one standard deviation
  away on either side of its mean (along this direction). These points serve 
  as the initial
  means for the two children generated due to splitting the parent component.
  Selecting the initial means in this manner
  ensures they are reasonably apart from each other and serves as a good
  starting point for optimizing the two-component sub-mixture.
  The memberships are initialized by allocating the data points to the closest
  of the two means. Once the means and the memberships are initialized,
  the covariance matrices of the two child components are computed. 

  There are conceivably several variations to how the two-component sub-mixture
  can be initialized. These include random initialization, selecting two data
  points as the initial component means, and many others. However, the reason for
  selecting the direction of maximum variance is to utilize the available 
  characteristic of data, \textit{i.e.,} the distribution within the component $\alpha$.
  For \emph{von Mises-Fisher} mixtures, the maximum variance
  strategy (as for Gaussian mixtures) cannot be easily adopted, 
  as the data is distributed on the hypershpere.
  Hence, in this work, we randomly allocate data memberships
  and compute the components' (initial) parameters.
  
  Once the parameters of the sub-mixture are initialized, an EM algorithm is carried out
  (just for the sub-mixture)
  with the following Maximization-step updates.
  Let $R^c=[r^c_{ik}]$ be the $N\times 2$ 
  responsibility matrix for the two-component sub-mixture. For $k \in \{1,2\}$,
  let $n_{\alpha}^{(k)}$ be the effective memberships of data belonging to 
  the two child components, let $w_{\alpha}^{(k)}$ be the weights of the 
  child components within the sub-mixture, and let $\Theta_{\alpha}^{(k)}$ 
  be the parameters describing the child components.
  \begin{itemize}
    \item The effective memberships are updated as given by Equation
    \eqref{eqn:split_update_mshp}.
    \begin{equation}
      n_{\alpha}^{(k)} = \sum_{i=1}^N r^c_{ik} 
      \quad\text{and}\quad 
      n_{\alpha}^{(1)}+n_{\alpha}^{(2)} = N
    \label{eqn:split_update_mshp}
    \end{equation}

    \item As the sub-mixture comprises of two child components, substitute
    $M=2$ in Equation \eqref{eqn:mml_weights} to obtain the updates for the
    weights. These are given by Equation \eqref{eqn:split_update_weights}.
    \begin{equation}
      w_{\alpha}^{(k)} = \frac{n_{\alpha}^{(k)}+\frac{1}{2}}{N+1} 
      \quad\text{and}\quad 
      w_{\alpha}^{(1)}+w_{\alpha}^{(2)} = 1
    \label{eqn:split_update_weights}
    \end{equation}

    \item For \emph{Gaussian} mixtures, the component parameters
    $\Theta_{\alpha}^{(k)} = (\hat{\boldsymbol{\mu}}_{\alpha}^{(k)},\hat{\mathbf{C}}_{\alpha}^{(k)})$ 
    are updated as follows:
    \begin{equation}
      \hat{\boldsymbol{\mu}}_{\alpha}^{(k)} = \frac{\displaystyle\sum_{i=1}^N r_{i{\alpha}} r^c_{ik} \mathbf{x}_i}{\displaystyle\sum_{i=1}^N r_{i{\alpha}} r^c_{ik}} 
      \quad\text{and}\quad
      \hat{\mathbf{C}}_{\alpha}^{(k)} = \frac{\displaystyle\sum_{i=1}^N r_{i{\alpha}} r^c_{ik}(\mathbf{x}_i-\boldsymbol{\hat{\mu}}_{\alpha}^{(k)}) (\mathbf{x}_i-\boldsymbol{\hat{\mu}}_{\alpha}^{(k)})^T
    }{\displaystyle\sum_{i=1}^N r_{i{\alpha}} r^c_{ik} - 1}  
    \label{eqn:split_update_gaussian_params}
    \end{equation}

    \item For \emph{von Mises-Fisher} mixtures, the component parameters
    $\Theta_{\alpha}^{(k)} = (\hat{\boldsymbol{\mu}}_{\alpha}^{(k)},\hat{\kappa}_{\alpha}^{(k)})$
    are updated as follows:
    \begin{equation}
      \hat{\boldsymbol{\mu}}_{\alpha}^{(k)} = \frac{\mathbf{R}_{\alpha}^{(k)}}{R_{\alpha}^{(k)}}
      \quad\text{where}\quad
      \mathbf{R}_{\alpha}^{(k)} = \sum_{i=1}^N r_{i{\alpha}} r^c_{ik}\mathbf{x}_i
    \label{eqn:split_update_vmf_params}
    \end{equation}
    $R_{\alpha}^{(k)}$ represents the magnitude of vector $\mathbf{R}_{\alpha}^{(k)}$.
    The update of the concentration parameter $\hat{\kappa}_{\alpha}^{(k)}$ 
    is obtained by solving $G(\hat{\kappa}_{\alpha}^{(k)}) = 0$ after 
    substituting $N \rightarrow \sum_{i=1}^N r_{i{\alpha}} r^c_{ik}$ and 
    $R \rightarrow R_{\alpha}^{(k)}$ in Equation \eqref{eqn:I_first_derivative}.
  \end{itemize}

  The difference between the EM updates in Equations \eqref{eqn:mml_gaussian_updates}, 
  \eqref{eqn:mml_vmf_mean_update} and Equations \eqref{eqn:split_update_gaussian_params}, 
  \eqref{eqn:split_update_vmf_params} is the presence of the coefficient 
  $r_{i{\alpha}} r^c_{ik}$ with each $\mathbf{x}_i$. Since we are 
  considering the sub-mixture, the original responsibility $r_{i{\alpha}}$ 
  is multiplied by the responsibility within the sub-mixture $r^c_{ik}$ to 
  quantify the influence of datum $\mathbf{x}_i$ to each of the child components.
  
  After the sub-mixture is locally optimized, it is integrated with the untouched
  $M-1$ components of $\fancym$ to result in a $M+1$ component mixture
  $\fancym'$. An EM is finally carried out on the combined $M+1$ components to 
  estimate the parameters of $\fancym'$ and result in an optimized
  $(M+1)$-component mixture as follows.\\ 

  \noindent \emph{EM initialization for $\fancym'$:}
  Usually, the EM is started by a random initialization of the members. 
  However, because the two-component sub-mixture is now optimal and the 
  $M-1$ components in $\fancym$ are also in an optimal state, we 
  exploit this situation to initialize the EM (for $\fancym'$) with a 
  reasonable starting point. As mentioned above, the component with index
  $\alpha$ with component weight $w_{\alpha}$ is split. Upon integration, 
  the (child) components that replaced component $\alpha$ will now 
  correspond to indices ${\alpha}$ and ${\alpha}+1$ in the new mixture 
  $\fancym'$. Let $R'=[r'_{ij}] \,\forall 1 \leq i \leq N, 1 \leq j \leq M+1$ 
  be the responsibility matrix for the new mixture $\fancym'$ and let
  $w_j'$ be the component weights in $\fancym'$.
  \begin{itemize}
    \item \emph{Component weights:} The weights are initialized as follows:
    \begin{align}
      w_j' &= w_j \quad\text{if}\quad j < \alpha \notag\\
      w'_{\alpha} = w_{\alpha} w_{\alpha}^{(1)}
      \quad&\text{and}\quad
      w'_{\alpha+1} = w_{\alpha} w_{\alpha}^{(2)} \notag\\
      w_j' &= w_{j-1} \quad\text{if}\quad j > \alpha+1
    \end{align}

    \item \emph{Memberships:} The responsibility matrix $R'$ is 
    initialized for all data $\mathbf{x}_i \, \forall 1 \le i \le N$ as 
    follows:
    \begin{align}
      r'_{ij} &= r_{ij} \quad\text{if}\quad j < \alpha \notag\\
      r'_{i\alpha} = r_{i\alpha} r^c_{i1}
      \quad&\text{and}\quad
      r'_{i\,\alpha+1} = r_{i\alpha} r^c_{i2} \notag\\
      r'_{ij} &= r_{i\,j-1} \quad\text{if}\quad j > \alpha+1 \notag\\
      \text{and}\quad
      n'_{j} &= \sum_{i=1}^N r'_{ij} \quad\forall \, 1 \le j \le M+1
    \end{align}
    where $n_j'$ are the effective memberships of the components in $\fancym'$ .
  \end{itemize}

  With these starting points, the parameters of $\fancym'$ are estimated
  using the traditional EM algorithm with updates in the Maximization-step given by
  Equations \eqref{eqn:mml_weights}, \eqref{eqn:mml_gaussian_updates}, and
  \eqref{eqn:mml_vmf_mean_update}. 
  The EM results in local convergence of the $(M+1)$-component mixture. If
  the resultant message length of encoding data using $\fancym'$ is 
  lower than that due to $\fancym$, that means the perturbation of 
  $\fancym$ because of splitting component $\alpha$ resulted in a
  new mixture $\fancym'$ that compresses the data better, and hence,
  is a better mixture model to explain the data.

  \item \emph{Delete (Line 10 in Algorithm~\ref{algm}):} The goal here is 
  to remove a component from the current mixture and check whether it 
  results in a better mixture model to explain the observed data. Assume 
  the component with index $\alpha$ and the
  corresponding weight $w_{\alpha}$ is to be deleted from $\fancym$ to
  generate a $M-1$ component mixture $\fancym'$. Once deleted, the
  data memberships of the component need to be redistributed between the
  remaining components. The redistribution of data results in a good
  starting point to employ the EM algorithm to estimate the parameters of
  $\fancym'$ as follows. \\ 

  \noindent \emph{EM initialization for $\fancym'$:}
  Let $R'=[r'_{ij}]$ 
  be the $N\times(M-1)$ responsibility matrix for the new mixture $\fancym'$ and let
  $w_j'$ be the weight of $j^{\text{th}}$ component in $\fancym'$.
  \begin{itemize}
    \item \emph{Component weights:} The weights are initialized as follows:
    \begin{align}
      w_j' &= \frac{w_j}{1-w_{\alpha}} \quad\text{if}\quad j < \alpha \notag\\
      w_j' &= \frac{w_{j+1}}{1-w_{\alpha}} \quad\text{if}\quad j \ge \alpha
    \end{align}
    It is to be noted that $w_{\alpha} \ne 1$ because the MML update
    expression in the M-step for the component weights always ensures 
    non-zero weights during every iteration of the EM algorithm (see 
    Equation \eqref{eqn:mml_weights}).

    \item \emph{Memberships:} The responsibility matrix $R'$ is 
    initialized for all data $\mathbf{x}_i \, \forall 1 \le i \le N$ as 
    follows:
    \begin{align}
      r'_{ij} &= \frac{r_{ij}}{1-r_{i\alpha}} \quad\text{if}\quad j < \alpha \notag\\
      r'_{ij} &= \frac{r_{i\,(j+1)}}{1-r_{i\alpha}} \quad\text{if}\quad j \ge \alpha \notag\\
      \text{and}\quad
      n'_{j} &= \sum_{i=1}^N r'_{ij} \forall \, 1 \le j \le M-1
    \end{align}
    where $n_j'$ are the effective memberships of the components in 
    $\fancym'$. It is possible for a datum $\mathbf{x}_i$ to have
    complete membership in component $\alpha$ (\textit{i.e.,} $r_{i\alpha} = 1$),
    in which case, its membership is equally distributed among the
    other $M-1$ components (\textit{i.e.,} $r'_{ij} = \dfrac{1}{M-1}, \,
    \forall \,j \in \{1,M-1\}$).
  \end{itemize}

  With these readjusted weights and memberships, and the constituent
  $M-1$ components, the traditional EM algorithm is used to estimate the
  parameters of the new mixture $\fancym'$. If the resultant message 
  length of encoding data using $\fancym'$ is lower than that due to 
  $\fancym$, that means the perturbation of $\fancym$ because of 
  deleting component $\alpha$ resulted in a new mixture $\fancym'$ with 
  better explanatory power, which is an improvement over the current
  mixture. 

  \item \emph{Merge (Line 14 in Algorithm~\ref{algm}):} The idea is to join 
  a pair of components of 
  $\fancym$ and determine whether the resulting $(M-1)$-component mixture 
  $\fancym'$ is any better than the current mixture $\fancym$. One 
  strategy to identify an improved mixture model would be to consider 
  merging all possible pairs of components and choose the one which results 
  in the greatest improvement. This would, however, lead to a runtime 
  complexity of $O(M^2)$, which could be significant for large values of 
  $M$. Another strategy is to consider merging components which are 
  ``close" to each other.
  For a given component, we identify its \emph{closest}
  component by computing the Kullback-Leibler (KL) distance with all others and selecting the
  one with the least value. This would result in a linear runtime complexity
  of $O(M)$ as computation of KL-divergence is a constant time operation.
  For every component in $\fancym$, its closest match is identified
  and they are merged to obtain a $M-1$ component mixture $\fancym'$. Merging
  the pair involves reassigning the component weights and the memberships. 
  An EM algorithm is then employed to optimize $\fancym'$.  

  Assume components with indices $\alpha$ and $\beta$ are merged. Let their
  weights be $w_{\alpha}$ and $w_{\beta}$; and their
  responsibility terms be $r_{i\alpha}$ and $r_{i\beta}, 1 \le i \le N$ respectively.
  The component that is formed by merging the pair is determined first. It is 
  then integrated with the $M-2$ remaining components of $\fancym$ to
  produce a $(M-1)$-component mixture $\fancym'$. \\

  \noindent \emph{EM initialization for $\fancym'$:}
  Let $w^{(m)}$ and $r^{(m)}_i$ be the weight and responsibility vector
  of the merged component $m$ respectively. They are given as follows:
  \begin{align}
    w^{(m)} &= w_{\alpha} + w_{\beta} \notag\\
    r^{(m)}_{i} &= r_{i\alpha} + r_{i\beta}, 1 \le i \le N
  \end{align}
  The parameters of this merged component are estimated as follows:
  \begin{itemize}
    \item \emph{Gaussian:} The parameters
    $\Theta^{(m)} = (\hat{\boldsymbol{\mu}}^{(m)},\hat{\mathbf{C}}^{(m)})$ 
    are:
    \begin{equation}
      \hat{\boldsymbol{\mu}}^{(m)} = \frac{\displaystyle\sum_{i=1}^N r^{(m)}_i \mathbf{x}_i}{\displaystyle\sum_{i=1}^N r^{(m)}_i}
      \quad\text{and}\quad
      \hat{\mathbf{C}}^{(m)} = \frac{\displaystyle\sum_{i=1}^N r^{(m)}_i(\mathbf{x}_i-\boldsymbol{\hat{\mu}}^{(m)}) (\mathbf{x}_i-\boldsymbol{\hat{\mu}}^{(m)})^T
    }{\displaystyle\sum_{i=1}^N r^{(m)}_i - 1}
    \label{eqn:merge_update_gaussian_params}
    \end{equation}

    \item \emph{von Mises-Fisher:} The parameters
    $\Theta^{(m)} = (\hat{\boldsymbol{\mu}}^{(m)},\hat{\kappa}^{(m)})$ are:
    \begin{equation}
      \hat{\boldsymbol{\mu}}^{(m)} = \frac{\mathbf{R}^{(m)}}{R^{(m)}}
      \quad\text{where}\quad
      \mathbf{R}^{(m)} = \sum_{i=1}^N r^{(m)}_i \mathbf{x}_i
    \label{eqn:merge_update_vmf_params}
    \end{equation}
    The concentration parameter $\hat{\kappa}^{(m)}$ is obtained by solving 
    $G(\hat{\kappa}^{(m)}) = 0$ after 
    substituting $N \rightarrow \sum_{i=1}^N r^{(m)}_i$ and 
    $R \rightarrow R^{(m)}$ in Equation \eqref{eqn:I_first_derivative}.
  \end{itemize}

  The merged component $m$ with weight $w^{(m)}$, responsibility vector
  $r_i^{(m)}$, and parameters $\Theta^{(m)}$ is then integrated with the
  $M-2$ components. The merged component and its associated memberships
  along with the $M-2$ other components serve as the starting point for 
  optimizing the new
  mixture $\fancym'$. If $\fancym'$ results in a lower message 
  length compared to $\fancym$ that means the perturbation of 
  $\fancym$ because of merging the pair of components  
  resulted in an improvement to the current mixture. 
\end{enumerate}

\subsection{Illustrative example of our search procedure} 

\begin{wrapfigure}{r}{0.35\textwidth}
\centering
\includegraphics[width=\textwidth]{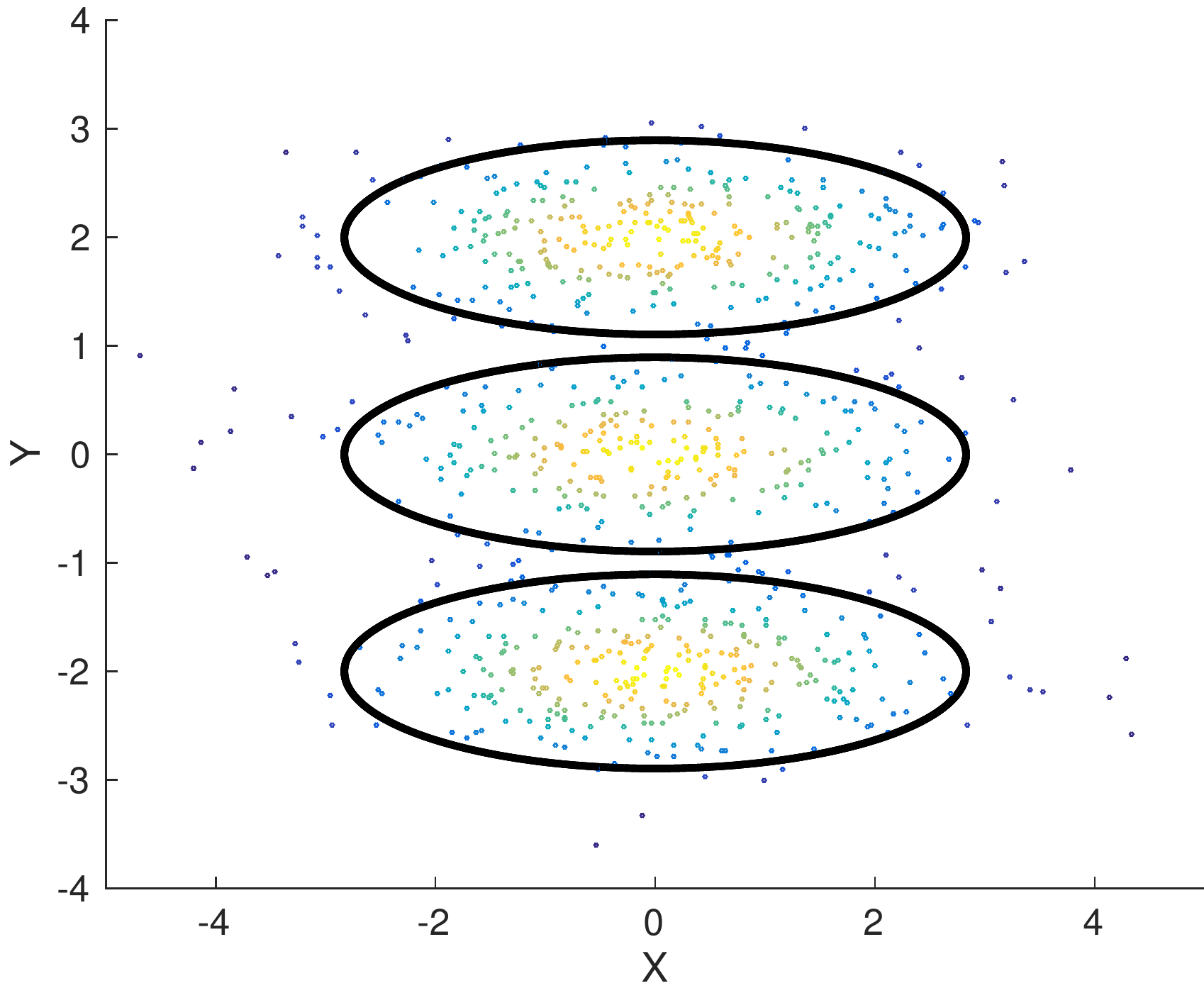}
\caption{Original mixture consisting of three components with equal mixing proportions.}
\label{fig:mix1}
\end{wrapfigure}
We explain the proposed inference of mixture components through 
the following example that was also considered by \cite{figueiredo2002unsupervised}.
Consider a bivariate Gaussian mixture shown in Fig.~\ref{fig:mix1}. 
The mixture has three
components with equal weights of 1/3 each and their means at (-2,0), (0,0), and (2,0).
The covariance matrices of the three components are the same and are equal to 
$\text{diag}\{2,0.2\}$. We simulate 900 data points from this mixture
(as done by \cite{figueiredo2002unsupervised})
and employ the proposed search strategy. 
The progression of the search method using various operations is detailed below.

\begin{figure}[htb]
  \centering
  \subfloat[]
  {
    \includegraphics[width=0.33\textwidth]{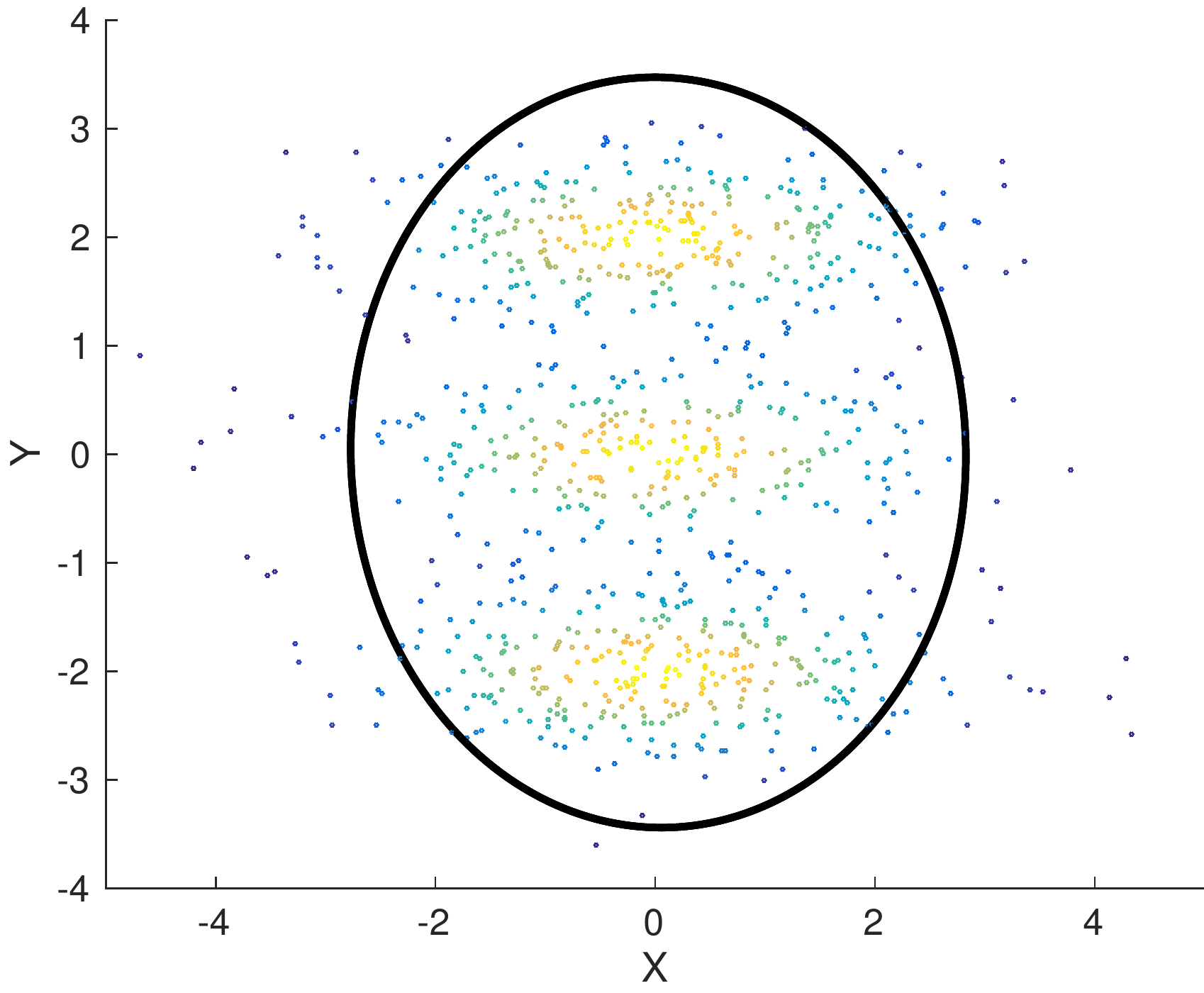}
  } 
  \subfloat[]
  {
    \includegraphics[width=0.33\textwidth]{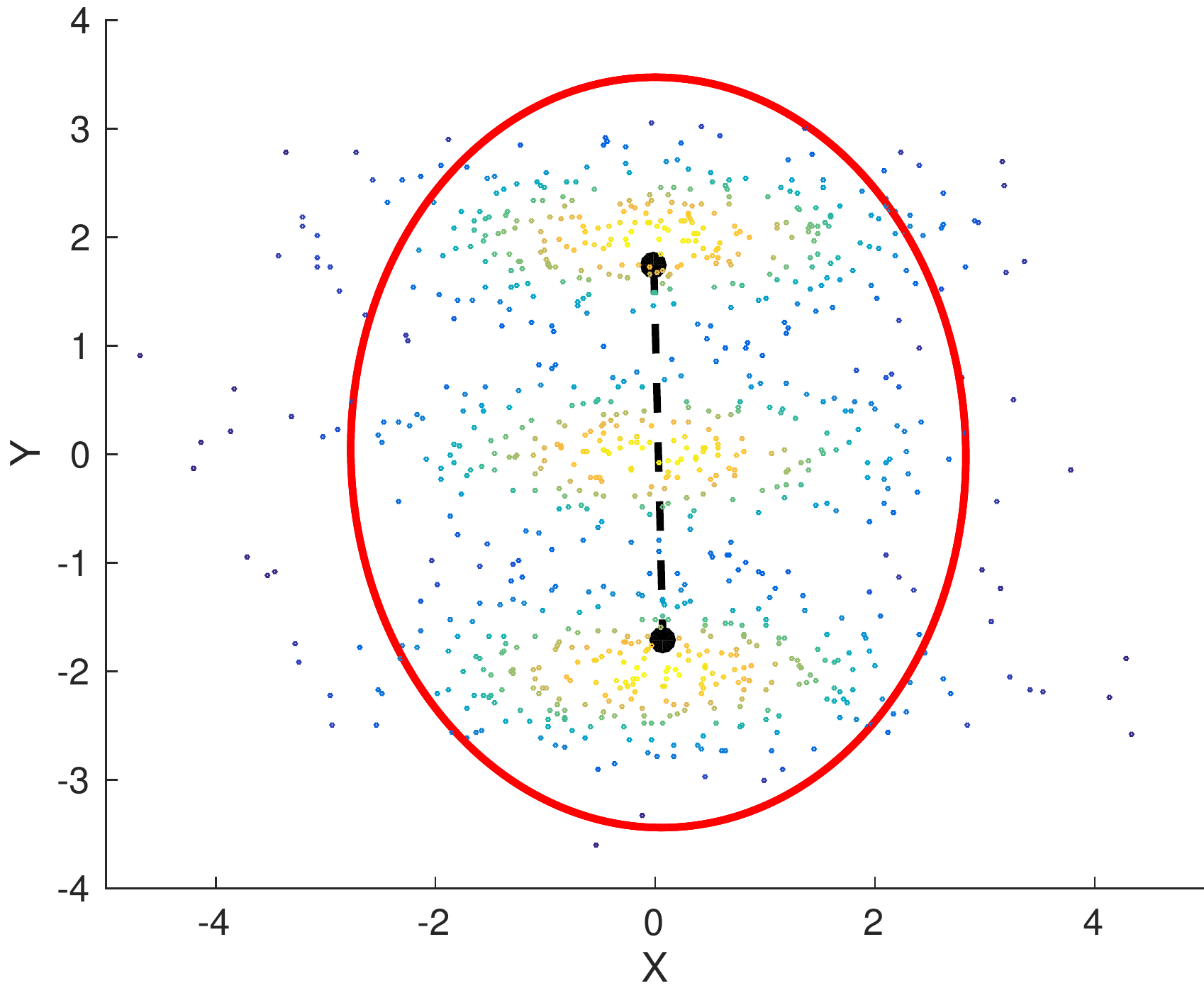}
  }
  \subfloat[]
  {
    \includegraphics[width=0.33\textwidth]{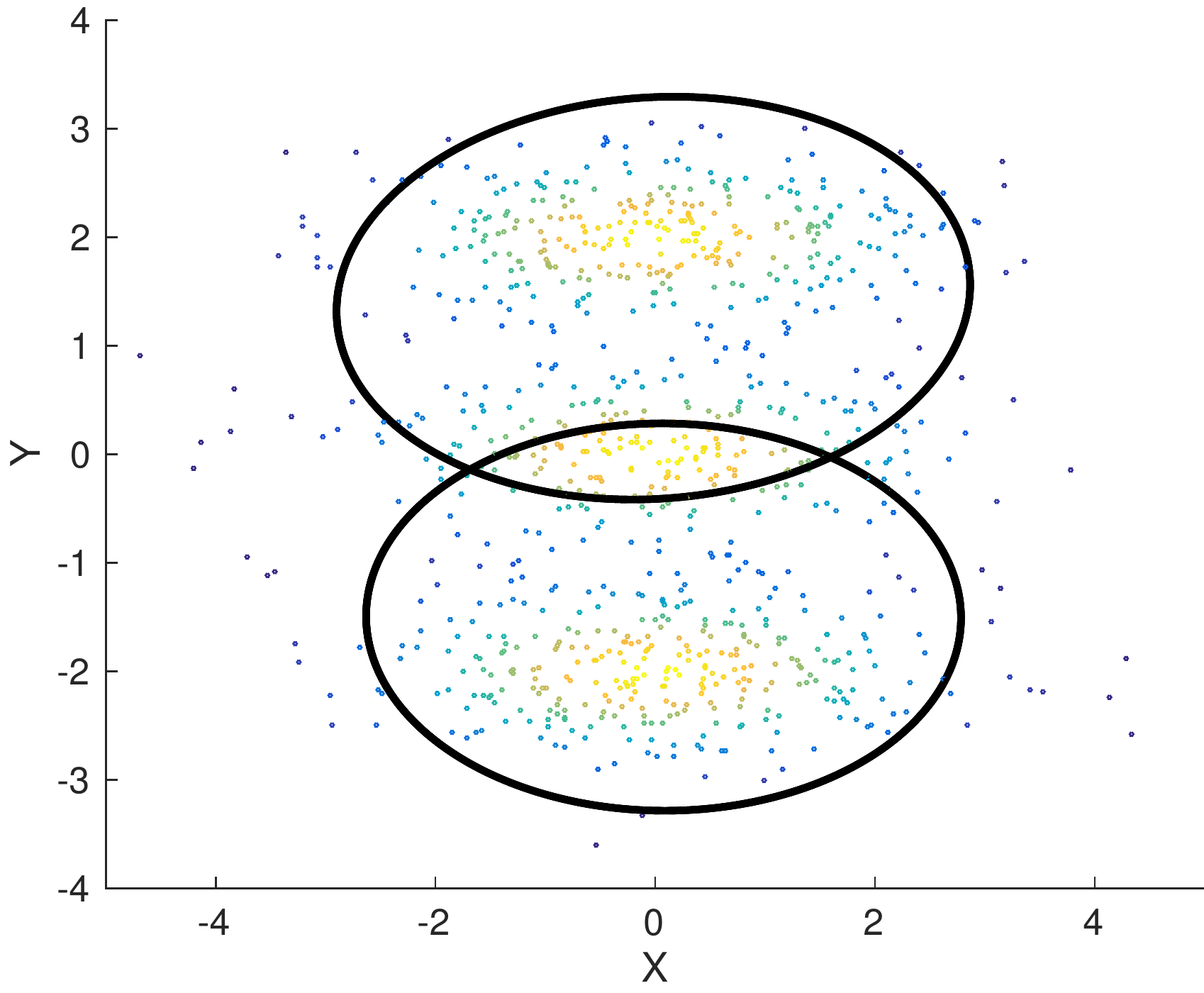}
  } 
  \caption{(a) $P_1$: the one-component mixture after the first iteration (message length $I = 22793$ bits) 
           (b) Red colour denotes the component being split. The dotted line is the direction
               of maximum variance. The black dots represent the initial  means of the 
               two-component sub-mixture
           (c) $P_2$: optimized mixture post-EM phase ($I = 22673$ bits) results in an improvement.
          }
  \label{fig:mix1_iter_1_splits}
\end{figure}

\noindent\emph{Search for the optimal mixture model:}
The method begins by inferring a one-component mixture $P_1$ (see Fig.~\ref{fig:mix1_iter_1_splits}(a)).
It then splits this component (as described in \emph{Split} step of 
Section~\ref{subsec:search_operations})
and checks whether there is an improvement in explanation.
The red ellipse in Fig~\ref{fig:mix1_iter_1_splits}(b) depicts the component being split. 
The direction of maximum variance (dotted black line) is first 
identified, and the means (shown by black dots at the end of the dotted line) 
are initialized. An EM algorithm is then used to
optimize the two children and this results
in a mixture $P_2$ shown in Fig~\ref{fig:mix1_iter_1_splits}(c). Since the new mixture
has a lower message length, the current is updated as $P_2$.

\begin{figure}[htb]
  \subfloat[]
  {
    \includegraphics[width=0.33\textwidth]{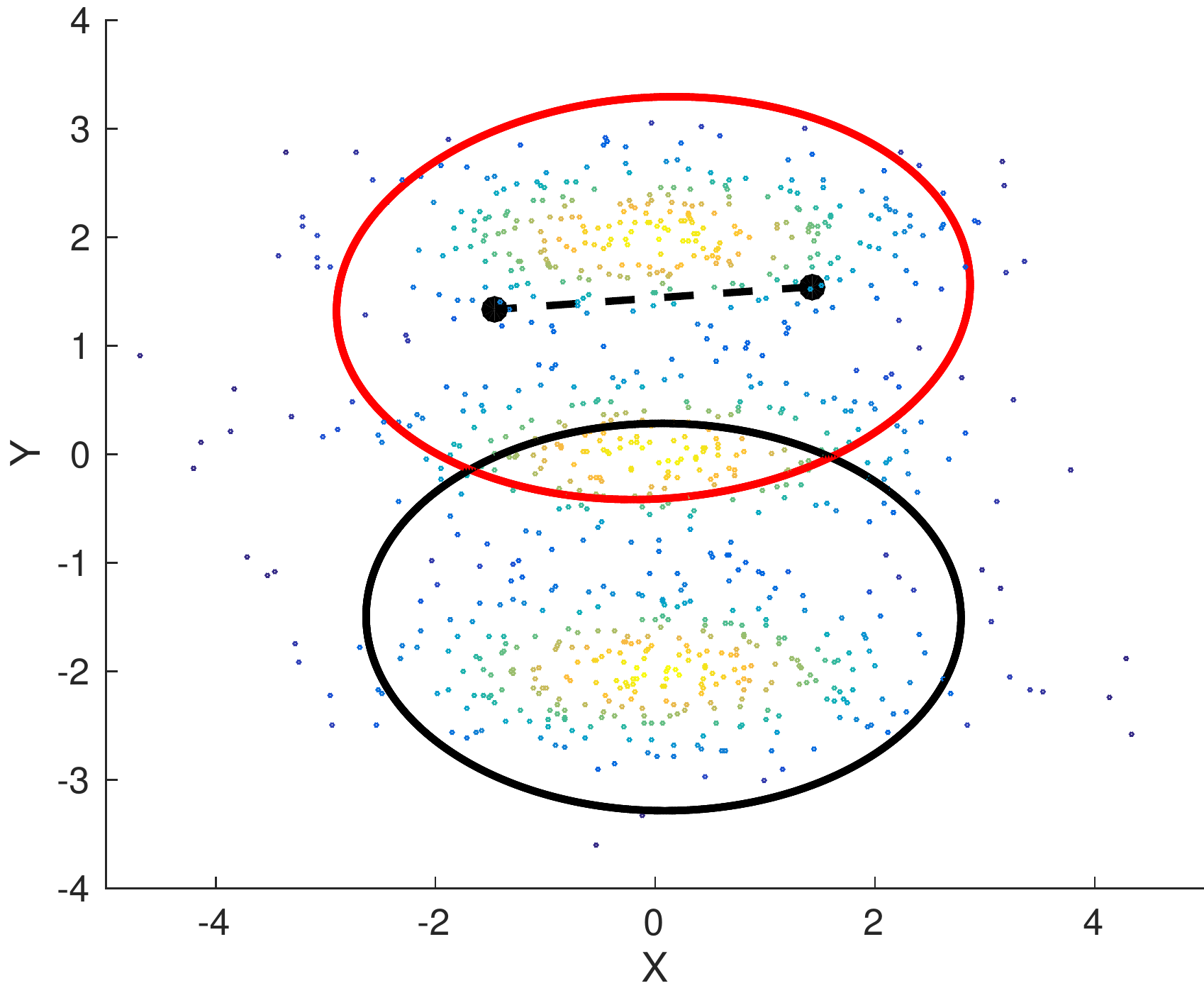}
  }
  \subfloat[]
  {
    \includegraphics[width=0.33\textwidth]{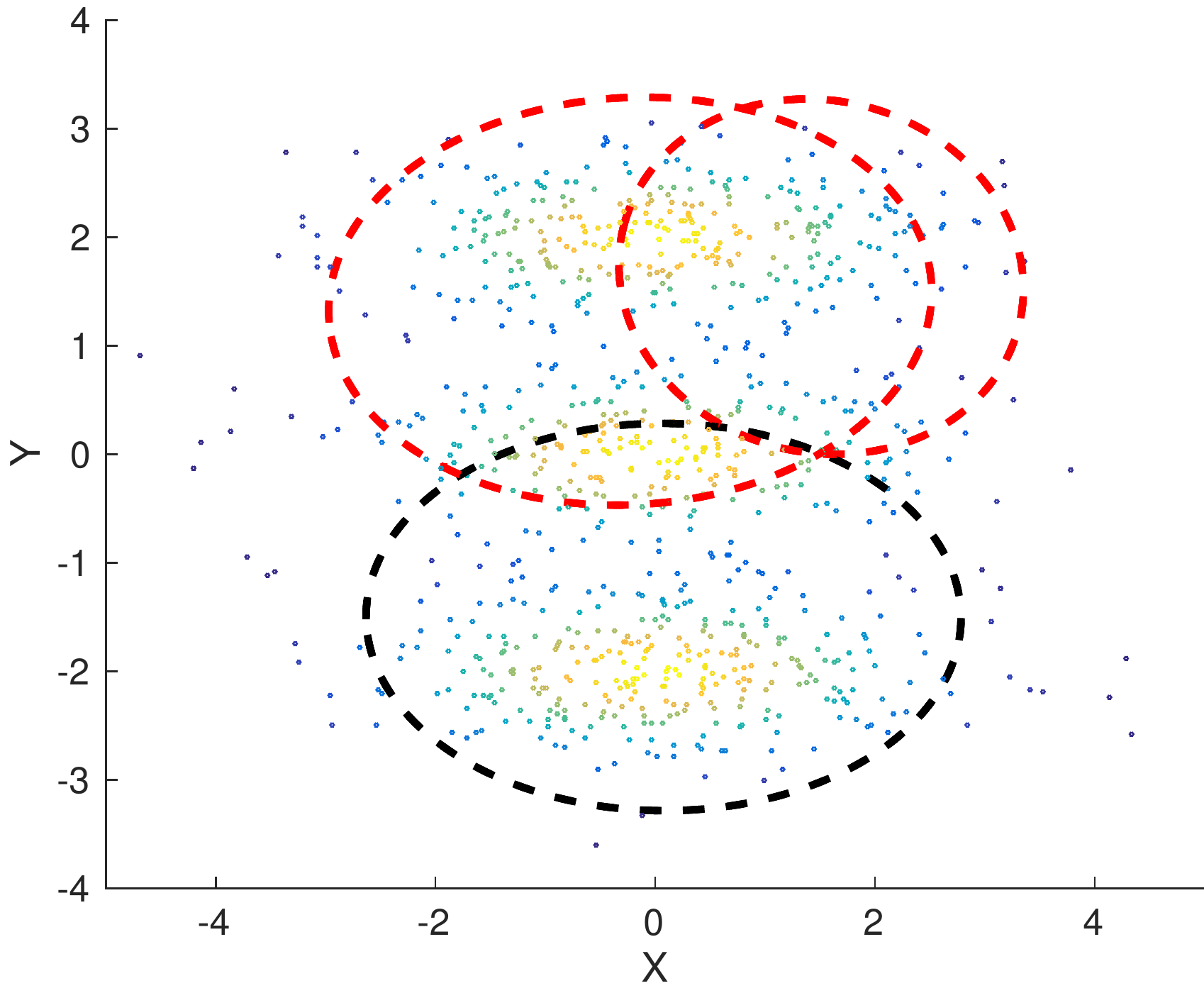}
  } 
  \subfloat[]
  {
    \includegraphics[width=0.33\textwidth]{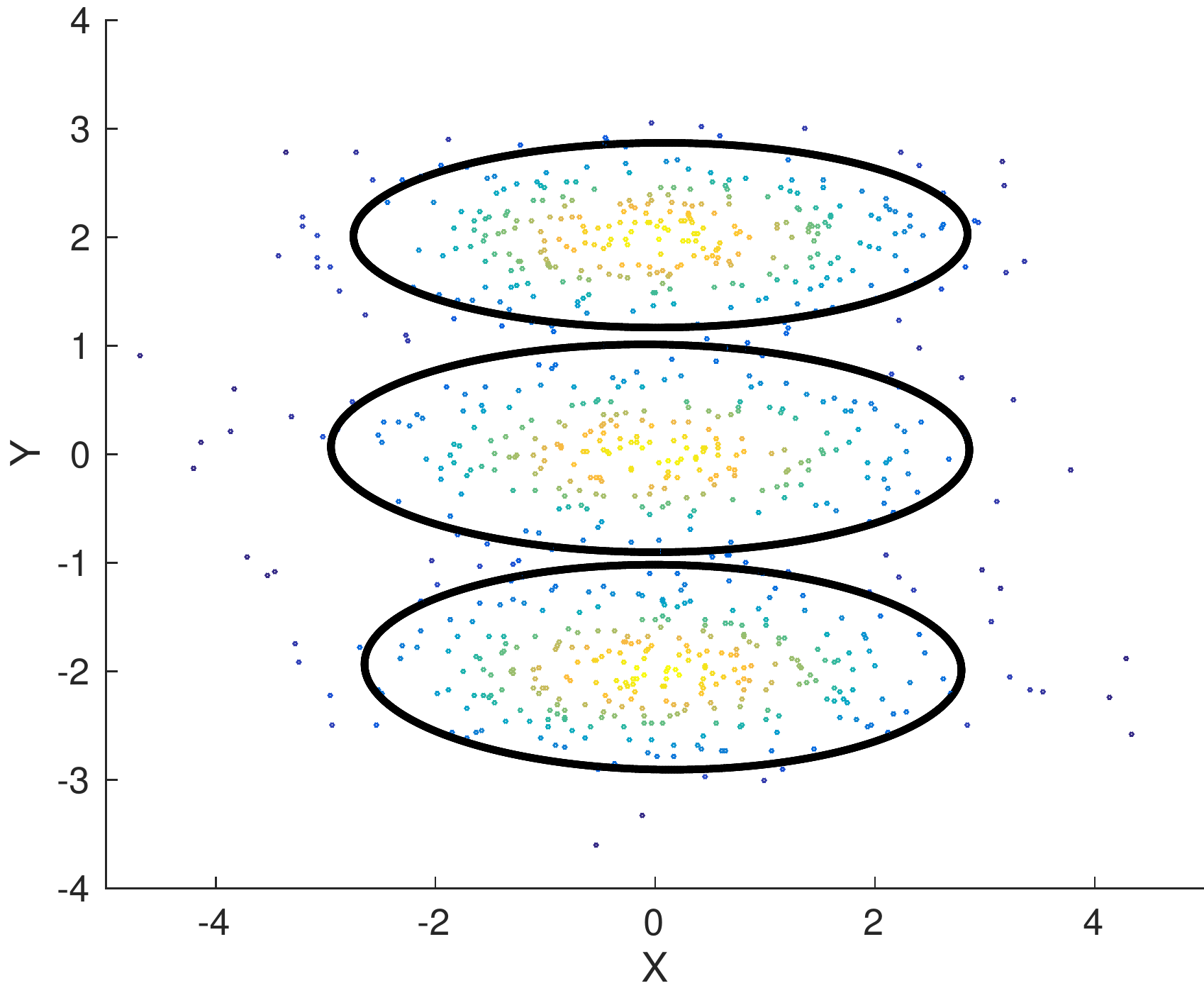}
  }
  \caption{Second iteration: \emph{splitting} the first component in $P_2$ ($I = 22673$ bits)
           (a) Initial means (shown by the black dots)
           (b) Optimized child mixture (denoted by red dashed lines)
               along with the second component of the parent (denoted by black dashes) ($I = 22691$ bits)
           (c) $P_3$: stabilized mixture post-EM phase ($I = 22460$ bits) results in a
               further improvement of message length.
          }
  \label{fig:mix1_iter_2_splits}
\end{figure}

\begin{figure}[htb]
  \subfloat[]
  {
    \includegraphics[width=0.33\textwidth]{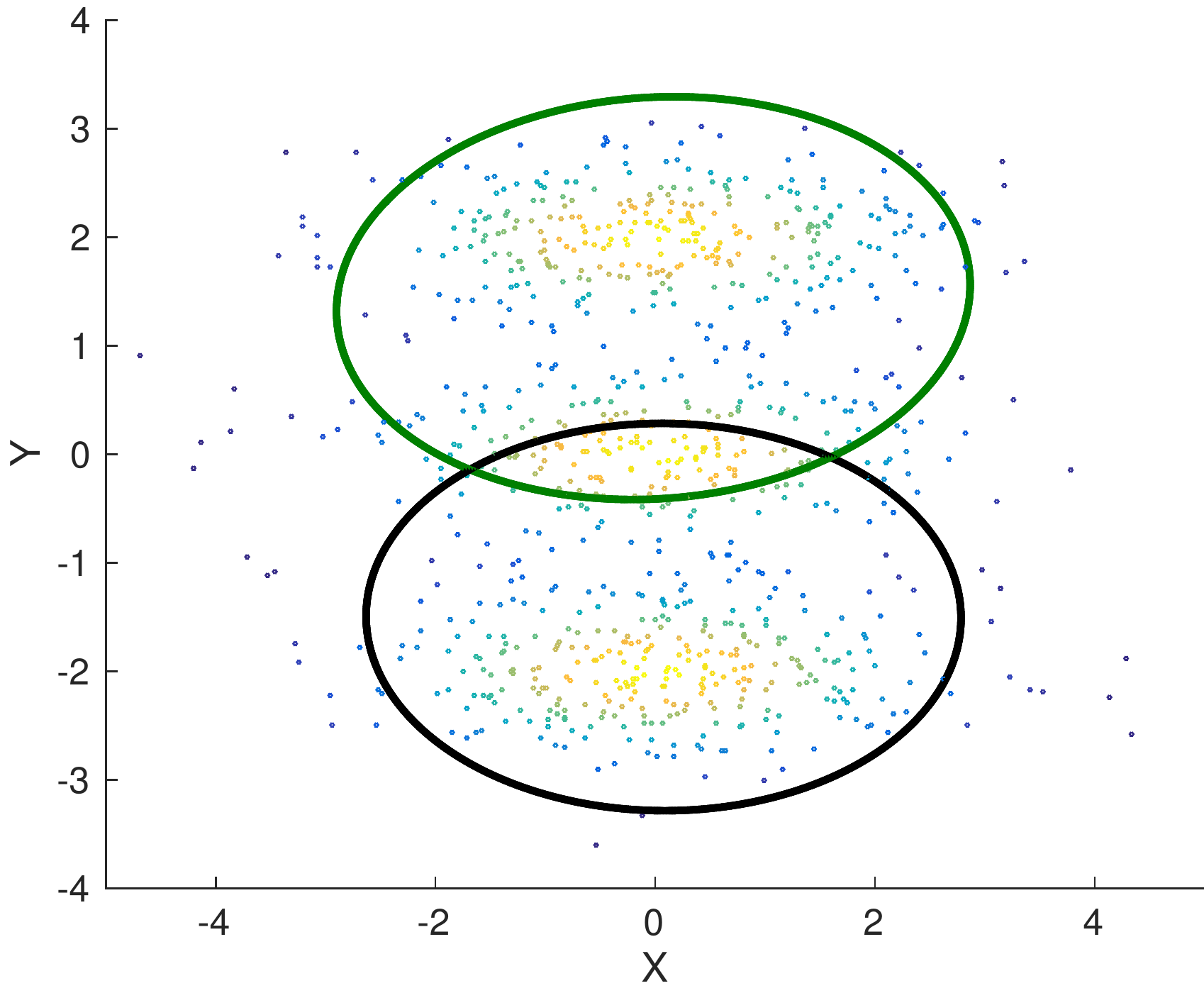}
  }
  \subfloat[]
  {
    \includegraphics[width=0.33\textwidth]{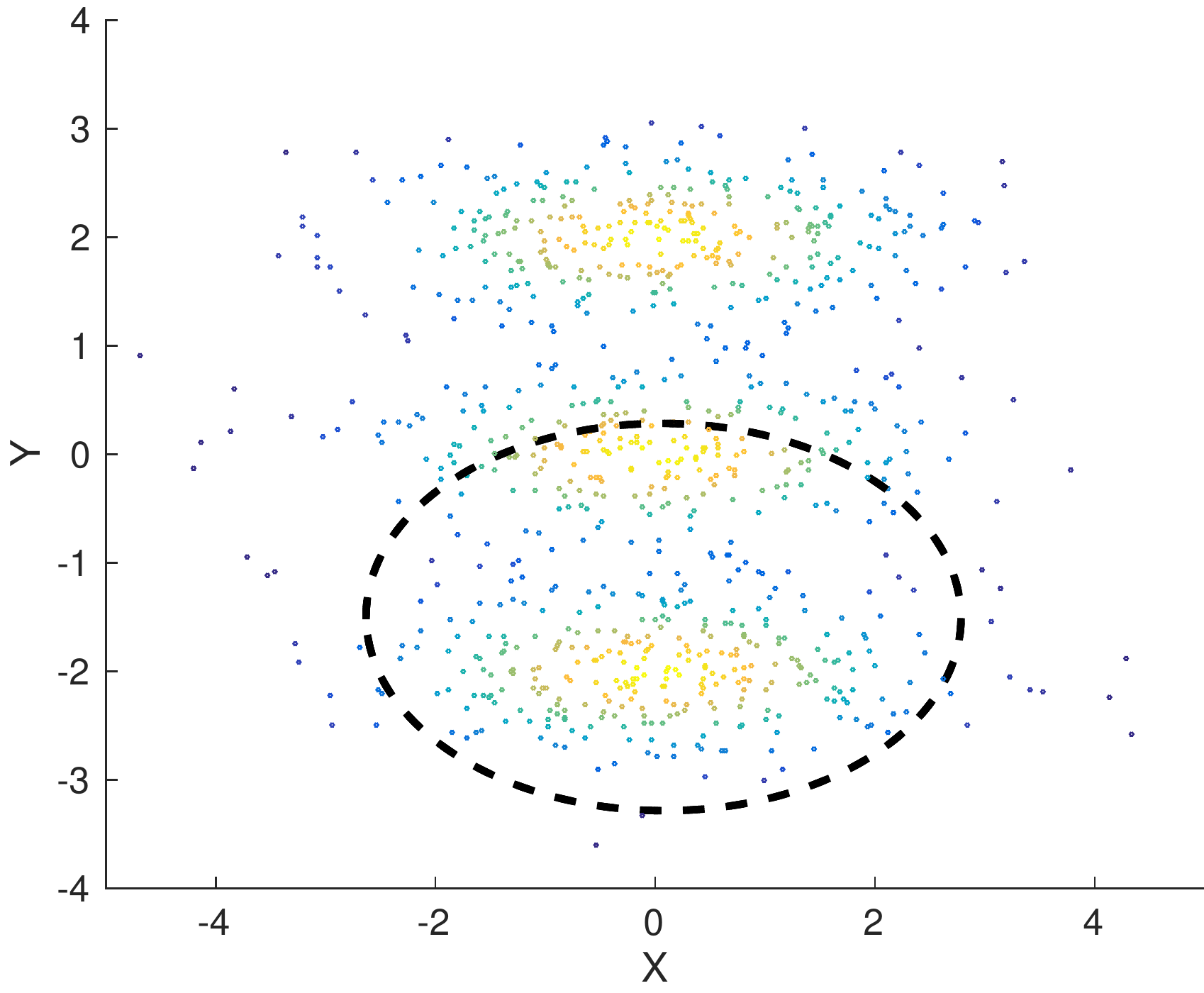}
  } 
  \subfloat[]
  {
    \includegraphics[width=0.33\textwidth]{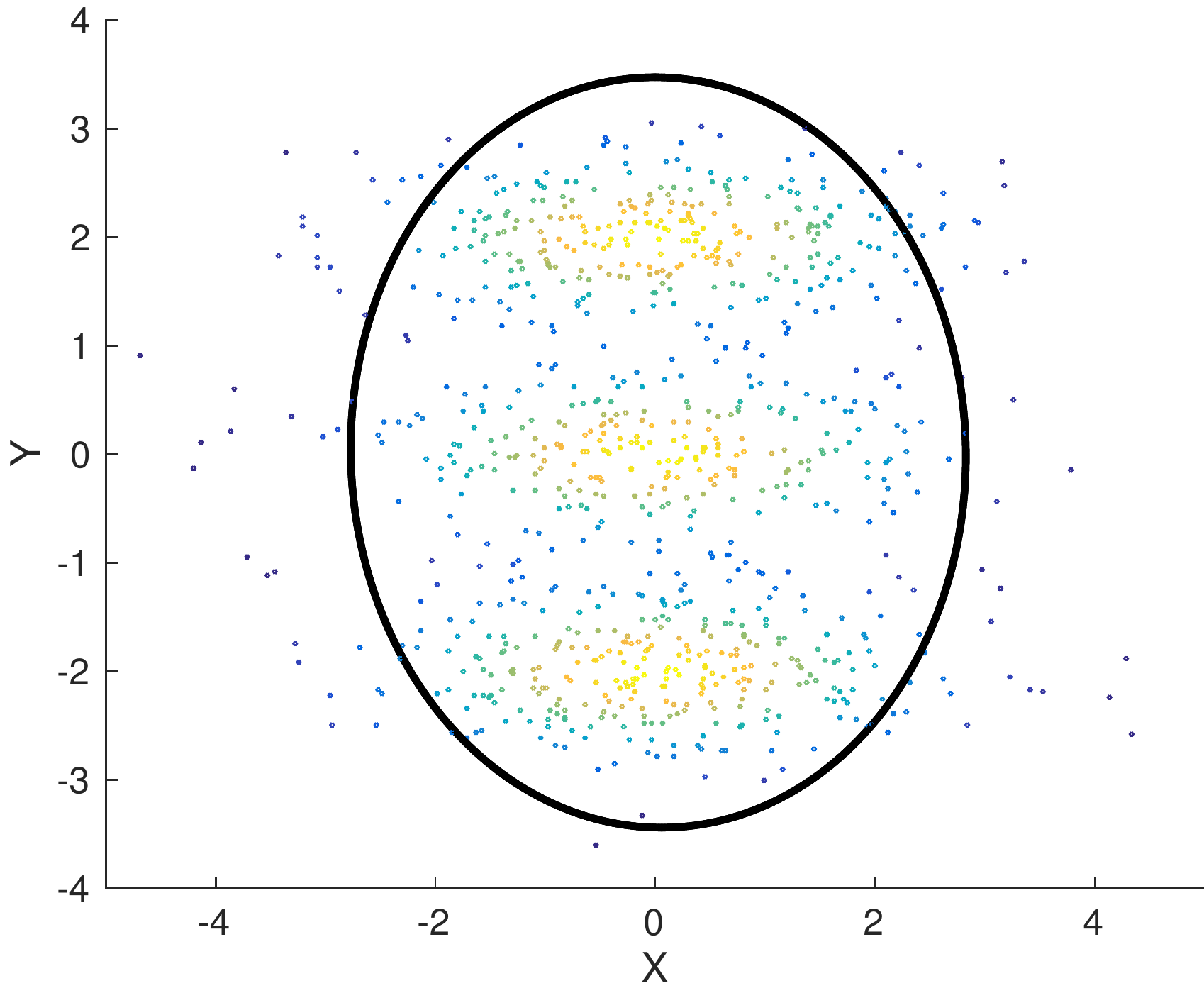}
  } 
  \caption{Second iteration: \emph{deleting} the first component in $P_2$
           (a) Green ellipse denotes the component being deleted
           (b) EM initialization with the remaining component ($I = 25599$ bits)
           (c) Resultant mixture after deletion and post EM ($I = 22793$ bits) -- no improvement.
          }
  \label{fig:mix1_iter_2_deletions}
\end{figure}

\begin{figure}[htb]
  \subfloat[]
  {
    \includegraphics[width=0.33\textwidth]{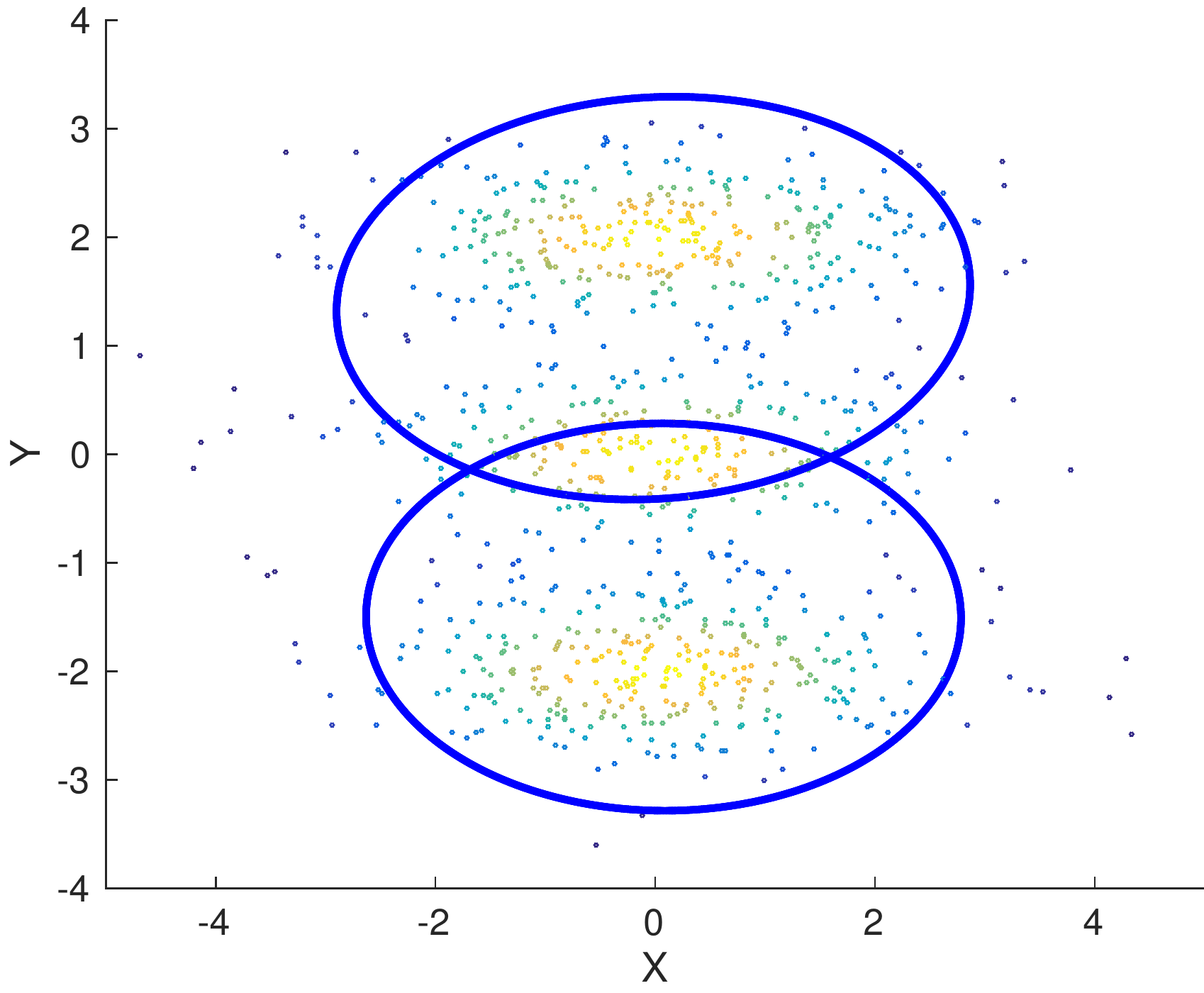}
  }
  \subfloat[]
  {
    \includegraphics[width=0.33\textwidth]{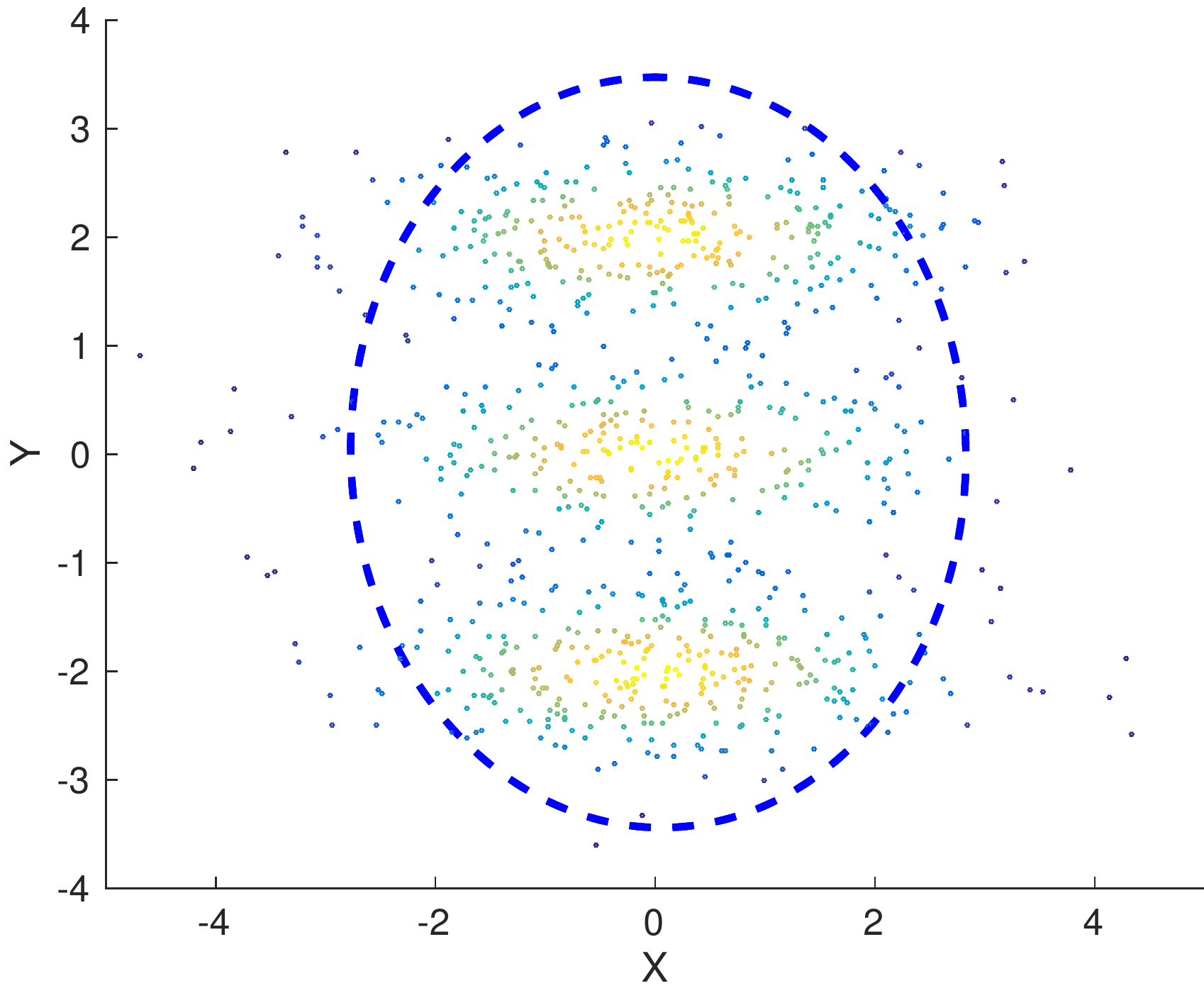}
  } 
  \subfloat[]
  {
    \includegraphics[width=0.33\textwidth]{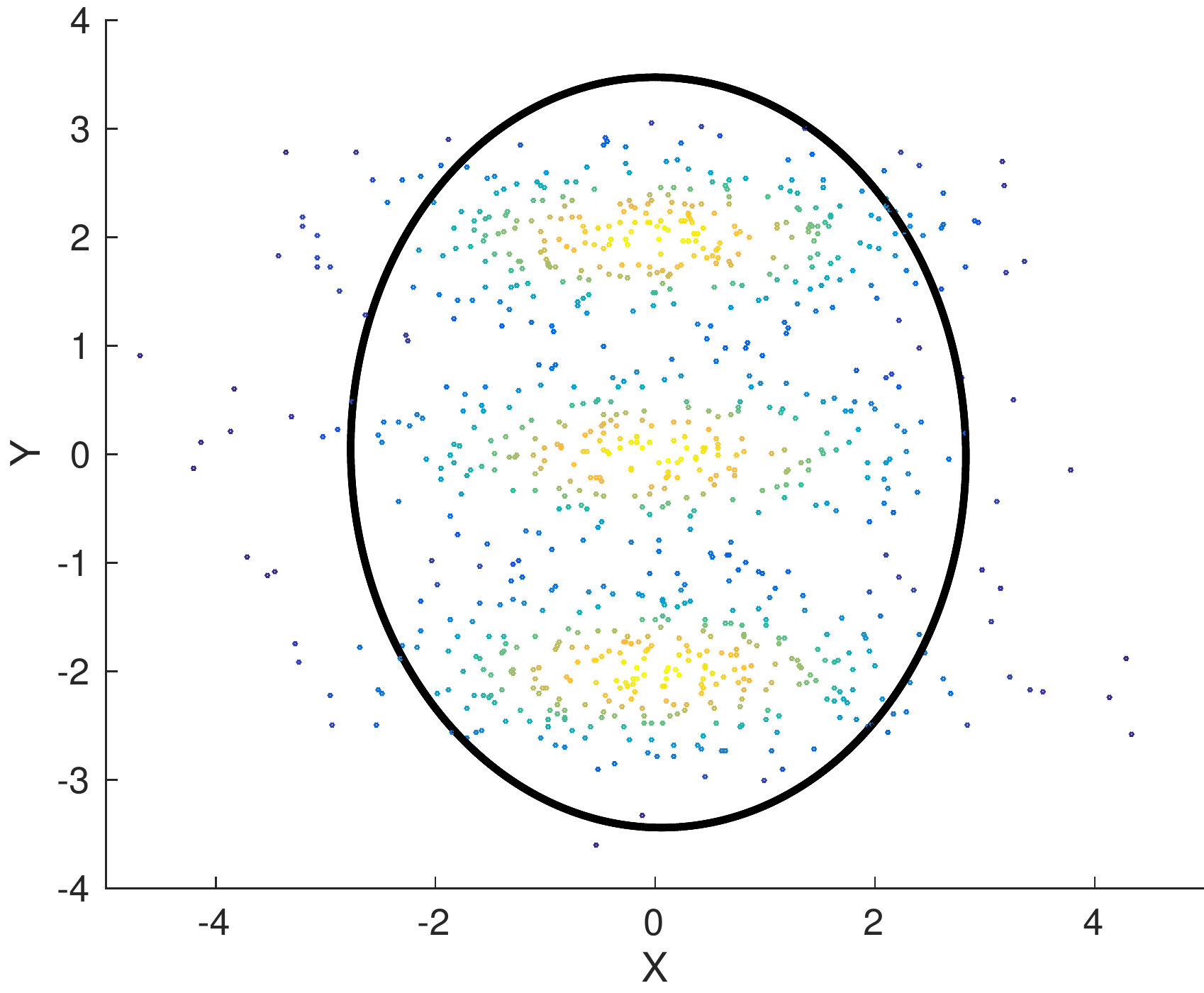}
  } 
  \caption{Second iteration: \emph{merging} the two components in $P_2$
           (a) Blue ellipses denote the components currently being merged
           (b) EM initialization with one merged component along with its parameters
           (c) Optimized mixture after merging ($I = 22793$ bits) -- no improvement
          }
  \label{fig:mix1_iter_2_merges}
\end{figure}

\begin{figure}[htb]
  \subfloat[$P_3$: parent ($I = 22460$ bits)]
  {
    \includegraphics[width=0.33\textwidth]{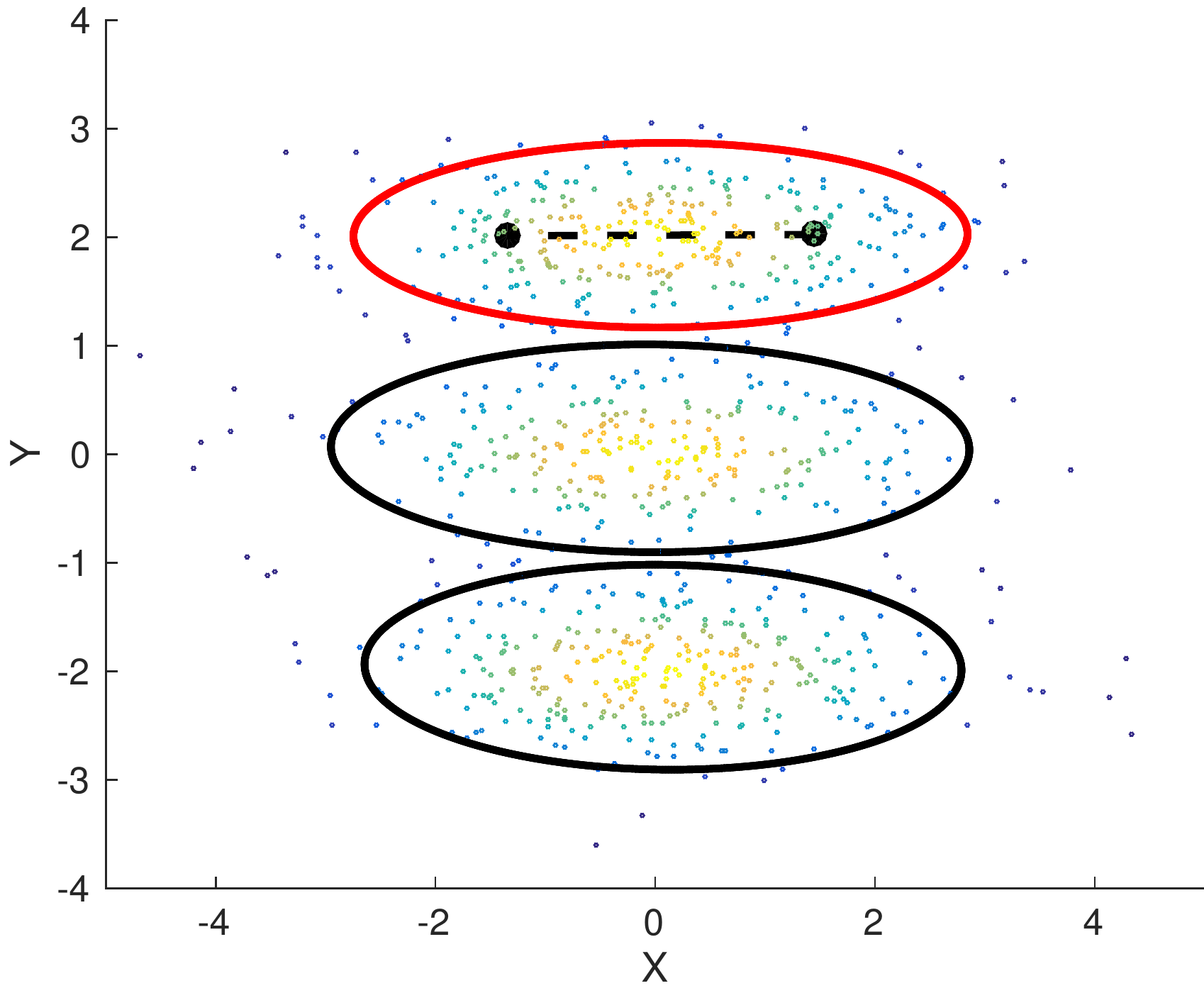}
  }
  \subfloat[Optimized children ($I = 22480$ bits)]
  {
    \includegraphics[width=0.33\textwidth]{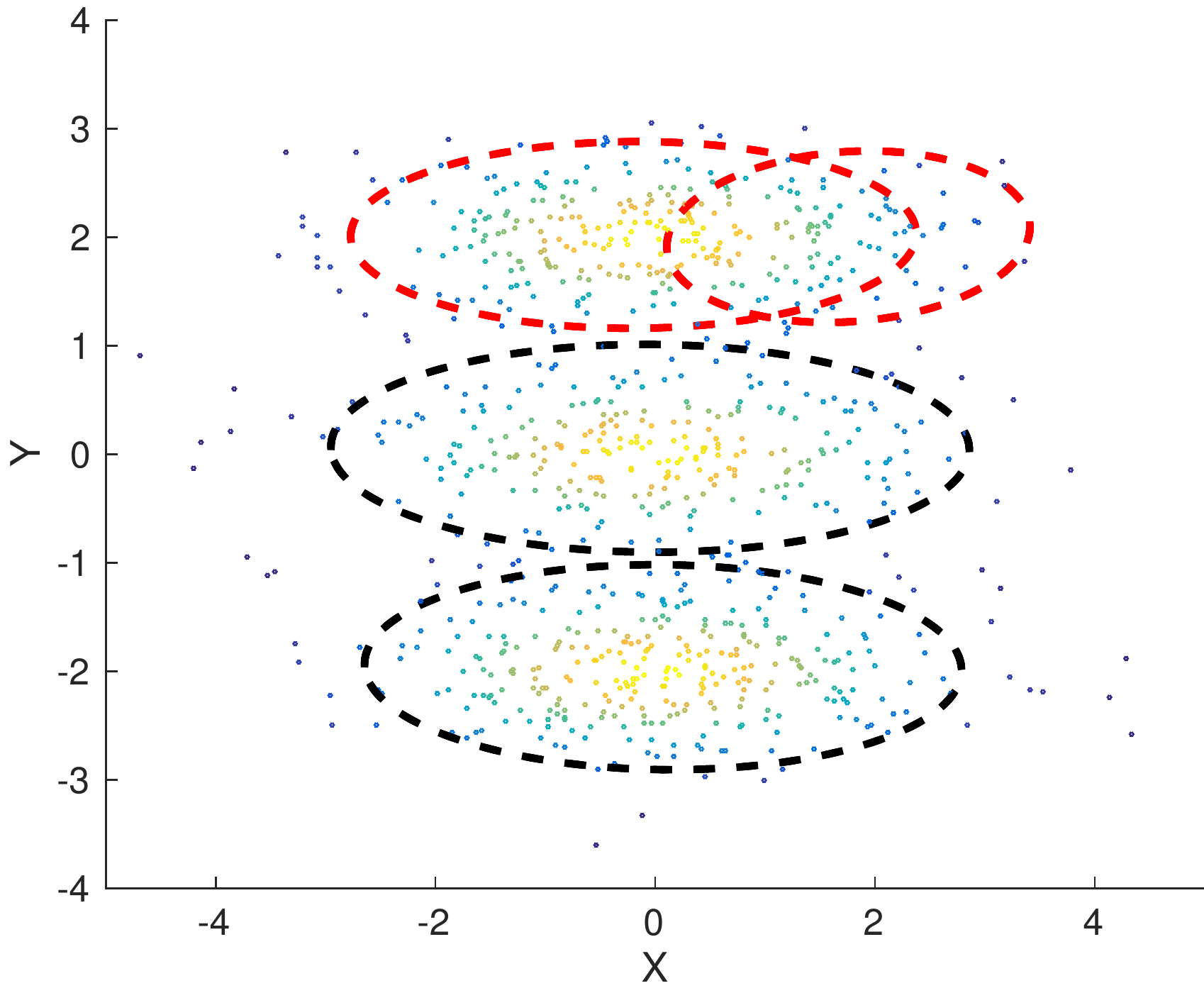}
  } 
  \subfloat[$P_4$: mixture post-EM ($I = 22474$ bits)]
  {
    \includegraphics[width=0.33\textwidth]{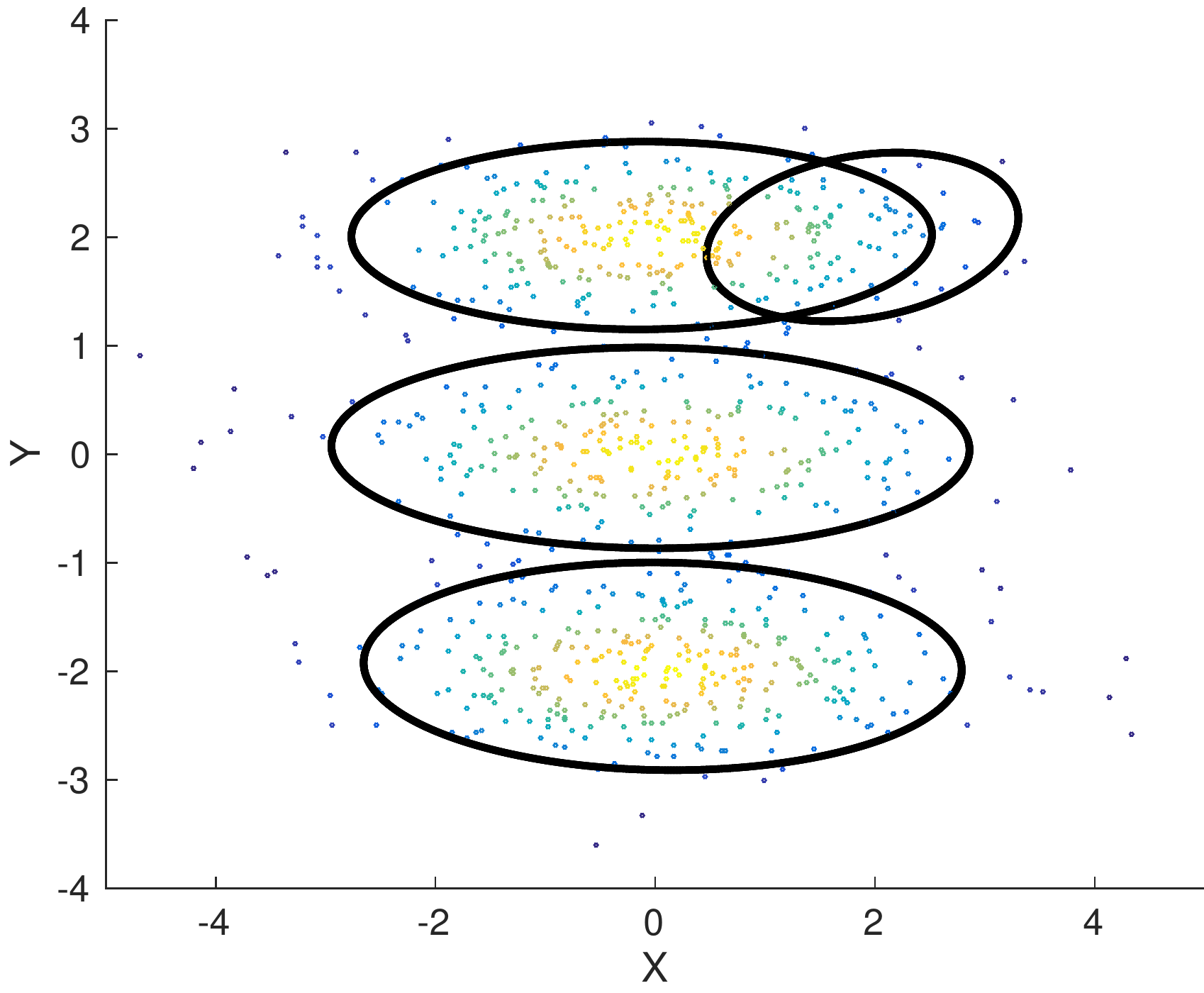}
  }\\
  \subfloat[]
  {
    \includegraphics[width=0.33\textwidth]{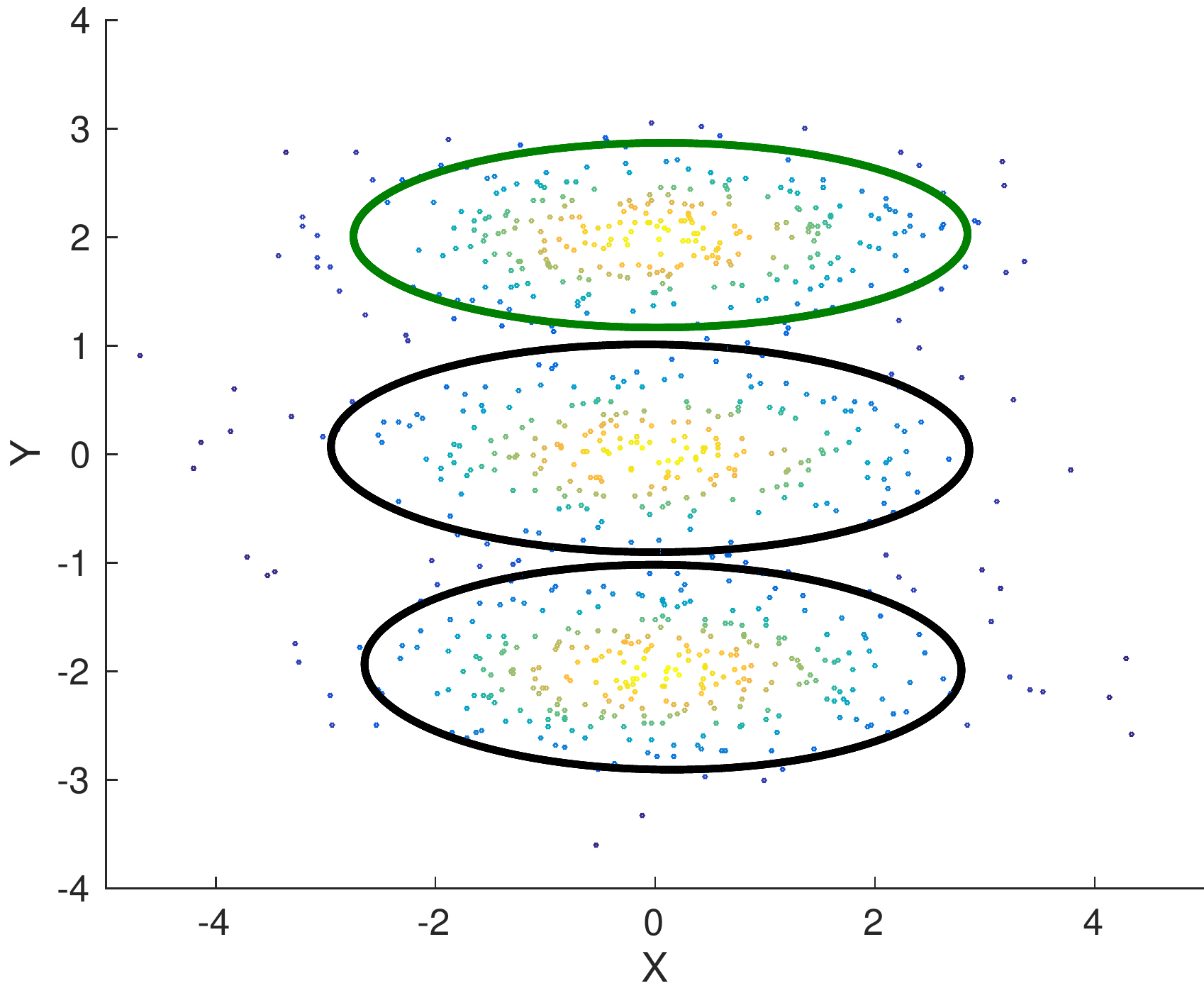}
  }
  \subfloat[Before optimizing ($I = 25767$ bits)]
  {
    \includegraphics[width=0.33\textwidth]{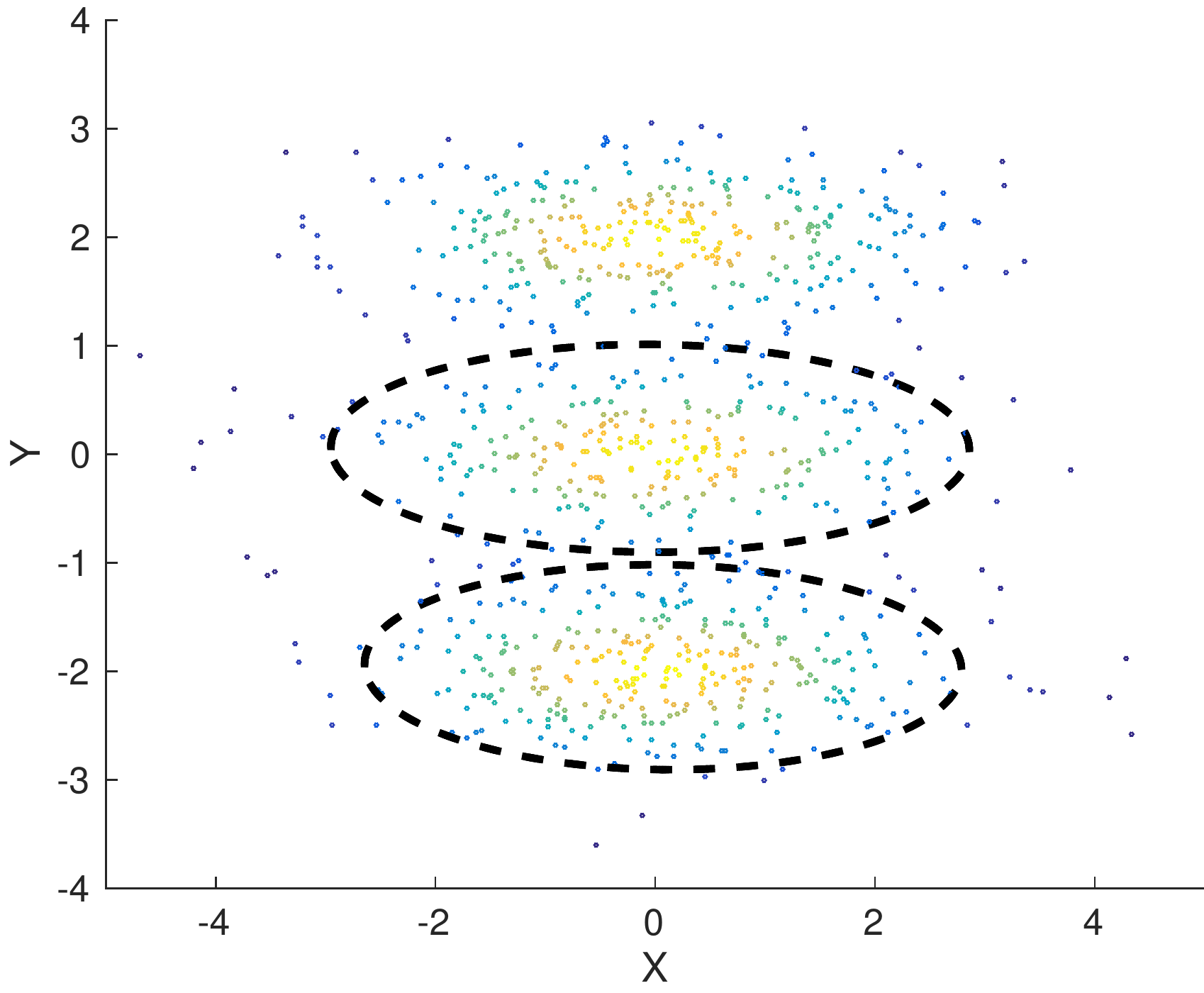}
  } 
  \subfloat[Post-EM ($I = 22610$ bits)]
  {
    \includegraphics[width=0.33\textwidth]{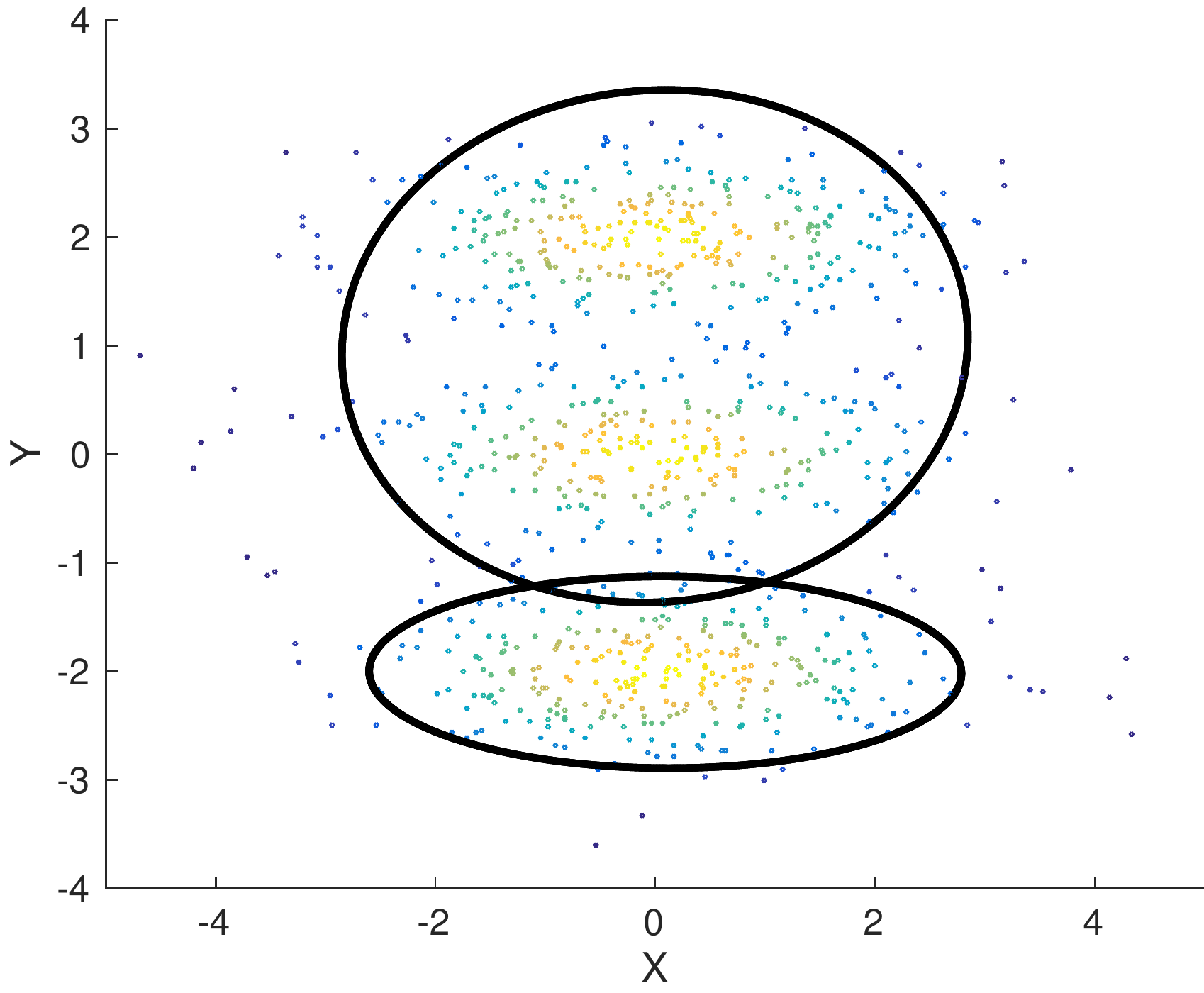}
  }\\ 
  \subfloat[]
  {
    \includegraphics[width=0.33\textwidth]{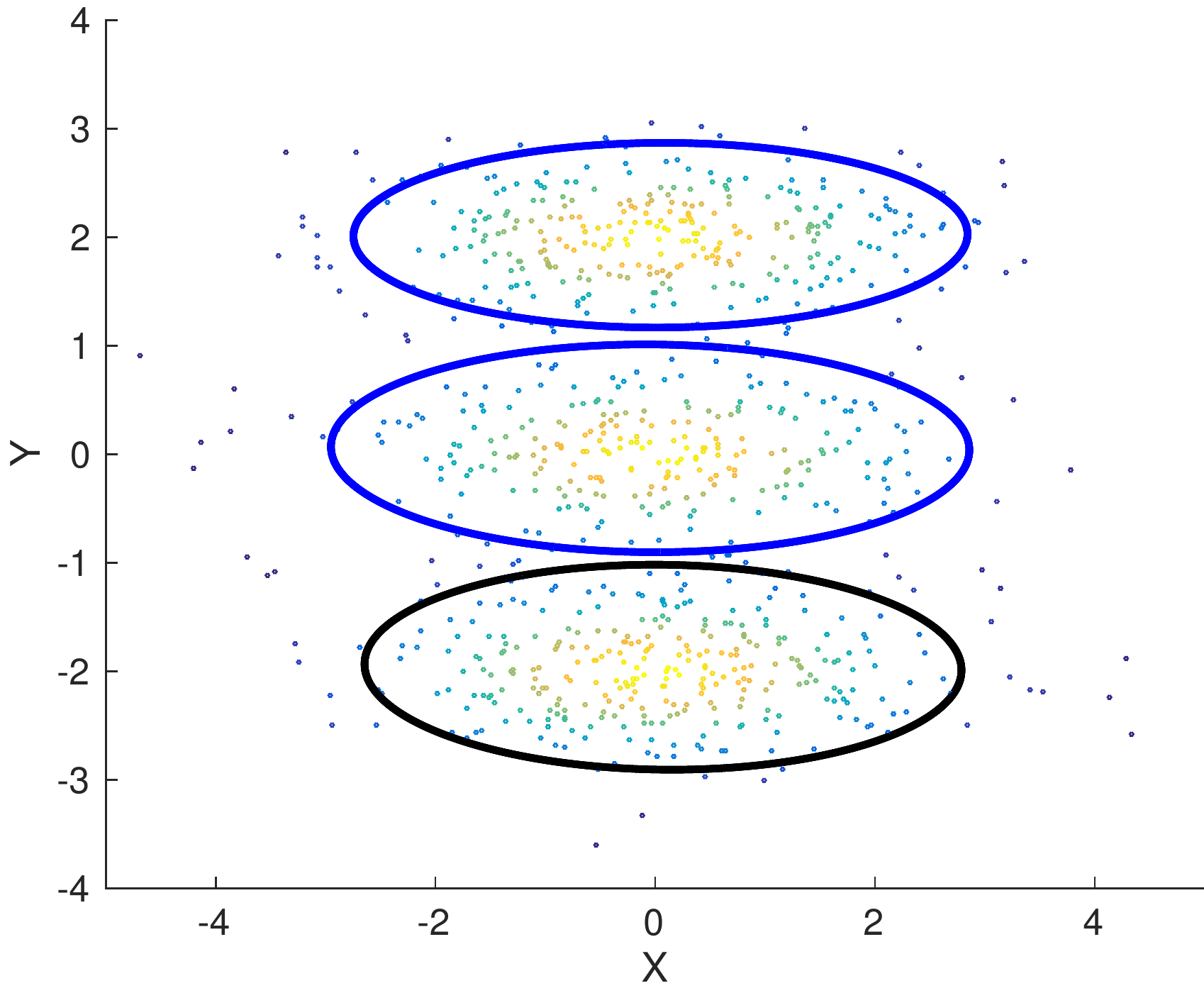}
  }
  \subfloat[Before optimizing ($I = 22617$ bits)]
  {
    \includegraphics[width=0.33\textwidth]{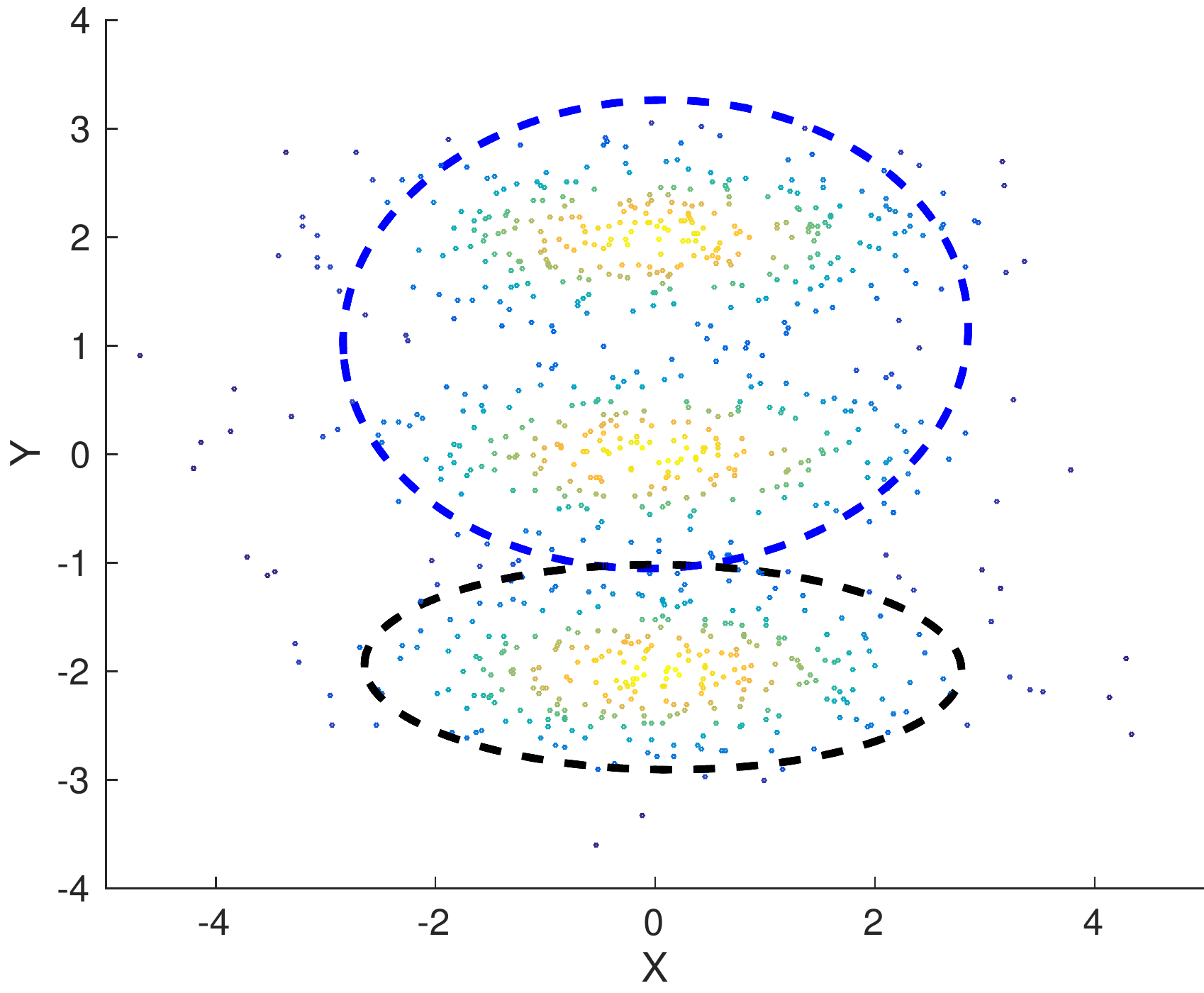}
  } 
  \subfloat[Post-EM ($I = 22610$ bits)]
  {
    \includegraphics[width=0.33\textwidth]{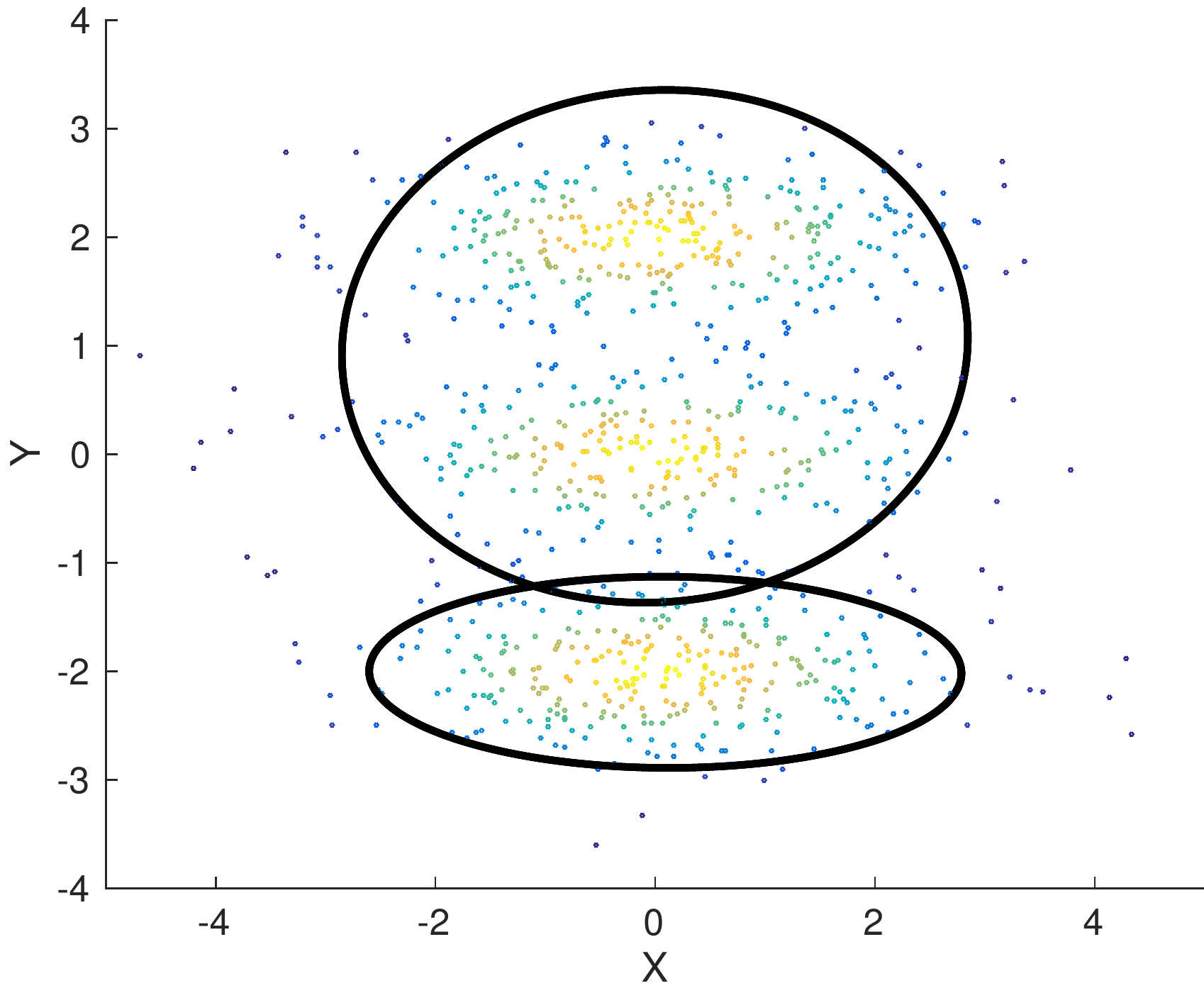}
  } 
  \caption{Third iteration: Operations involving the first component 
           (a)-(c) denote the \emph{splitting} process,
           (d)-(f) denote the \emph{deletion} process, and
           (g)-(i) shows the \emph{merging} of the first component with its closest component.
          }
  \label{fig:mix1_iter_3}
\end{figure}

In the second iteration, each component in $P_2$ is iteratively split, deleted, and merged.
Fig.~\ref{fig:mix1_iter_2_splits} shows the splitting (red) of the first component.
On splitting, the new mixture $P_3$ results in a lower message length.
Deletion of the first component is shown in Fig.~\ref{fig:mix1_iter_2_deletions}.
Before merging the first component, we identify its closest component (the one
with the least KL-divergence) (see Fig.~\ref{fig:mix1_iter_2_merges}). Deletion and merging
operations, in this case, do not result in an improvement. These two
operations have different intermediate EM initializations (Figures~\ref{fig:mix1_iter_2_deletions}(b)
and \ref{fig:mix1_iter_2_merges}(b)) but result in the same optimized one-component
mixture. The same set of operations are performed on the second component in $P_2$.
In this particular case, splitting results in an improved mixture (same as $P_3$).
$P_3$ is updated as the new parent and the series of split, delete, and merge
operations are carried out on all components in $P_3$. Fig.~\ref{fig:mix1_iter_3}
shows these operations on the first component. We see that splitting the first
component in $P_3$ results in $P_4$ (see Fig.~\ref{fig:mix1_iter_3}(c)). However, $P_4$
is not an improvement over $P_3$ as seen by the message lengths and is, therfore, discarded.
Similarly, deletion and merging of the components do not yield improvements to $P_3$.
The operations are carried out on the remaining two components in $P_3$ (not shown in 
the figure) too. These perturbations do not produce improved mixtures
in terms of the total message length.
Since the third iteration does not result in any further improvement, the search
terminates and the parent $P_3$ is considered to be the best mixture.

In different stages of the search method, we have different intermediate
mixtures. EM is a gradient
descent technique and it can get trapped in a local optimum. By employing 
the suggested search, we are exhaustively
considering the possible options, and aiming to reduce the
possibility of the EM getting stuck in a local optimum. 
The proposed method infers a mixture by balancing the tradeoff
due to model complexity and the fit to the data.
This is particularly useful when there is no 
prior knowledge pertaining to the nature of the data. In such a case, this 
method provides an objective way to infer a mixture with suitable components
that best explains the data through lossless compression.
Another example is shown in Appendix \ref{subsec:appendix_mix2},
where the evolution of 
the inferred mixture is explored in the case of a mixture with overlapping components.\\

\noindent\emph{Variation of the two-part message length:}
The search method infers three components and terminates. 
In order to demonstrate that the inferred number of components
is the optimum number, we infer mixtures with increasing
number of components (until $M=15$ as an example) and plot their
resultant message lengths. For each $M > 3$, the standard EM
algorithm (Section~\ref{subsec:em_mml}) is employed to infer the mixture parameters.

\begin{wrapfigure}{r}{0.5\textwidth}
\centering
\includegraphics[width=\textwidth]{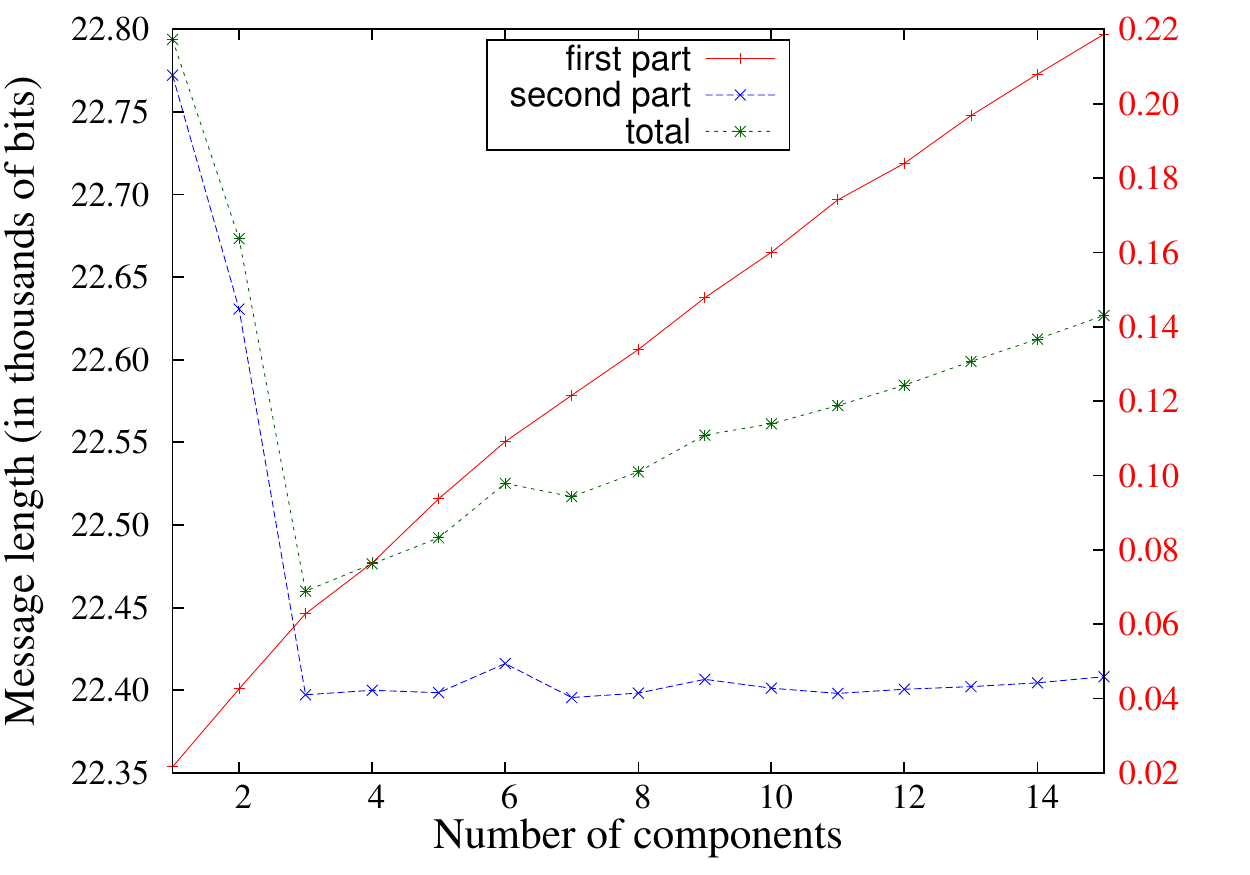}
\caption{Variation of the individual parts of the total message length 
         with increasing number of components (note the two Y-axes have 
         different scales -- the first part follows the right side Y-axis;
         the second part and total message lengths
         follow the left side Y-axis)
        }
\label{fig:individual_msglens_gaussian}
\end{wrapfigure}
Fig.~\ref{fig:individual_msglens_gaussian}
shows the total message lengths 
to which the EM algorithm converges for varying number of
components $M$. As expected, the total message length (green curve) drastically
decreases initially until $M=3$ components are inferred.
Starting from $M=4$, the total message length gradually increases, 
clearly suggesting that the inferred
models are over-fitting the data with increasing statement cost to
encode the additional parameters of these (more complex) models. 
We further elaborate on the reason for the initial decrease and subsequent increase
in the total message length. As per MML evaluation criterion, the message length comprises
of two parts -- statement cost for the parameters and the cost for stating the data
using those parameters.  
The model complexity (which corresponds to the mixture parameters) increases with
increasing $M$. Therefore, the first part of the message to encode parameters
increases with an increase in the number of parameters.
This behaviour is illustrated by the red curve in
Fig.~\ref{fig:individual_msglens_gaussian}. 
The first part message lengths are shown in red on the right side Y-axis in the figure.
As the mixture model becomes increasingly more complex, the error of fitting 
the data decreases. This corresponds to the second part of the message 
in the MML encoding framework. This behaviour is consistent with what is 
observed in Fig.~\ref{fig:individual_msglens_gaussian} (blue curve). 
There is a sharp fall until $M=3$;
then onwards increasing the model complexity does not lower the error significantly.
The error saturates and there is minimal gain with regards to encoding the data
(the case of overfitting).
However, the model complexity dominates after $M=3$. 
The optimal balance is achieved when $M=3$. 
In summary, the message length at $M=3$ components was rightly observed to be the 
optimum for this example. 
We note that for a fixed number of mixture components, the EM algorithm for the MML metric is
monotonically decreasing. However, while searching for the number of components, MML 
continues to decrease until some optimum is found and then steadily increases
as illustrated through this example.

\section{Experiments with Gaussian mixtures} 
\label{sec:gaussian_experiments}
We compare our proposed inference methodology against the widely cited
method of \cite{figueiredo2002unsupervised}. 
The performance of their method is compared against that of 
Bayesian Information Criterion (BIC), Integrated Complete Likelihood (ICL), 
and approximate Bayesian (LEC) methods
(discussed in Section \ref{sec:mixture_existing_methods}).
It was shown that the method of \cite{figueiredo2002unsupervised} was
far superior than BIC, ICL and LEC (using Gaussian mixtures). 
In the following sections, we demonstrate through a series of experiments
that our proposed approach to infer mixtures fares better when
compared to that of \cite{figueiredo2002unsupervised}.
The experimental setup is as follows: we use a Gaussian mixture $\fancym^t$ (true distribution),
generate a random sample from it, and infer the mixture using the data.
This is repeated 50 times and we compare the performance of our method
against that of \cite{figueiredo2002unsupervised}. 
As part of our analysis, we compare the number of inferred mixture components
as well as the quality of mixtures.

\subsection{Methodolgies used to compare the mixtures inferred by our proposed
approach and FJ's method}
\label{subsec:comparison_mixtures}
\noindent\emph{Comparing message lengths:}
The MML framework allows us to objectively compare mixture models by
computing the total message length used to encode the data.
The difference in message lengths gives the log-odds posterior ratio
of any two mixtures
(Equation \eqref{eqn:compare_models}). Given some observed data,
and any two mixtures, one can determine which of the two
best explains the data.
Our search methodology uses the scoring function ($I_{MML}$) defined in Equation \eqref{eqn:mixture_msglen}.
As elaborated in Section~\ref{subsec:fj}, \cite{figueiredo2002unsupervised} use an
approximated MML-like scoring function ($I_{FJ}$) given by Equation \eqref{eqn:fj}. 

We employ our search method and the method of \cite{figueiredo2002unsupervised} 
to infer the mixtures using the same data;
let the inferred mixtures be $\fancym^*$ and $\fancym^{FJ}$ respectively.
We compute two quantities:
\begin{align}
\Delta I_{MML} &= I_{MML}(\fancym^{FJ}) - I_{MML}(\fancym^*) \notag\\
\text{and}\quad\Delta I_{FJ} &= I_{FJ}(\fancym^{FJ}) - I_{FJ}(\fancym^*) \label{eqn:comparisons_mixtures}
\end{align}
We use the two different scoring functions to compute
the differences in message lengths of the resulting mixtures $\fancym^{FJ}$ and $\fancym^*$.
Since the search method used to obtain $\fancym^*$ optimizes the scoring function $I_{MML}$,
it is expected that $I_{MML}(\fancym^*) < I_{MML}(\fancym^{FJ})$ and consequently $\Delta I_{MML} > 0$.
This implies that our method is performing better using our defined objective function.
However, if $I_{FJ}(\fancym^*) < I_{FJ}(\fancym^{FJ})$, this indicates that our inferred
mixture $\fancym^*$ results in a lower value of the scoring function that is defined by 
\cite{figueiredo2002unsupervised}. Such an evaluation not only demonstrates
the superior performance of our search (leading to $\fancym^*$) using 
our defined scoring function
but also proves it is better using the scoring function as 
defined by \cite{figueiredo2002unsupervised}.

\noindent\emph{Kullback Leibler (KL) divergence:}
In addition to using message length based evaluation criterion, we also compare the 
mixtures using KL-divergence \citep{kullback1951information}. The metric gives a measure
of the similarity between two distributions (the lower the value, the more similar the distributions). 
For a mixture probability distribution,
there is no analytical form to compute the metric. However, one can
calculate its empirical value (which asymptotically converges to the KL-divergence). 
In experiments relating to mixture simulations, we know the true mixture $\fancym^t$
from which the data $\{\mathbf{x}_i\}, 1 \le i \le N$ is being sampled. The KL-divergence is given by the following
expression: 
\begin{align} 
D_{KL}(\fancym^t\,||\,\fancym) = E_{\fancym^t}\left[\log\frac{\Pr(\mathbf{x},\fancym^t)}{\Pr(\mathbf{x},\fancym)} \right]
             \approx \frac{1}{N} \sum_{i=1}^N\log\frac{\Pr(\mathbf{x}_i,\fancym^t)}{\Pr(\mathbf{x}_i,\fancym)}
\end{align}
where $\fancym$ is a mixture distribution ($\fancym^*$ or $\fancym^{FJ}$) whose 
\emph{closeness} to the true mixture $\fancym^t$ is to be determined.

\subsection{Bivariate mixture simulation}\label{subsec:bivariate_mix_simulation}

An experiment conducted by \cite{figueiredo2002unsupervised} was to 
randomly generate
$N=800$ data points from a two-component (with equal mixing proportions) bivariate mixture 
$\fancym^t$ whose means are at 
$\boldsymbol{\mu}_1 = (0,0)^T$ and $\boldsymbol{\mu}_2 = (\delta,0)^T$,
and equal covariance matrices: $\mathbf{C}_1 = \mathbf{C}_2 = \mathbf{I}$ (the identity matrix),
and compare the number of inferred components. 
We repeat the same experiment here and compare with the results of \cite{figueiredo2002unsupervised}.
The separation $\delta$ between the means
is gradually increased and the percentage of the correct selections
(over 50 simulations) as determined by the two search methods is plotted.
\begin{figure}[htb]
  \centering
  \subfloat[]
  {
    \includegraphics[width=0.5\textwidth]{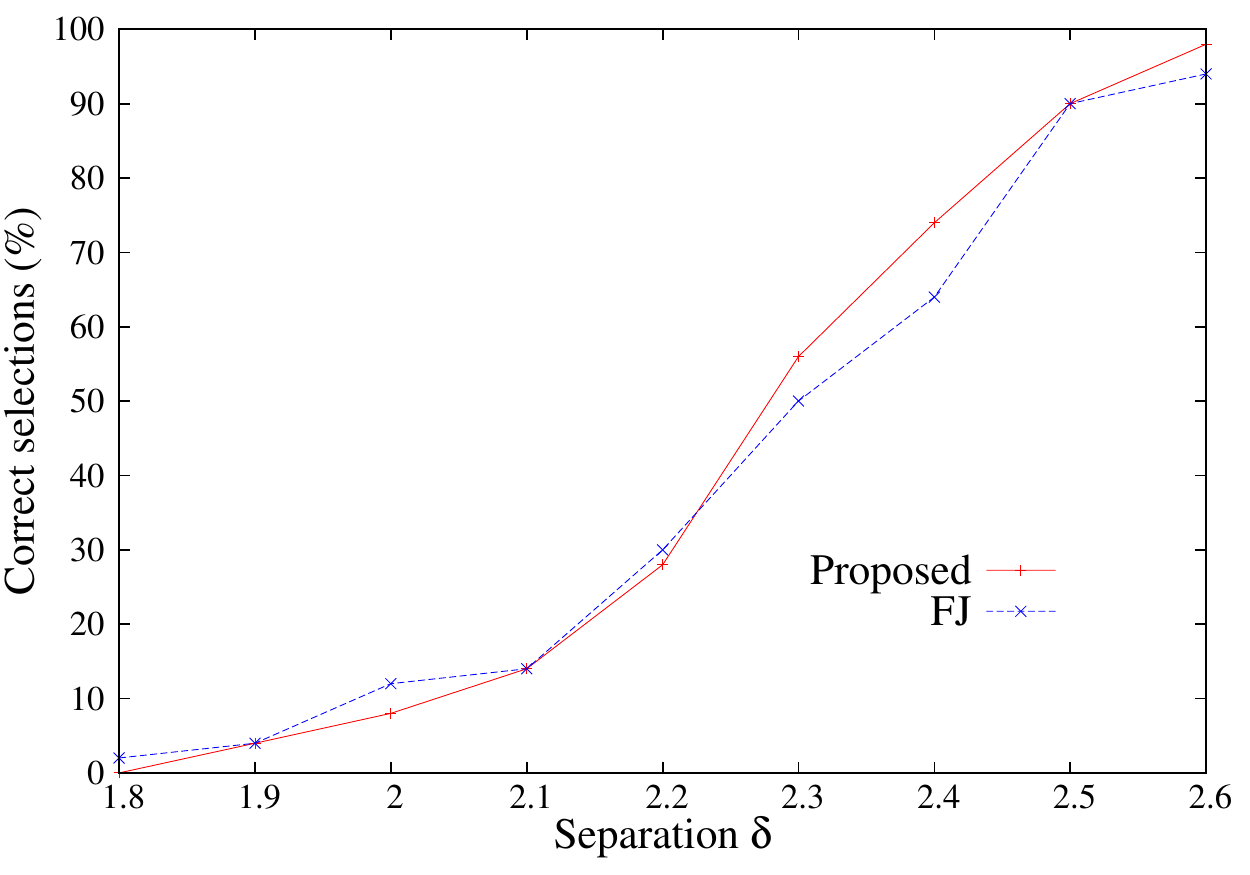}
  }
  \subfloat[]
  {
    \includegraphics[width=0.5\textwidth]{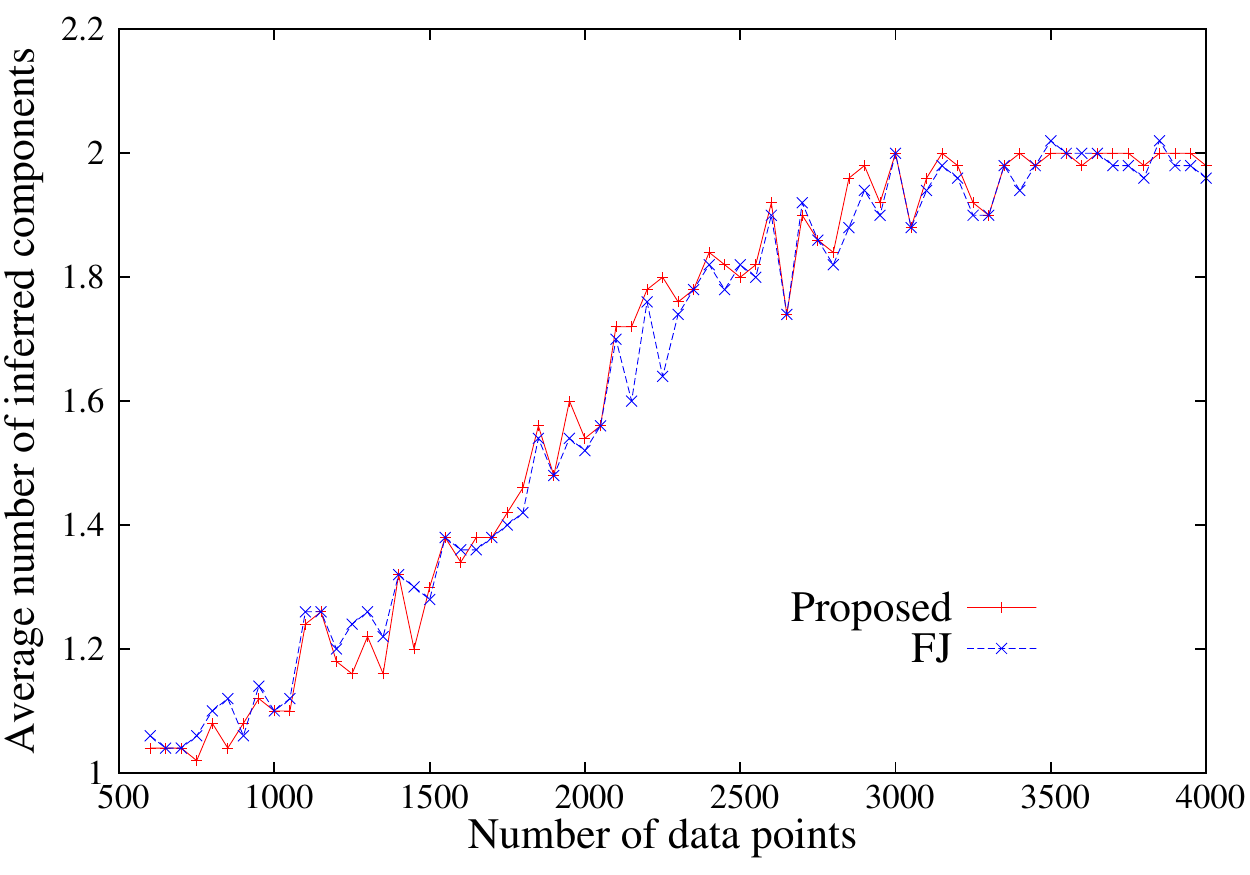}
  } 
  \caption{2-dimensional mixture (a) Percentage of correct selections with varying separation
            for a fixed sample size of $N=800$
           (b) Average number of inferred mixture components with different sample sizes
              and a fixed separation of $\delta = 2.0$ between component means.}
  \label{fig:gaussian_2d_mixture_inference}
\end{figure}
Fig.~\ref{fig:gaussian_2d_mixture_inference}(a) shows the results of this experiment.
As the separation between the component means is increased, the number
of correctly inferred components increases.
We conducted another experiment where we fix the separation between the two
components and increase the amount of data being sampled from the mixture.
Fig.~\ref{fig:gaussian_2d_mixture_inference}(b) illustrates the results 
for a separation of $\delta=2.0$.
As expected, increasing the sample size results in an increase in the number
of correct selections. Both the search methods eventually infer the 
true number of components at sample size $> 3500$. We note that in both these
cases, the differences in message lengths $\Delta I_{MML}$ and $\Delta I_{FJ}$
are close to zero. The KL-divergences for the mixtures inferred by the two search
methods are also the same. Therefore, for this experimental setup, the performance of 
both the methods is roughly similar.

\begin{figure}
  \CenterFloatBoxes
  \begin{floatrow}
  \ffigbox
  {
    \includegraphics[width=0.45\textwidth]{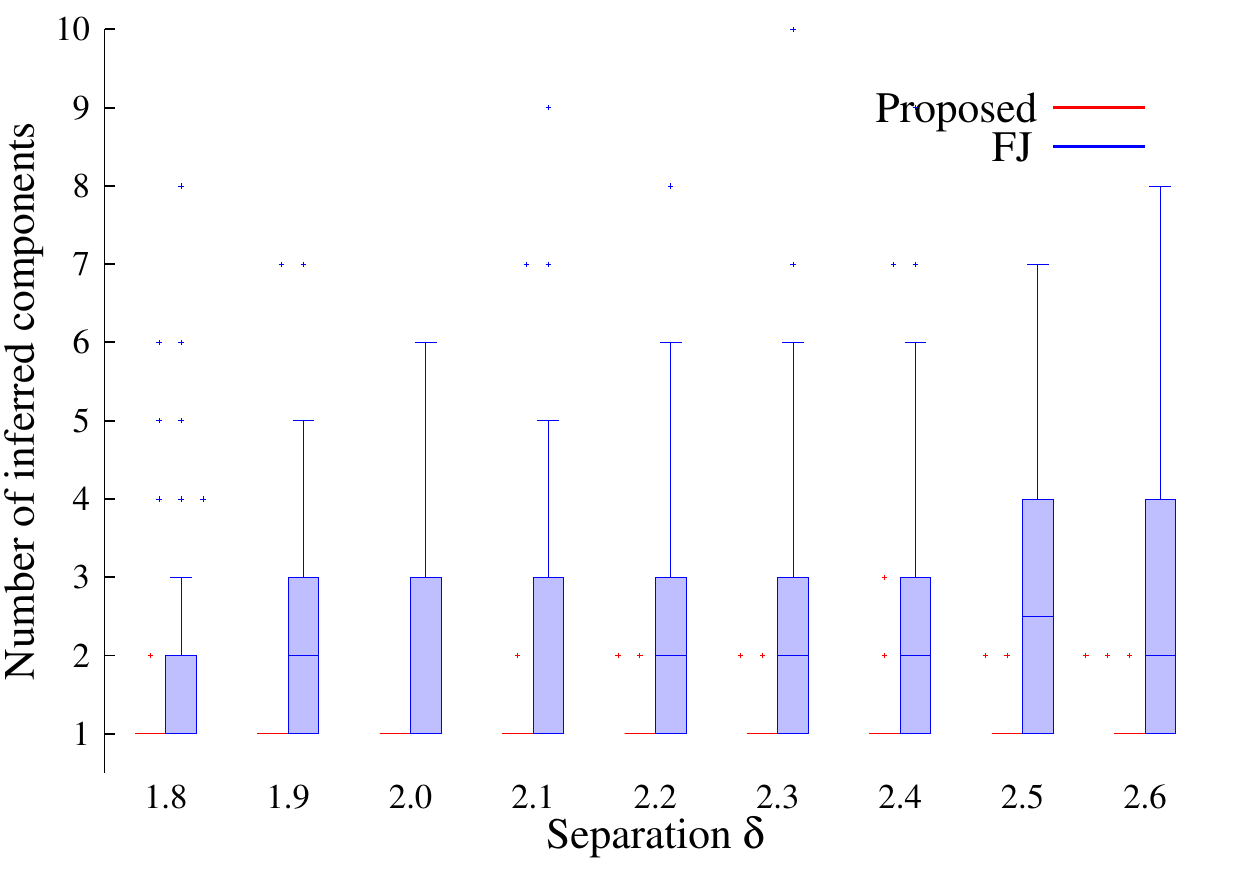}
  }
  {
    \caption{Box-whisker plot showing the variability in the number of inferred components
             ($N=100$ and over 50 simulations).}
    \label{fig:gaussian_2d_mixture_inference_boxplot}
  }
  \killfloatstyle
  \ttabbox
  {
    \begin{tabular}{|c|c|c|c|c|}
      \hline
      Separation & \multicolumn{2}{c|}{Proposed} & \multicolumn{2}{c|}{FJ} \\ \cline{2-5}
       $\delta$  & Mean & Variance & Mean & Variance \\ \hline
        1.8  & 1.02 & 0.020 &  1.98 & 2.673 \\
        1.9  & 1.00 & 0.000 &  2.26 & 2.482 \\
        2.0  & 1.00 & 0.000 &  2.04 & 2.325 \\ 
        2.1  & 1.02 & 0.020 &  2.20 & 3.510 \\
        2.2  & 1.04 & 0.039 &  2.20 & 2.285 \\
        2.3  & 1.06 & 0.057 &  2.44 & 3.639 \\
        2.4  & 1.06 & 0.098 &  2.54 & 3.967 \\
        2.5  & 1.04 & 0.039 &  2.98 & 3.203 \\
        2.6  & 1.10 & 0.092 &  2.42 & 2.942 \\
      \hline
    \end{tabular}
  }
  {
    \caption{The mean and variance of the number of inferred components 
             for each $\delta$ value ($N=100$ and over 50 simulations).}
    \label{tab:gaussian_2d_mixture_inference_boxplot}
  }
  \end{floatrow}
\end{figure}

As the difference between the two search methods is not apparent from these experiments,
we wanted to investigate the behaviour of the methods with smaller sample sizes.
We repeated the experiment similar to that shown in
Fig.~\ref{fig:gaussian_2d_mixture_inference}(a) but with a sample size of $N=100$.
Our search method results in a mean value close to 1 for different
values of $\delta$ (see Table~\ref{tab:gaussian_2d_mixture_inference_boxplot}).
The mean value of the number of inferred components using the search
method of \cite{figueiredo2002unsupervised} fluctuates between 2 and 3.
However, there is significant variance in the number of inferred components 
(see Table~\ref{tab:gaussian_2d_mixture_inference_boxplot}). These results are also
depicted through a boxplot (Fig.~\ref{fig:gaussian_2d_mixture_inference_boxplot}).
There are many instances where the number of inferred
components is more than 3.
The results indicate that the search method (FJ) is overfitting
the data. 

\begin{wrapfigure}{r}{0.5\textwidth}
  \centering
  \includegraphics[width=\textwidth]{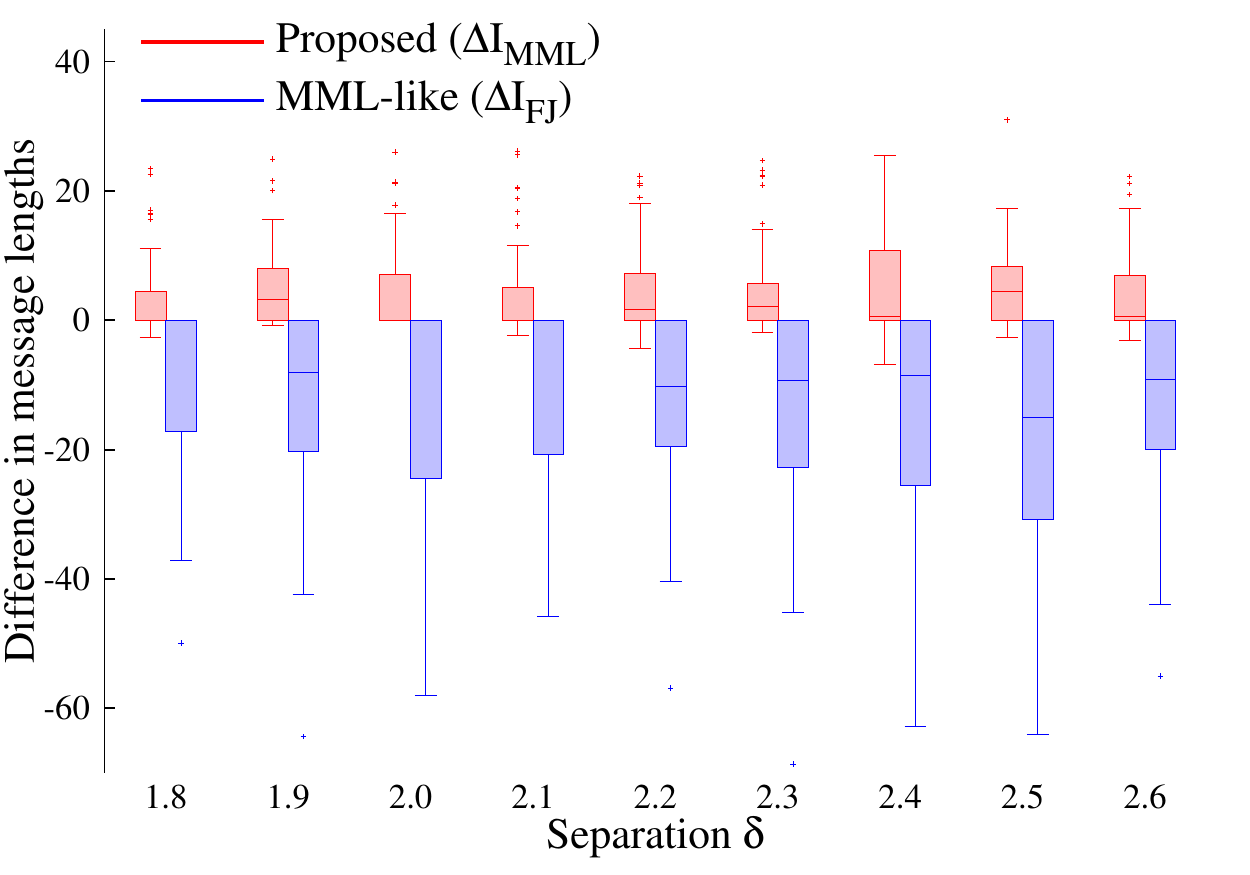}
  \caption{Bivariate mixture ($N=100$): difference in message lengths computed using the two different
           scoring functions (see Equation~\eqref{eqn:comparisons_mixtures}).}
  \label{fig:gaussian_2d_msglen_comparisons}
\end{wrapfigure}
Further, we evaluate the correctness of the mixtures
inferred by the two search methods by comparisons using the message length formulations
and KL-divergence.
Fig.~\ref{fig:gaussian_2d_msglen_comparisons} shows the boxplot of the difference
in message lengths of the mixtures $\fancym^*$ inferred using our proposed search method 
and the mixtures $\fancym^{FJ}$ inferred using that of \cite{figueiredo2002unsupervised}.
$\Delta I_{MML} > 0$ across all values of $\delta$ for the 50 simulations.
As per Equation~\eqref{eqn:comparisons_mixtures}, we have $I_{MML}(\fancym^*) < I_{MML}(\fancym^{FJ})$.
This implies that $\fancym^*$ has a lower message length compared to $\fancym^{FJ}$
when evaluated using our scoring function.
Similarly, we have $\Delta I_{FJ} < 0$, \textit{i.e.,} $I_{FJ}(\fancym^*) > I_{FJ}(\fancym^{FJ})$.
This implies that $\fancym^{FJ}$ has a lower message length compared to $\fancym^*$
when evaluated using FJ's scoring function.
These results are not surprising as $\fancym^*$ and $\fancym^{FJ}$ are obtained
using the search methods which optimize the respective MML and MML-like scoring functions.

\begin{figure}[htb]
  \centering
  \subfloat[]
  {
    \includegraphics[width=0.5\textwidth]{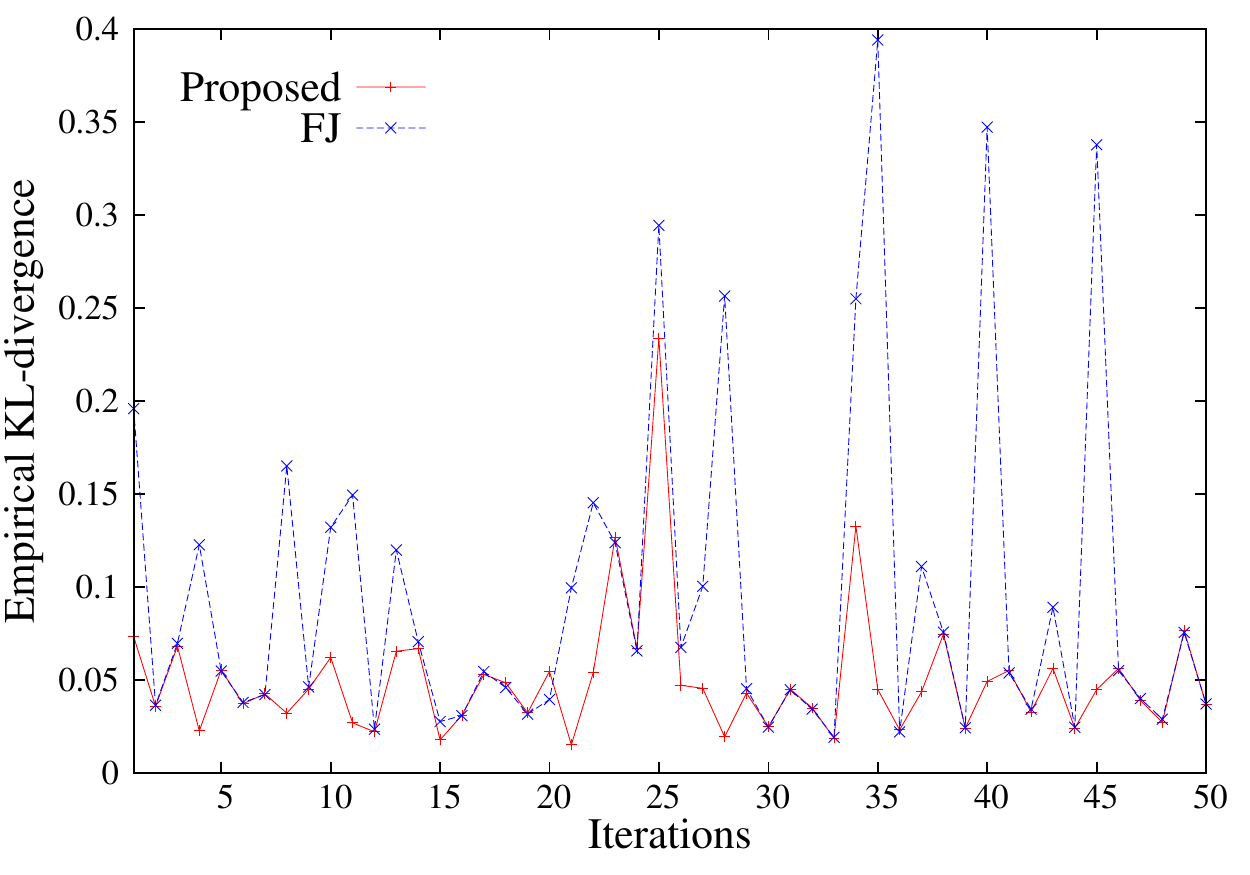}
  }
  \subfloat[]
  {
    \includegraphics[width=0.5\textwidth]{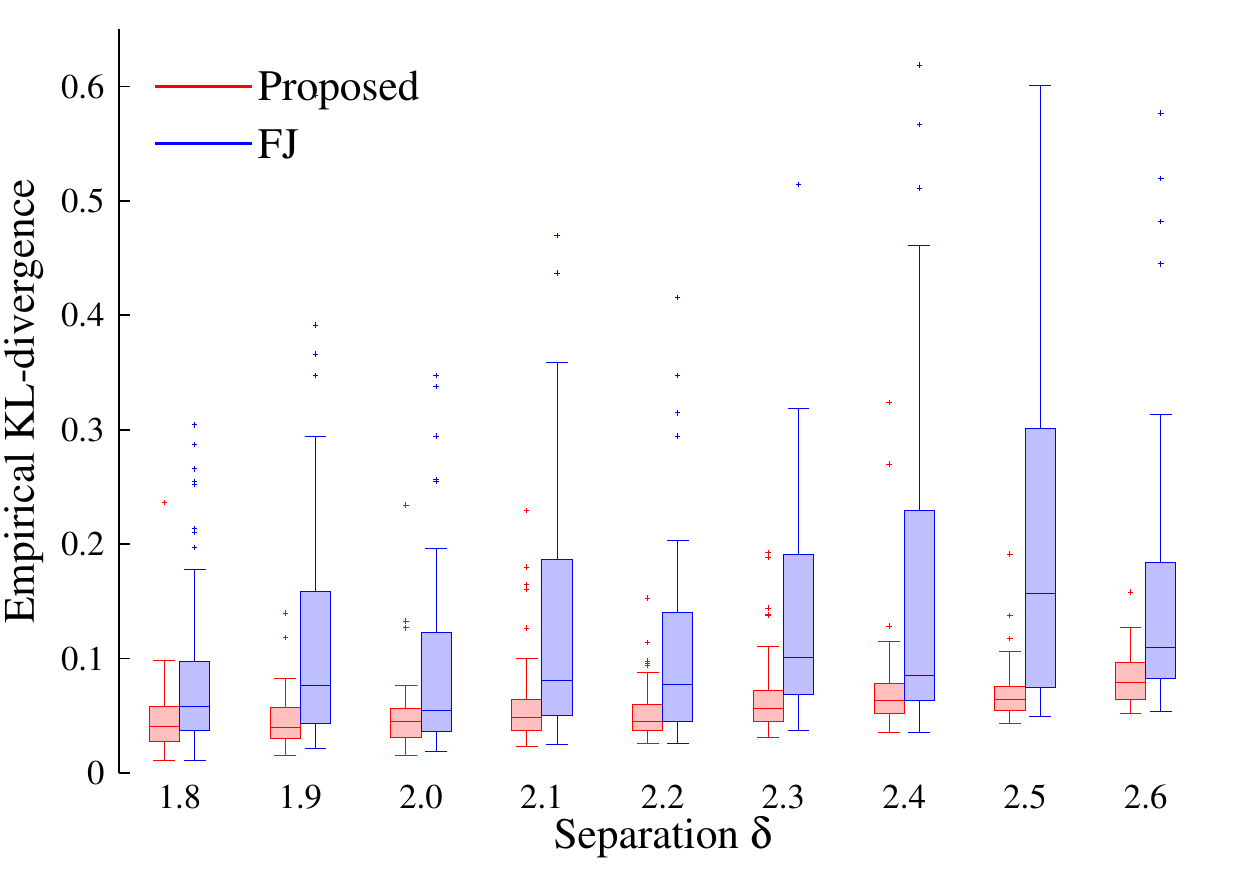}
  } 
  \caption{Comparison of inferred mixtures using KL-divergence (bivariate example with $N = 100$ and 50 simulations)
          (a) Particular case of $\delta = 2.0$ 
          (b) For all values of $\delta \in \{1.8,\dots,2.6\}$.}
  \label{fig:gaussian_2d_kldiv_comparisons}
\end{figure}

We then analyzed the KL-divergence of $\fancym^*$ and $\fancym^{FJ}$ with respect
to the true bivariate mixture $\fancym^t$ over all 50 simulations and across 
all values of $\delta$. Ideally, the KL-divergence should be close to zero.
Fig.~\ref{fig:gaussian_2d_kldiv_comparisons}(a) shows the KL-divergence
of the mixtures inferred using the two search methods with respect to $\fancym^t$
when the separation is $\delta=2.0$. The proposed search method infers
mixtures whose KL-divergence (denoted by red lines) is close to zero,
and more importantly less than the KL-divergence of mixtures inferred 
by the search method of \cite{figueiredo2002unsupervised} (denoted by blue lines).
The same type of behaviour is noticed with other values of $\delta$.
Fig.~\ref{fig:gaussian_2d_kldiv_comparisons}(b) compares the KL-divergence
for varying values of $\delta$. The median value of the KL-divergence
due to the proposed search method is close to zero with not much variation.
The search method of \cite{figueiredo2002unsupervised} always result in KL-divergence
higher than that of ours. The results suggest that, in this case,
mixtures $\fancym^{FJ}$ inferred by employing the search method of \cite{figueiredo2002unsupervised} 
deviate significantly from the true mixture distribution $\fancym^t$. This can also be explained
by the fact that there is a wide spectrum of the number of inferred components 
(Fig.~\ref{fig:gaussian_2d_mixture_inference_boxplot}). This suggests
that the MML-like scoring function is failing to control
the tradeoff between complexity and quality of fit, and hence,
is selecting overly complex mixture models.

\subsection{Simulation of 10-dimensional mixtures} \label{subsec:10d_mix_simulation}
\begin{figure}[htb]
  \centering
  \subfloat[]
  {
    \includegraphics[width=0.5\textwidth]{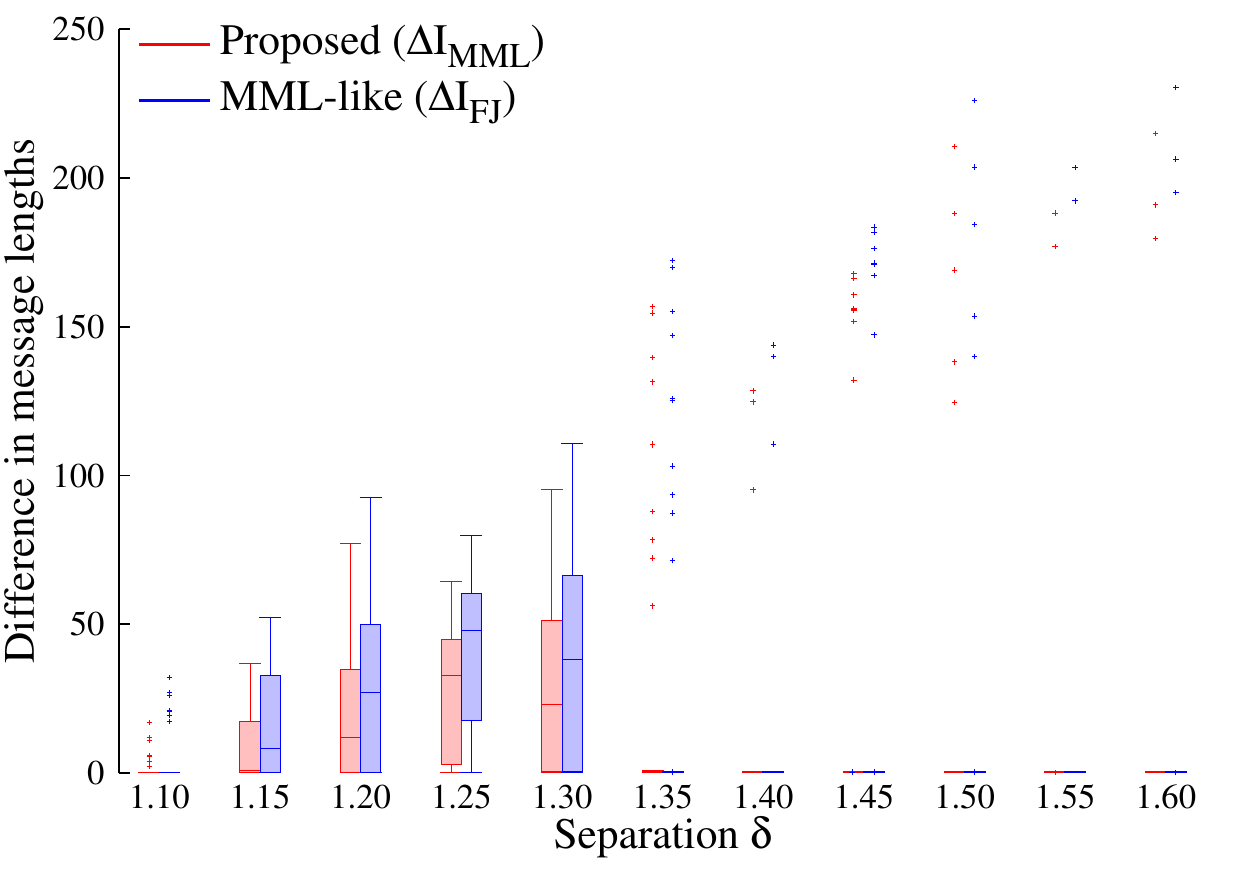}
  }
  \subfloat[]
  {
    \includegraphics[width=0.5\textwidth]{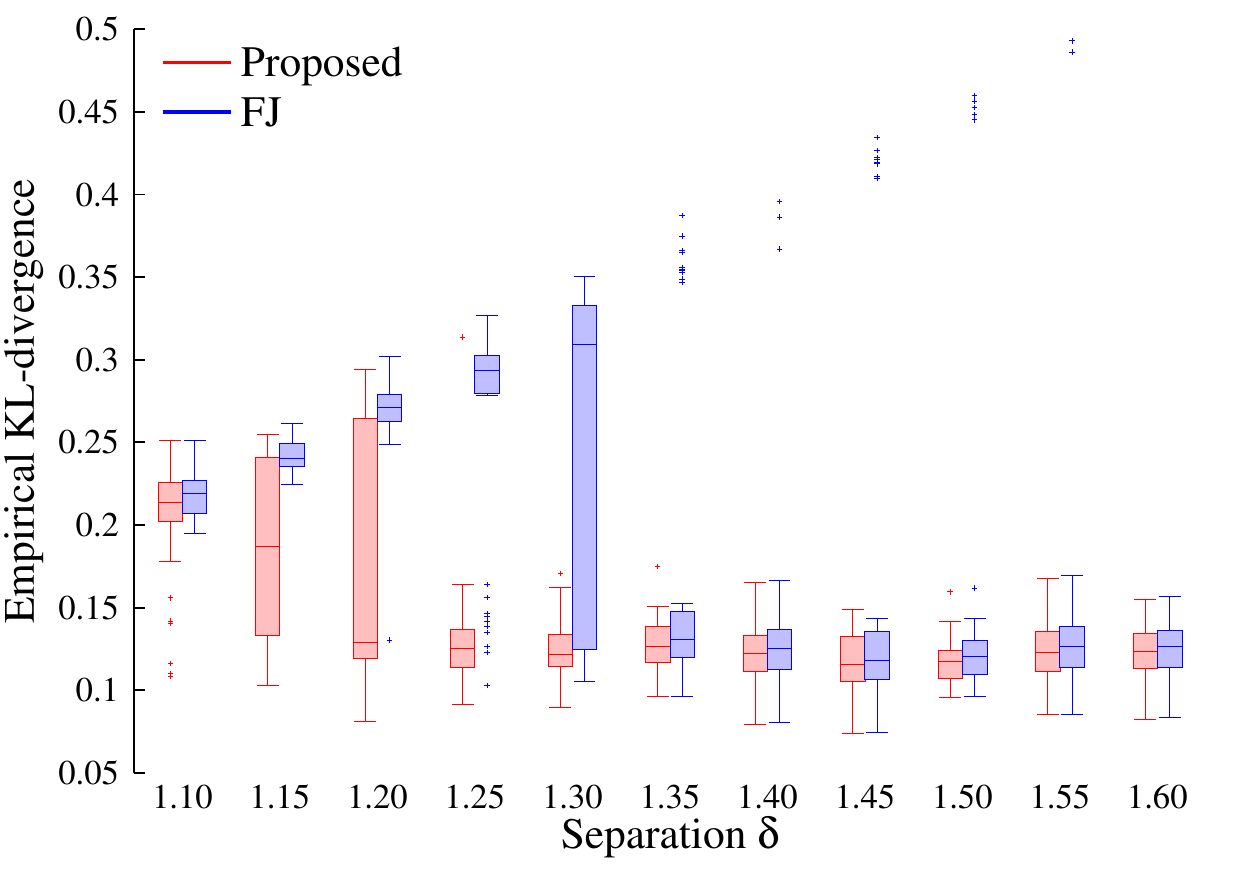}
  } 
  \caption{Comparison of mixtures with respect to the 10-variate true mixture and $N=800$
          (a) Difference in message lengths computed using the two scoring functions
          (b) Box-whisker plot of KL-divergence 
          }
  \label{fig:gaussian_10d_comparisons}
\end{figure}

Along the lines of the previous experiment, \cite{figueiredo2002unsupervised}
conducted another experiment for a 10-variate two-component mixture $\fancym^t$ with equal
mixing proportions. The means are at $\boldsymbol{\mu}_1=(0,\ldots,0)^T$ and $\boldsymbol{\mu}_2=(\delta,\ldots,\delta)^T$,
so that the Euclidean distance between them is $\delta \sqrt{10}$. The covariances of the
two components are $\mathbf{C}_1 = \mathbf{C}_2 = \mathbf{I}$ (the identity matrix).
Random samples of size $N=800$ were generated from the mixture and the number of
inferred components are plotted. The experiment is repeated for different values of $\delta$
and over 50 simulations.
\begin{figure}[htb]
  \centering
  \subfloat[]
  {
    \includegraphics[width=0.5\textwidth]{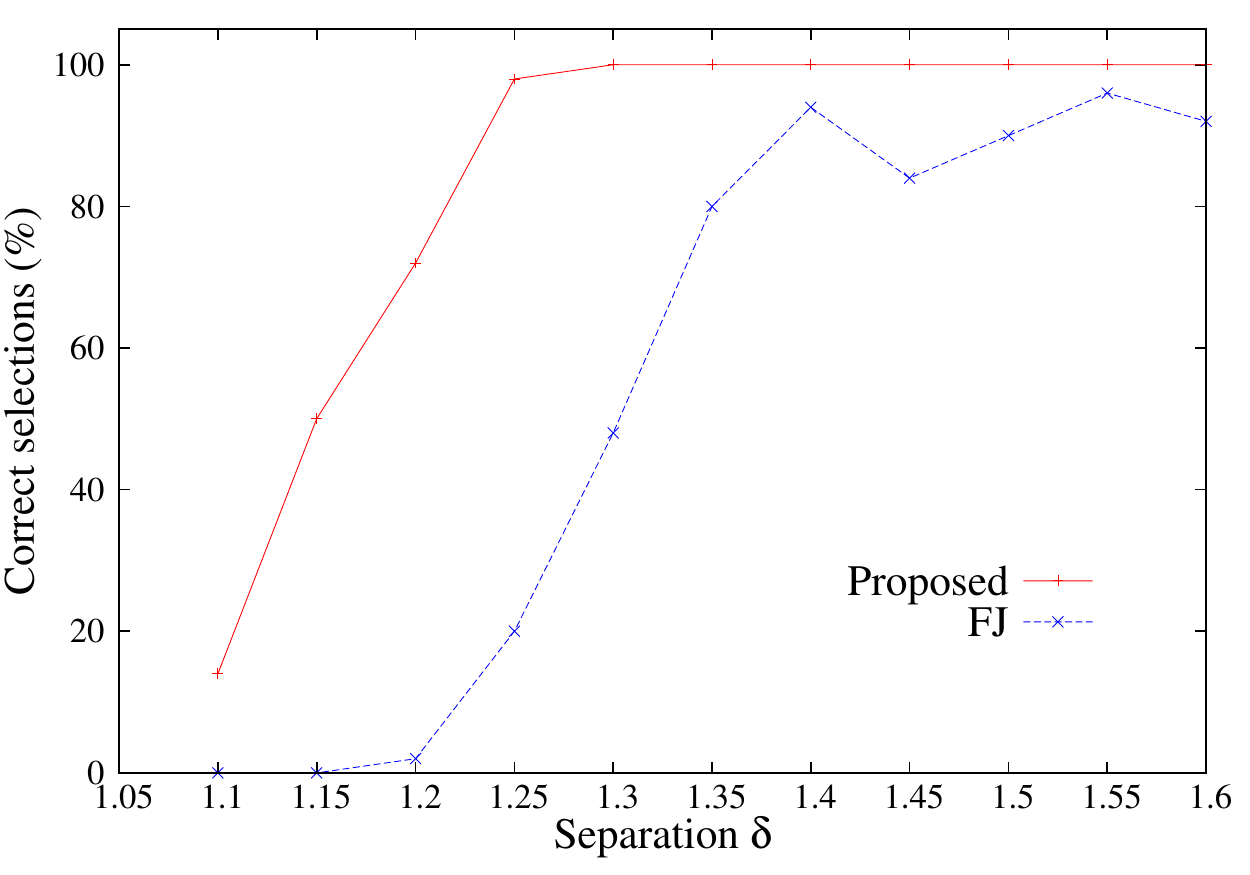}
  }
  \subfloat[]
  {
    \includegraphics[width=0.5\textwidth]{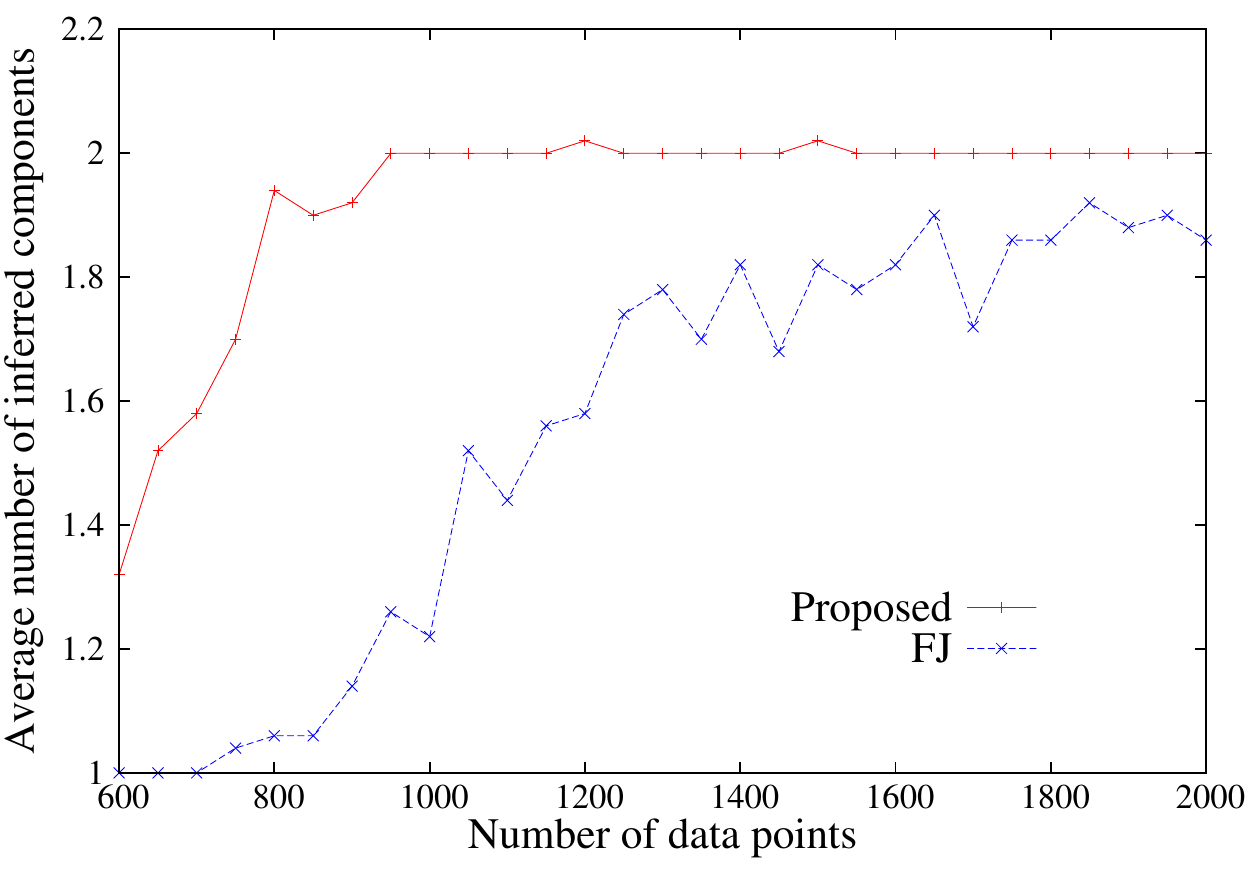}
  } 
  \caption{10-dimensional mixture: (a) Percentage of correct selections with varying $\delta$
            for a fixed sample size of $N=800$ (separation between the means is $\delta \sqrt{10}$)
           (b) Average number of inferred mixture components with different sample sizes
              and $\delta = 1.20$ between component means.
          }
  \label{fig:gaussian_10d_mixture_inference}
\end{figure}
Fig.~\ref{fig:gaussian_10d_mixture_inference}(a) shows the number of inferred components
using the two search methods. At lower values of $\delta$, the components are close
to each other, and hence, it is relatively more difficult to correctly infer the true number of components.
We observe that our proposed method performs clearly better than that of \cite{figueiredo2002unsupervised}
across all values of $\delta$. We also compared the quality of these inferred mixtures 
by calculating the difference in message lengths
using the two scoring functions and the KL-divergence with respect to $\fancym^t$ . 
For all values of $\delta$, $\Delta I_{MML} > 0$, \textit{i.e.,}
our inferred mixtures $\fancym^*$ have a lower message length compared to $\fancym^{FJ}$
when evaluated using our scoring function. More interestingly, we also note that
$\Delta I_{FJ} > 0$ (see Fig.~\ref{fig:gaussian_10d_comparisons}(a)). 
This reflects that $\fancym^*$ have a lower message length compared 
to $\fancym^{FJ}$ when evaluated using the scoring function of \cite{figueiredo2002unsupervised}.
This suggests that their search method results in a sub-optimal mixture $\fancym^{FJ}$
and fails to infer the better $\fancym^*$.

In addition to the message lengths, we analyze the mixtures using KL-divergence.
Similar to the bivariate example in Fig.~\ref{fig:gaussian_2d_kldiv_comparisons}(a),
the KL-divergence of our inferred mixtures $\fancym^*$ is lower than $\fancym^{FJ}$, the mixtures
inferred by \cite{figueiredo2002unsupervised}.
Fig.~\ref{fig:gaussian_10d_comparisons}(b) shows the boxplot of 
KL-divergence of the inferred mixtures $\fancym^*$ and $\fancym^{FJ}$.
At higher values of $\delta >= 1.45$, 
the median value of KL-divergence is close to zero, as the number of correctly
inferred components (Fig.~\ref{fig:gaussian_10d_mixture_inference}(a)) is more than 90\%.
However, our method always infers mixtures $\fancym^*$ with lower KL-divergence 
compared to $\fancym^{FJ}$. These experimental results demonstrate 
the superior performance of our proposed search method.

Another experiment was carried out where 
the $\delta=1.20$ was held constant (extremely close components), gradually increased 
the sample size $N$,
and plotted the average number of inferred components by running 50 simulations for
each $N$. Fig.~\ref{fig:gaussian_10d_mixture_inference}(b) shows the results for
the average number of inferred components as the amount of data increases.
Our search method, on average, infers the true mixture when the sample 
size is $\sim 1000$.
However, the search method of \cite{figueiredo2002unsupervised} requires
larger amounts of data; even with a sample size of 2000, the average number of inferred
components is $\sim 1.9$. In Fig.~\ref{fig:gaussian_10d_mixture_inference}(b),
the red curve reaches the true number of 2 and saturates more rapidly 
than the blue curve.

\subsection{The impact of weight updates as formulated by \cite{figueiredo2002unsupervised}}
\label{subsec:fj_weight_updates_exp2b}
One of the drawbacks associated with the search method of \cite{figueiredo2002unsupervised}
is due to the form of the updating expression for the component weights 
(Equation~\eqref{eqn:fj_weight_update}). As discussed in Section~\ref{subsubsec:fj_search_drawbacks},
a particular instance of wrong inference is bound to happen when the
net membership of a (valid) component is less than $N_p/2$, where $N_p$
is the number of free parameters per component. In such a case,
the component weight is updated as zero, and it is eliminated,
effectively reducing the mixture size by one. 

We conducted the following experiment to demonstrate this behaviour: we considered the
two-component 10-variate mixture $\fancym^t$ as before and randomly
generate samples of size 50 from the mixture. Since the constituent components of $\fancym^t$
have equal weights, on average, each component has a membership of 25.
We used $\delta=\{10,100,1000\}$, so that the two components are well 
apart from each other. For each $\delta$, we run 50 simulations
and analyze the number of inferred components.
As expected, the search method of \cite{figueiredo2002unsupervised}
always infer a mixture with one component regardless of the separation $\delta$.
Our method always infers the correct number of components.
In order to test the validity of mixtures inferred by our proposed method,
we analyze the resultant mixtures by comparing the message lengths as discussed in 
Section \ref{subsec:comparison_mixtures}. 
\begin{figure}[htb]
  \centering
  \subfloat[$\Delta I_{MML}$]
  {
    \includegraphics[width=0.5\textwidth]{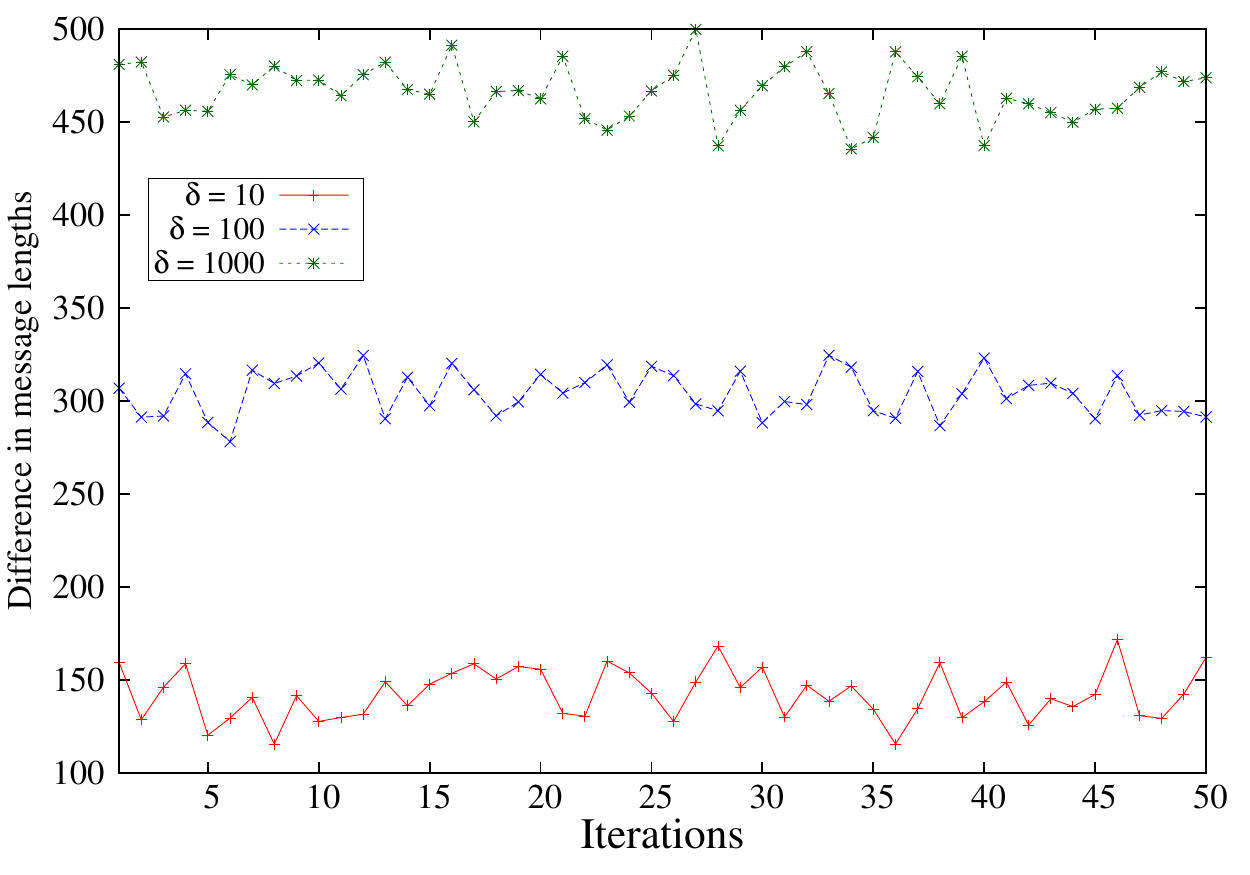}
  }
  \subfloat[$\Delta I_{FJ}$]
  {
    \includegraphics[width=0.5\textwidth]{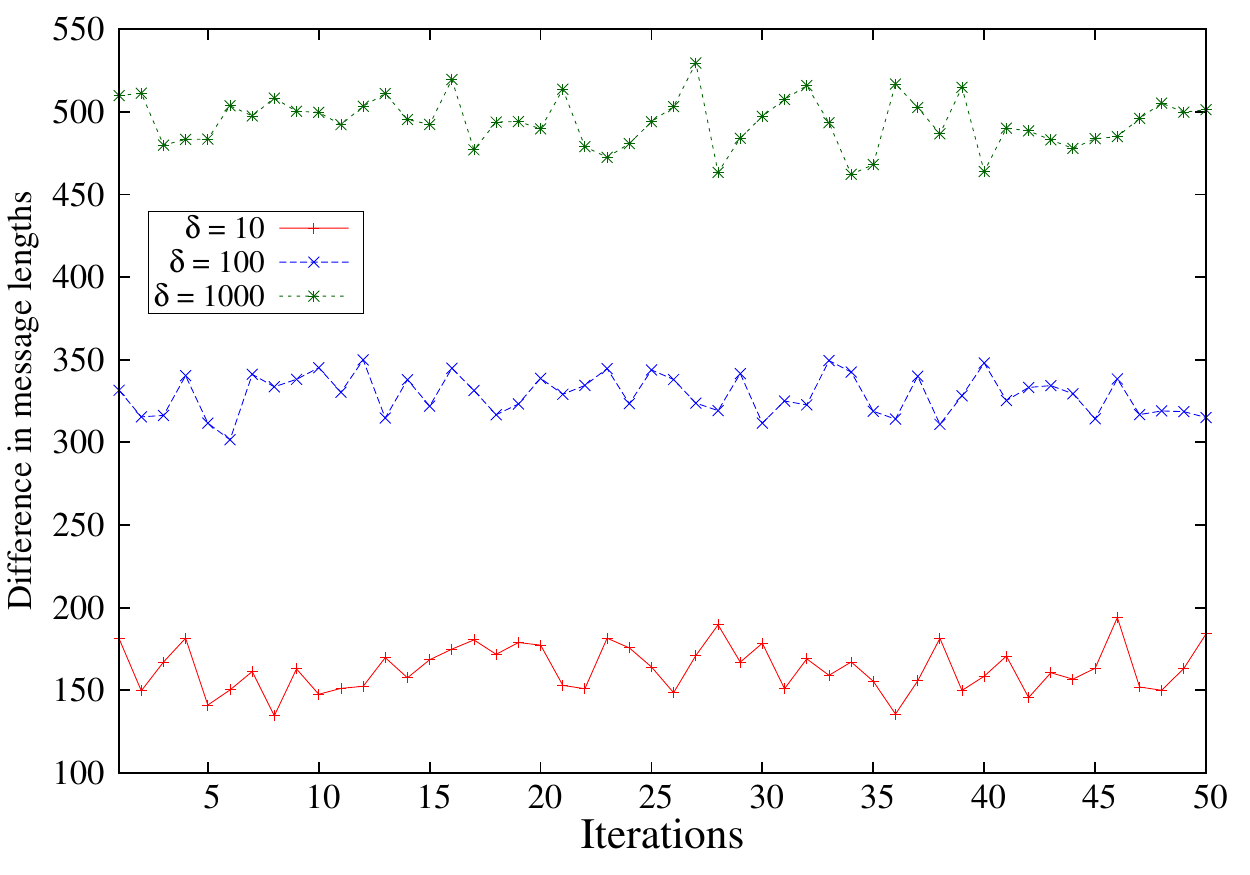}
  } 
  \caption{Evaluation of the quality of inferred mixtures 
           by comparing the difference in message lengths as computed using the two
           scoring functions. Positive difference indicates that the mixtures inferred
           by our search method have lower message lengths 
           (see Equation~\eqref{eqn:comparisons_mixtures}).}
  \label{fig:gaussian_10d_msglens_comparisons_exp2b}
\end{figure}
Fig.~\ref{fig:gaussian_10d_msglens_comparisons_exp2b}(a)
shows the difference in message lengths $\Delta I_{MML}$ 
given in Equation \eqref{eqn:comparisons_mixtures}. We observe that $\Delta I_{MML} > 0$ for 
all $\delta$. This demonstrates that our search based mixtures $\fancym^*$ 
have lower message lengths compared to mixtures $\fancym^{FJ}$ using our scoring function. 
The same phenomenon is observed when using the MML-like scoring function of 
\cite{figueiredo2002unsupervised}. In Fig.~\ref{fig:gaussian_10d_msglens_comparisons_exp2b}(b),
we observe that $\Delta I_{FJ} > 0$, which means
our search based mixtures $\fancym^*$ have lower message lengths compared
to mixtures $\fancym^{FJ}$ when evaluated using their scoring function. 

\begin{wrapfigure}{r}{0.45\textwidth}
\centering
\vspace{-3mm}
\includegraphics[width=\textwidth]{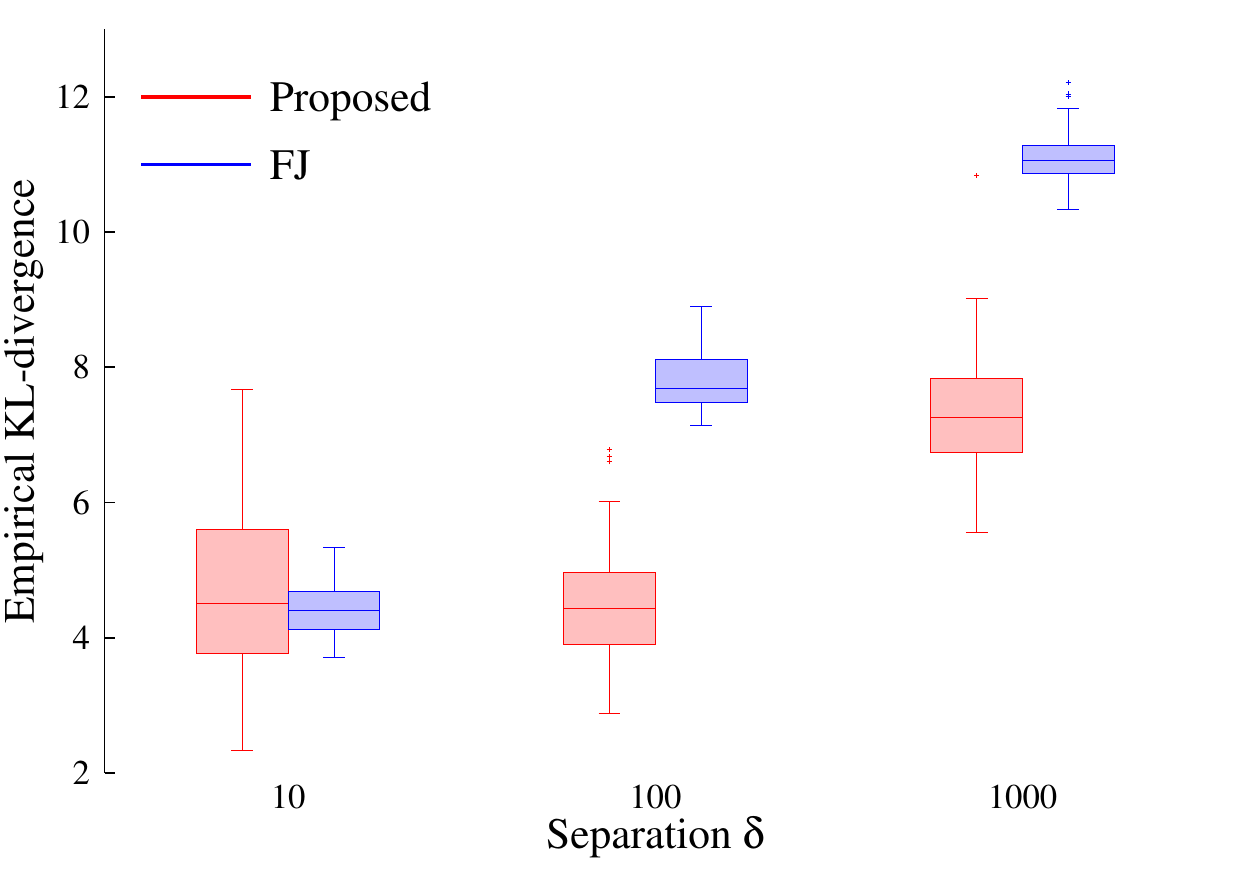}
\caption{Box-whisker plot of KL-divergence of mixtures inferred by the two search methods. 
         A random sample of size $N=50$ is generated for each $\delta$ and this
         is repeated 50 times.}
\label{fig:gaussian_10d_kldiv_comparisons_exp2b}
\vspace{-3mm}
\end{wrapfigure}
This demonstrates that $\fancym^*$ is a better mixture as compared to $\fancym^{FJ}$ 
and their search method is unable to infer it. We also note that the differences
in message lengths increases with increasing $\delta$. This is because 
for the one-component inferred mixture $\fancym^{FJ}$, the second part of the message
(Equation~\eqref{eqn:fj}) which corresponds to the negative log-likelihood term
increases because of poorer fit to the data. 
The two modes in the data
become increasingly pronounced as the separation
between constituent components of the true mixture increases, and hence, modelling such
a distribution using a one-component mixture results in a poorer fit. This is
clearly an incorrect inference.

We further strengthen our case by comparing the KL-divergence of 
the inferred mixtures $\fancym^*$ and $\fancym^{FJ}$ with respect to the true mixture.
Fig.~\ref{fig:gaussian_10d_kldiv_comparisons_exp2b} illustrates the results.
As $\delta$ increases, the blue coloured plots 
shift higher. These correspond to mixtures
$\fancym^{FJ}$ inferred by \cite{figueiredo2002unsupervised}. Our search method,
however, infers mixtures $\fancym^*$ which have lower KL-divergence. The figure
indicates that the inferred mixtures $\fancym^*$ are more similar to the true
distribution as compared to mixtures $\fancym^{FJ}$. 

These experiments demonstrate the ability of our search
method to perform better than the widely used method of \cite{figueiredo2002unsupervised}.
We compared the resulting mixtures using our proposed MML formulation
and the MML-like formulation of \cite{figueiredo2002unsupervised},
showing the advantages of the former over the latter.
We also used a neutral metric, KL-divergence, to establish the
closeness of our inferred mixtures to the true distributions.
We will now illustrate the behaviour of our search method on two real world datasets. 

\subsection{Analysis of the computational cost}
\begin{figure}[htb]
  \centering
  \subfloat[Bivariate mixture ($N=800$)]
  {
    \includegraphics[width=0.33\textwidth]{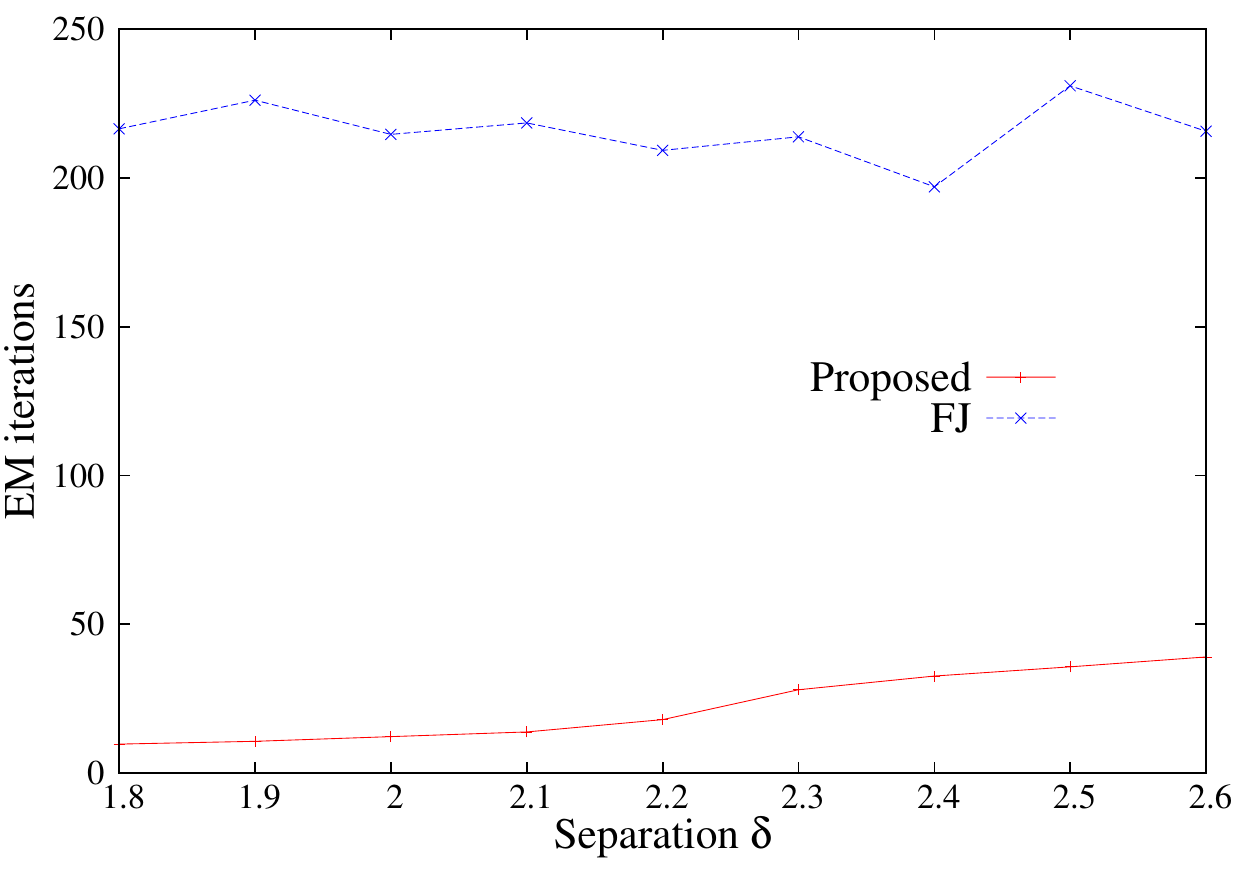}
  }
  \subfloat[Bivariate mixture ($N=100$)]
  {
    \includegraphics[width=0.33\textwidth]{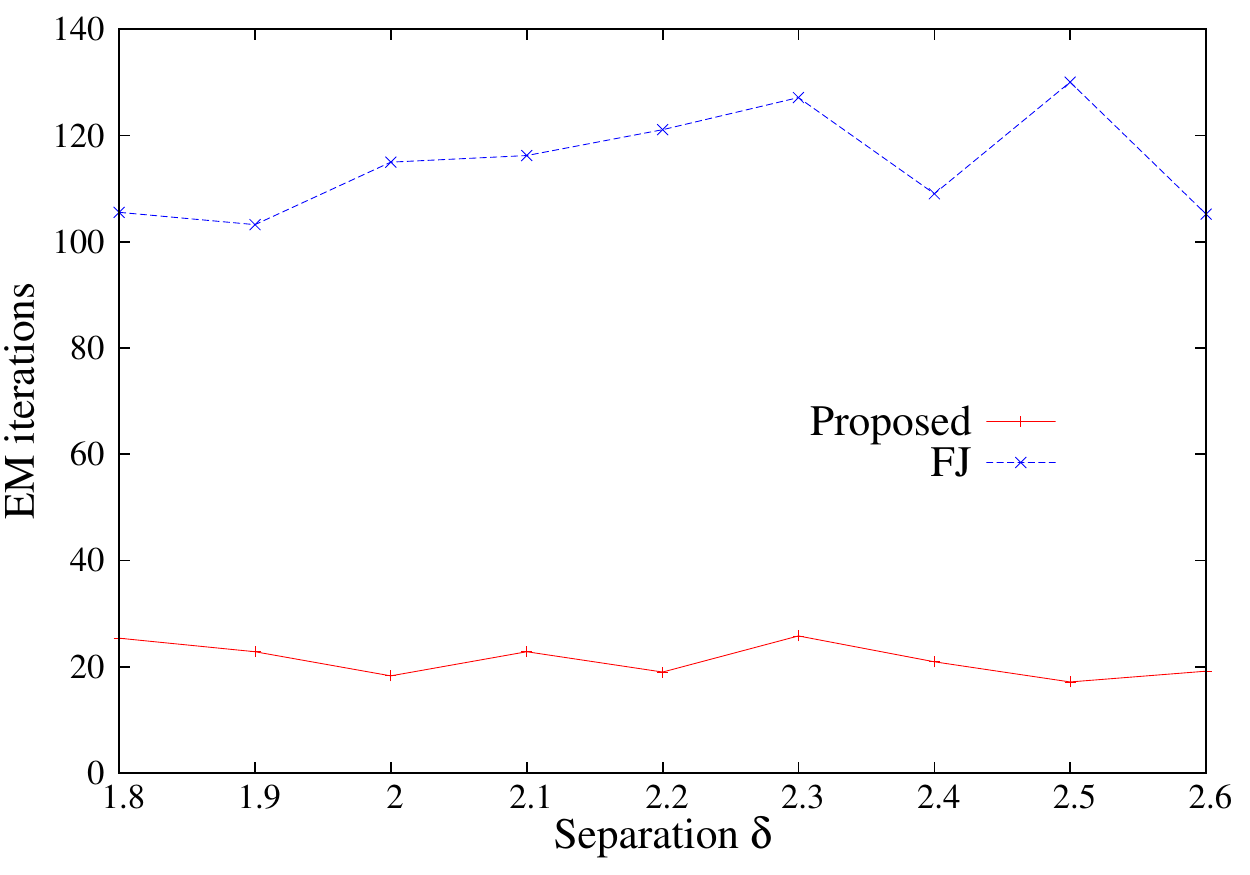}
  }
  \subfloat[10-variate mixture ($N=800$)]
  {
    \includegraphics[width=0.33\textwidth]{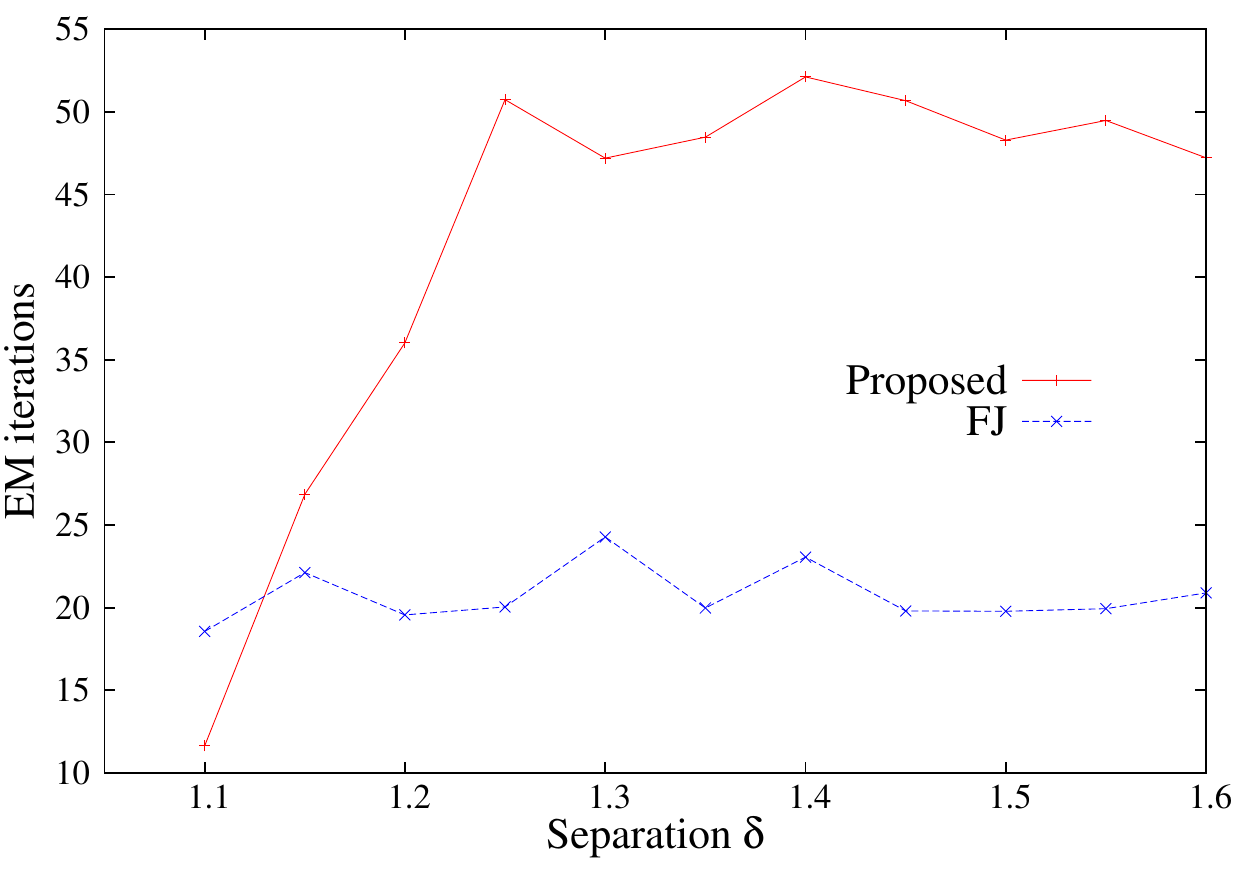}
  }
  \caption{Number of EM iterations performed during the mixture simulations discussed in Sections
~\ref{subsec:bivariate_mix_simulation} and ~\ref{subsec:10d_mix_simulation}.
          }
  \label{fig:em_iterations}
\end{figure}

At any intermediate stage of the search procedure, a \emph{current} mixture 
with $M$ components requires $M$
number of split, delete, and merge operations before it is updated. 
Each of the perturbations involve performing an EM
to optimize the corresponding mixture parameters. 
To determine the convergence of EM, we used a threshold of $10^{-5}$
which was the same as used by \cite{figueiredo2002unsupervised}.
FJ's method also requires to start from an initial large number of components.
We used 25 as an initial number based on what was suggested in \cite{figueiredo2002unsupervised}.
We investigate the number of times the EM routine is called and compare
it with that of \cite{figueiredo2002unsupervised}. 
We examine with respect to 
the simulations that were carried out previously. For the bivariate mixture
discussed in Section~\ref{subsec:bivariate_mix_simulation}, the number of resulting
EM iterations when the sample sizes are $N=800$ and $N=100$
are compared in Fig.~\ref{fig:em_iterations}(a), (b) respectively.

As per the discussion in Section~\ref{subsec:bivariate_mix_simulation},
at $N=800$, the average number of components inferred by the two methods
are about the same (Fig.~\ref{fig:gaussian_2d_mixture_inference}(a)).
However, the number of EM iterations required by FJ's method is greater than
200 across all values of $\delta$ (Fig.~\ref{fig:em_iterations}(a)). In
contrast, the proposed method, on average, requires fewer than 50 iterations.
In this case, both methods produce a similar result with FJ's method
requiring more number of EM iterations. When the bivariate mixture simulation
is carried out using $N=100$, the number of EM iterations required by FJ's method,
on average, is greater than 100, while the proposed method requires fewer
than 40 iterations (Fig.~\ref{fig:em_iterations}(b)). 
In this case, the proposed method not only infers
better mixtures (as discussed in Section~\ref{subsec:bivariate_mix_simulation}) 
but is also conservative with respect to computational cost.

For the simulation results corresponding to the 10-variate mixtures 
in Section~\ref{subsec:10d_mix_simulation}, the proposed method requires
close to 50 iterations on average, while FJ's method requires about 20
(Fig.~\ref{fig:em_iterations}(c)).
However, the mixtures inferred by the proposed method fare better when
compared to that of FJ (Figs.~\ref{fig:gaussian_10d_mixture_inference},
\ref{fig:gaussian_10d_comparisons}). Furthermore, for the simulation results
explained in Section~\ref{subsec:fj_weight_updates_exp2b}, FJ's method stops
after 3 EM iterations. This is because their program does not accommodate
components when the memberships are less than $N_p/2$. The proposed method
requires 18 EM iterations on average and infers the correct mixture components.
In these two cases, our method infers better quality mixtures, with no 
significant overhead with regard to the computational cost.

\subsection{Acidity data set \citep{richardson1997bayesian,mclachlan1997contribution}}
\begin{figure}[htb]
  \centering
  \subfloat[Mixture $\fancym^*$ inferred using our proposed search method.]
  {
    \includegraphics[width=0.5\textwidth]{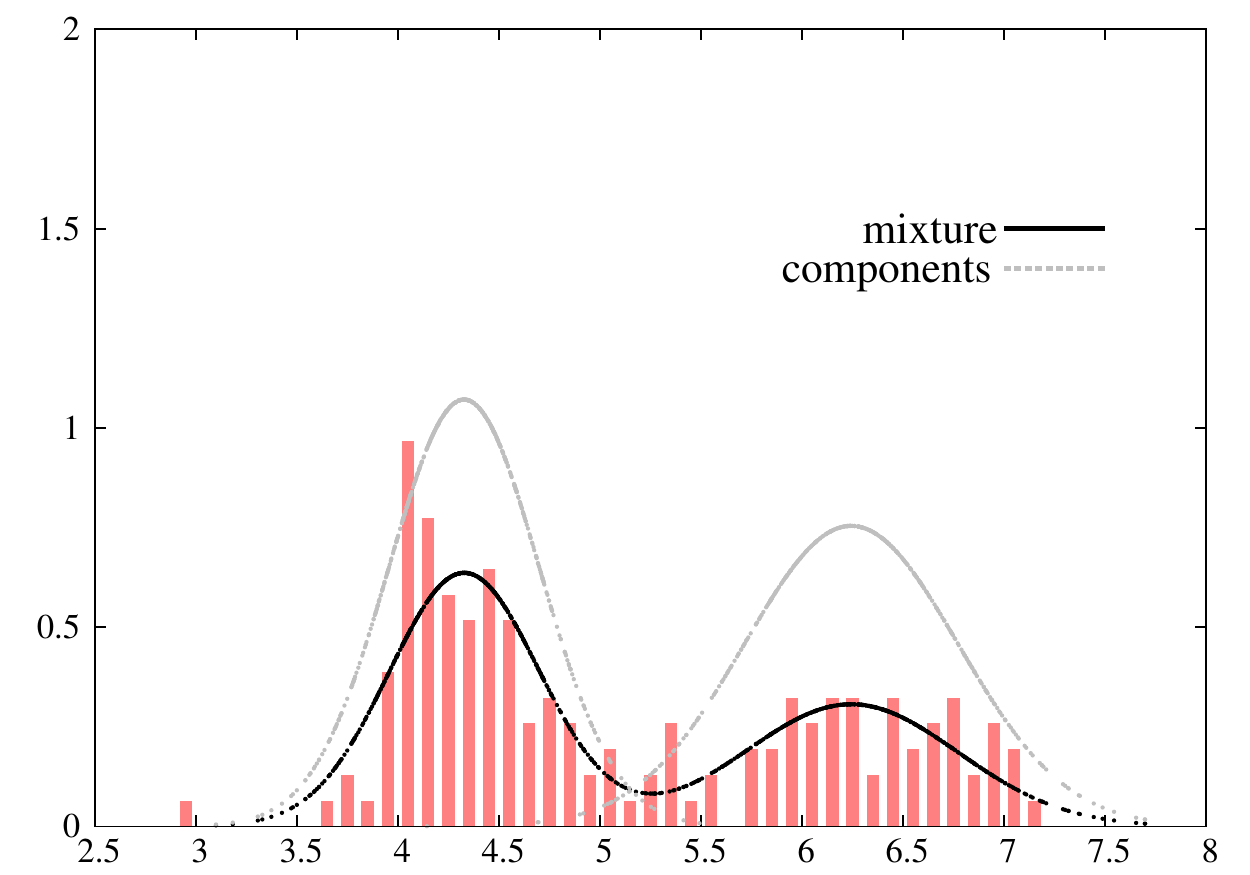}
  }
  \subfloat[Mixture $\fancym^{FJ}$ inferred using FJ's search method.]
  {
    \includegraphics[width=0.5\textwidth]{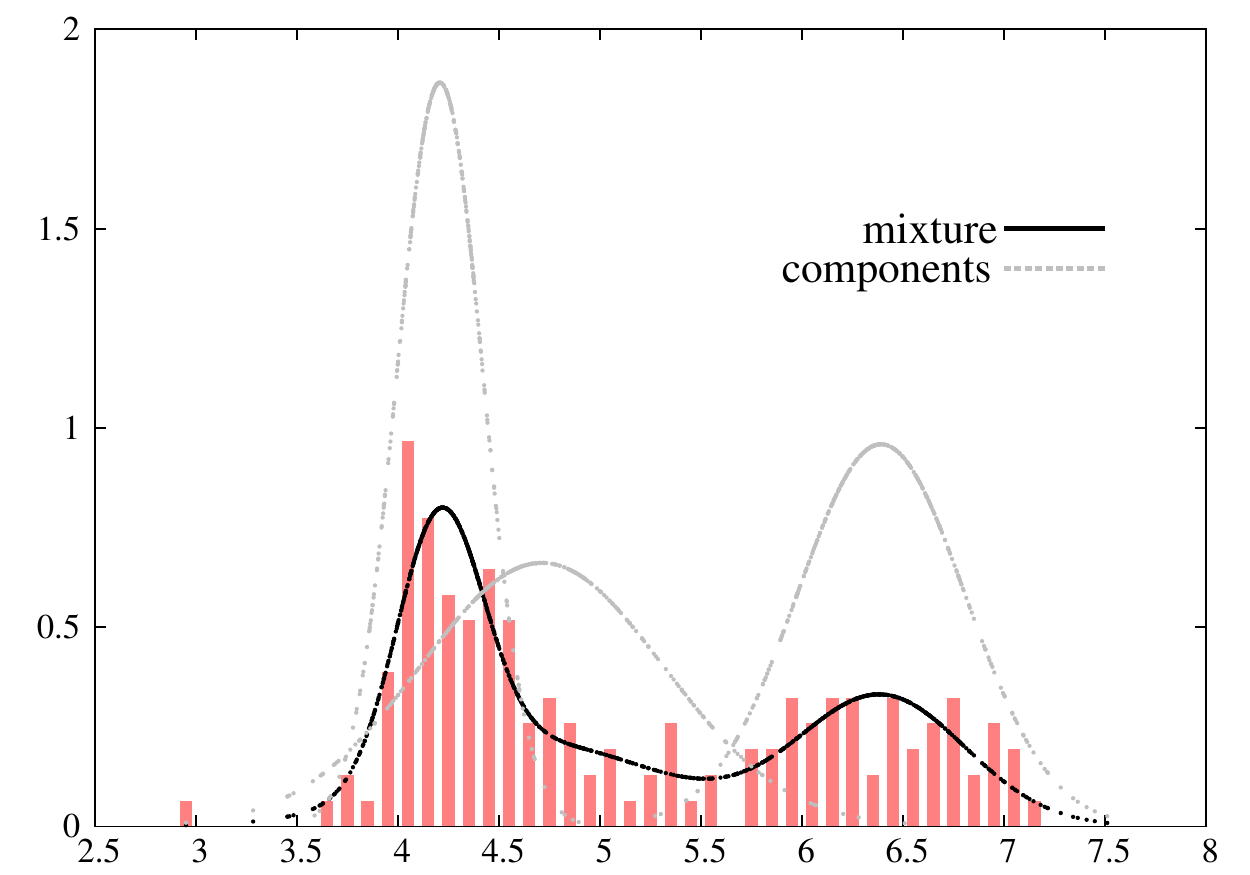}
  } 
  \caption{Mixtures inferred by the two search methods using the acidity data set.
           See Table~\ref{tab:acidity_mixtures_inference} for the corresponding parameter 
           estimates.}
  \label{fig:acidity}
\end{figure}

The first example is the univariate \emph{acidity} 
data set which contains 155 points. 
Our proposed search method infers a mixture $\fancym^*$ with 2 components
whereas the search method of \cite{figueiredo2002unsupervised} infers
a mixture $\fancym^{FJ}$ with 3 components.
The inferred mixtures are shown in Fig.~\ref{fig:acidity} and their corresponding
parameter estimates are given in Table~\ref{tab:acidity_mixtures_inference}.
In order to compare the mixtures inferred by the two search methods,
we compute the message lengths of the inferred mixtures 
using our complete MML and the approximated MML-like scoring functions.

When evaluated using our MML scoring function, our inferred mixture results in a
gain of $\sim 4$ bits (see Table~\ref{tab:acidity_comparison}).
Based on the MML framework, our two-component mixture $\fancym^*$ is $2^4$ times more likely  
than the three-component mixture $\fancym^{FJ}$ (as per Equation~\eqref{eqn:compare_models}).
Furthermore, when the inferred mixtures are
evaluated as per the MML-like scoring function, $\fancym^*$
is still considered better ($\sim 298$ bits) than $\fancym^{FJ}$ ($\sim 320$ bits).
Thus, using both forms of scoring function,
$\fancym^*$ is the better mixture model of this data set.

\begin{figure}[H]
  \begin{minipage}{0.6\textwidth}
      \begin{table}[H]
        \caption{The parameters of the inferred mixtures shown in Fig.~\ref{fig:acidity}}
        \subfloat[Proposed]
        {
          \begin{tabular}{|c|c|c|}
            \hline
            Component &   \multirow{2}{*}{Weight}      & Parameters       \\ 
            index     &                                & ($\mu,\sigma^2$) \\ \hline
            1         & 0.41  & 6.24, 0.28 \\
            2         & 0.59  & 4.33, 0.14 \\
            \hline
          \end{tabular}
        }
        \subfloat[FJ]
        {
          \begin{tabular}{|c|c|c|}
            \hline
            Component &   \multirow{2}{*}{Weight}      & Parameters       \\ 
            index     &                                & ($\mu,\sigma^2$) \\ \hline
            1         & 0.34  & 6.39, 0.17 \\
            2         & 0.35  & 4.21, 0.05 \\
            3         & 0.31  & 4.71, 0.36 \\
            \hline
          \end{tabular}
        }
        \label{tab:acidity_mixtures_inference}
      \end{table}
  \end{minipage}
  \quad
  \begin{minipage}{0.37\textwidth}
      \begin{table}[H]
        \caption{Message lengths (measured in bits) of the mixtures 
                 (in Fig.~\ref{fig:acidity}) as evaluated using the 
                 MML and MML-like scoring functions.
                }
        \begin{tabular}{|c|c|c|}
          \cline{2-3}
          \multicolumn{1}{c|}{}    & \multicolumn{2}{c|}{Inferred mixtures} \\\hline
          Scoring   &   Proposed         & FJ \\ 
          functions &   ($\fancym^*$)    & ($\fancym^{FJ}$) \\ \hline
          MML       &  \textbf{1837.61}  & 1841.69 \\ 
          MML-like  &  \textbf{298.68}   & 320.02 \\
          \hline
        \end{tabular}
        \label{tab:acidity_comparison}
      \end{table}
  \end{minipage}
\end{figure}

\subsection{Iris data set \citep{anderson1935irises,fisher1936iris}}
\begin{figure}[htb]
  \centering
  \subfloat[Mixture $\fancym^*$ inferred using our proposed search method.]
  {
    \includegraphics[width=0.5\textwidth]{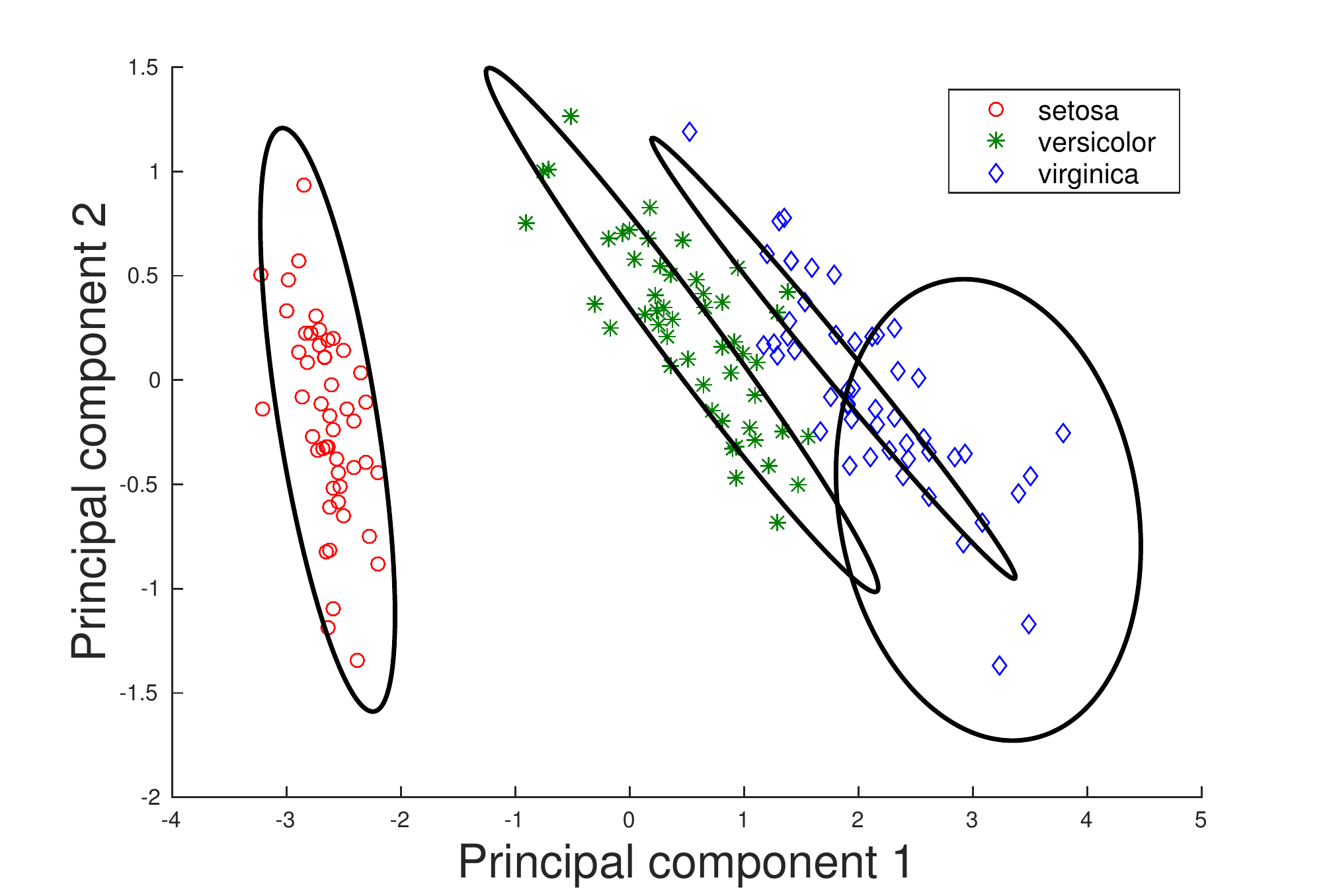}
  }
  \subfloat[Mixture $\fancym^{FJ}$ inferred using FJ's search method.]
  {
    \includegraphics[width=0.5\textwidth]{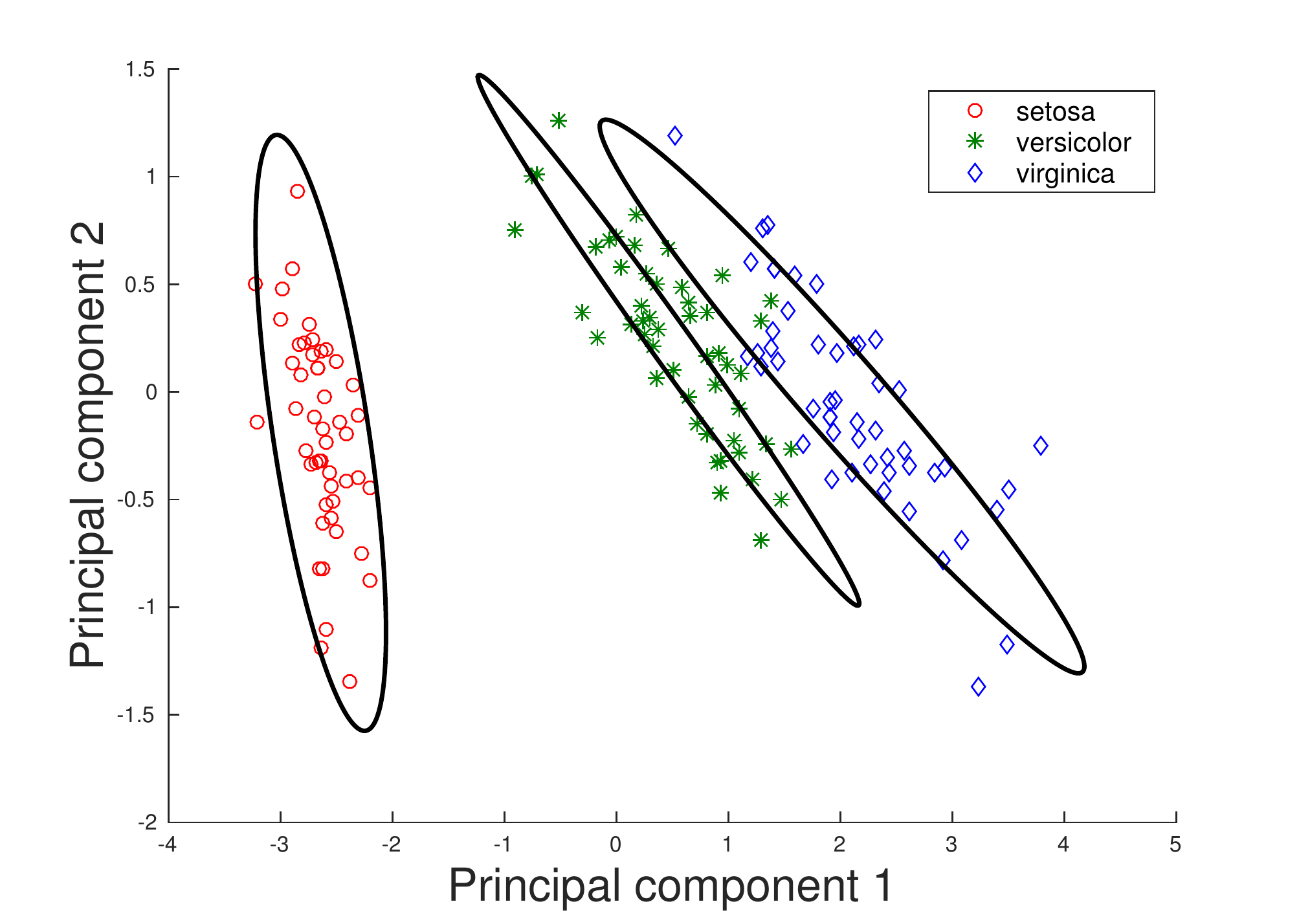}
  } 
  \caption{Mixtures inferred by the two search methods using the Iris data set.
           The data is projected onto the two principal components.}
  \label{fig:iris}
\end{figure}

The second example is the popular Iris data set. The data
is 4-dimensional and comes from three Iris species namely,
\emph{Iris-setosa, Iris-versicolor,} and \emph{Iris-virginica}.
The data size is 150 with each class (species) comprising of 50 representative elements.
Our search method infers a 4 component mixture $\fancym^*$
and the search method of \cite{figueiredo2002unsupervised} infers
a 3 component mixture $\fancym^{FJ}$ (see Fig.~\ref{fig:iris}).
Table~\ref{tab:iris_mixture_inference} shows the memberships of the 150 elements 
in each of the components in the inferred mixtures.
We notice an additional component M4 in $\fancym^*$ which has a net membership 
of 9.51, that is $\sim 6\%$ of the entire data set. It appears that the component
M2 in $\fancym^{FJ}$ (Table~\ref{tab:iris_mixture_inference}(b)) is split into
two components M2 and M4 in $\fancym^*$ (Table~\ref{tab:iris_mixture_inference}(a)).
The quality of the inferred mixtures is determined by comparing their
message lengths using the MML and MML-like scoring functions.
Table~\ref{tab:iris_comparison} shows the values obtained using the two formulations.
When evaluated using our complete MML formulation, our inferred mixture $\fancym^*$ 
results in extra compression of $\sim 1$ bit, which makes it twice as likely as
$\fancym^{FJ}$ -- it is a closely competing model compared to ours.
When evaluated using the MML-like scoring function, our inferred
mixture still has a lower message length compared to $\fancym^{FJ}$.
In both the cases, the mixture $\fancym^*$ inferred by our search method
is preferred.

\begin{figure}[H]
  \begin{minipage}{0.65\textwidth}
  \begin{table}[H]
    \caption{Memberships of Iris data as using the inferred mixtures in Fig.~\ref{fig:iris}
             (a) Distribution of data using $\fancym^*$
             (b) Distribution of data using $\fancym^{FJ}$
}
    \subfloat[Data distribution using 4 components]
    {
      \begin{tabular}{|c|c|c|c|c|}
        \hline
        Species           &  M1   & M2    & M3    & M4  \\ \hline
        \emph{setosa}     &  50   & 0     & 0     & 0   \\
        \emph{versicolor} &  0    & 5.64  & 44.36 & 0   \\
        \emph{virginica}  &  0    & 40.29 & 0.20  & 9.51\\
        \hline
      \end{tabular}
    }
    \subfloat[Data distribution using 3 components]
    {
      \begin{tabular}{|c|c|c|c|}
        \hline
        Species           &  M1   & M2   & M3    \\ \hline
        \emph{setosa}     &  50   & 0    & 0     \\
        \emph{versicolor} &  0    & 5.55 & 44.45  \\
        \emph{virginica}  &  0    & 49.78& 0.22 \\
        \hline
      \end{tabular}
    }
    \label{tab:iris_mixture_inference}
  \end{table}
  \end{minipage}
  \quad
  \begin{minipage}{0.3\textwidth}
      \begin{table}[H]
        \caption{Message lengths (measured in bits) of the mixtures 
                 (in Fig.~\ref{fig:iris}) as evaluated using the 
                 MML and MML-like scoring functions.
                }
        \begin{tabular}{|c|c|c|}
          \cline{2-3}
          \multicolumn{1}{c|}{}    & \multicolumn{2}{c|}{Inferred mixtures} \\\hline
          Scoring   &   Proposed         & FJ \\ 
          functions &   ($\fancym^*$)    & ($\fancym^{FJ}$) \\ \hline
          MML       &  \textbf{6373.01}  & 6374.27 \\ 
          MML-like  &  \textbf{323.31}   & 342.57 \\
          \hline
        \end{tabular}
        \label{tab:iris_comparison}
      \end{table}
  \end{minipage}
\end{figure}

\section{Experiments with von Mises-Fisher distributions} \label{sec:vmf_experiments}
We compare our MML-based parameter inference with the current state of the art
vMF estimators (discussed in Section~\ref{sec:vmf_existing_methods}). 
Tests include the analysis of the MML estimates of the concentration parameter: 
$\kappa_{MN}$ is the approximation of MML estimate using Newton's method and
$\kappa_{MH}$ is the approximation using Halley's method
(see Equations \eqref{eqn:mml_newton_approx} and \eqref{eqn:mml_halley_approx})
against the traditionally used approximations.
Estimation of the vMF mean direction is the same across all these methods.
Estimation of $\kappa$, however, differs and hence, the corresponding results
are presented. Through these experiments, we demonstrate that the MML
estimates perform better than its competitors. 
These are followed by experiments demonstrating how these estimates
aid in the inference of vMF mixtures. These experiments illustrate the 
application of the proposed search method to
infer vMF mixtures using empirical studies and on real world datasets.

\subsection{MML-based parameter estimation for a vMF distribution}
For different values of dimensionality $d$ and concentration parameter $\kappa$,
data of sample size $N$ are randomly generated 
from a vMF distribution using the algorithm proposed by \cite{wood1994simulation}.
The parameters of a vMF distribution are estimated using the
previously mentioned approximations. Let $\hat{\kappa} = \{\kappa_T,\kappa_N,
\kappa_H,\kappa_{MN},\kappa_{MH}\}$ denote the estimate of $\kappa$ due to the 
respective methods. \\

\begin{figure}[htb]
  \centering
  \subfloat[Tanabe estimate ($\kappa_T$)]
  {
    \includegraphics[width=0.33\textwidth]{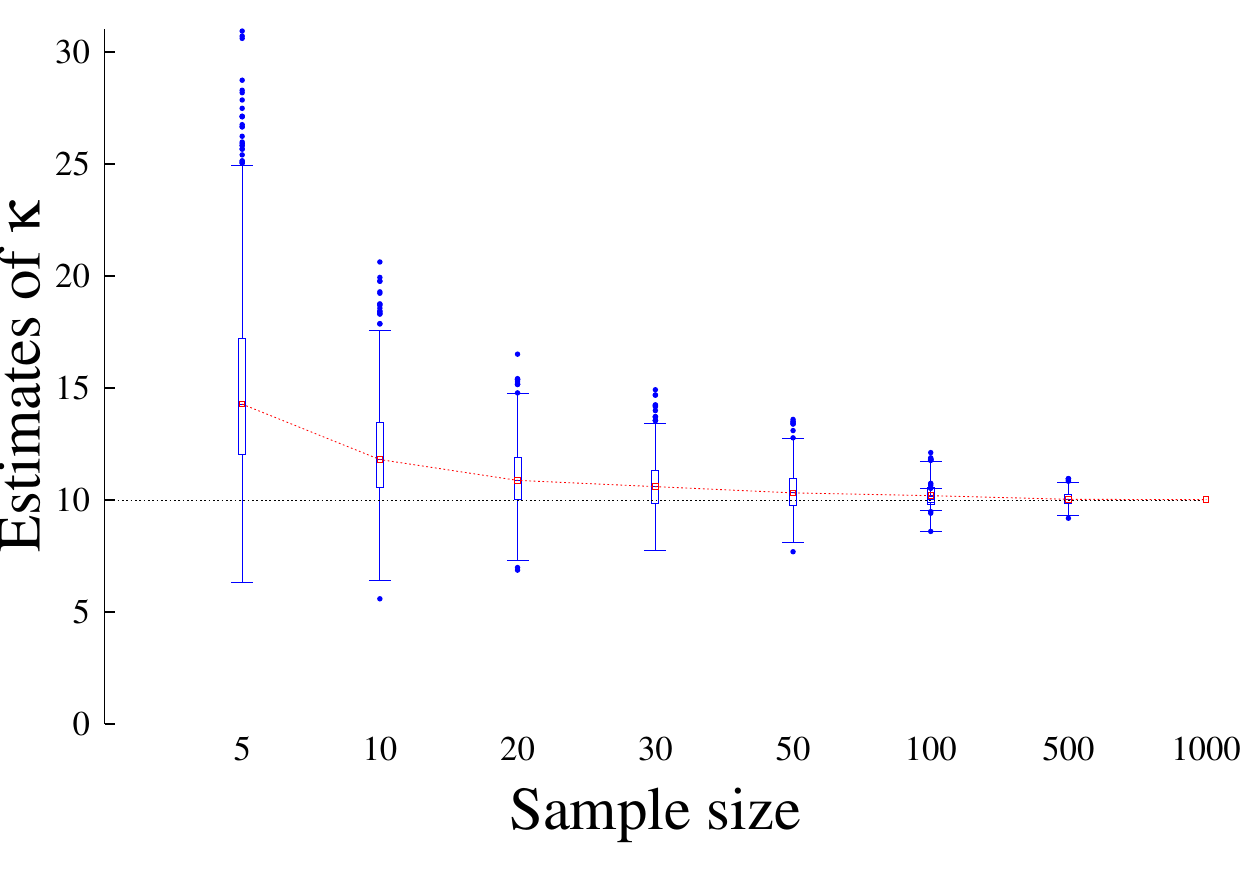}
  }
  \subfloat[Sra estimate ($\kappa_N$)]
  {
    \includegraphics[width=0.33\textwidth]{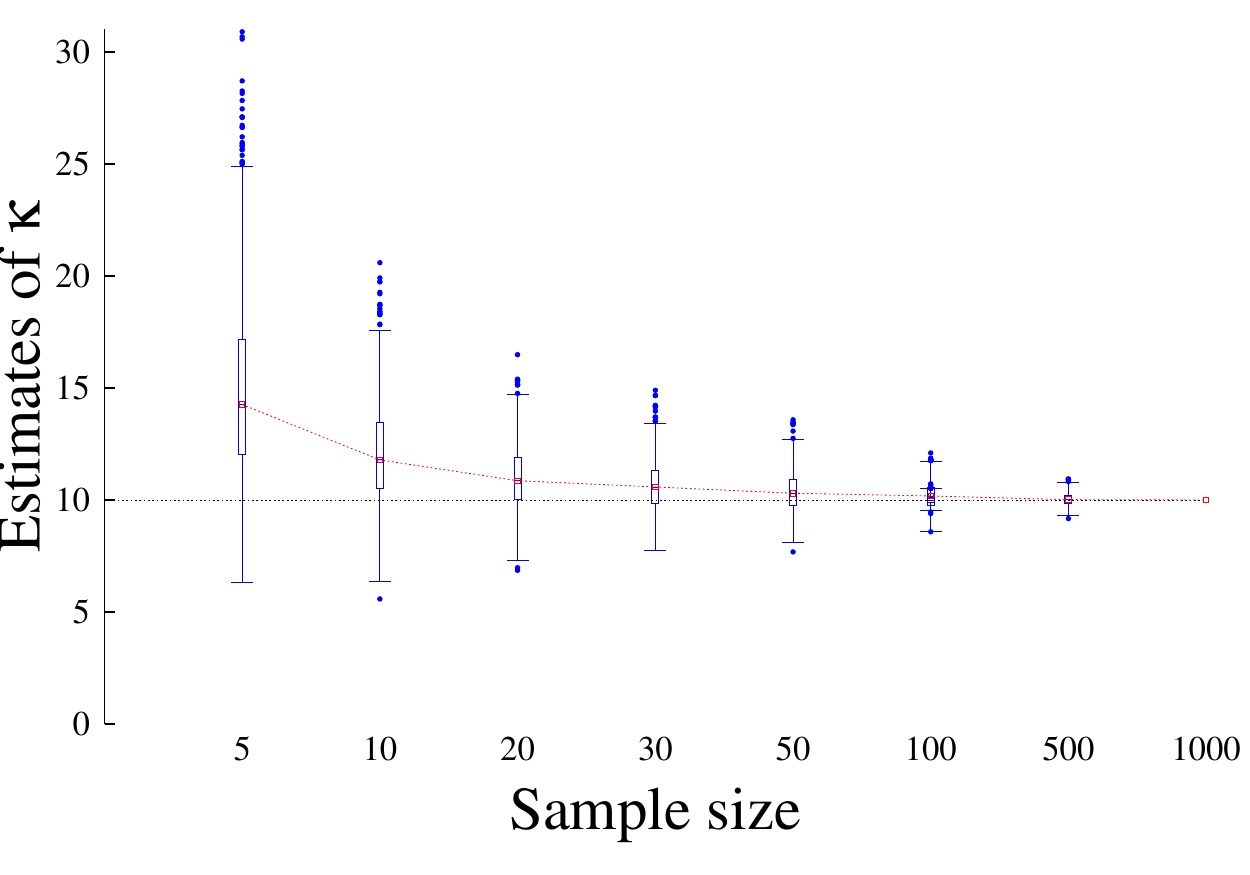}
  }
  \subfloat[Song estimate ($\kappa_H$)]
  {
    \includegraphics[width=0.33\textwidth]{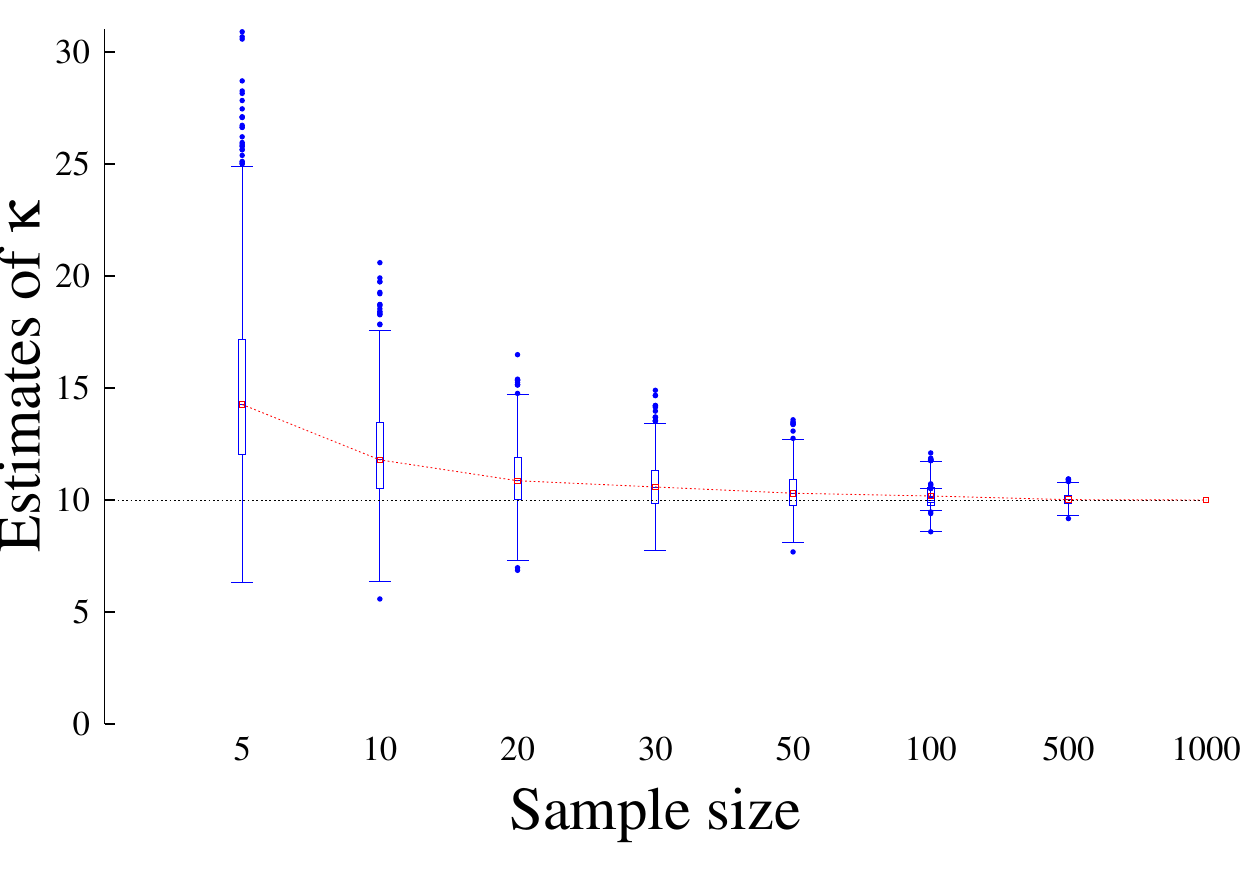}
  }\\
  \subfloat[MML (Newton) ($\kappa_{MN}$)]
  {
    \includegraphics[width=0.33\textwidth]{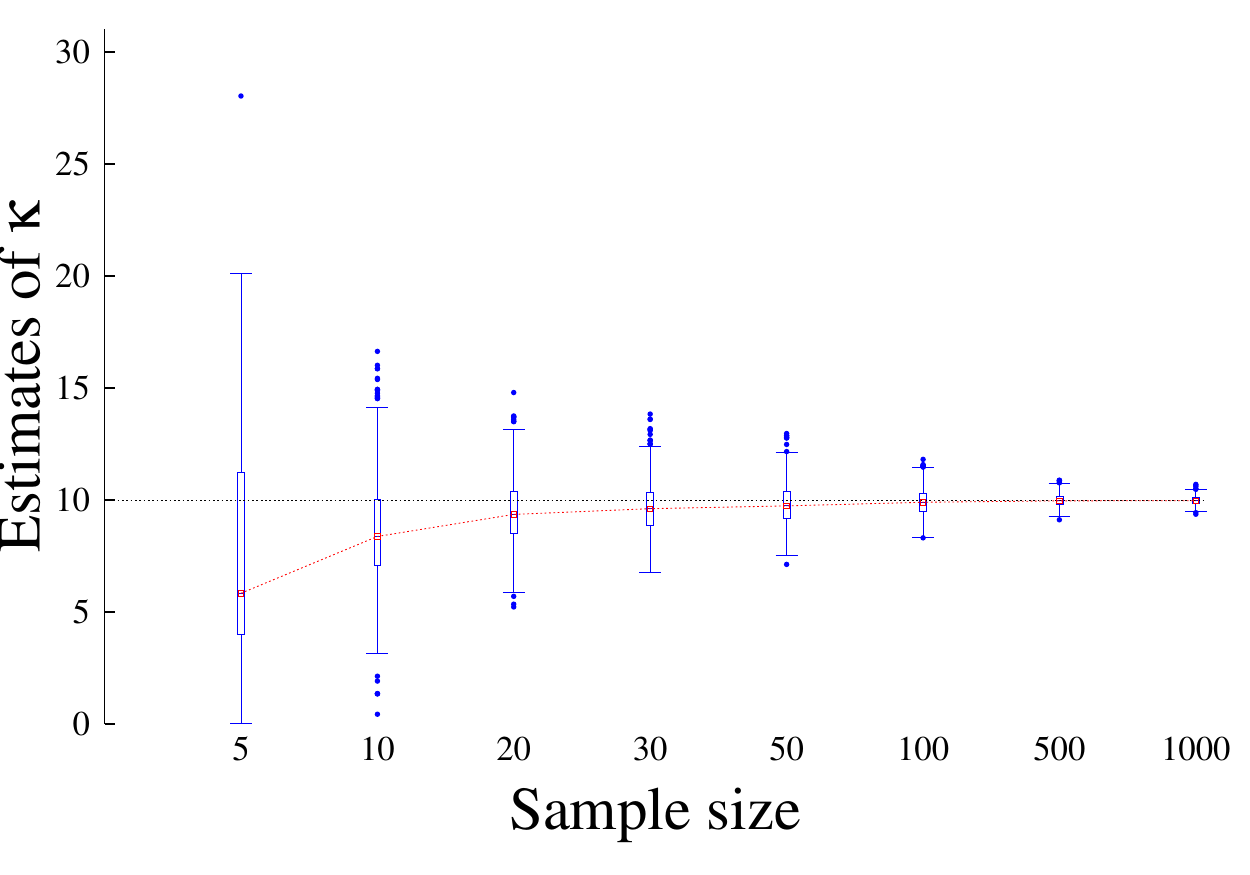}
  } 
  \subfloat[MML (Halley) ($\kappa_{MH}$)]
  {
    \includegraphics[width=0.33\textwidth]{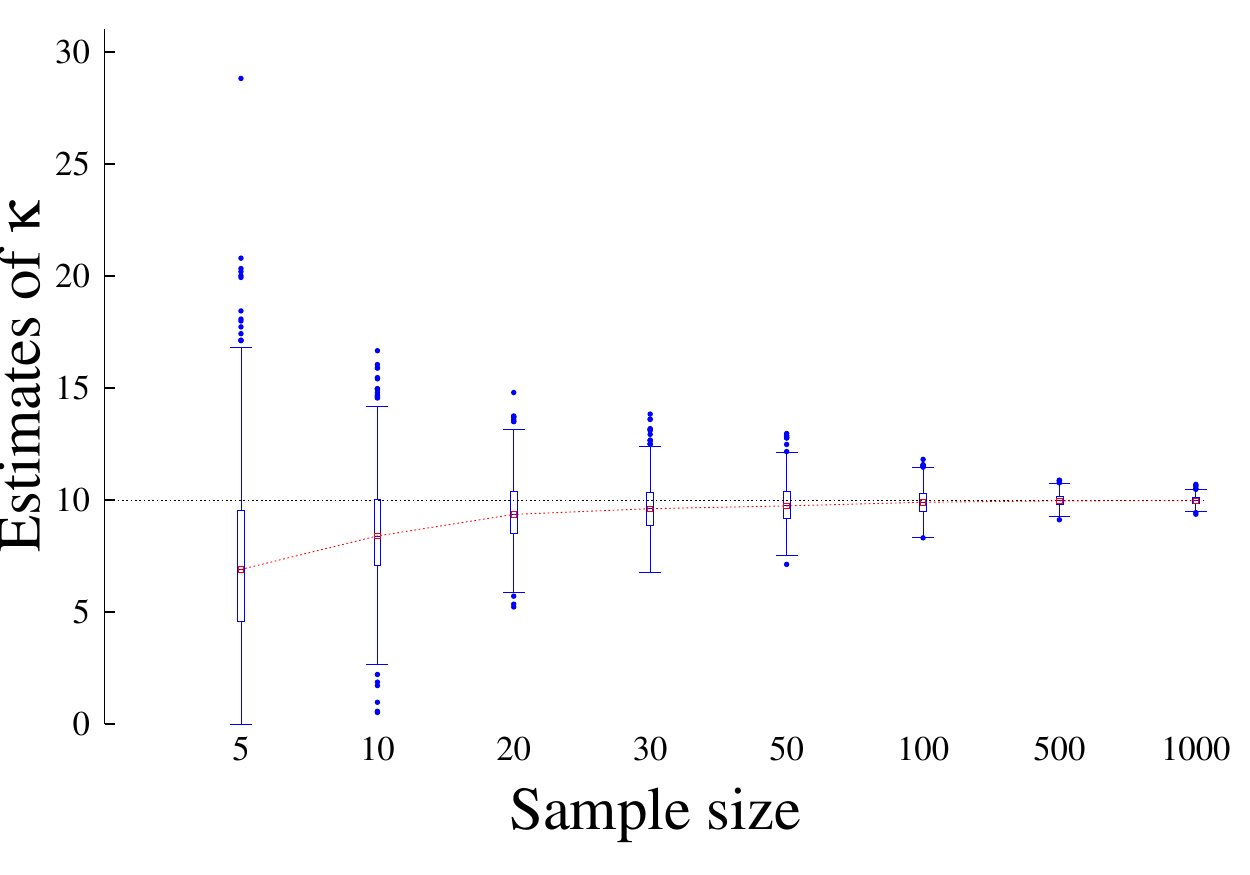}
  } 
  \caption{Box-whisker plots illustrating the $\kappa$ estimates as the sample size
           is gradually increased. True distribution is a 10-dimensional vMF with 
           $\kappa = 10$. 
           The plots are also indicative of the bias due to the estimates. 
          }
  \label{fig:bias_comparison}
\end{figure}

\noindent\emph{Errors in $\kappa$ estimation:}
We first report the errors in $\kappa$ estimation by calculating 
the absolute error $|\hat{\kappa} - \kappa|$ and the squared error 
$(\hat{\kappa} - \kappa)^2$ averaged over 1000 simulations.
The relative error $\frac{|\hat{\kappa} - \kappa|}{\kappa}$ can be 
used to measure the percentage error in $\kappa$ estimation.
The following observations are
made based on the results shown in Table~\ref{tab:kappas_errors}.
\begin{table}[htb]
  \caption{Errors in $\kappa$ estimation. 
           The averages are reported over 1000 simulations for each $(N,d,\kappa)$ triple.
           } 
  \centering
  \begin{tabular}{|l|c|c|c|c|c||c|c|c|c|c|}
  \hline
  \multirow{3}{*}{$(N, d,\kappa)$} & \multicolumn{5}{c||}{Mean absolute error}
                                  & \multicolumn{5}{c|}{Mean squared error} \\ \cline{2-11}
              &   Tanabe      &    Sra        &     Song      & \multicolumn{2}{c||}{MML} &      Tanabe    &     Sra       &     Song      &   \multicolumn{2}{c|}{MML}  \\ \cline{5-6}\cline{10-11}
              & $\kappa_T$& $\kappa_N$& $\kappa_H$& $\kappa_{MN}$& $\kappa_{MH}$& $\kappa_T$& $\kappa_N$& $\kappa_H$& $\kappa_{MN}$& $\kappa_{MH}$\\
  \hline
     10,10,10       & 2.501e+0	&	2.486e+0	&	2.486e+0	&	\textbf{2.008e+0}	&	2.012e+0	          &	1.009e+1	&	9.984e+0	&	9.984e+0	&	\textbf{5.811e+0}	&	5.850e+0	\\
     10,10,100      & 1.879e+1	&	1.877e+1	&	1.877e+1	&	\textbf{1.316e+1}	&	\textbf{1.316e+1}	  &	5.930e+2	&	5.920e+2	&	5.920e+2	&	\textbf{2.800e+2}	&	2.802e+2	\\
     10,10,1000     & 1.838e+2	&	1.838e+2	&	1.838e+2	&	\textbf{1.289e+2}	&	\textbf{1.289e+2}	  &	5.688e+4	&	5.687e+4	&	5.687e+4	&	\textbf{2.721e+4}	&	2.724e+4	\\	
     10,100,10      & 2.716e+1	&	2.716e+1	&	2.716e+1	&	2.708e+1	&	\textbf{1.728e+1}	          &	7.464e+2	&	7.464e+2	&	7.464e+2	&	7.414e+2	&	\textbf{4.102e+2}	\\
     10,100,100     & 2.014e+1	&	2.014e+1	&	2.014e+1	&	1.274e+1	&	\textbf{1.265e+1}	          &	4.543e+2	&	4.543e+2	&	4.543e+2	&	2.069e+2	&	\textbf{2.049e+2}	\\
     10,100,1000    & 1.215e+2	&	1.215e+2	&	1.215e+2	&	3.873e+1	&	\textbf{3.870e+1}	          &	1.760e+4	&	1.760e+4	&	1.760e+4	&	2.338e+3	&	\textbf{2.337e+3}	\\	
     10,1000,10     & 3.415e+2	&	3.415e+2	&	3.415e+2	&	3.415e+2	&	\textbf{1.386e+2}	          &	1.167e+5	&	1.167e+5	&	1.167e+5	&	1.167e+5	&	\textbf{2.220e+4}	\\
     10,1000,100    & 2.702e+2	&	2.702e+2	&	2.702e+2	&	2.702e+2	&	\textbf{1.652e+2}	          &	7.309e+4	&	7.309e+4	&	7.309e+4	&	7.309e+4	&	\textbf{3.101e+4}	\\
     10,1000,1000   & 1.991e+2	&	1.991e+2	&	1.991e+2	&	1.232e+2	&	\textbf{1.222e+2}	          &	4.014e+4	&	4.014e+4	&	4.014e+4	&	1.570e+4	&	\textbf{1.547e+4}	\\ \hline
     100,10,10      & 5.092e-1	&	5.047e-1	&	5.047e-1	&	\textbf{4.906e-1}	&	\textbf{4.906e-1}   &	4.097e-1	&	4.022e-1	&	4.022e-1	&	\textbf{3.717e-1}	&	\textbf{3.717e-1} \\
     100,10,100     & 3.921e+0	&	3.915e+0	&	3.915e+0	&	\textbf{3.813e+0}	&	\textbf{3.813e+0}	  &	2.457e+1	&	2.450e+1	&	2.450e+1	&	\textbf{2.278e+1}	&	\textbf{2.278e+1}		\\
     100,10,1000    & 3.748e+1	&	3.747e+1	&	3.747e+1	&	\textbf{3.669e+1}	&	\textbf{3.669e+1}	  &	2.320e+3	&	2.319e+3	&	2.319e+3	&	\textbf{2.174e+3}	&	\textbf{2.174e+3}	\\	
     100,100,10     & 4.223e+0	&	4.223e+0	&	4.223e+0	&	3.674e+0	&	\textbf{3.414e+0}	          &	1.862e+1	&	1.862e+1	&	1.862e+1	&	\textbf{1.403e+1}	&	1.420e+1	\\	
     100,100,100    & 2.187e+0	&	2.186e+0	&	2.186e+0	&	\textbf{1.683e+0}	&	\textbf{1.683e+0}	  &	7.071e+0	&	7.067e+0	&	7.067e+0	&	\textbf{4.395e+0}	&	\textbf{4.395e+0}	\\	
     100,100,1000   & 1.447e+1	&	1.447e+1	&	1.447e+1	&	\textbf{1.129e+1}	&	\textbf{1.129e+1}	  &	3.226e+2	&	3.226e+2	&	3.226e+2	&	\textbf{2.027e+2}	&	\textbf{2.027e+2}	\\
     100,1000,10    & 9.150e+1	&	9.150e+1	&	9.150e+1	&	9.146e+1	&	\textbf{8.251e+1}	          &	8.377e+3	&	8.377e+3	&	8.377e+3	&	8.370e+3	&	\textbf{6.970e+3}		\\
     100,1000,100   & 4.299e+1	&	4.299e+1	&	4.299e+1	&	4.882e+1	&	\textbf{4.080e+1}	          &	1.856e+3	&	1.856e+3	&	1.856e+3	&	2.659e+3	&	\textbf{1.738e+3}		\\
     100,1000,1000  & 1.833e+1	&	1.833e+1	&	1.833e+1	&	\textbf{8.821e+0}	&	\textbf{8.821e+0}	  &	3.728e+2	&	3.728e+2	&	3.728e+2	&	\textbf{1.060e+2}	&	\textbf{1.060e+2}  \\
  \hline
  \end{tabular}
  \label{tab:kappas_errors}
\end{table}
\begin{itemize}
  \item For $N=10,d=10,\kappa=10$, the average relative error of 
   $\kappa_T,\kappa_N,\kappa_H$ is $\sim 25\%$; for $\kappa_{MN},\kappa_{MH}$,
   it is $\sim 20\%$. When $N$ is increased to 100, the average relative error
   of $\kappa_T$ is $5.09\%$, $\kappa_N,\kappa_H$ is $5.05\%$, and
   $\kappa_{MN},\kappa_{MH}$ is $4.9\%$.
   We note that increasing $N$ while holding $d$ and $\kappa$ 
   reduces the error rate across all estimation methods and for all
   tested combinations of $d,\kappa$. This is expected because as more data
   becomes available, the inference becomes more accurate.
   The plots shown in Figure~\ref{fig:bias_comparison} reflect this behaviour.
   The mean error at lower values of $N=5,10,20,30$ is noticeable. However, as $N$
   is increased to 1000, there is a drastic drop in the error. We note that this
   behaviour is consistent across all the different estimation methods. 

  \item For fixed $N$ and $d$, increasing $\kappa$ increases the mean absolute
  error. However, the average relative error decreases. As an example, for
  $N=100,d=100,\kappa=10$, the average relative error of
  $\kappa_T,\kappa_N,\kappa_H$ is $\sim 42\%$; for $\kappa_{MN},\kappa_{MH}$, it
  is $36.7\%$ and $34\%$ respectively. When $\kappa$ is increased to 100, the
  error rate for $\kappa_T,\kappa_N,\kappa_H$ drops to $2.18\%$ and for 
  $\kappa_{MN},\kappa_{MH}$, it drops to $1.68\%$. Further increasing $\kappa$
  by an order of magnitude to 1000 results in average relative errors of
  $1.4\%$ for $\kappa_T,\kappa_N,\kappa_H$ and $1.1\%$ for $\kappa_{MN},\kappa_{MH}$.
  This indicates that as the data becomes more concentrated, the errors in 
  parameter estimation decrease.

  \item There does not appear to be a clear pattern of the variation in error
  rates when $d$ is changed keeping $N$ and $\kappa$ fixed. However, in any case,
  MML-based approximations have the least mean absolute and mean squared error.
\end{itemize}
\noindent\emph{KL-divergence and message lengths of the estimates:}
The quality of parameter inference is further determined by computing the
KL-divergence and the message lengths associated with the parameter estimates. 
The analytical expression to calculate the KL-divergence of any two vMF 
distributions is derived in the Appendix. The KL-divergence is computed
between the estimated parameters and the true vMF parameters. The minimum message
length expression for encoding data using a vMF distribution is previously derived
in Equation~\eqref{eqn:I_kappa}. Table~\ref{tab:kappas_kldiv_msglens} lists
the average values of both the metrics. The MML 
estimates of $\kappa$ result in the least value of KL-divergence across all
combinations of $N,d,\kappa$. Also, the message lengths associated with the MML
based estimates are the least. From Table~\ref{tab:kappas_kldiv_msglens}, we
notice that when $N=10$, $\kappa_{MN}$ and $\kappa_{MH}$ clearly have lower
message lengths. For $N=10,d=10,\kappa=10$, $\kappa_{MN},\kappa_{MH}$ result
in extra compression of $\sim 1.5$ bits over $\kappa_T,\kappa_N,\kappa_H$,
which makes the MML estimates $2^{1.5}$ times more likely than the others
(as per Equation~\eqref{eqn:compare_models}). \\
\begin{table}[htb]
  \caption{Comparison of the $\kappa$ estimates using KL-divergence and message length formulation (both metrics are measured in bits).} 
  \centering
  \begin{tabular}{|l|c|c|c|c|c||c|c|c|c|c|}
  \hline
  \multirow{3}{*}{$(N, d,\kappa)$} & \multicolumn{5}{c||}{Average KL-divergence}
                                  & \multicolumn{5}{c|}{Average message length} \\ \cline{2-11}
              &   Tanabe      &    Sra        &     Song      & \multicolumn{2}{c||}{MML} &      Tanabe    &     Sra       &     Song      &   \multicolumn{2}{c|}{MML}  \\ \cline{5-6}\cline{10-11}
              & $\kappa_T$& $\kappa_N$& $\kappa_H$& $\kappa_{MN}$& $\kappa_{MH}$& $\kappa_T$& $\kappa_N$& $\kappa_H$& $\kappa_{MN}$& $\kappa_{MH}$\\
  \hline
     10,10,10       & 8.777e-1	&	8.750e-1	&	8.750e-1	&	\textbf{6.428e-1}	&	6.445e-1	          &	9.285e+2	&	9.285e+2	&	9.285e+2	&	\textbf{9.269e+2}	&	\textbf{9.269e+2}	\\
     10,10,100      & 8.803e-1	&	8.798e-1	&	8.798e-1	&	\textbf{7.196e-1} &	7.199e-1	          &	8.214e+2	&	8.214e+2	&	8.214e+2	&	\textbf{8.208e+2}	&	\textbf{8.208e+2}	\\	
     10,10,1000     & 9.006e-1	&	9.005e-1	&	9.005e-1	&	\textbf{7.443e-1}	&	7.446e-1	          &	6.925e+2	&	6.925e+2	&	6.925e+2	&	\textbf{6.919e+2}	&	\textbf{6.919e+2}	\\	
     10,100,10      & 8.517e+0	&	8.517e+0	&	8.517e+0	&	8.479e+0	&	\textbf{5.321e+0}	          &	8.633e+3	&	8.633e+3	&	8.633e+3	&	8.633e+3	&	\textbf{8.585e+3}	\\	
     10,100,100     & 8.444e+0	&	8.444e+0	&	8.444e+0	&	\textbf{6.007e+0}	&	6.009e+0	          &	8.428e+3	&	8.428e+3	&	8.428e+3	&	\textbf{8.414e+3}	&	\textbf{8.414e+3} \\	
     10,100,1000    & 8.472e+0	&	8.472e+0	&	8.472e+0	&	\textbf{7.118e+0}	&	7.120e+0	          &	7.274e+3	&	7.274e+3	&	7.274e+3	&	\textbf{7.269e+3}	&	\textbf{7.269e+3}	\\
     10,1000,10     & 8.433e+1	&	8.433e+1	&	8.433e+1	&	8.433e+1	&	\textbf{1.777e+1}	          &	7.030e+4	&	7.030e+4	&	7.030e+4	&	7.030e+4	&	\textbf{6.925e+4} \\
     10,1000,100    & 8.430e+1	&	8.430e+1	&	8.430e+1	&	8.430e+1	&	\textbf{4.697e+1}	          &	7.030e+4	&	7.030e+4	&	7.030e+4	&	7.030e+4	&	\textbf{6.989e+4}	\\	
     10,1000,1000   & 8.451e+1	&	8.451e+1	&	8.451e+1	&	\textbf{5.976e+1}	&	5.977e+1	          &	6.825e+4	&	6.825e+4	&	6.825e+4	&	\textbf{6.811e+4}	&	\textbf{6.811e+4}	\\ \hline
     100,10,10      & 7.409e-2	&	7.385e-2	&	7.385e-2	&	\textbf{7.173e-2}	&	\textbf{7.173e-2}	  &	\textbf{9.115e+3}	&	\textbf{9.115e+3}	&	\textbf{9.115e+3}	&	\textbf{9.115e+3}	&	\textbf{9.115e+3}	\\	
     100,10,100     & 7.539e-2	&	7.535e-2	&	7.535e-2	&	\textbf{7.411e-2}	&	\textbf{7.411e-2}	  &	\textbf{7.858e+3}	&	\textbf{7.858e+3}	&	\textbf{7.858e+3}	&	\textbf{7.858e+3}	&	\textbf{7.858e+3}	\\	
     100 10,1000    & 7.271e-2	&	7.271e-2	&	7.271e-2	&	\textbf{7.161e-2}	&	\textbf{7.161e-2}	  &	\textbf{6.403e+3}	& \textbf{6.403e+3}	& \textbf{6.403e+3}	& \textbf{6.403e+3}	& \textbf{6.403e+3}	\\
     100,100,10     & 7.270e-1	&	7.270e-1	&	7.270e-1	&	\textbf{6.146e-1}	&	6.208e-1	          &	8.615e+4	&	8.615e+4	&	8.615e+4	&	\textbf{8.614e+4}	&	\textbf{8.614e+4}	\\	
     100,100,100    & 7.357e-1	&	7.357e-1	&	7.357e-1	&	\textbf{7.117e-1}	&	\textbf{7.117e-1}	  &	\textbf{8.299e+4}	&	\textbf{8.299e+4}	&	\textbf{8.299e+4}	&	\textbf{8.299e+4}	&	\textbf{8.299e+4}\\	
     100,100,1000   & 7.330e-1	&	7.330e-1	&	7.330e-1	&	\textbf{7.210e-1}	&	\textbf{7.210e-1}	  &	\textbf{6.976e+4}	& \textbf{6.976e+4}	& \textbf{6.976e+4}	& \textbf{6.976e+4}	& \textbf{6.976e+4} \\	
     100,1000,10    & 7.324e+0	&	7.324e+0	&	7.324e+0	&	7.318e+0	&	\textbf{6.201e+0}	          &	7.024e+5	&	7.024e+5	&	7.024e+5	&	7.024e+5	&	\textbf{7.023e+5}	\\	
     100,1000,100   & 7.302e+0	&	7.302e+0	&	7.302e+0	&	\textbf{7.045e+0}	&	7.106e+0	          &	7.022e+5	&	7.022e+5	&	7.022e+5	&	\textbf{7.019e+5}	&	7.022e+5	\\	
     100,1000,1000  & 7.340e+0	&	7.340e+0	&	7.340e+0	&	\textbf{7.097e+0}	&	\textbf{7.097e+0}	  &	\textbf{6.707e+5}& \textbf{6.707e+5}& \textbf{6.707e+5}& \textbf{6.707e+5}& \textbf{6.707e+5}	\\
  \hline
  \end{tabular}
  \label{tab:kappas_kldiv_msglens}
\end{table}

\noindent\emph{Bias of the parameter estimates:}
The maximum likelihood estimate of $\kappa$ is known to have significant bias
\citep{schou1978estimation,best1981bias,cordeiro1999theory}. 
Our goal here is to demonstrate
that MML-based parameter approximations result in estimates with reduced bias. 
The mean squared error in Table~\ref{tab:kappas_errors} can be 
decomposed into the sum of bias and variance terms as
shown below \citep{taboga2012lectures}.
\begin{equation*}
\text{mean squared error} = \mathrm{E}[(\hat{\kappa}-\kappa)^2] = 
\underbrace{(\mathrm{E}[\hat{\kappa}]-\kappa)^2}_{\text{Bias}^2(\hat{\kappa})} 
+ \underbrace{\mathrm{E}[(\hat{\kappa} - \mathrm{E}[\hat{\kappa}])^2]}_{\text{Variance}(\hat{\kappa})} 
\label{eqn:mse_decomp}
\end{equation*}
where $\mathrm{E}[.]$ denotes the expectation of the related quantity.
Table~\ref{tab:kappas_bias_variance} shows the bias-variance of the
estimated concentration parameter $\hat{\kappa}$ in the above simulations.
The bias of $\kappa_{MN}$ and $\kappa_{MH}$ is lower 
compared to the other estimates. The variance of the MML estimates,
however, is not always the least, as observed in Table~\ref{tab:kappas_bias_variance}.
The combination of bias and variance, which is the mean squared error,
is empirically demonstrated to be the least for the MML estimates.\\
\begin{table}[htb]
  \caption{Bias-variance decomposition of the squared error $(\hat{\kappa} - \kappa)^2$.} 
  \centering
  \begin{tabular}{|l|c|c|c|c|c||c|c|c|c|c|}
  \hline
  \multirow{3}{*}{$(N, d,\kappa)$} & \multicolumn{5}{c||}{Bias (squared)}
                                  & \multicolumn{5}{c|}{Variance} \\ \cline{2-11}
              &   Tanabe      &    Sra        &     Song      & \multicolumn{2}{c||}{MML} &      Tanabe    &     Sra       &     Song      &   \multicolumn{2}{c|}{MML}  \\ \cline{5-6}\cline{10-11}
              & $\kappa_T$& $\kappa_N$& $\kappa_H$& $\kappa_{MN}$& $\kappa_{MH}$& $\kappa_T$& $\kappa_N$& $\kappa_H$& $\kappa_{MN}$& $\kappa_{MH}$\\
  \hline
     10,10,10       & 5.609e+0	&	5.520e+0	&	5.520e+0	&	1.299e+0	&	\textbf{1.269e+0}	          &	4.476e+0	&	\textbf{4.464e+0}	&	\textbf{4.464e+0}	&	4.512e+0	&	4.581e+0	\\
     10,10,100      & 2.298e+2	&	2.288e+2	&	2.288e+2	&	4.986e-3	&	\textbf{2.577e-4}	          &	3.632e+2	&	3.632e+2	&	3.632e+2	&	\textbf{2.800e+2}	&	2.802e+2	\\
     10,10,1000     & 2.157e+4	&	2.156e+4	&	2.156e+4	&	\textbf{2.764e+1}	&	3.193e+1	          &	3.531e+4	&	3.531e+4	&	3.531e+4	&	\textbf{2.718e+4}	&	2.720e+4	\\	
     10,100,10      & 7.378e+2	&	7.378e+2	&	7.378e+2	&	7.333e+2	&	\textbf{2.875e+2}	          &	8.660e+0	&	8.660e+0	&	8.660e+0	&	\textbf{8.066e+0}	&	1.226e+2		\\
     10,100,100     & 4.054e+2	&	4.053e+2	&	4.053e+2	&	1.546e+2	&	\textbf{1.522e+2}	          &	\textbf{4.894e+1}	&	\textbf{4.894e+1}	&	\textbf{4.894e+1}	&	5.231e+1	&	5.273e+1	\\	
     10,100,1000    & 1.473e+4	&	1.473e+4	&	1.473e+4	&	2.207e+1	&	\textbf{1.994e+1}	          &	2.870e+3	&	2.870e+3	&	2.870e+3	&	\textbf{2.316e+3}	&	2.317e+3	\\	
     10,1000,10     & 1.166e+5	&	1.166e+5	&	1.166e+5	&	1.166e+5	&	\textbf{1.921e+4}	          &	\textbf{8.090e+1}	&	\textbf{8.090e+1}	&	\textbf{8.090e+1}	&	\textbf{8.090e+1}	&	2.983e+3	\\
     10,1000,100    & 7.301e+4	&	7.301e+4	&	7.301e+4	&	7.300e+4	&	\textbf{2.728e+4}	          &	8.685e+1	&	8.685e+1	&	8.685e+1	&	\textbf{8.635e+1}	&	3.735e+3		\\
     10,1000,1000   & 3.964e+4	&	3.964e+4	&	3.964e+4	&	1.517e+4	&	\textbf{1.493e+4}	          &	\textbf{4.969e+2}	&	\textbf{4.969e+2}	&	\textbf{4.969e+2}	&	5.306e+2	&	5.342e+2		\\\hline
     100,10,10      & 4.129e-2	&	3.528e-2	&	3.528e-2	&	8.132e-3	&	\textbf{8.129e-3}           &	3.684e-1	&	3.669e-1	&	3.669e-1	&	\textbf{3.636e-1}	&	\textbf{3.636e-1}	\\	
     100,10,100     & 1.280e+0	&	1.206e+0	&	1.206e+0	&	5.505e-2	&	\textbf{5.504e-2}           &	2.329e+1	&	2.329e+1	&	2.329e+1	&	\textbf{2.273e+1}	&	\textbf{2.273e+1}		\\
     100,10,1000    & 9.796e+1	&	9.728e+1	&	9.728e+1	&	6.620e+0	&	\textbf{6.619e+0}           &	2.222e+3	&	2.222e+3	&	2.222e+3	&	2.168e+3	&	\textbf{2.168e+3}		\\
     100,100,10     & 1.783e+1	&	1.783e+1	&	1.783e+1	&	\textbf{4.661e+0}	&	6.202e+0	          &	\textbf{7.807e-1}	&	\textbf{7.807e-1}	&	\textbf{7.807e-1}	&	9.369e+0	&	8.003e+0		\\
     100,100,100    & 3.371e+0	&	3.367e+0	&	3.367e+0	&	7.147e-1	&	\textbf{7.146e-1}	          &	3.700e+0	&	3.700e+0	&	3.700e+0	&	\textbf{3.681e+0}	&	\textbf{3.681e+0}		\\
     100,100,1000   & 1.161e+2	&	1.161e+2	&	1.161e+2	&	\textbf{3.504e-1}	&	\textbf{3.504e-1}	  &	2.065e+2	&	2.065e+2	&	2.065e+2	&	\textbf{2.023e+2}	&	\textbf{2.023e+2}		\\
     100,1000,10    & 8.372e+3	&	8.372e+3	&	8.372e+3	&	8.364e+3	&	\textbf{6.809e+3}	          &	5.385e+0	&	5.385e+0	&	5.385e+0	&	\textbf{5.200e+0}	&	1.614e+2		\\
     100,1000,100   & 1.848e+3	&	1.848e+3	&	1.848e+3	&	\textbf{5.143e+2}	&	1.628e+3	          &	\textbf{7.656e+0}	&	\textbf{7.656e+0}	&	\textbf{7.656e+0}	&	2.145e+3	&	1.099e+2 \\
     100,1000,1000  & 3.359e+2	&	3.359e+2	&	3.359e+2	&	6.926e+1	&	\textbf{6.925e+1}	          &	3.692e+1	&	3.692e+1	&	3.692e+1	&	\textbf{3.674e+1}	&	\textbf{3.674e+1}		\\
  \hline
  \end{tabular}
  \label{tab:kappas_bias_variance}
\end{table}

\noindent\emph{Statistical hypothesis testing:}
There have been several goodness-of-fit methods proposed in the literature 
to test the null hypothesis of a vMF distribution against some alternative
hypothesis \citep{kent1982fisher,mardia1984goodness,mardia-book}. Recently, 
\cite{figueiredo2012goodness} suggested tests for the specific case of 
concentrated vMF distributions. Here, we examine the behaviour of $\kappa$
estimates for generic vMF distributions as proposed by \cite{mardia1984goodness}.
They derived a likelihood ratio test for the null hypothesis
of a vMF distribution ($H_0$) against the alternative of a Fisher-Bingham distribution ($H_a$).
The asymptotically equivalent Rao's score statistic \citep{rao73}
was used to test the hypothesis. 

The score statistic $\mathcal{W}$, in this case, is a  function of the
concentration parameter. It has an asymptotic $\chi^2(p)$ distribution 
(with degrees of freedom $p = \frac{1}{2}d(d+1) - 1$)
under $H_0$ as the sample size $N\to\infty$. 
For $d = \{10,100,1000\}$,
the critical values at 5\% significance level are given in Table~\ref{tab:hypothesis_test_1}.  
If the computed test statistic exceeds the critical value, then the null
hypothesis of a vMF distribution is rejected. 
We conduct a simulation study where we generate random samples of 
size $N=1$ million from a vMF distribution with known
mean and $\kappa=\{10,100,1000\}$. For each inferred estimate $\hat{\kappa}$,
we compute the test statistic and compare it with the corresponding critical value.
The results are shown in Table~\ref{tab:hypothesis_test_1}. For $d=10$, the approximation
$\kappa_T$ has a significant effect as its test statistic exceeds the critical value
and consequently the p-value is close to zero. This implies that the null hypothesis
of a vMF distribution is rejected by using the estimate $\kappa_T$. However, this is incorrect as the data
was generated from a vMF distribution. The p-values due to the estimates 
$\kappa_N,\kappa_H,\kappa_{MN},\kappa_{MH}$ are all greater than 0.05 
(the significance level) which implies that the null hypothesis is accepted.
For $d=\{100,1000\}$, the p-values corresponding to the different 
estimates are greater than 0.05. In these cases, the use of
all the estimates lead to the same conclusion
of accepting the null hypothesis of a vMF distribution.
As the amount of data increases, the error due to all the estimates
decreases. This is further exemplified below.
\begin{table}[htb]
  \caption{Goodness-of-fit tests for the null hypothesis $H_0:$ vMF distribution
           and the alternate hypothesis $H_a:$ Fisher-Bingham distribution.
           Critical values of the test statistic correspond to a significance of 5\%. 
          }
  \centering
  \begin{tabular}{|l|c|c|c|c|c|c||c|c|c|c|c|}
  \hline
  \multirow{3}{*}{$(d,\kappa)$} & Critical & \multicolumn{5}{c||}{Test statistic} & \multicolumn{5}{c|}{p-value of the test} \\ \cline{3-12}
     & value  &   Tanabe      &    Sra        &     Song      & \multicolumn{2}{c||}{MML} &      Tanabe    &     Sra       &     Song      &   \multicolumn{2}{c|}{MML}  \\ \cline{6-7}\cline{11-12}
              & $\chi^2(p)$ & $\kappa_T$& $\kappa_N$& $\kappa_H$& $\kappa_{MN}$& $\kappa_{MH}$& $\kappa_T$& $\kappa_N$& $\kappa_H$& $\kappa_{MN}$& $\kappa_{MH}$\\
  \hline
     10,10      & 7.215e+1 & 1.850e+2	&	5.353e+1	&	5.353e+1	&	5.353e+1	&	5.353e+1	&	0.000e+0	&	5.258e-1	&	5.258e-1	&	5.260e-1	&	5.260e-1	\\	
     10,100     & 7.215e+1 & 1.698e+3	&	4.949e+1	&	4.949e+1	&	4.945e+1	&	4.945e+1	&	0.000e+0	&	6.247e-1	&	6.247e-1	&	6.267e-1	&	6.267e-1	\\	
     10,1000    & 7.215e+1 & 1.950e+3	&	4.811e+1	&	4.811e+1	&	5.060e+1	&	5.060e+1	&	0.000e+0	&	6.571e-1	&	6.571e-1	&	5.724e-1	&	5.724e-1	\\	
     100,10     & 5.215e+3 & 5.090e+3	&	5.090e+3	&	5.090e+3	&	5.090e+3	&	5.090e+3	&	3.739e-1	&	3.739e-1	&	3.739e-1	&	3.741e-1	&	3.741e-1	\\	
     100,100    & 5.215e+3 & 5.010e+3	&	5.010e+3	&	5.010e+3	&	5.010e+3	&	5.010e+3	&	6.103e-1	&	6.127e-1	&	6.127e-1	&	6.125e-1	&	6.125e-1	\\	
     100,1000   & 5.215e+3 & 5.025e+3	&	5.018e+3	&	5.018e+3	&	5.022e+3	&	5.022e+3	&	5.427e-1	&	5.597e-1	&	5.597e-1	&	5.517e-1	&	5.517e-1	\\	
     1000,10    & 5.021e+5 & 5.006e+5	&	5.006e+5	&	5.006e+5	&	5.006e+5	&	5.006e+5	&	4.682e-1	&	4.682e-1	&	4.682e-1	&	4.687e-1	&	4.687e-1	\\
     1000,100   & 5.021e+5 & 5.005e+5	&	5.005e+5	&	5.005e+5	&	5.005e+5	&	5.005e+5	&	5.050e-1	&	5.050e-1	&	5.050e-1	&	5.057e-1	&	5.057e-1	\\	
     1000,1000  & 5.021e+5 & 5.007e+5	&	5.007e+5	&	5.007e+5	&	5.007e+5	&	5.007e+5	&	4.283e-1	&	4.283e-1	&	4.283e-1	&	4.196e-1	&	4.196e-1	\\	
  \hline
  \end{tabular}
  \label{tab:hypothesis_test_1}
\end{table}

\begin{figure}[htb]
  \centering
  \subfloat[Variation in error with increasing $\kappa$]
  {
    \includegraphics[width=0.5\textwidth]{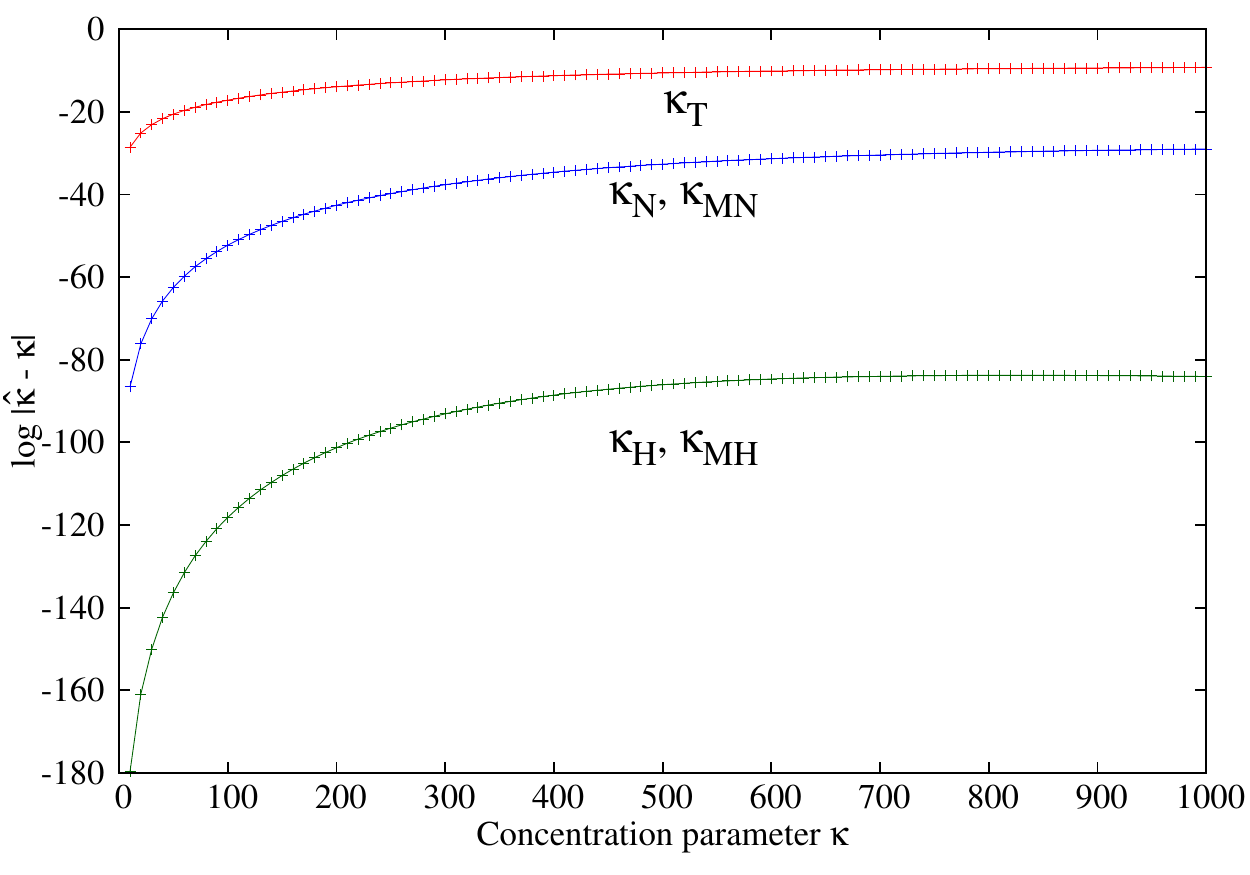}
  }
  \subfloat[Variation in error with increasing $\bar{R}$]
  {
    \includegraphics[width=0.5\textwidth]{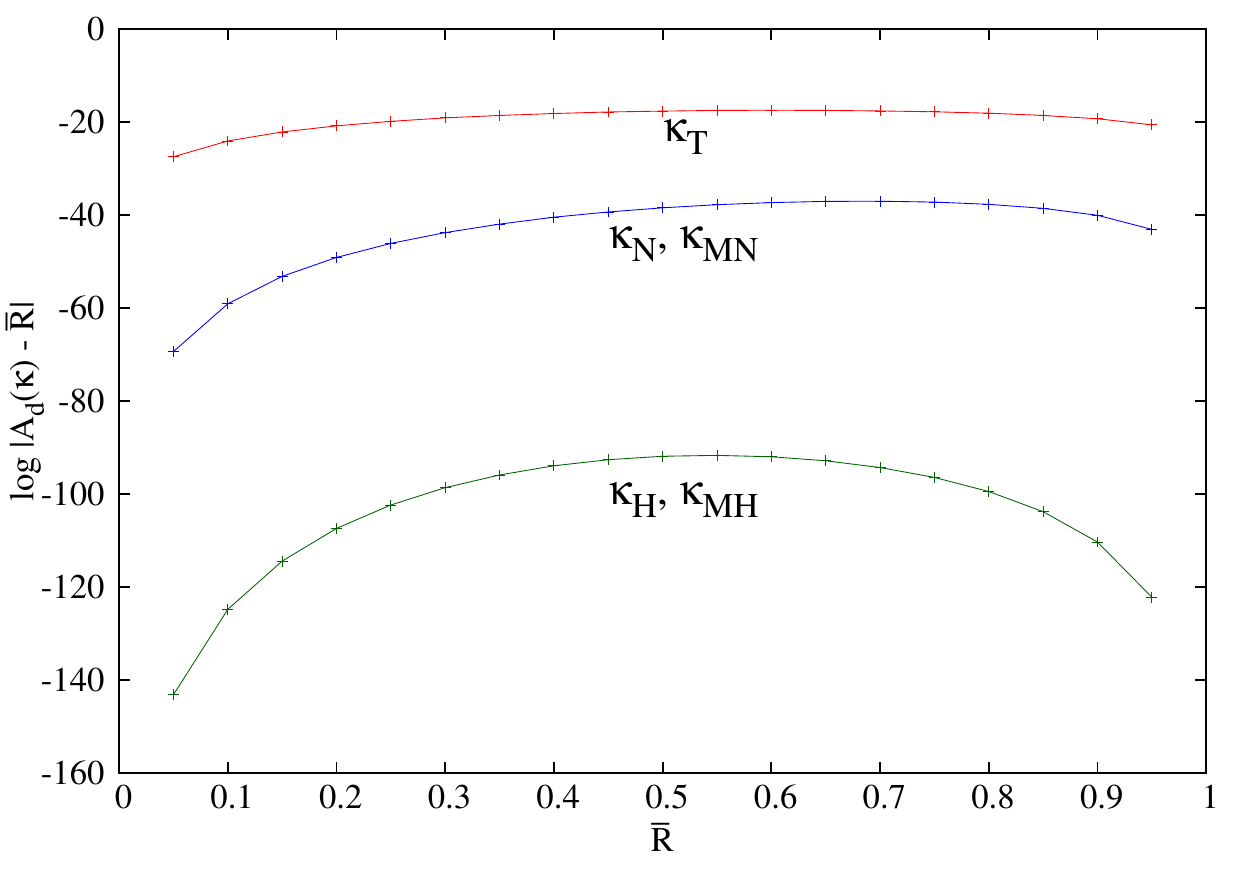}
  } 
  \caption{Errors in $\kappa$ estimation for $d = 1000$ as the sample size $N \to\infty$.}
  \label{fig:asymptotic_fixed_d}
\end{figure}

\noindent\emph{Asymptotic behaviour of MML estimates:}
Based on the empirical tests, we have so far seen that MML estimates fare better 
when compared to the other approximations.
We now discuss the behaviour of the MML estimates in the limiting case.
For large sample sizes ($N\to\infty$), we plot the errors in $\kappa$ estimation.
\cite{song2012high} demonstrated that their approximation $\kappa_H$ results in
the lowest error in the limiting case.
We compute the variation in error in two scenarios with fixed $d = 1000$ and:
\begin{enumerate}
  \item \emph{increasing $\kappa$}: Fig.~\ref{fig:asymptotic_fixed_d}(a)
  illustrates the behaviour of the absolute error with increasing $\kappa$.
  The first observation is that irrespective of the estimation procedure,
  the error continues to increase with increasing $\kappa$ values (which
  corroborates our observations in the empirical tests) and then saturates.
  According to \cite{song2012high}, their estimate $\kappa_H$ produces
  the lowest error which we can see in the figure. Further, 
  our MML Newton approximation $\kappa_{MN}$ actually performs worse
  than Song's approximation $\kappa_H$. However, we note that the errors due to MML Halley's
  approximation $\kappa_{MH}$ are identical to those produced by $\kappa_H$.
  This suggests that asymptotically, the approximations achieved by $\kappa_H$
  and $\kappa_{MH}$ are more accurate (note that the errors in the limiting case
  are extremely low).

  \item \emph{increasing $\bar{R}$}:  The maximum likelihood estimate of $\kappa$ aims
  to achieve $F(\hat{\kappa}) \equiv A_d(\hat{\kappa}) - \bar{R} = 0$ (Equation~\ref{eqn:ml_estimates}).
  Hence, $\log|A_d(\kappa)-\bar{R}|$ gives a measure of the error corresponding to an estimate of $\kappa$.
  Figure~\ref{fig:asymptotic_fixed_d}(b) depicts the variation of this error with increasing $\bar{R}$ .
  We observe that $\kappa_H$ and $\kappa_{MH}$ produce the least error. We also note that
  the error produced due to $\kappa_{MN}$ is greater than that produced by $\kappa_{MH}$.
  However, we highlight the fact that MML-based parameter inference aims to achieve $G(\hat{\kappa}) \equiv 0$ 
  (Equation~\ref{eqn:I_first_derivative}), a fundamentally different objective function compared
  to the maximum likelihood based one. 
\end{enumerate}
The asymptotic results are shown here by assuming a value of $N=10^{200}$
(note the corresponding extremely low error rates).
In the limiting case, the MML estimate $\kappa_{MH}$ coincides with the ML estimate $\kappa_H$.
However, $\kappa_H$'s performance is better compared to the MML
Newton's approximation $\kappa_{MN}$. The same behaviour is observed 
for when $\kappa$ is fixed and the dimensionality is increased.
For \emph{enormous} amount of data, the ML approximations converge
to the MML ones.

\subsection{MML-based inference of mixtures of vMF distributions}
\begin{figure}[htb]
  \centering
  \subfloat[$\kappa_1 = 10$, $\kappa_2 = 100$]
  {
    \includegraphics[width=0.50\textwidth]{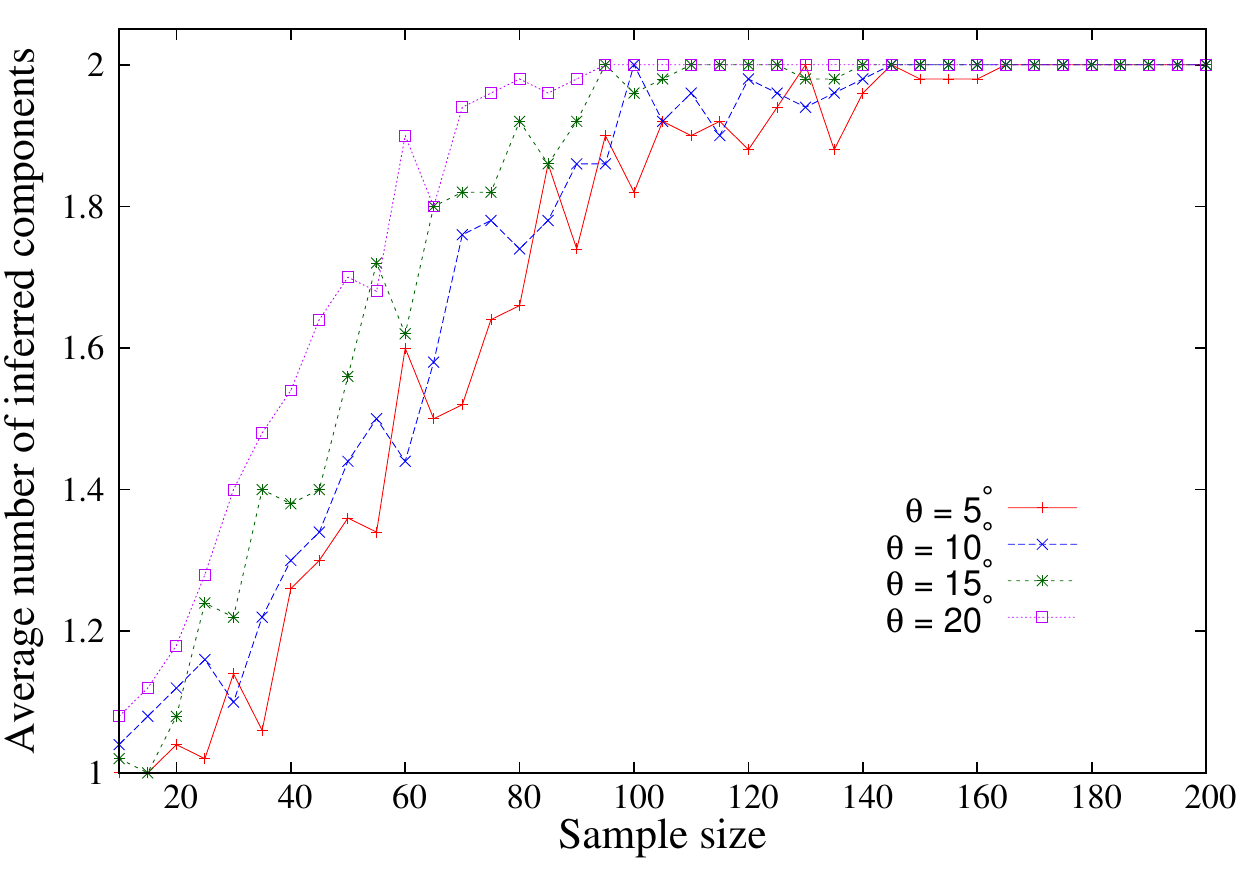}
  }
  \subfloat[$\kappa_1=\kappa_2=100$]
  {
    \includegraphics[width=0.50\textwidth]{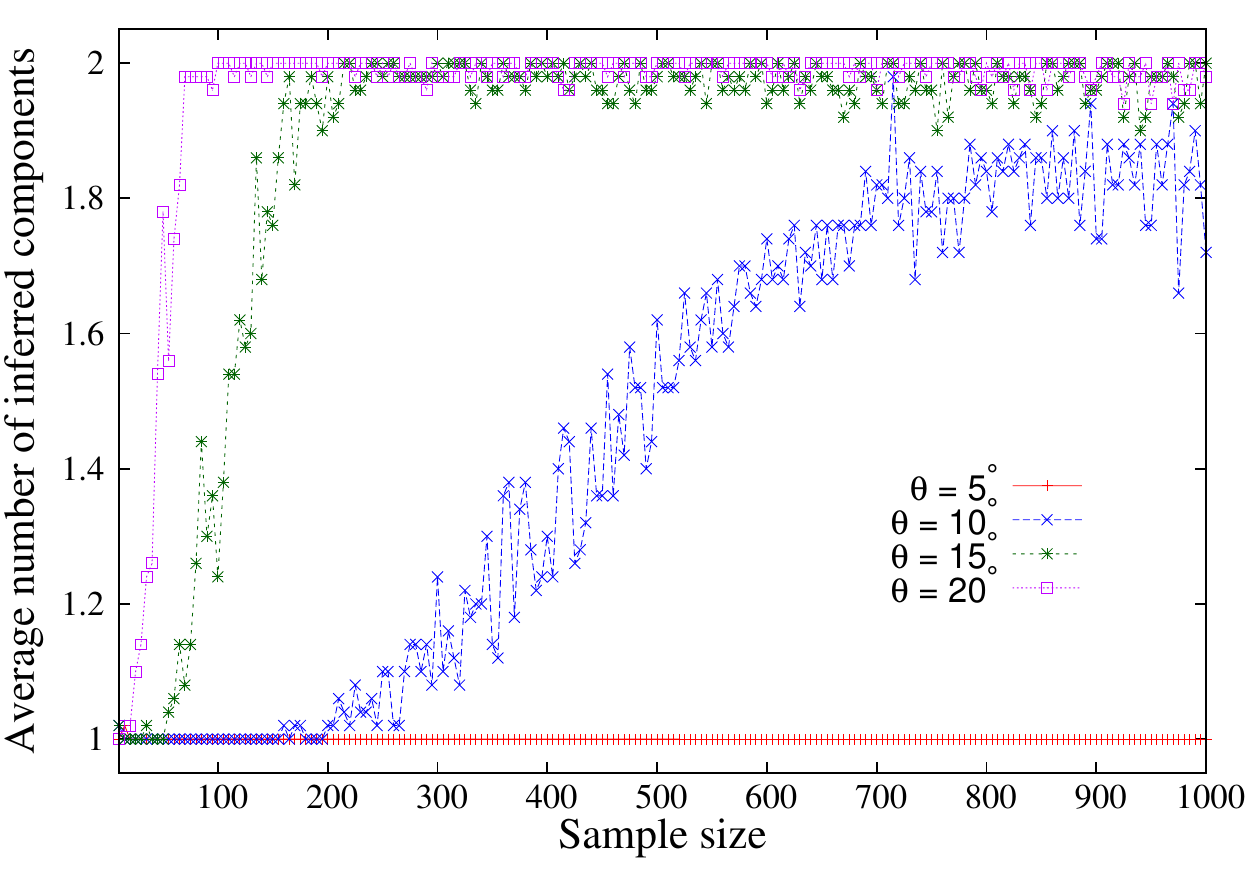}
  } 
  \caption{Case 1: Average number of components inferred for the two-component mixture 
           whose means are separated by $\theta^{\circ}$.
          }
  \label{fig:diff_means}
\end{figure}

An empirical study is carried out where the proposed search method is employed
to infer a mixture using data sampled from a known
mixture distribution. The amount of data is gradually increased; 
for each sample size $N$, the simulation is repeated 50 times
and the number of inferred components is plotted (we used MML Halley's
approximation in all the discussed results).
The various case studies are discussed below.
\begin{enumerate}
\item The true components in the mixture have \emph{different} mean
directions (separated by an angle $\theta$).
\item The true components in the mixture have the \emph{same} mean
direction but different concentration parameters. 
\end{enumerate}
\noindent\emph{Case 1:}
The true mixture has two components with equal mixing proportions.
We consider the case when $d=3$. 
The mean of one of the vMF components is aligned with the Z-axis. 
The mean of the other component is chosen such that the angle between
the two means is $\theta^{\circ}$. 
Fig.~\ref{fig:diff_means}(a) illustrates the scenario when 
the concentration parameters of the constituent components are different.
Fig.~\ref{fig:diff_means}(b) shows the variation in the number
of inferred components when the true vMF components
have the same concentration parameter.
In both scenarios, as the angular separation is increased, the components become
more distinguishable and hence, less amount of data is required to
correctly identify them.

When the concentration parameters of the constituent components are different,
the inference of the mixture is relatively easier compared to the case when the
concentration parameters are same.
In Fig.~\ref{fig:diff_means}(a), for all angular separations,
the true number of components is correctly inferred at a sample
size of $N=200$.
When $\theta=20^{\circ}$, the search method converges faster at $N\sim 100$ as compared 
to $\theta=5^{\circ}$, when the convergence is at $N\sim180$.
In Fig.~\ref{fig:diff_means}(b), when $\theta=5^{\circ}$,
the search method infers only one-component as the true mixture components are 
hardly distinguishable. When $\theta=10^{\circ}$, even at $N\sim1000$, the
average number of inferred components is $\sim1.8$. When $\theta=15^{\circ}$,
the search method converges at $N\sim300$ as compared to $N\sim120$ in Fig.~\ref{fig:diff_means}(a).
Clearly, when the component means are different, it is easier to correctly infer
mixtures whose components have different concentration parameters. 

\noindent\emph{Case 2:}
In this case, multivariate ($d=\{2,3,10\}$) vMF mixtures with equal mixing proportions 
and same component means are considered. The simulation results of true mixtures with
two and three components are presented here.

\begin{figure}[H]
  \centering
  \subfloat[2-component mixture]
  {
    \includegraphics[width=0.50\textwidth]{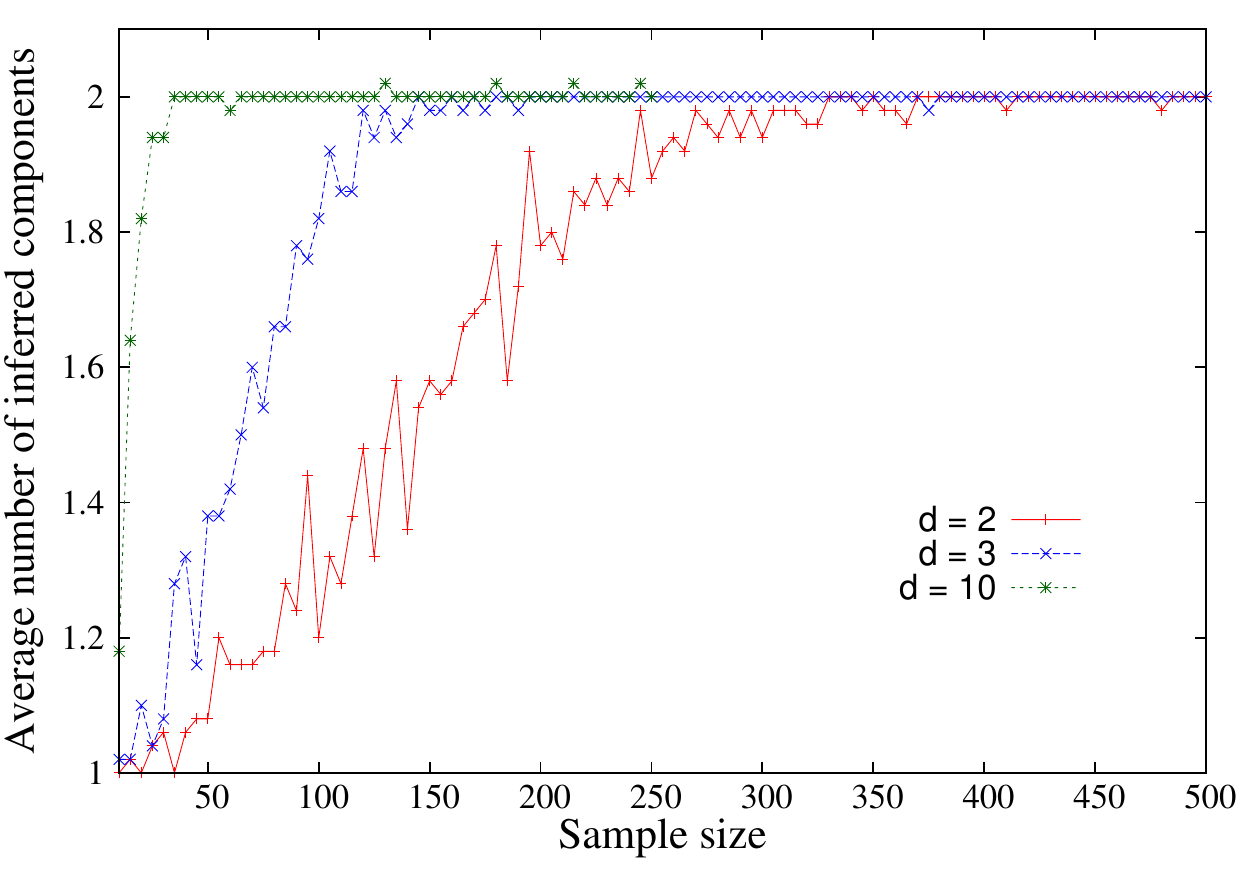}
  }
  \subfloat[3-component mixture]
  {
    \includegraphics[width=0.50\textwidth]{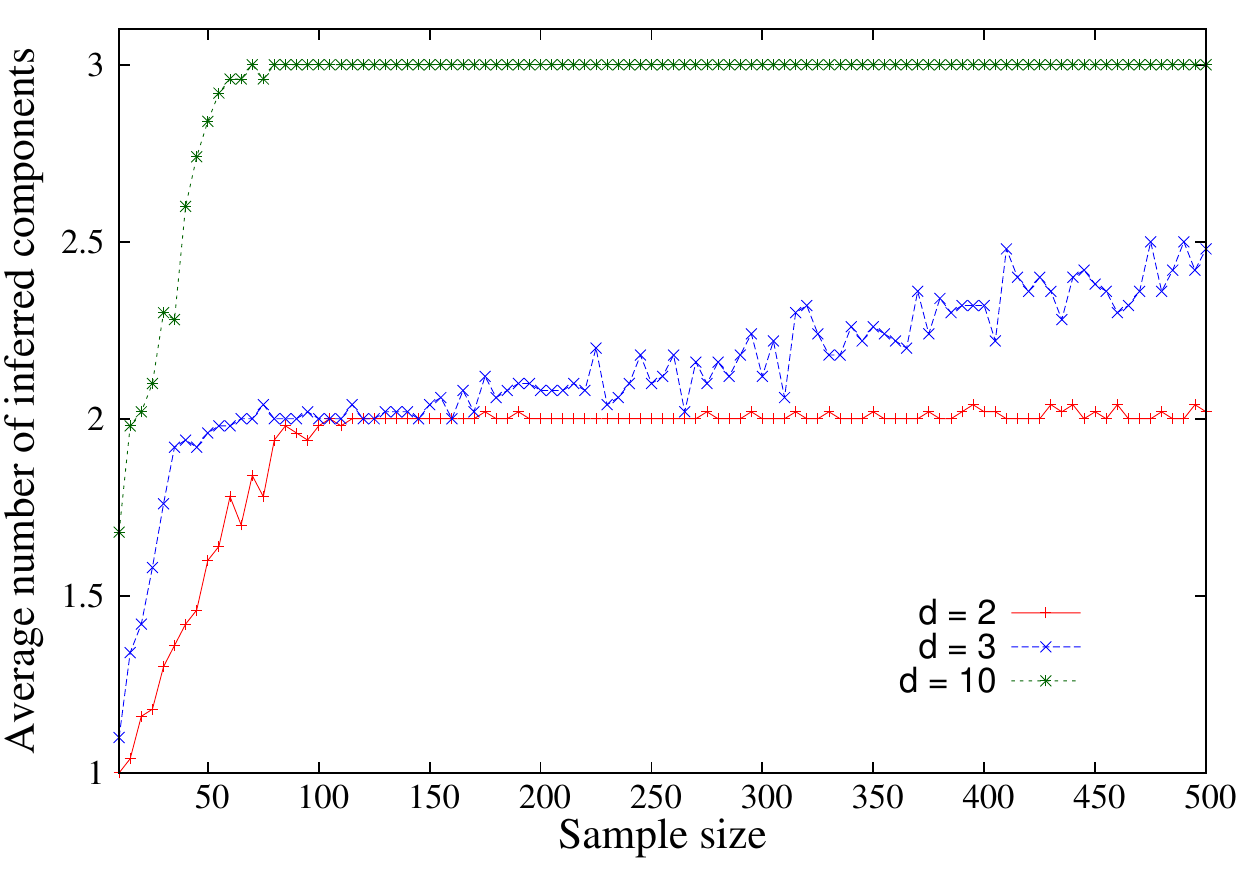}
  } 
  \caption{Case 2: Average number of components inferred when the true mixture has components 
with the same mean directions but different concentration parameters. 
(a) 2-component mixture ($\kappa_1 = 10, \kappa_2 = 100$) (b) 3-component mixture ($\kappa_1 = 10, \kappa_2 = 100, \kappa_3 = 1000$)}
  \label{fig:same_mu_diff_kappa}
\end{figure}

Fig.~\ref{fig:same_mu_diff_kappa}(a) shows the average 
number of components inferred for a two-component mixture whose
concentration parameters are $\kappa_1 = 10$ and $\kappa_2 = 100$.
For each value of $d$, as the sample size increases,
the average number of inferred components gradually increases until it saturates and 
reaches the true value (2 in this case). Increasing the sample size
beyond this does not impact the number of inferred mixture components.
The results for a 3-component mixture with identical means and concentration parameters
$\kappa_1 = 10, \kappa_2 = 100$, and $\kappa_3 = 1000$ are shown in
Fig.~\ref{fig:same_mu_diff_kappa}(b). As expected, 
the average number of inferred components increases in light 
of greater evidence. However, we note that it requires
greater amount of data to correctly infer the three mixture components
as compared to the two-component case.
In the two-component case (Fig.~\ref{fig:same_mu_diff_kappa}(a)),
at around $N=450$, all the three curves converge to the right number of components.
For the three-component mixture in Fig.~\ref{fig:same_mu_diff_kappa}(b),
up until $N=500$, there is no convergence for $d=2,3$.
For $d=10$, the average number of inferred components 
converges quickly (at $N\sim 25$) for the 2-component mixture when compared
with $N\sim 100$ for the 3-component mixture.
This is expected as correctly distinguishing three components (with same means)
requires far more data. 

\begin{wrapfigure}{r}{0.45\textwidth}
\centering
\includegraphics[width=\textwidth]{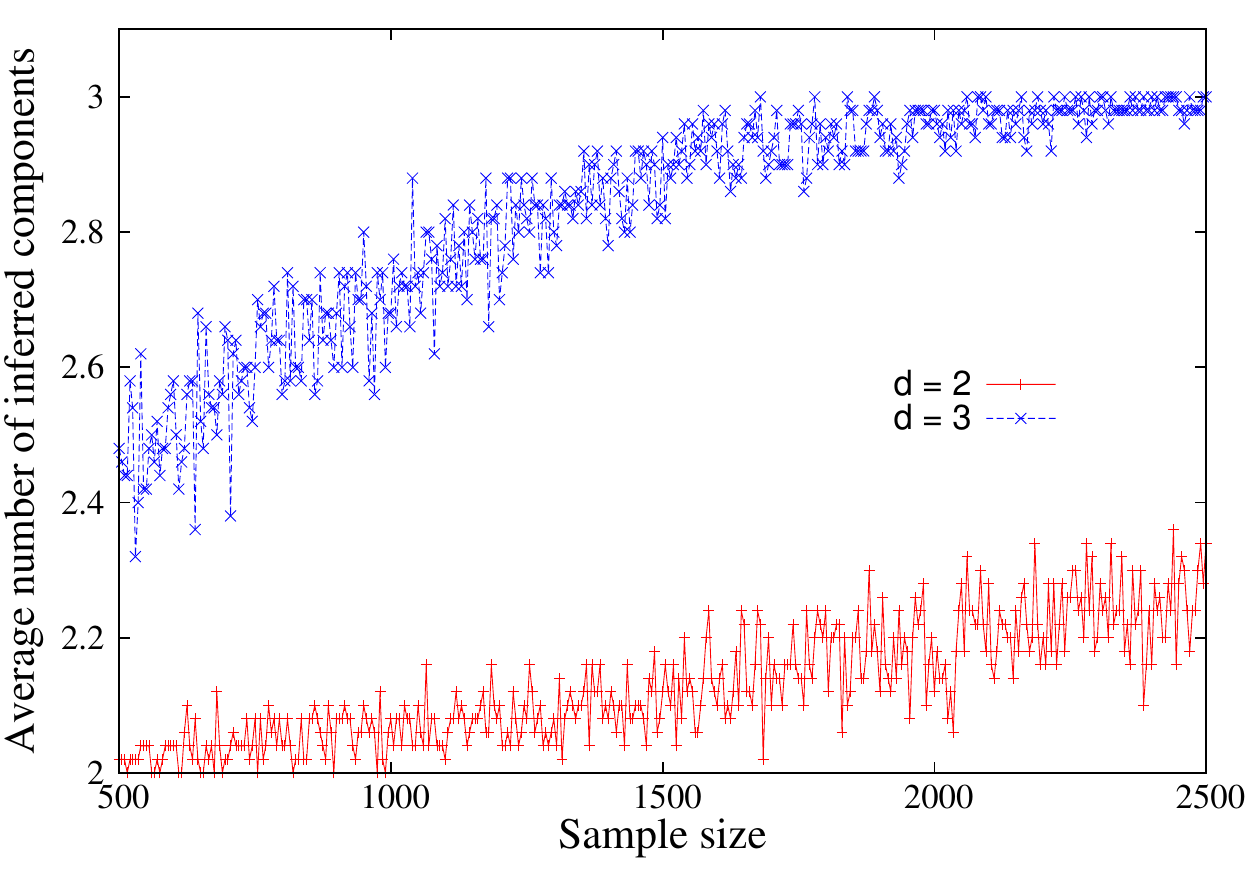}
\caption{Gradual increase in the average inferred components for the 3-component mixture 
         in Fig.~\ref{fig:same_mu_diff_kappa}(b).}
\label{fig:mean_components_k_3_extra}
\end{wrapfigure}

It is also interesting to note that for $d=2$ in Fig.~\ref{fig:same_mu_diff_kappa}(b),
the average number of inferred components saturates at 2, while the actual number of mixture components is 3.
However, as the amount of available data increases, the (almost) horizontal line
shows signs of gradual increase in its slope. Fig.~\ref{fig:mean_components_k_3_extra}
shows the increase in the average number for $d=2,3$ as the data is increased beyond $N=500$.
The blue curve representing $d=3$ stabilizes at $N\sim 2500$. However, the red curve ($d=2$)
slowly starts to move up but the average is still well below the true number.
This demonstrates the relative difficulty in estimating the true mixture
when the means coincide especially at lower dimensions.

One can imagine that when the means coincide, it would require greater
amount of data to accurately infer the true mixture components. However,
the amount of data required also depends on the dimensionality
in consideration. It appears that as the dimensionality increases,
we need smaller amounts of data (as can be seen for the $d=10$ case).
In $d$-dimensional space, each datum comprises of $d$ real values with the constraint
that it should lie on the \emph{unit} hypersphere. So the estimation of the mean direction
requires the estimation of $(d-1)$ values and one value for $\kappa$. 
When there is $N$ data, we actually have $n_d = N \times (d-1)$
values available for estimating the $d$ free parameters. For instance, given a sample of 
size $N$,  for $d=2, n_2 = N$ and for $d=10, n_{10} = 9N$. 
We conjecture that this could be
a possible reason for faster convergence in high dimensional space.
Through these experiments, we have demonstrated the ability of our
search method to infer appropriate mixtures in situations 
with varying difficulty levels.

\section{Applications of vMF mixtures}  \label{sec:vmf_applications}
\subsection{Application to text clustering}
The use of vMF mixtures in modelling high dimensional text data
has been investigated by \cite{Banerjee:clustering-hypersphere}.
To compute the similarity between text documents requires their
representation in some vector form. The elements of the 
vectors are typically a function of the word and document frequencies in a given collection.
These vector representations are commonly used in clustering
textual data with cosine based similarity metrics being central to
such analyses \citep{strehl2000impact}. 
There is a strong argument for transforming the vectors into points on a unit hypersphere 
\citep{salton1983introduction,salton1988term}. Such a normalized representation 
of text data (which compensates for different document lengths)
motivates their modelling using vMF mixture distributions.

\cite{Banerjee:clustering-hypersphere} used their proposed approximation (Equation~\eqref{eqn:banerjee_approx})
to estimate the parameters of a mixture with \emph{known} number of components.
They did not, however, propose a method to search for the optimal number of
mixture components. 
We not only derived MML estimates which fare better compared to the previous
approximations but also apply them to devise a search method to infer
the optimal mixtures. 
Ideally, the search is continued until there is no further improvement
in the message length (Algorithm~\ref{algm}).
For practial purposes, the search is terminated when the improvement
due to the intermediate split, delete and merge operations during the 
search process is less than $0.01\%$.
Our proposed method to infer mixtures was employed on the datasets that were used 
in the analysis by \cite{Banerjee:clustering-hypersphere}. The parameters of the intermediate mixtures are
estimated using the MML Halley's estimates (Equation~\eqref{eqn:mml_halley_approx})
for the component vMF distributions.
\cite{Banerjee:clustering-hypersphere} use mutual information (MI) to assess 
the quality of clustering. For given cluster assignments $X$ and the (known) class labels $Y$,
MI is defined as: $E\left[ \log \dfrac{\Pr(X,Y)}{\Pr(X)\Pr(Y)} \right]$. Along with 
the message lengths, we use MI as one other evaluation criterion in our analysis.
We also compute the average F-measure when the number of clusters is same 
as the number of actual classes.

For each of the datasets, in the preprocessing step,
we generate feature vectors using the most frequently occuring words and 
generating a TF-IDF score for each feature (word) based on Okapi BM25 score \citep{bm25}.
These feature vectors are then normalized to generate unit vectors in
some $d$-dimensional space. Using this as directional data on a hypersphere, 
a suitable mixture model was inferred using the greedy search proposed in 
Section~\ref{sec:search_method}.

\subsubsection{Classic3 dataset}

The dataset\footnote{\url{http://www.dataminingresearch.com/index.php/2010/09/classic3-classic4-datasets/}}
contains documents from three distinct categories: 1398 Cranfield (aeronautical related),
1033 Medline (medical journals) and 1460 Cisi (information retrieval related) documents.
The processed data has $d = 4358$ features. 

\noindent\emph{Optimal number of clusters:} In this example, it is known that there are three 
distinct categories.
However, this information is not usually known in real world setting
(and we do not know if they are from three vMF distributions).
Assuming no knowledge of the nature of the data, the search
method infers a mixture with 16 components.
The corresponding assignments are shown in Table~\ref{tab:conf_matrix_16}.
A closer look at the generated assignments
illustrate that each category of documents is represented by more than one component.
The three categories are split to possibly represent specialized sub-categories.
The Cisi category is distributed among 6 main components (M4 -- M9).
The Cranfield documents
are distributed among M6, M10 -- M15 components and
the Medline category is split into M0 -- M3, and M6 components.
We observe that all but three components are non-overlapping;
only M6 has representative documents from all three categories.
\begin{table}[htb]
  \caption{Confusion matrix for 16-component assignment (MML Halley).}
  \centering
    \begin{tabular}{|c|c|c|c|c|c|c|c|c|c|c|c|c|c|c|c|c|}
    \hline
           & M0 & M1 & M2 & M3 & M4 & M5 & M6 & M7 & M8 & M9 & M10 & M11 & M12 & M13 & M14 & M15  \\ \hline
      cisi & 0   &      0 &        4 &        0 &      288 &      133 &       28  &     555 &      197  &    255  &       0  &       0 &        0    &     0  &       0 &        0 \\
      cran & 0   &      0 &        0 &        0 &        2 &        0 &      362  &       1 &        0  &      0  &      58  &     144 &      135    &   175  &     223 &      298 \\
      med  & 9   &    249 &      376 &      138 &        2 &        0 &        9  &       0 &        0  &      0  &       0  &       0 &        0    &     0  &       0 &        0 \\
    \hline
    \end{tabular}
    \label{tab:conf_matrix_16}
\end{table}

The 16-component mixture inferred by the search method 
is a finer segregation of the data when compared to modelling using a 3-component mixture.
The parameters of the 3-component mixture are estimated
using an EM algorithm (Section~\ref{subsec:em_mml}). 
Table~\ref{tab:conf_matrix_3} shows the confusion matrices obtained for the cluster
assignments using the various estimation methods. We see that all the estimates perform
comparably with each other; there is not much difference in the assignments
of data to the individual mixture components.
\begin{table}[htb]
  \caption{Confusion matrices for 3-cluster assignment.
   (Sra's confusion matrix is omitted as it is same as that of Tanabe)}
  \subfloat[Banerjee]
  {
    \centering
    \begin{tabular}{|c|c|c|c|}
    \hline
           & cisi  & cran & med \\ \hline
      cisi & 1441  &  0   & 19 \\
      cran & 22    & 1293  & 83 \\
      med  & 8    &  0   & \textbf{1025} \\
    \hline
    \end{tabular}
  }
  \subfloat[Tanabe]
  {
    \centering
    \begin{tabular}{|c|c|c|c|}
    \hline
           & cisi  & cran & med \\ \hline
      cisi & 1449  &  0   & 11 \\
      cran & 24    & 1331 & 43 \\
      med  & 13    &  0   & 1020 \\
    \hline
    \end{tabular}
  }
  \subfloat[Song]
  {
    \centering
    \begin{tabular}{|c|c|c|c|}
    \hline
           & cisi  & cran & med \\ \hline
      cisi & \textbf{1450}  &  0   & 10 \\
      cran & 24    & \textbf{1339} & 35 \\
      med  & 14    &  0   & 1019 \\
    \hline
    \end{tabular}
  }
  \subfloat[MML (Halley)]
  {
    \centering
    \begin{tabular}{|c|c|c|c|}
    \hline
           & cisi  & cran & med \\ \hline
      cisi & \textbf{1450}  &  0   & 10 \\
      cran & 24    & 1331 & 43 \\
      med  & 13    &  0   & 1020 \\
    \hline
    \end{tabular}
  }
  \label{tab:conf_matrix_3}
\end{table}

The collection is comprised of
documents that belong to dissimilar categories and hence, the clusters obtained
are wide apart. This can be seen from the extremely high F-measure scores
(Table~\ref{tab:classic3_evaluation}). For the 3-component mixture, all the five different estimates 
result in high F-measure values
with Song being the best with an average F-measure of 0.978 and a MI of 0.982. 
MML (Halley's) estimate are close with an average F-measure of 0.976 and a MI of 0.976. 
However, based on the
message length criterion, the MML estimate results in the least message length
($\sim 190$ bits less than Song's).
The mutual information score using MML estimate
is 1.04 (for 16 components) compared to 0.976 for 3 components.
Also, the message length is lower for the 16 component case.
However, Song's estimator results in a MI score of 1.043, 
very close to the score of 1.040 obtained using MML estimates.
\begin{table}[htb]
  \caption{Clustering performance on Classic3 dataset.}
  \centering
  \begin{tabular}{|c|c|c|c|c|c|c|}
  \hline
   Number of clusters    & Evaluation metric & Banerjee   & Tanabe     &  Sra      & Song      &   MML (Halley)   \\\hline 
   \multirow{3}{*}{3} &   Message length      & 100678069  & 100677085  & 100677087 & 100677080 & \textbf{100676891} \\
                      &  Avg. F-measure       & 0.9644     & 0.9758     & 0.9758    & \textbf{0.9780}    & 0.9761    \\
                      &   Mutual Information  & 0.944     & 0.975     & 0.975    & \textbf{0.982}    & 0.976    \\\hline
   \multirow{2}{*}{16} &   Message length      & 100458153& 100452893  & 100439983 & 100444649 & \textbf{100427178} \\
                      &   Mutual Information  & 1.029     &  1.036    & 0.978    & \textbf{1.043} & 1.040    \\\hline
  \end{tabular}
  \label{tab:classic3_evaluation}
\end{table}

For the Classic3 dataset, \cite{Banerjee:clustering-hypersphere} analyzed mixtures with
greater numbers of components than the ``natural" number of clusters.
They report that a 3-component mixture is not necessarily a good model
and more number of clusters may be preferred for this example.
As part of their observations, they suggest to ``generate greater number
of clusters and combine them appropriately".
However, this is subjective and
requires some background information about the likely number of clusters.
Our search method in conjunction with the inference framework is able to resolve this dilemma
and determine the optimal number of mixture components in a completely unsupervised setting.

\subsubsection{CMU\_Newsgroup}

The dataset\footnote{\url{http://archive.ics.uci.edu/ml/datasets/Twenty+Newsgroups}}
contains documents from 20 different news categories each  containing
1000 documents. Preprocessing of the data, as discussed above, resulted in feature vectors
of dimensionality $d=6448$. The data is first modelled using a mixture
containing 20 components. 
The evaluation metrics are shown in Table~\ref{tab:newsgroup_evaluation}.
The average F-measure is 0.509 for MML-based estimation,
slightly better than Banerjee's score of 0.502. The low F-measure values
are indicative of the difficulty in accurately distinguishing the news categories.
The mutual information score for MML case is 1.379 which is lower than that of Sra's.
However, the total message length is lower for MML mixture compared to that of others.
\begin{table}[htb]
  \caption{Clustering performance on CMU\_Newsgroup dataset.}
  \centering
  \begin{tabular}{|c|c|c|c|c|c|c|}
  \hline
   Number of clusters    & Evaluation metric & Banerjee   & Tanabe     &  Sra      & Song      &   MML (Halley)   \\\hline 
   \multirow{3}{*}{20} &   Message length    & 728666702  & 728545471  & 728585441 & 728536451 & \textbf{728523254} \\
                      &  Avg. F-measure      & 0.502     & 0.470     & 0.487    & 0.435    & \textbf{0.509}    \\   
                      &   Mutual Information & 1.391     & 1.383     & \textbf{1.417}    & 1.244    & 1.379    \\\hline
   \multirow{2}{*}{21} &   Message length    &  728497453 & 728498076 & 728432625 & 728374429 &\textbf{728273820} \\
                      &   Mutual Information & 1.313 & 1.229  & \textbf{1.396} & 1.377 & 1.375 \\ \hline
  \end{tabular}
  \label{tab:newsgroup_evaluation}
\end{table}

\noindent\emph{Optimal number of clusters:}
The proposed search method when applied to this dataset infers a mixture with
21 components. This is close to the ``true'' number of 20 (although there
is no strong reason to believe that each category corresponds to a vMF component).
The mutual information 
for the 21-cluster assignment is highest for Sra's mixture with a score of 1.396
and for MML mixture, it is 1.375 (Table~\ref{tab:newsgroup_evaluation}).
However, the total message length is the least for the MML-based mixture.

The analysis of vMF mixtures by \cite{Banerjee:clustering-hypersphere}
for both the datasets considered here
shows a continued increase in the MI scores even beyond the true number 
of clusters. As such, using the MI evaluation metric for different number
of mixture components does not aid in the inference of an optimal mixture model.
Our search method balances the tradeoff between using a certain
mixture and its ability to explain the observed data, and thus, objectively aids in
inferring mixtures to model the normalized vector representations
of a given collection of text documents.

A mixture modelling problem of this kind where there is some information
available regarding the nature of the data can be studied by 
altering the proposed search method.
We provide some alternate strategies where the mixture modelling can be
done in a semi-supervised setting.
\begin{itemize}
\item The priors on the number of components and their parameters can be
modelled based on the background knowledge.
\item If the true number of clusters are known, only splits may be
carried out until we near the true number (each split being the best one
given the current mixture). As the mixture size approaches
the true number, all the three operations (split, delete, and merge) can be resumed until convergence.
This increases the chance that the inferred mixture would have about 
the same number of components as the true model.
\item Another variant could be to start from a number close to the
true number and prefer delete/merge operations over the splits.
We cannot ignore splits completely because a component after splitting
may be merged at a later iteration if there would be an improvement to
the message length.
\item Another strategy could be to employ the EM algorithm and infer a
mixture with the true number of components. This mixture can then 
be perturbed using split, delete, and merge operations until convergence.
\end{itemize}

\subsection{Mixture modelling of protein coordinate data} \label{subsec:protein_mixture_modelling}
The following application concerns the vMF mixture modelling of directional data
arising from the orientation of main chain carbon atoms
in protein structures. 
The structures that proteins adopt are largely dictated by the interactions
between the constituent atoms. These chemical interactions
impose constraints on the orientation of atoms
with respect to one another. The directional nature of the protein data
and the (almost constant) bond length between the main chain carbon atoms
motivate the modelling using vMF mixtures. Further, structural modelling
tasks such as generating random protein chain conformations, three-dimensional 
protein structure alignment, secondary structure assignment, 
and representing protein folding patterns using concise protein fragments
require efficient encoding of protein data 
\citep{konagurthu-sst,konagurthu2013statistical,collier2014new}.
As part of our results, we demonstrate that vMF mixtures offer a better 
means of encoding and can potentially
serve as strong candidate models to be used in such varied tasks.

The dataset considered here is a collection of 8453 non-redundant experimentally determined
protein structures from the publicly available ASTRAL SCOP-40 (version 1.75) database \citep{murzin1995scop}.
For each  protein structure, the coordinates of the central carbon, $C_\alpha$,
of successive residues (amino acids) are considered. 
Protein coordinate data is transformed into
directional data and each direction vector is characterized by $(\theta,\phi) = $
(co-latitude, longitude), where $\theta\in[0,180^{\circ}]$ and $\phi\in[0,360^{\circ}]$.
Note that these $(\theta,\phi)$ values
have to be measured in a consistent, canonical manner. 

\begin{wrapfigure}{r}{0.45\textwidth}
\centering
\includegraphics[scale=0.15]{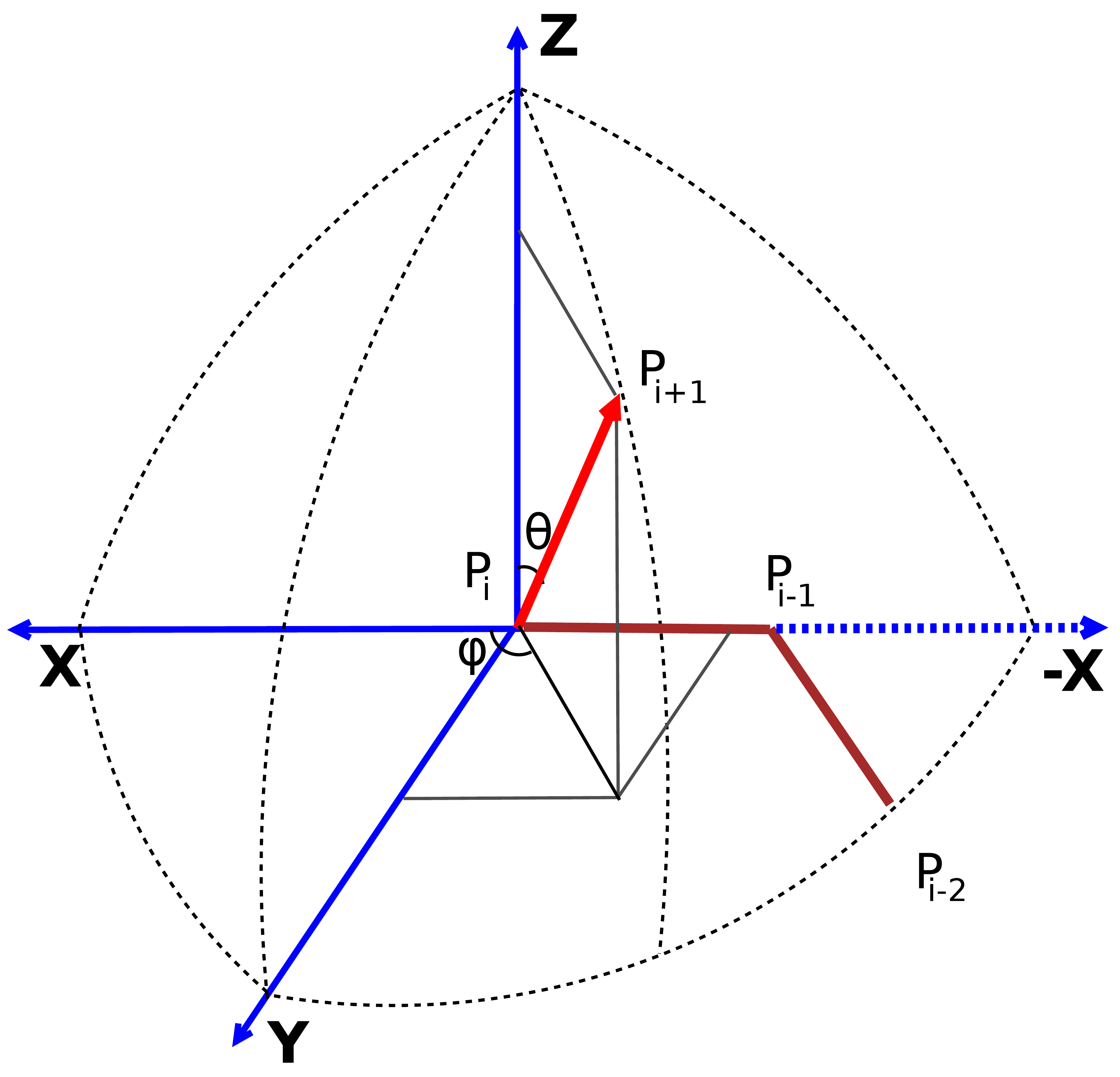}
\caption{Canonical orientation used to determine the directional data corresponding to protein coordinates.}
\label{fig:canonical_orientation} 
\end{wrapfigure}
To compute
$(\theta,\phi)$ corresponding to the point $P_{i+1}$ associated to residue $i+1$, 
we consider this point in the context of 3
preceding points, forming a 4-mer comprising of the points $P_{i-2}, P_{i-1},
P_{i}$, and $P_{i+1}$. This 4-mer is orthogonally transformed into a canonical
orientation (Fig.~\ref{fig:canonical_orientation}) in the following steps:
\begin{itemize}
  \item translate the 4-mer such that $P_i$ is at the origin.
  \item rotate the resultant 4-mer so that $P_{i-1}$ lies on the negative X-axis.
  \item rotate further so that $P_{i-2}$ lies in the XY plane such that the angle 
  between the vector $\mathbf{P_{i-2}-P_{i-1}}$ and the positive Y-axis is acute.
\end{itemize}
The transformation yields a canonical orientation for $P_{i+1}$ with respect to its previous 3 coordinates. 
Using the transformed coordinates of $P_{i+1}$, the direction $(\theta,\phi)$ of $P_{i+1}$ is computed.
We repeat this transformation for every successive set of 4-mers in the protein
chain, over all possible source structures in our collection. The data collected  
in this way resulted in a total of $\sim 1.3$ million $(\theta,\phi)$ pairs for all
the 8453 structures in the database.

Protein data is an example where the number of mixture components are not known a priori.
Hence, we use the method outlined in Section~\ref{sec:search_method} to infer 
suitable mixture models. 
The original dataset comprises of 7 different categories of proteins.
The proposed search method using MML (Halley's) estimates infers a mixture 
containing 13 vMF components. 
Further, each protein category can be individually modelled using a mixture.
As an example, for the ``$\beta$ class'' proteins which contains 
1802 protein structures and
251,346 corresponding $(\theta,\phi)$ pairs, our search method 
terminates after inferring 11 vMF components. 
We compare the MML-based mixtures with those inferred by the standalone EM
algorithm (Section~\ref{subsec:em_mml}) using other estimates.
These values are presented in Table~\ref{tab:protein_mixture}.
We observe that the mixtures inferred using the MML estimates result in a message length
lower than that obtained using the other estimates.

\begin{figure}[htb]
  \centering
  \subfloat[]
  {
    \includegraphics[width=0.5\textwidth]{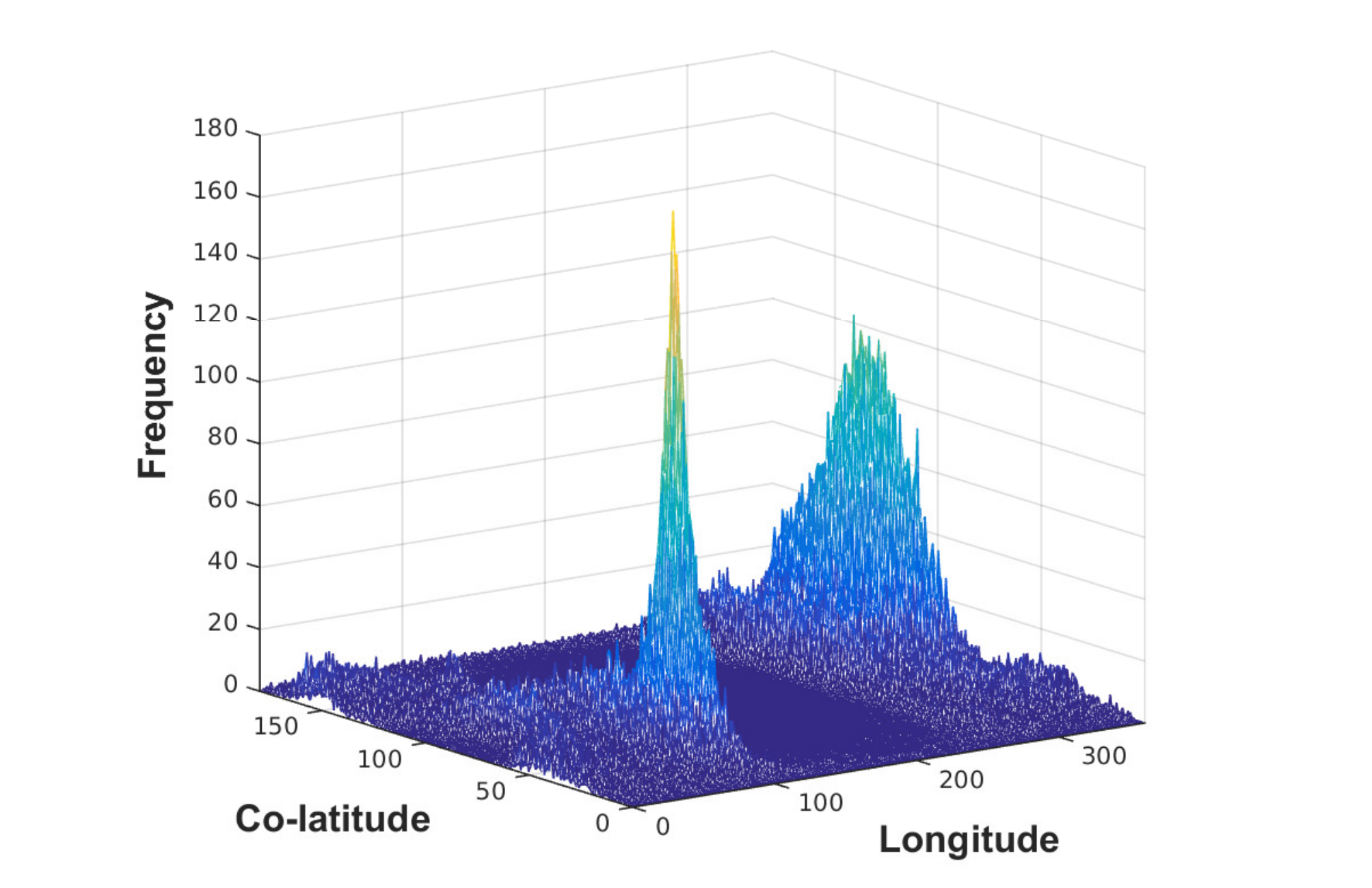}
  }
  \subfloat[]
  {
    \includegraphics[width=0.5\textwidth]{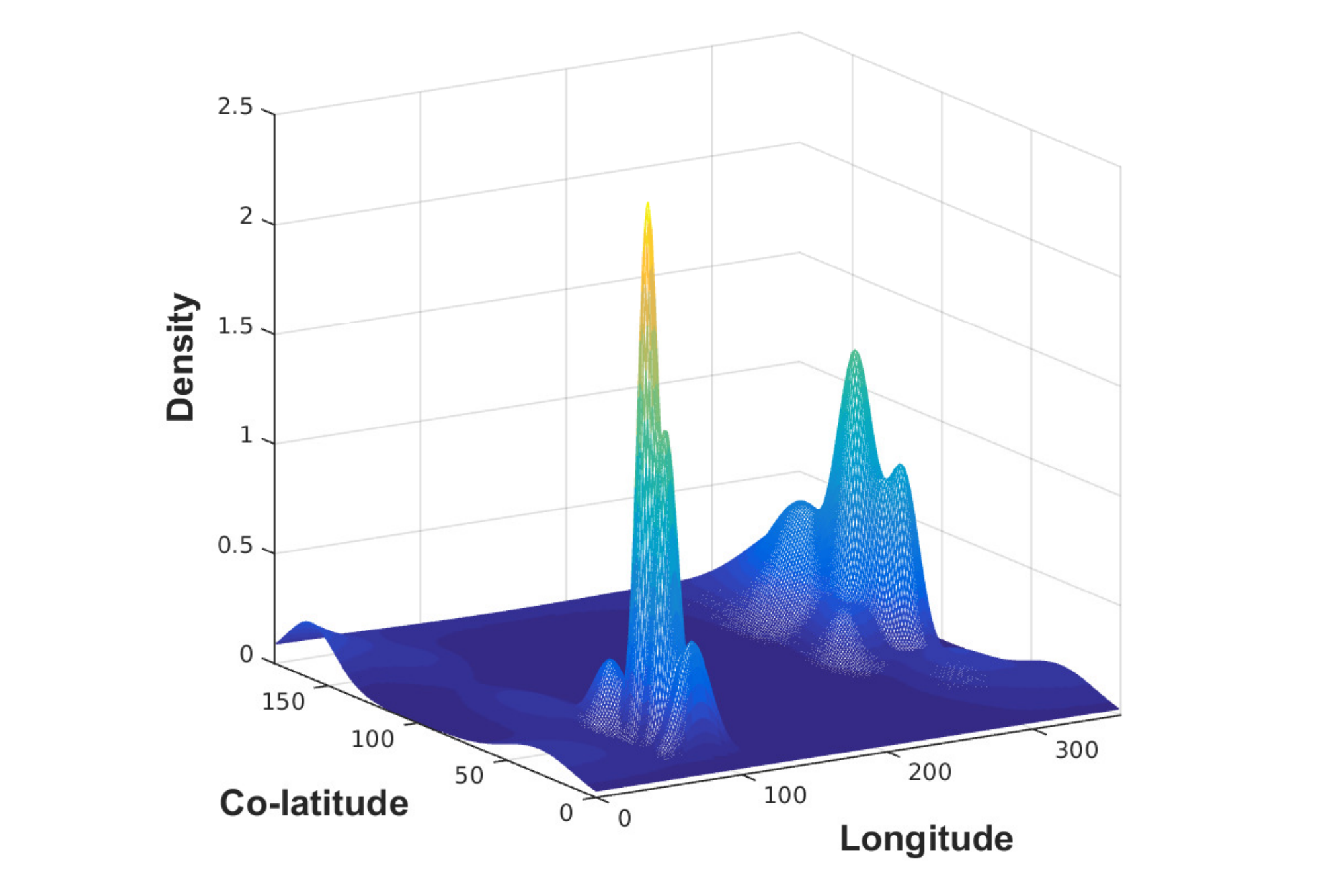}
  } 
  \caption{
    Distribution of directional data of $\beta$ class $C_\alpha$ atoms 
    (a) Empirical distribution
    (b) Mixture density corresponding to the 11 inferred vMF components
    }
  \label{fig:protein_dist}
\end{figure}

Fig.~\ref{fig:protein_dist}(a) shows the empirical distribution of directional data 
of $C_\alpha$ coordinates belonging to $\beta$ class.
The $(\theta,\phi)$ values with their corresponding frequencies
are plotted in Fig.~\ref{fig:protein_dist}(a). 
Fig.~\ref{fig:protein_dist}(b) is a plot illustrating 
the 11-component vMF mixture density as inferred for this class of proteins.
Notice that the two major modes 
in the figure correspond to commonly observed local secondary structural bias
of residues towards, helices and strands of sheet. Also notice the empty
region in the  middle which corresponds to physically unrealizable directions in the local chain,
excluded in the observed samples due to steric hindrance of the atoms in
proteins. If we were to model such data using truncated distributions,
the regions of zero probability will be modelled using an infinite
code length. As an example, at $(\theta,\phi)=(100^{\circ},200^{\circ})$,
the truncated distribution would have zero probability and consequently an
\emph{infinite} code length. However, when the same point is explained using
the 11-component mixture, it would have a
probability of $\Pr=3.36\times10^{-12}$ and a corresponding code
length of $-\log_2\Pr=38.11$ bits. 
For protein data, it is possible to have such (rare exceptional) observations,
due to reasons such as experimental error, noise, or the conformation
of the protein itself.
Hence, although the empirical 
distribution has distinct modes, it is better off modelled as a vMF
mixture distribution, rather than by truncated distributions.\\

\noindent\emph{Compressibility of protein structures:}
The explanatory framework of MML allows for testing competing hypotheses.
Recently, \cite{konagurthu-sst} developed a null model description
of protein coordinate data as part of the statistical inference of protein
secondary structure. A null model gives a baseline for transmitting the raw
coordinate data. Each $C_{\alpha}$ atom is described using the distance and orientation
with respect to the preceding $C_{\alpha}$ atoms.
Because the distance between successive $C_{\alpha}$ atoms is highly
constrained, compression can only be gained in describing the orientation of a $C_{\alpha}$
atom with respect to its previous one.

\cite{konagurthu-sst} describe their null hypothesis
by discretizing the surface of a 3D-sphere into chunks of equal areas 
(of $\epsilon^2$, where $\epsilon$ is the accuracy of measurement of coordinate data).
This results in $4\pi r^2/\epsilon^2$ cells distributed uniformly on the surface of the sphere
of radius $r$ (the distance between successive $C_{\alpha}$ coordinates).
The cells are uniquely numbered. To encode $C_{\alpha}^{i+1}$ with respect to $C_{\alpha}^{i}$,
the location of $C_{\alpha}^{i+1}$ on the surface is identified and the corresponding cell index
is encoded. Using this description, the stated null model results in 
a message length expression~\cite{konagurthu-sst} given by
\begin{equation}
\text{Uniform Null}= -\log_2\left(\frac{\epsilon^2}{4\pi r^2}\right) = \log_2(4\pi) - 2 \log_2\left(\frac{\epsilon}{r}\right)  \quad{\text{bits.}}\label{eqn:null_uniform}
\end{equation}

The null model of \cite{konagurthu-sst} assumes a uniform distribution of orientation angles on the surface of the sphere.
However, this is a crude assumption and one can leverage the directional
properties of protein coordinates to build an efficient null model.
To this effect, we explore the use of vMF mixtures as null model descriptors for protein
structures. Using vMF mixtures, we encode the co-latitude ($\theta$) and longitude ($\phi$) angles (described in Section~\ref{subsec:protein_mixture_modelling}).
The message length expression to encode the orientation angles (using Equation~\ref{eqn:mixture}) is then given by
\begin{equation}
\text{vMF Null} = -\log_2 \left(\sum_{j=1}^M w_j f_j(\mathbf{x};\Theta_j)\right)- 2 \log_2\left(\frac{\epsilon}{r}\right) \quad{\text{bits.}}\label{eqn:null_vmf}
\end{equation}
where $\mathbf{x}$ corresponds to a unit vector described by $(\theta,\phi)$ on the surface of the sphere.
Equations (\ref{eqn:null_uniform}) and (\ref{eqn:null_vmf}) are two competing null models.
These are used to encode the directional data corresponding to the 8453 
protein structures in the ASTRAL SCOP-40 database. 
The results are shown in Table~\ref{tab:null_models_comparison}. 

\begin{figure}[htb]
  \begin{minipage}{0.5\textwidth}
    \begin{table}[H]
      \centering
      \caption{Message lengths (in bits) computed for the inferred protein mixtures
               using various methods (`All' refers to all the 7 protein categories).}
      \begin{tabular}{|c|c|c|c|c|}
      \hline
       Category      &  Tanabe  & Sra     & Song    & MML (Halley)  \\ \hline
       $\beta$       & 5514800  & 5518679 & 5520073 &\textbf{5513507} \\
        All          & 27818524 & 27833704 & 27839802 & \textbf{27803427} \\ 
        \hline
      \end{tabular}
      \label{tab:protein_mixture}
    \end{table}
  \end{minipage}
  \quad\quad
  \begin{minipage}{0.45\textwidth}
    \begin{table}[H]
      \centering
      \caption{Comparison of the uniform and vMF null model encoding schemes.}
      \begin{tabular}{|c|c|c|}
      \hline
      Null model & Total message length (in bits) & Bits per residue \\ \hline
      Uniform & 36119900  & 27.437 \\
      vMF    & \textbf{32869700} & \textbf{24.968}  \\
      \hline
      \end{tabular}
      \label{tab:null_models_comparison}
    \end{table}
  \end{minipage}
\end{figure}

The per residue statistic is calculated by dividing the total message length by the 
sample size (the number of $(\theta,\phi)$ pairs).
This statistic shows that close to 2.5 bits 
can be saved (on average) if the protein data is encoded using the vMF null model.
The vMF null model thus supercedes the naive model of encoding. This can potentially
improve the accuracy of statistical inference that is central to the various protein 
modelling tasks briefly introduced above.

\section{Conclusion}
We presented a statistically robust approach for inferring mixtures of
(i)~multivariate Gaussian distributions, and
(ii)~von Mises-Fisher distributions for $d$-dimensional directional data.
It is based on the
general information-theoretic framework of minimum message length inference.
This provides an objective tradeoff between the hypothesis
complexity and the quality of fit to the data.
An associated search procedure for an optimal mixture model of given data
chooses the number of component distributions, $M$, by minimizing the total message length. 
We established the better performance of the proposed search 
algorithm by comparing with a popularly used search method \citep{figueiredo2002unsupervised}.
We demonstrated the effectiveness of our approach through extensive experimentation
and validation of our results.
We also applied our method to real-world high dimensonal text data and to directional
data that arises from protein chain conformations.
The experimental results demonstrate that our proposed
method fares better when compared with the current state of the art techniques.

\begin{acknowledgements}
The authors would like to thank Arun Konagurthu for discussions pertaining
to protein data clustering, and Maria Garcia de la Banda for stimulating
discussions and providing interesting insights. The authors would also like 
to acknowledge Wray Buntine's inputs with regard to text clustering.
\end{acknowledgements}


\section{Appendix} \label{sec:appendix}
\subsection{Supporting derivations required for evaluating $\kappa_{MN}$ and $\kappa_{MH}$} \label{subsec:appendix_derivations}
For brevity, we represent $A_d(\kappa), A'_d(\kappa), A_d''(\kappa),$ and $A_d'''(\kappa)$ as $A,A',A'',$ and $A'''$ respectively.
Expressions to evaluate $A,A',$ and $A''$ are given in Equations (\ref{eqn:ratio_bessels}), (\ref{eqn:ratio_first_derivative}), and (\ref{eqn:ratio_second_derivative}) respectively.
We require $A'''$ for its use in the remainder of the derivation and we provide its expression below:
\begin{equation}
\frac{A'''}{A'} = -\frac{2 A A''}{A'} - 2 A' - \frac{(d-1)}{\kappa}\frac{A''}{A'} - \frac{2(d-1)}{\kappa^3}\frac{A}{A'} + \frac{2(d-1)}{\kappa^2} \label{eqn:ratio_third_derivative}
\end{equation}

Now we discuss the derivation of $G'(\kappa)$ and $G''(\kappa)$
that are required for computing the MML estimate $\kappa_M$ (Equations (\ref{eqn:mml_newton_approx}) and (\ref{eqn:mml_halley_approx})).
Differentiating Equation~\ref{eqn:I_first_derivative}, we have
\begin{equation}
G'(\kappa) = \frac{(d-1)}{2\kappa^2} + (d+1)\frac{(1-\kappa^2)}{(1+\kappa^2)^2} + \frac{(d-1)}{2} \frac{\partial}{\partial \kappa}\left( \frac{A'}{A} \right)
+ \frac{1}{2}\frac{\partial}{\partial \kappa} \left(\frac{A''}{A'}\right) + n A' \label{eqn:I_second_derivative}
\end{equation}
\begin{equation}
\text{and}\quad G''(\kappa) = -\frac{(d-1)}{\kappa^3} + (d+1)\frac{2\kappa(\kappa^2-3)}{(1+\kappa^2)^3} + \frac{(d-1)}{2} \frac{\partial^2}{\partial \kappa^2}\left( \frac{A'}{A} \right)
+ \frac{1}{2}\frac{\partial^2}{\partial \kappa^2} \left(\frac{A''}{A'}\right) + n A''\label{eqn:I_third_derivative}
\end{equation}

Using Equations (\ref{eqn:ratio_bessels}) and (\ref{eqn:ratio_first_derivative}), we have
\begin{align}
\frac{A'}{A} &= \frac{1}{A} - A - \frac{(d-1)}{\kappa}  \notag \\
\frac{\partial}{\partial \kappa}\left( \frac{A'}{A}\right) &= -\frac{A'}{A^2} - A' + \frac{(d-1)}{\kappa^2} \notag \\
\frac{\partial^2}{\partial \kappa^2}\left( \frac{A'}{A} \right) &= 2\frac{(A')^2}{A^3} - \frac{A''}{A^2} - A'' - \frac{2(d-1)}{\kappa^3} \label{eqn:tmp1}
\end{align}

Using Equations (\ref{eqn:ratio_bessels}), (\ref{eqn:ratio_first_derivative}), (\ref{eqn:ratio_second_derivative}), and (\ref{eqn:ratio_third_derivative}) we have
\begin{align}
\frac{A''}{A'} &= -2 A - \frac{(d-1)}{\kappa} + \frac{(d-1)}{\kappa^2}\frac{A}{A'} \notag\\
\frac{\partial}{\partial \kappa}\left( \frac{A''}{A'}\right) &= -2 A' + \frac{2(d-1)}{\kappa^2} - \frac{(d-1)}{\kappa^3}\frac{A}{A'}\left(\frac{\kappa A''}{A'}+2\right) \notag \\
\frac{\partial^2}{\partial \kappa^2}\left( \frac{A''}{A'} \right) &= -2 A'' - \frac{4(d-1)}{\kappa^3} 
- (d-1) \frac{\partial}{\partial \kappa}\left(\frac{A A''}{\kappa^2 A'^2}  \right) -2 (d-1)\frac{\partial}{\partial \kappa}\left(\frac{A}{\kappa^3 A'} \right) \label{eqn:tmp2}
\end{align}
\begin{align*}
\text{where}\quad\frac{\partial}{\partial \kappa}\left(\frac{A A''}{\kappa^2 A'^2} \right) &= \frac{\kappa A A' A''' + \kappa A'^2 A'' - 2\kappa A A''^2 -2 A A' A''}{\kappa^3 A'^3} \\
\text{and}\quad\frac{\partial}{\partial \kappa}\left(\frac{A}{\kappa^3 A'} \right) &= \frac{1}{\kappa^3} - \frac{A}{\kappa^4 A'^2}(\kappa A'' + 3 A')
\end{align*}

Equations (\ref{eqn:tmp1}) and (\ref{eqn:tmp2}) can be used to evaluate $G'(\kappa)$ and $G''(\kappa)$ 
which can then be used to approximate the MML estimates $\kappa_{MN}$ and $\kappa_{MH}$ 
(Equations (\ref{eqn:mml_newton_approx}) and (\ref{eqn:mml_halley_approx}) respectively).

\subsection{Derivation of the weight estimates in MML mixture modelling} \label{subsec:wts_mml}
\begin{equation*}
\text{As per Equation~\eqref{eqn:mixture_msglen}, we have}\quad
I(\boldsymbol{\Phi},D) = -\frac{1}{2} \sum_{j=1}^M w_j -\sum_{i=1}^N \log \sum_{j=1}^M w_j f_j(\mathbf{x}_i;\Theta_j)
+ \text{terms independent of } w_j
\end{equation*}
To obtain the optimal weights under the constraint $\sum_{j=1}^M w_j = 1$, 
the above equation is optimized using the \emph{Lagrangian} objective 
function defined below using some \emph{Lagrangian multiplier} $\lambda$.
\begin{equation*}
L(\boldsymbol{\Phi},D,\lambda) = I(\boldsymbol{\Phi},D) + \lambda \left(\sum_{j=1}^M w_j-1\right) 
\end{equation*}
For some $k\in\{1,M\}$, the equation resulting from computing the partial derivative of $L$ with respect to $w_k$
and equating it to zero gives the optimal weight $w_k$.
\begin{equation}
\frac{\partial L}{\partial w_k} = 0 \implies \lambda = \frac{1}{2w_k} 
+ \sum_{i=1}^{N} \frac{f_k(\mathbf{x}_i;\Theta_k)}{\sum_{j=1}^M w_j f_j(\mathbf{x}_i;\Theta_j)} \label{eqn:lagrangian_opt}
\end{equation}
\begin{equation*}
\text{We have}\quad \sum_{i=1}^{N} \frac{f_k(\mathbf{x}_i;\Theta_k)}{\sum_{j=1}^M w_j f_j(\mathbf{x}_i;\Theta_j)}
= \frac{1}{w_k}\displaystyle\sum_{i=1}^{N} \frac{w_kf_k(\mathbf{x}_i;\Theta_k)}{\sum_{j=1}^M w_j f_j(\mathbf{x}_i;\Theta_j)}
=\frac{1}{w_k}\displaystyle\sum_{i=1}^{N} r_{ik}
= \frac{n_k}{w_k}
\end{equation*}
where $r_{ik}$ and $n_k$ are the responsibility and effective membership
terms given
as per Equations~\eqref{eqn:responsibility} and \eqref{eqn:comp_eff_mshp} respectively.
Substituting the above value in Equation~\eqref{eqn:lagrangian_opt},
we have
\begin{equation}
\lambda = \frac{1}{2w_k} + \frac{n_k}{w_k}
\implies \lambda w_k = n_k + \frac{1}{2}
\label{eqn:lagrangian_proof}
\end{equation}
There are $M$ equations similar to Equation~\eqref{eqn:lagrangian_proof}
for values of $k\in\{1,M\}$.
Adding all these equations together, we get
\begin{equation*}
\lambda \sum_{j=1}^M w_j = \sum_{j=1}^M n_j + \frac{M}{2}
\implies \lambda = N + \frac{M}{2}
\end{equation*}
Substituting the above value of $\lambda$ in Equation~\eqref{eqn:lagrangian_proof},
we get $w_k = \dfrac{n_k+\frac{1}{2}}{N+\frac{M}{2}}$

\subsection{Derivation of the Kullback-Leibler (KL) distance between two von Mises-Fisher distributions} \label{subsec:vmf_kldiv}
The closed form expression to calculate the KL divergence between two vMF distributions is presented below.
Let $f(\mathbf{x}) = C_d(\kappa_1) e^{\kappa_1 \boldsymbol{\mu_1}^T\mathbf{x}}$ and $g(\mathbf{x}) = C_d(\kappa_2) e^{\kappa_2 \boldsymbol{\mu_2}^T\mathbf{x}}$ be
two von Mises-Fisher distributions with mean directions $\boldsymbol{\mu_1}, \boldsymbol{\mu_2}$ and concentration parameters $\kappa_1, \kappa_2$. 
The KL distance between any two distributions is given by
\begin{equation} 
D_{KL}(f||g) = \int_{\mathbf{x}} f(\mathbf{x})\log\frac{f(\mathbf{x})}{g(\mathbf{x})} d\mathbf{x}
             = \mathrm{E}_f\left[\log\frac{f(\mathbf{x})}{g(\mathbf{x})} \right] \label{eqn:kldiv} 
\end{equation}
where $\mathrm{E}_f[.]$ is the expectation of the quantity $[.]$ using the probability density function $f$.
\begin{equation*}
\log\frac{f(\mathbf{x})}{g(\mathbf{x})} = \log\frac{C_d(\kappa_1)}{C_d(\kappa_2)} + (\kappa_1\boldsymbol{\mu_1} -\kappa_2\boldsymbol{\mu_2})^T \mathbf{x}
\end{equation*}
Using the fact that $\mathrm{E}_f[\mathbf{x}] = A_d(\kappa_1)\boldsymbol{\mu_1}$ \citep{mardia1984goodness,fisher1993statistical}, we have the following expression:
\begin{align}
\mathrm{E}_f\left[\log\frac{f(\mathbf{x})}{g(\mathbf{x})} \right] &= \log\frac{C_d(\kappa_1)}{C_d(\kappa_2)} + (\kappa_1\boldsymbol{\mu_1} -\kappa_2\boldsymbol{\mu_2})^T A_d(\kappa_1) \boldsymbol{\mu_1} \notag\\
D_{KL}(f||g) &= \log\frac{C_d(\kappa_1)}{C_d(\kappa_2)} + A_d(\kappa_1) (\kappa_1 -\kappa_2\boldsymbol{\mu_1}^T\boldsymbol{\mu_2}) \label{eqn:vmf_kldiv}
\end{align}

\pagebreak
\subsection{An example showing the evolution of the mixture model} \label{subsec:appendix_mix2}

\begin{wrapfigure}{r}{0.35\textwidth}
  \centering
  \includegraphics[width=\textwidth]{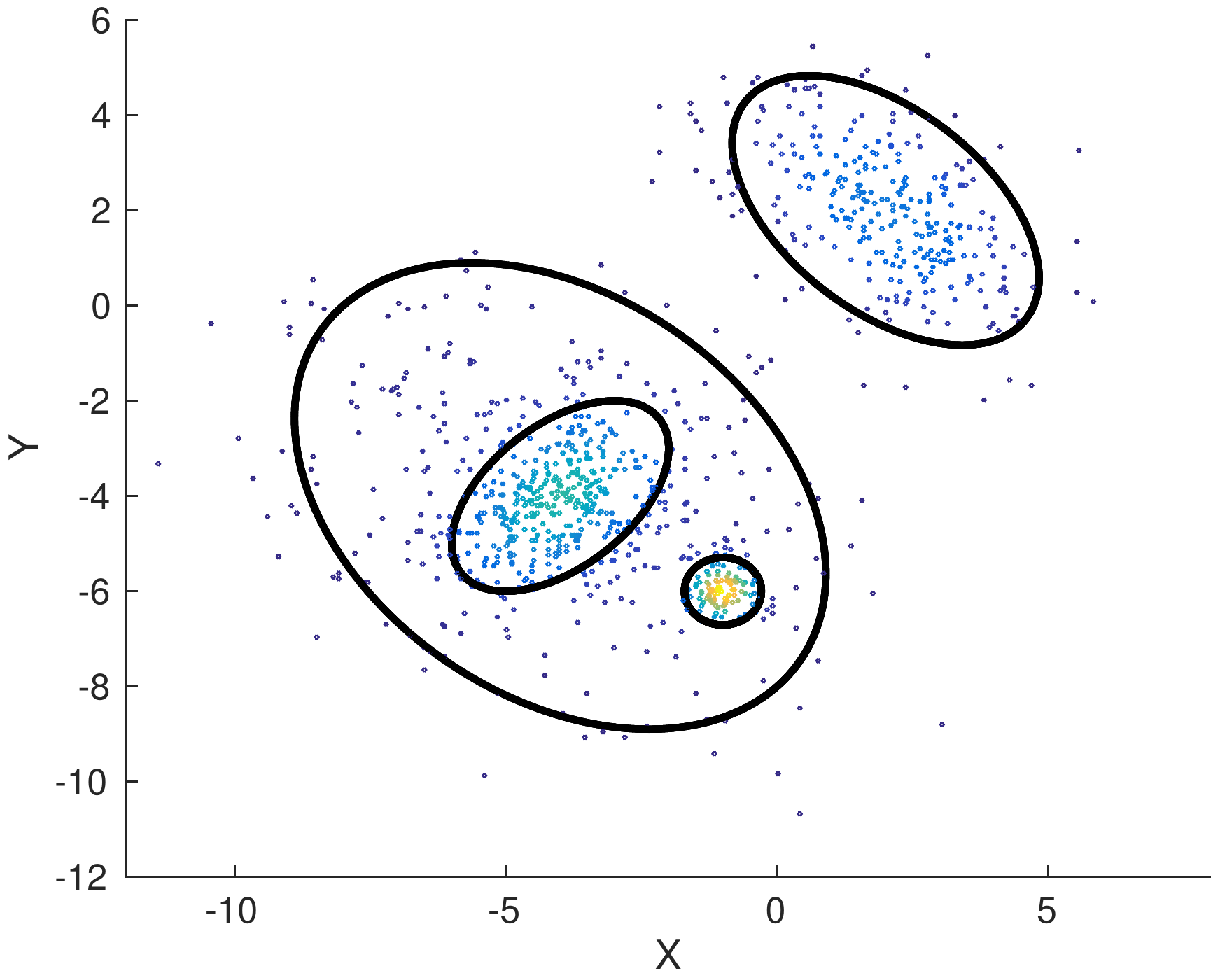}
  \caption{Original mixture consisting of overlapping components.}
  \label{fig:mix2} 
\end{wrapfigure}
We consider an example of a mixture with overlapping components and employ the
proposed search method to determine the number of mixture components. 
\cite{figueiredo2002unsupervised} considered the mixture shown in Fig.~\ref{fig:mix2}
which is described as follows: let $w_i,\boldsymbol{\mu}_i,\mathbf{C}_i, i \in \{1,4\}$
be the weight, mean, and covariance matrix of the $i^{\text{th}}$ component respectively.
Then, the mixture parameters are given as
\begin{gather*}
w_1 = w_2 = w_3 = 0.3, w_4 = 0.1\\
\boldsymbol{\mu}_1 = \boldsymbol{\mu}_2 = (-4,-4)^T, \,\boldsymbol{\mu}_3 = (2,2)^T, \,\boldsymbol{\mu}_4 = (-1,-6)^T\\
\mathbf{C}_1 =  \begin{bmatrix}
                  1   & 0.5       \\[0.3em]
                  0.5 & 1           
                \end{bmatrix},
\mathbf{C}_2 =  \begin{bmatrix}
                  6   & -2       \\[0.3em]
                  -2  & 6           
                \end{bmatrix},\,
\mathbf{C}_3 =  \begin{bmatrix}
                  2  & -1       \\[0.3em]
                  -1  & 2           
                \end{bmatrix},\,
\mathbf{C}_4 =  \begin{bmatrix}
                  0.125   & 0       \\[0.3em]
                  0  & 0.125           
                \end{bmatrix}
\end{gather*}

We generate a random sample of size 1000 (similar to \cite{figueiredo2002unsupervised}) 
and infer mixtures using the proposed heuristic.
Below are diagrams which portray the splitting of the individual
components. We show a selection of operations which result in improved mixtures.
\begin{figure}[htb]
  \centering
  \subfloat[$P_1$: one-component mixture]
  {
    \includegraphics[width=0.33\textwidth]{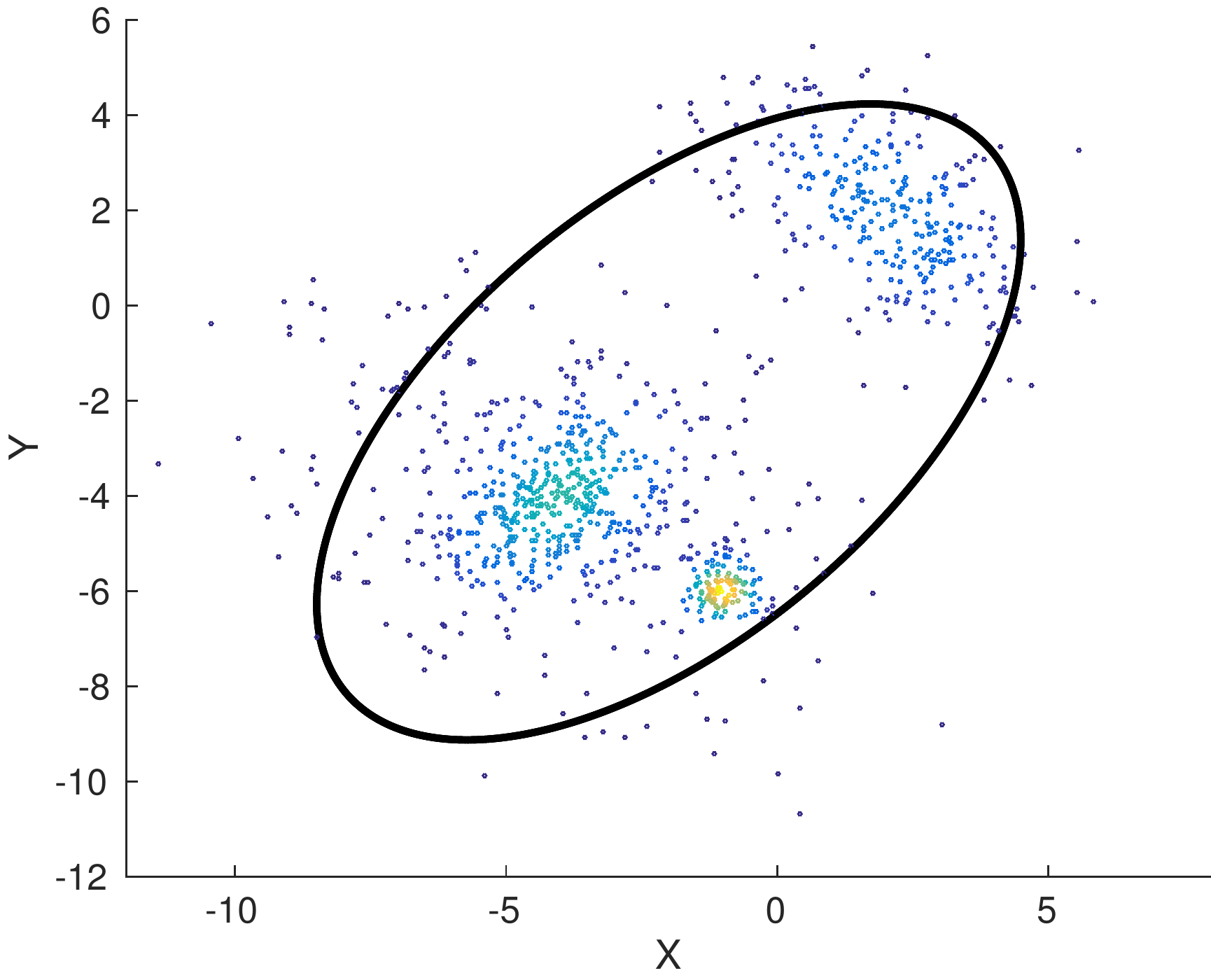}
  } 
  \subfloat[Prior to splitting: initialization of means]
  {
    \includegraphics[width=0.33\textwidth]{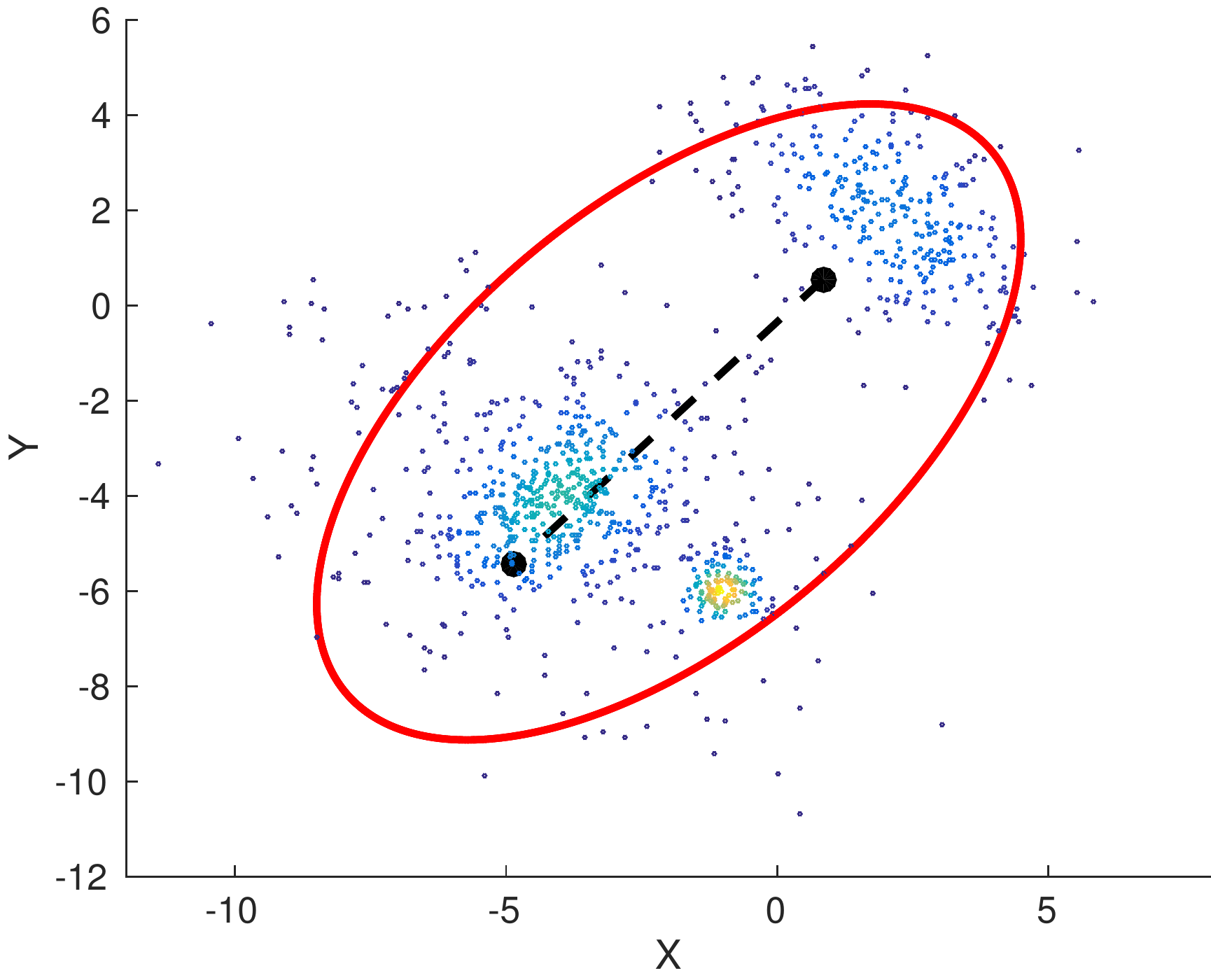}
  }
  \subfloat[$P_2$: after EM optimization]
  {
    \includegraphics[width=0.33\textwidth]{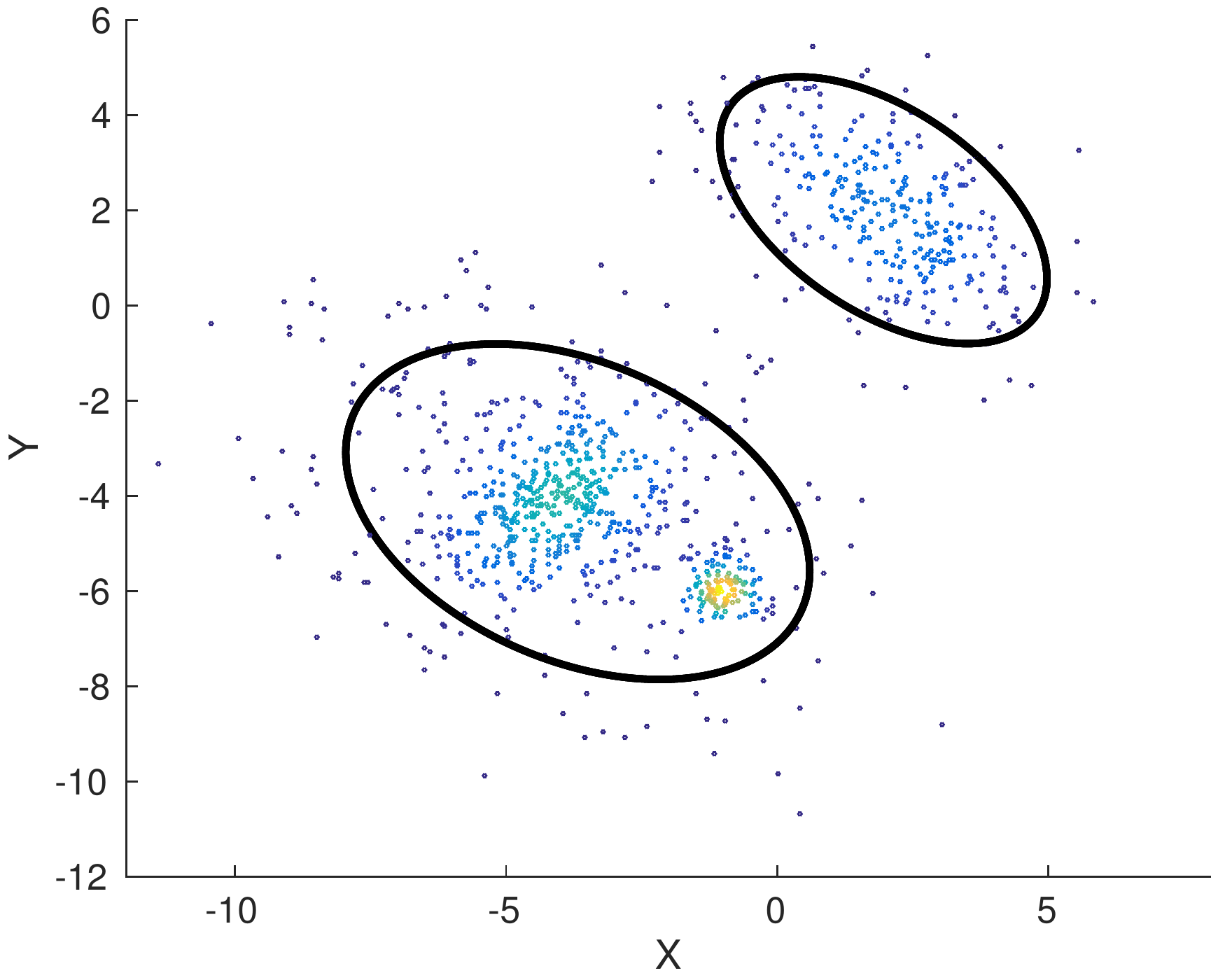}
  } 
  \caption{First iteration (a) $P_1$: the one-component mixture ($I = 27199$ bits)
           (b) Red colour denotes the component being split. The dotted line is the direction
               of maximum variance. The black dots represent the initial  means of the 
               two-component sub-mixture
           (c) $P_2$: optimized mixture post-EM ($I = 26479$ bits) results in an improvement.
          }
  \label{fig:mix_example2_iter_1_splits}
\end{figure}

\begin{figure}[htb]
  \subfloat[Splitting a component in $P_2$]
  {
    \includegraphics[width=0.33\textwidth]{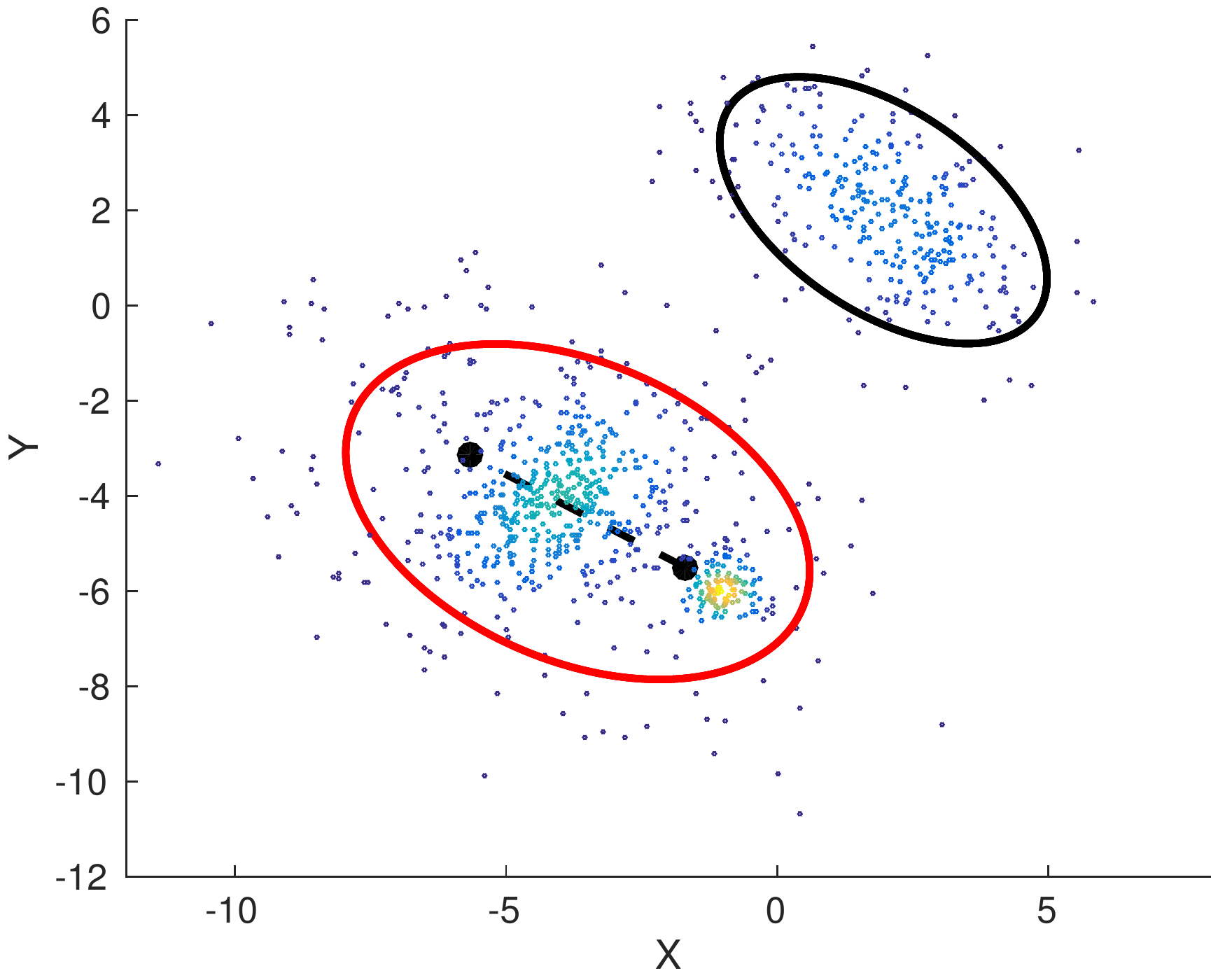}
  }
  \subfloat[Optimized child sub-mixture]
  {
    \includegraphics[width=0.33\textwidth]{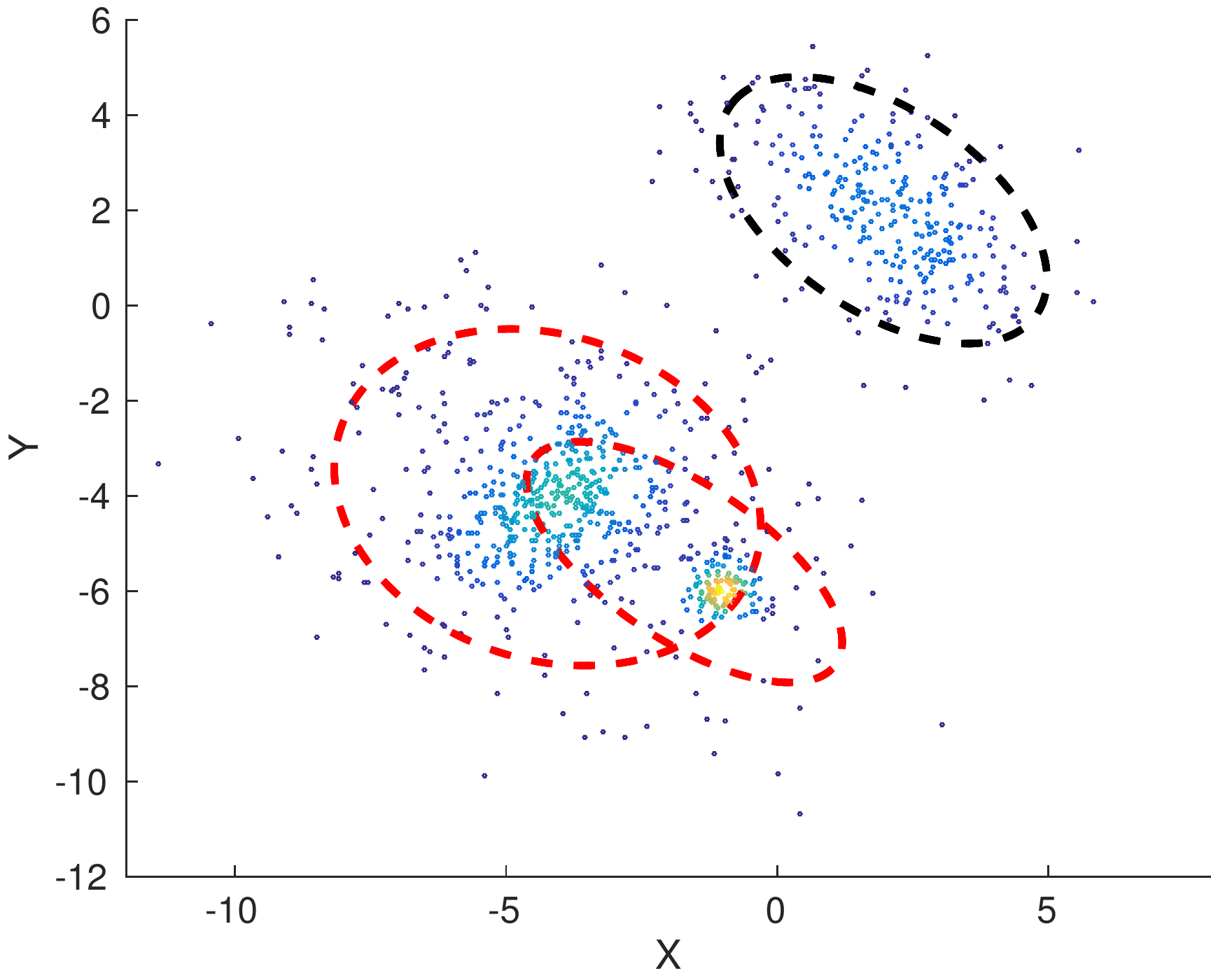}
  } 
  \subfloat[$P_3$: post-EM improved mixture]
  {
    \includegraphics[width=0.33\textwidth]{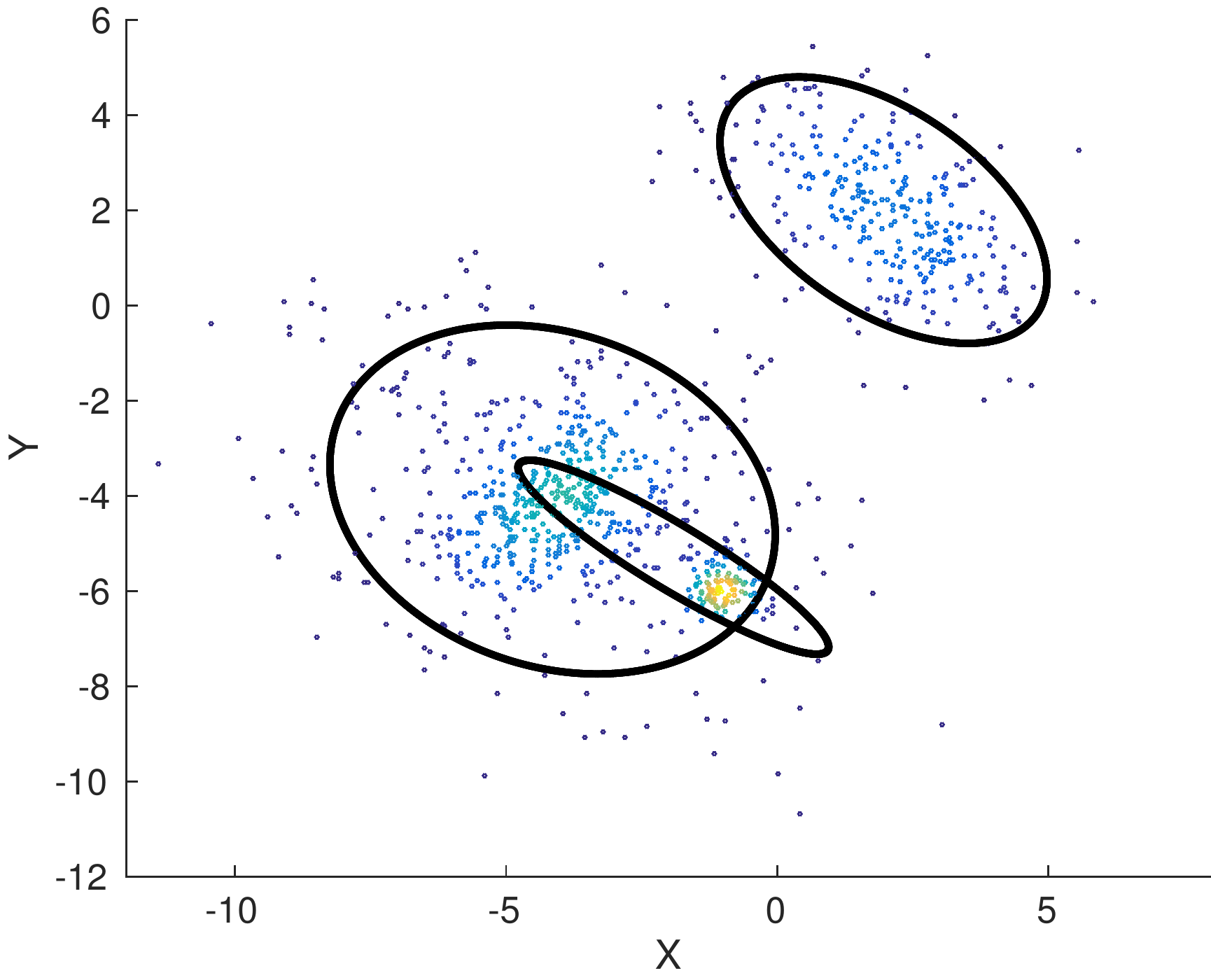}
  } 
  \caption{Second iteration: \emph{splitting} the (red) component in parent $P_2$ ($I = 26479$ bits) results in an improvement
           (a) Initial means of the child components
           (b) Optimized child mixture (denoted by red dashed lines)
               along with the second component of $P_2$ (denoted by black dashes) ($I = 26467$ bits)
           (c) $P_3$: stabilized mixture post-EM ($I = 26418$ bits) results in an improvement.
          }
  \label{fig:mix_example2_iter_2_splits}
\end{figure}

\begin{figure}[htb]
  \subfloat[Splitting a component in $P_3$]
  {
    \includegraphics[width=0.33\textwidth]{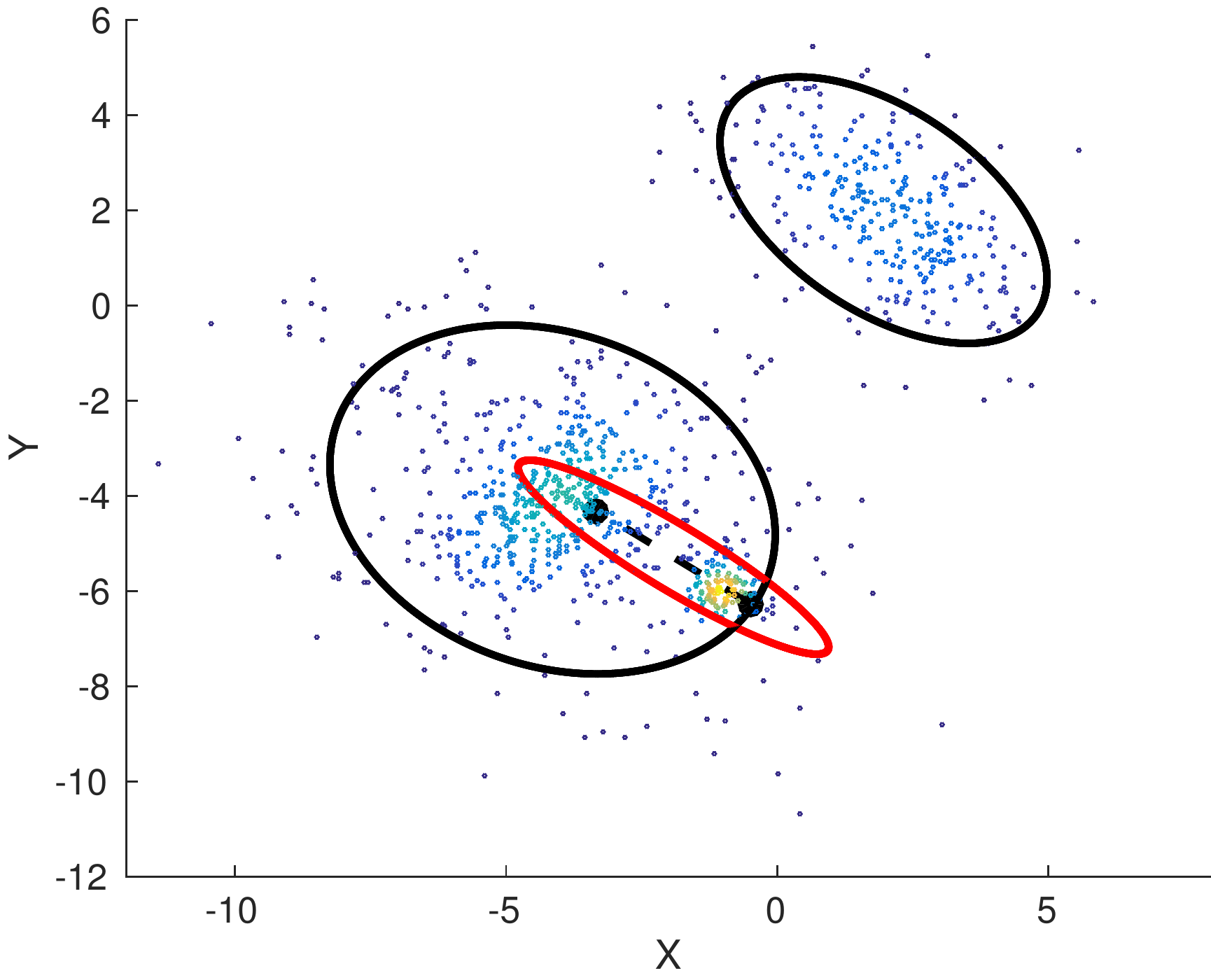}
  }
  \subfloat[Optimized children ($I = 26424$ bits)]
  {
    \includegraphics[width=0.33\textwidth]{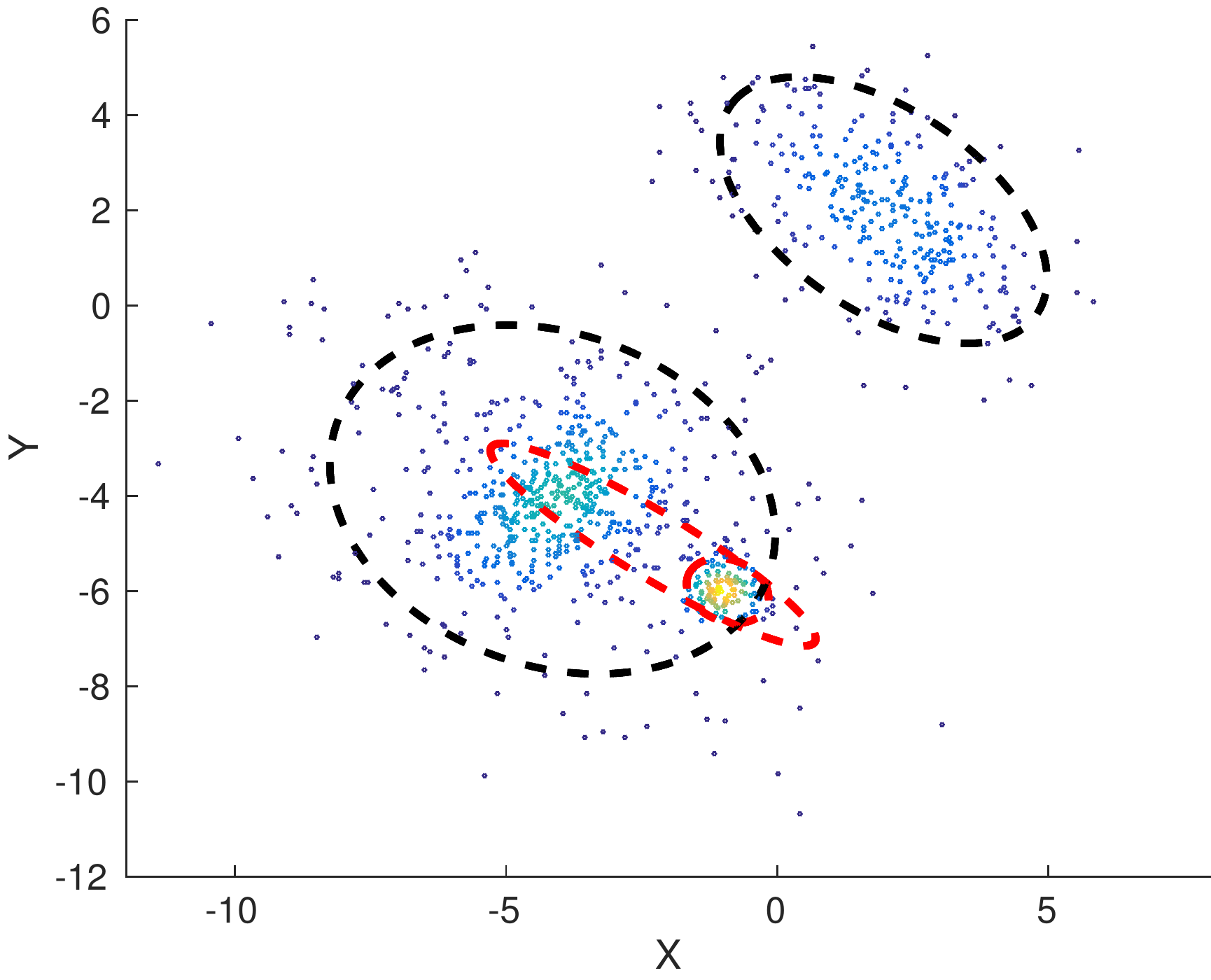}
  } 
  \subfloat[$P'_4$: an improved mixture ($I = 26358$ bits)]
  {
    \includegraphics[width=0.33\textwidth]{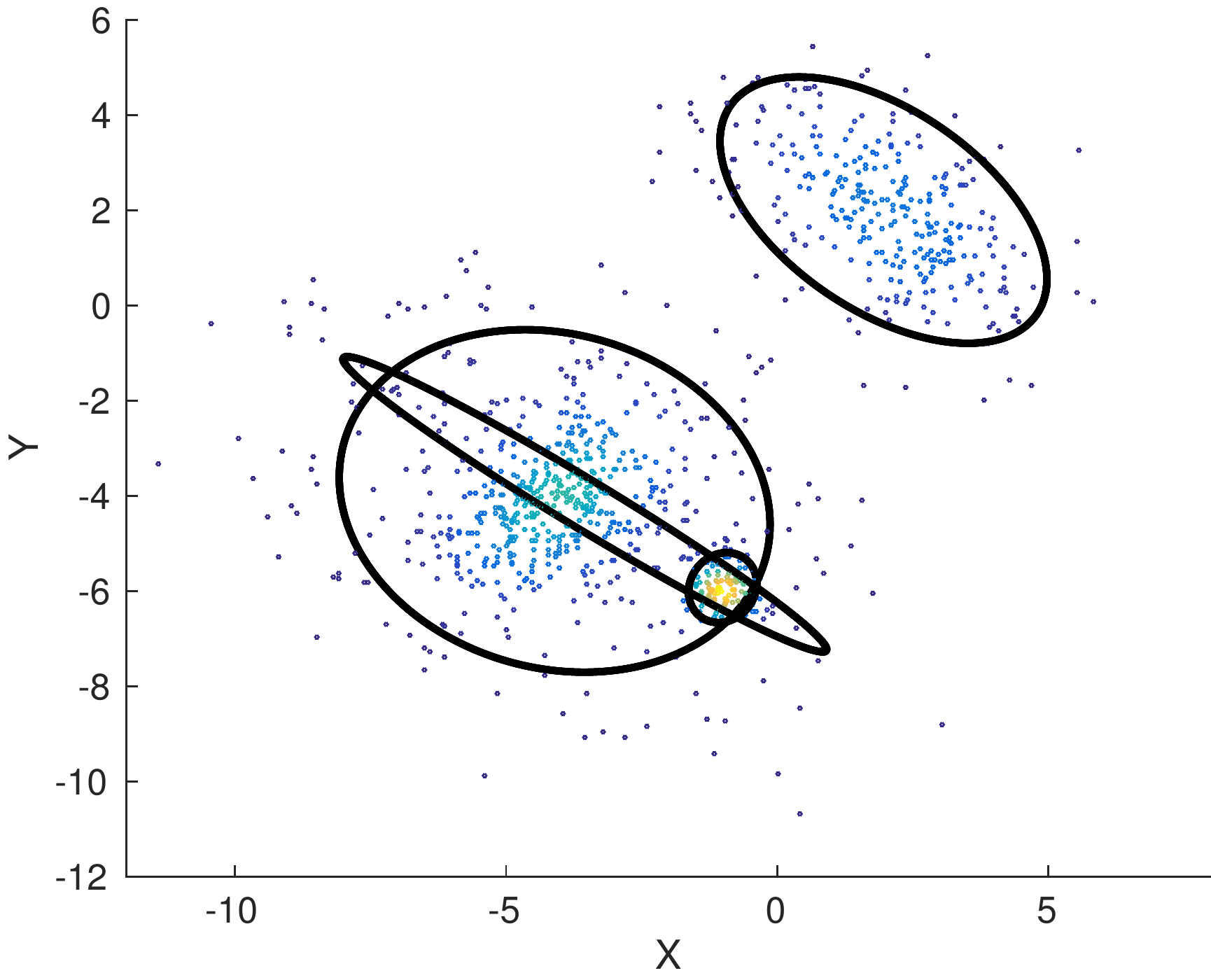}
  }\\
  \subfloat[Splitting a component in $P_3$]
  {
    \includegraphics[width=0.33\textwidth]{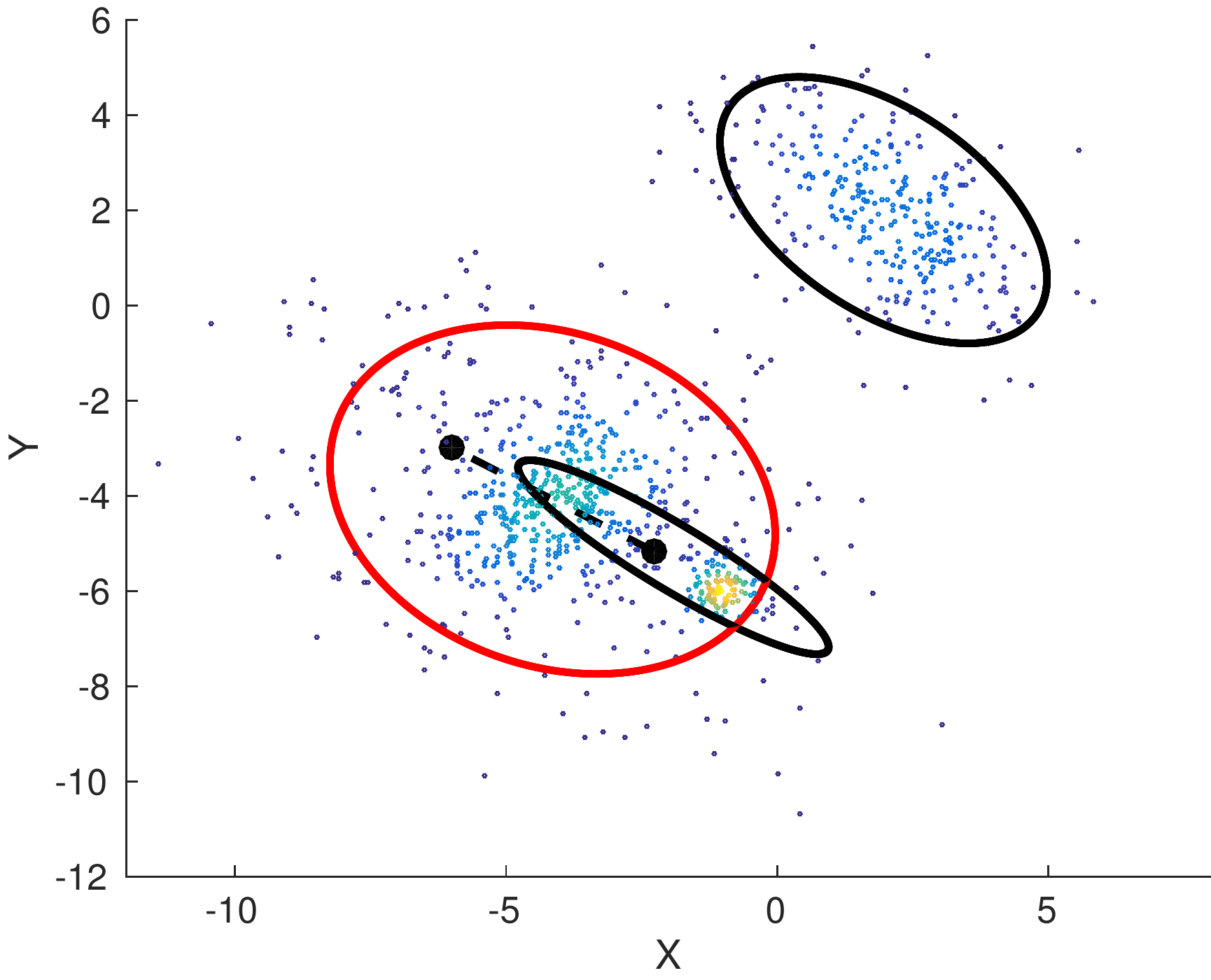}
  }
  \subfloat[Optimized children ($I = 26451$ bits)]
  {
    \includegraphics[width=0.33\textwidth]{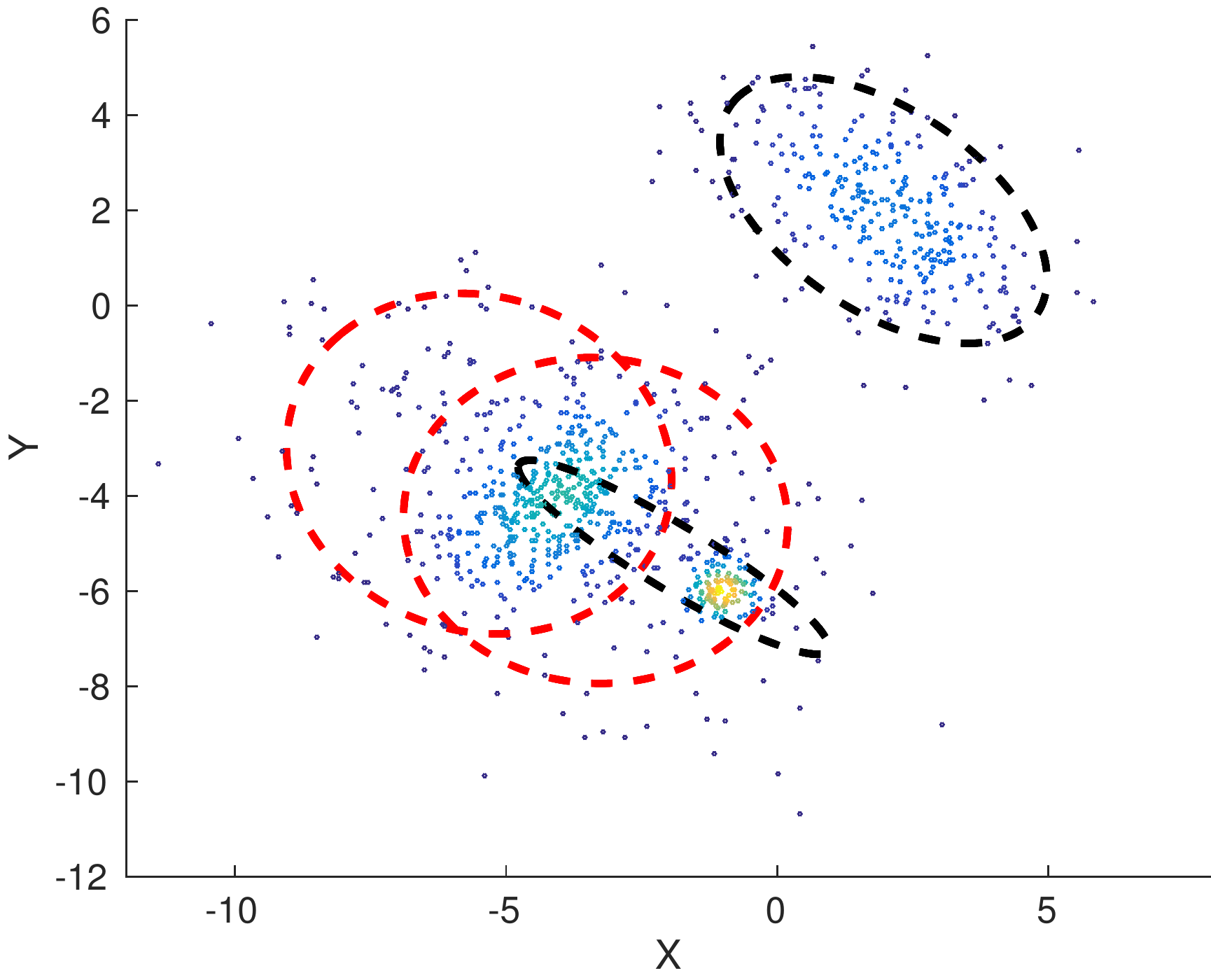}
  } 
  \subfloat[$P_4$: also an improved mixture ($I = 26266$ bits)]
  {
    \includegraphics[width=0.33\textwidth]{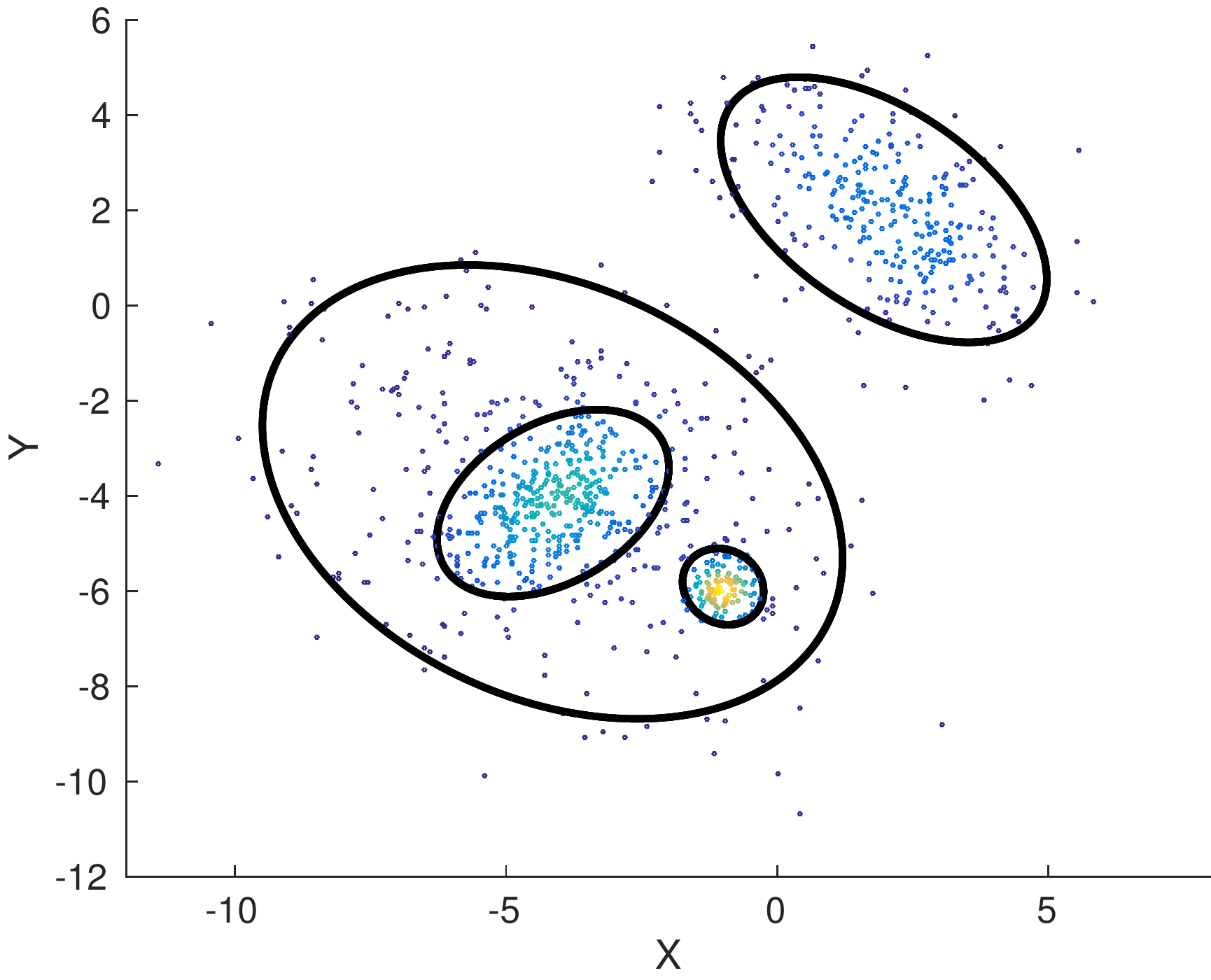}
  }
  \caption{Third iteration: \emph{splitting} the (red) components in parent $P_3$ ($I = 26418$ bits).
           We see that there are two splits which result in improved mixtures $P'_4$ and $P_4$.
           We select $P_4$ as the new parent as it has a lower message length compared to $P'_4$.
          }
  \label{fig:mix_example2_iter_3_splits}
\end{figure}

\begin{figure}[htb]
  \subfloat[$I = 26277$ bits]
  {
    \includegraphics[width=0.33\textwidth]{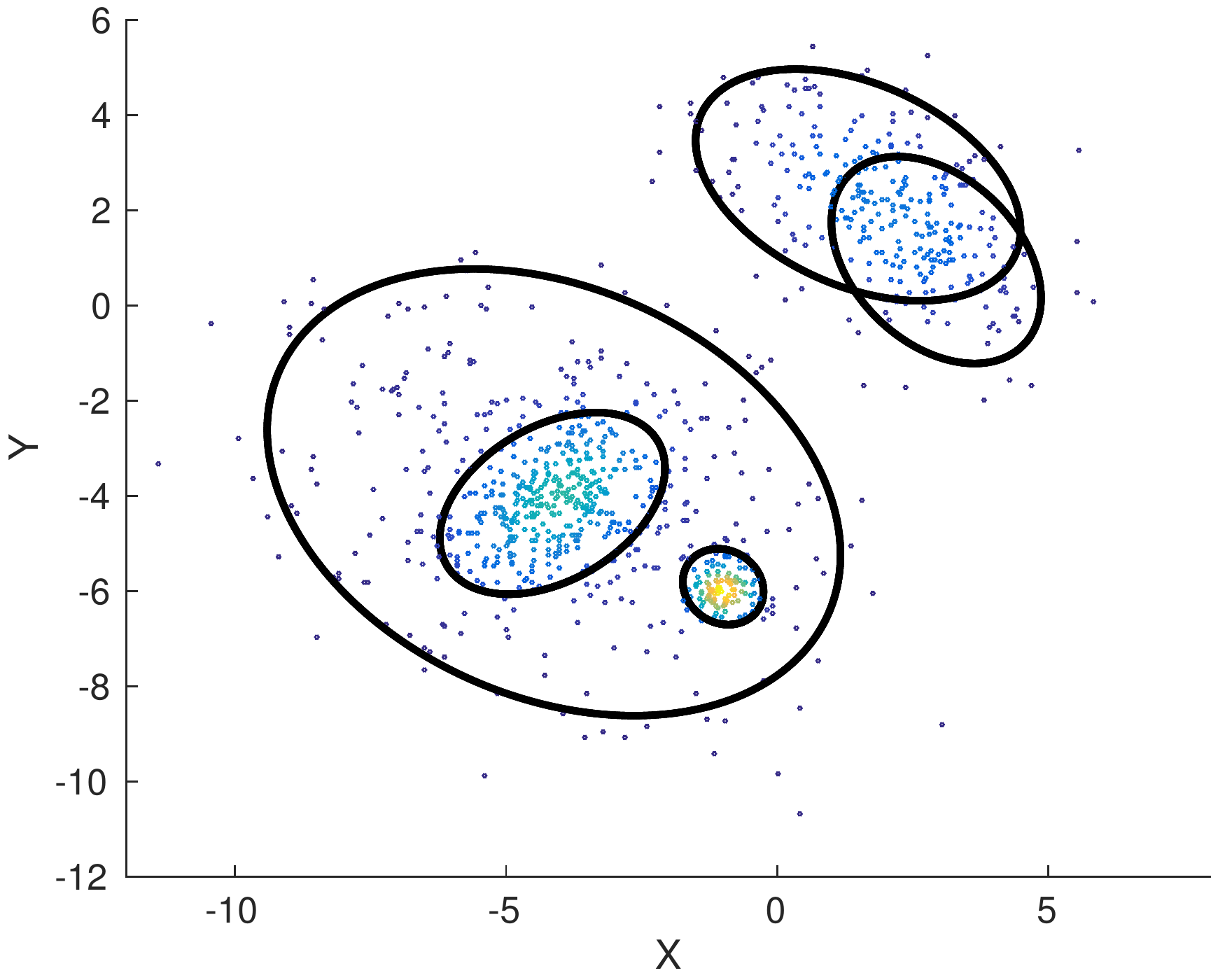}
  }
  \subfloat[$I = 26276$ bits]
  {
    \includegraphics[width=0.33\textwidth]{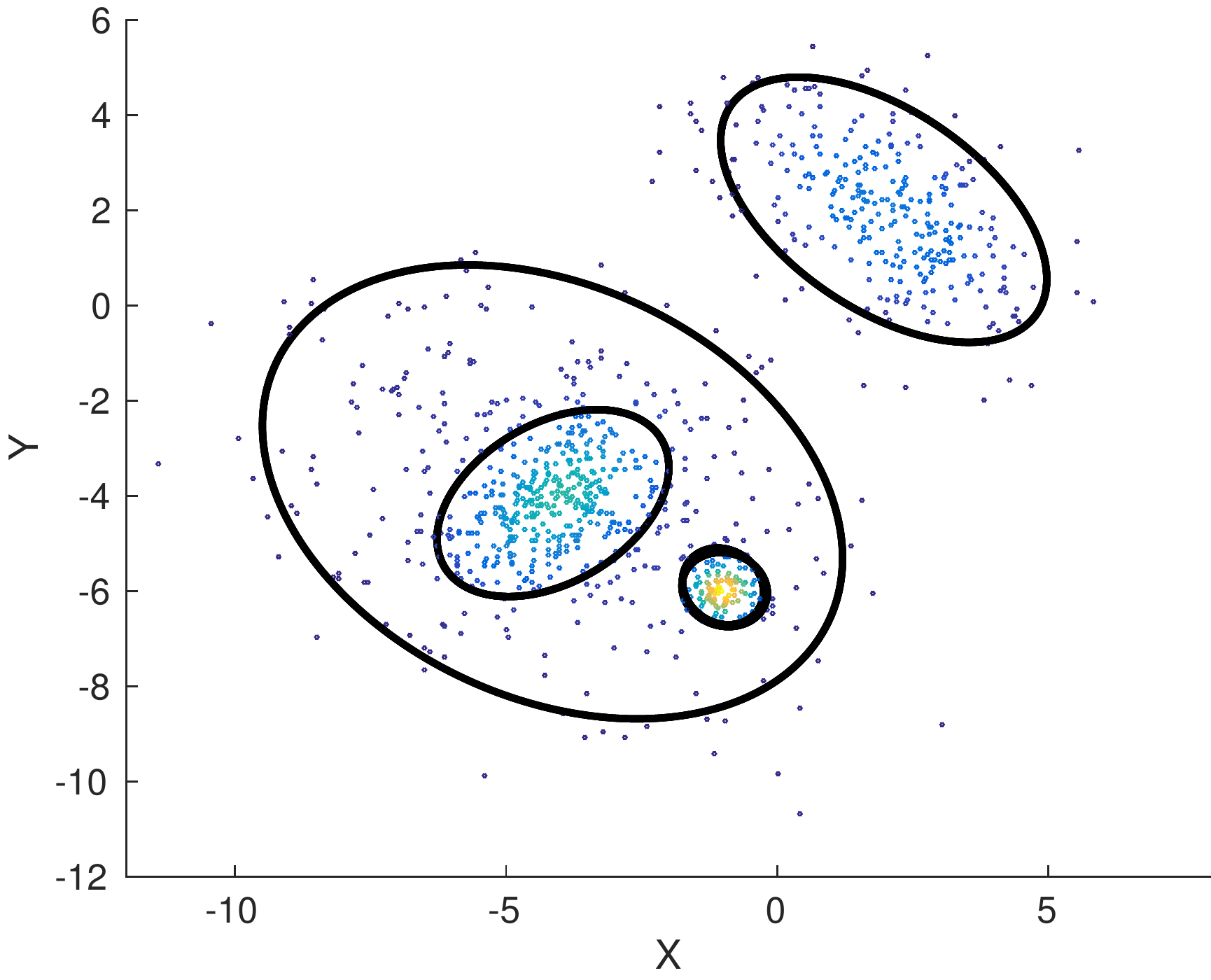}
  }\\
  \subfloat[$I = 26283$ bits]
  {
    \includegraphics[width=0.33\textwidth]{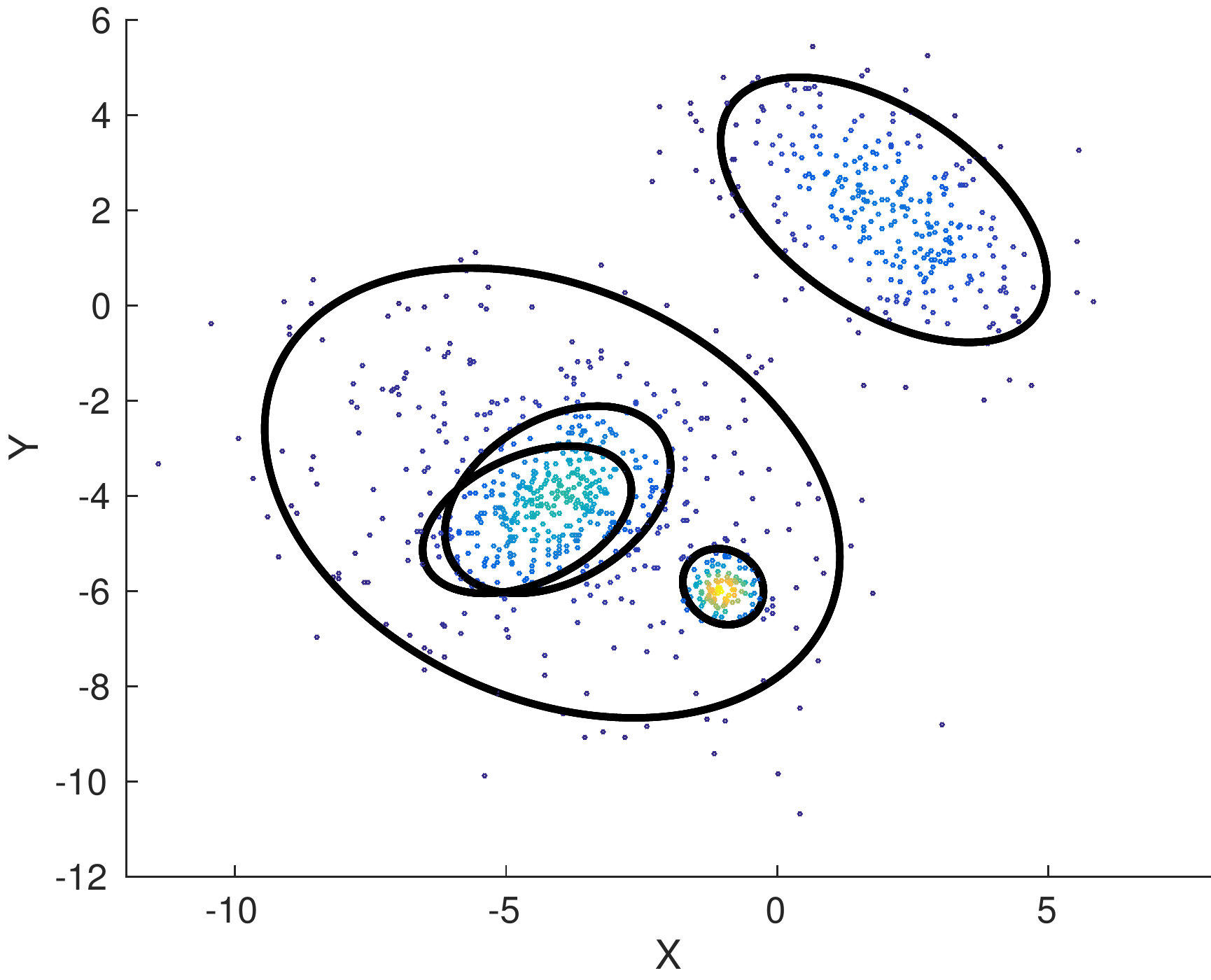}
  }
  \subfloat[$I = 26279$ bits]
  {
    \includegraphics[width=0.33\textwidth]{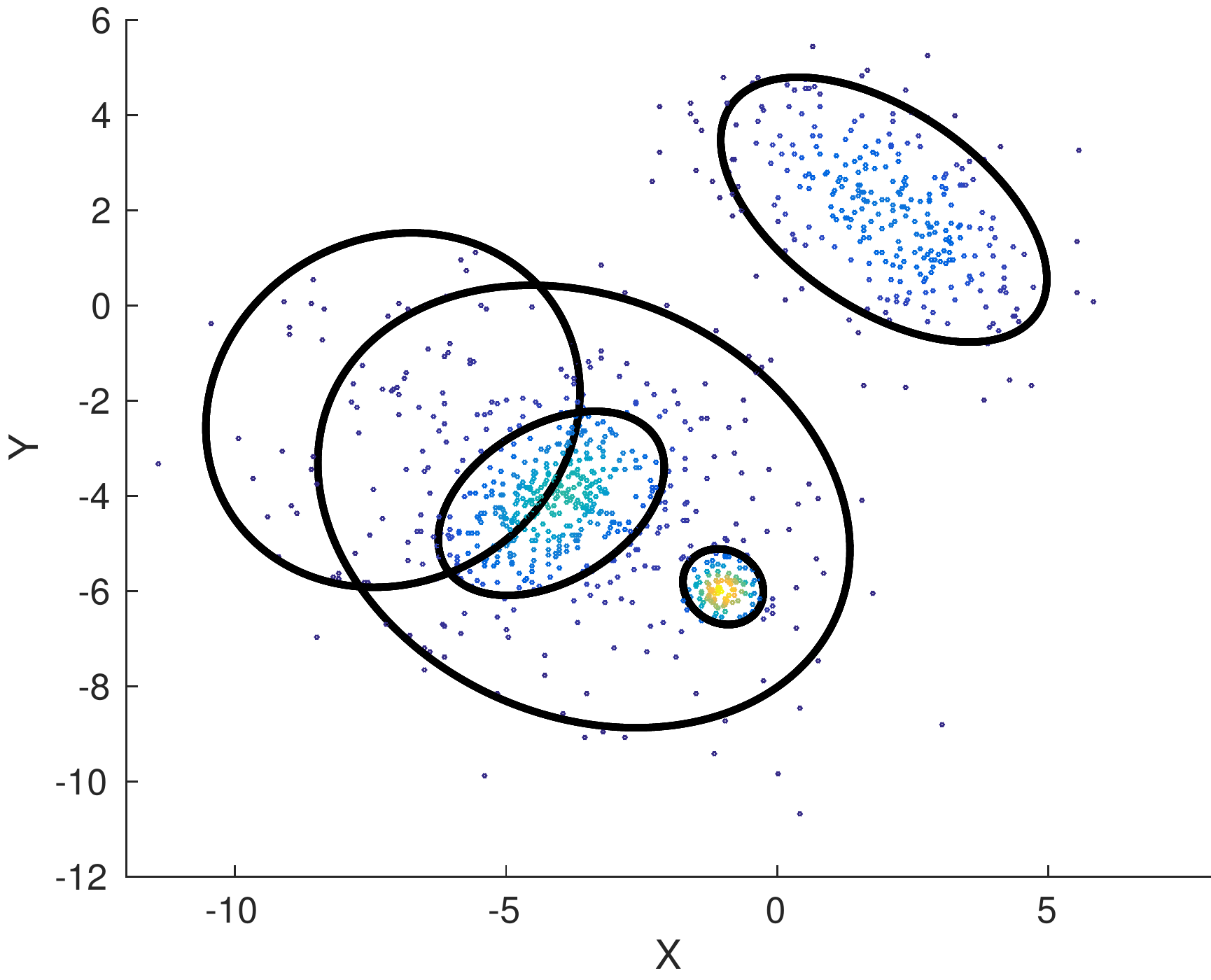}
  }
  \caption{Fourth iteration: mixtures resulting from \emph{splitting} each of the components in parent $P_4$ ($I = 26266$ bits).
           We see that none of the 5-component mixtures result in further reduction of the message length.
           The search terminated and $P_4$ is considered the best mixture.
          }
  \label{fig:mix_example2_iter_4_splits}
\end{figure}

\end{document}